\documentclass[a4paper, 11pt, oneside]{Thesis} 
\graphicspath{Figures/} 

\usepackage[square, numbers, comma, sort&compress]{natbib}  
\usepackage{verbatim} 
\usepackage{vector} 
\usepackage {tikz}
\usetikzlibrary {positioning}
\hypersetup{urlcolor=blue, colorlinks=true}  
\usepackage[numbers]{natbib}
\usepackage{bibentry}
\usepackage{algorithmic}
\usepackage[ruled,vlined]{algorithm2e}
\usepackage{subfloat}
\usepackage{multirow}
\usepackage{pdfpages}

\begin{document}
\frontmatter      % Begin Roman style (i, ii, iii, iv...) page numbering

% Set up the Title Page
\title  {Exploiting Contextual Information with Deep Neural Networks}
\authors  {\texorpdfstring
            {\href{your web site or email address}{Ismail Elezi}}
            {848027}
            }
\addresses  {\groupname\\\deptname\\\univname}  % Do not change this here, instead these must be set in the "Thesis.cls" file, please look through it instead
\date       {\today}
\subject    {}

\addresses  {\groupname\\\deptname\\\univname}  % Do not change this here, instead these must be set in the "Thesis.cls" file, please look through it instead

\maketitle
%% ----------------------------------------------------------------

\setstretch{1.3}  % It is better to have smaller font and larger line spacing than the other way round

% Define the page headers using the FancyHdr package and set up for one-sided printing
\fancyhead{}  % Clears all page headers and footers
\rhead{\thepage}  % Sets the right side header to show the page number
\lhead{}  % Clears the left side page header

\pagestyle{fancy}  % Finally, use the "fancy" page style to implement the FancyHdr headers

%% ----------------------------------------------------------------

%% ----------------------------------------------------------------
% The "Funny Quote Page"
\pagestyle{empty}  % No headers or footers for the following pages

\null\vfill
% Now comes the "Funny Quote", written in italics
\textit{Dedicated to my father, Sejdi Elezi}

\vfill\vfill\vfill\vfill\vfill\vfill\null
\clearpage  % Funny Quote page ended, start a new page
%% ----------------------------------------------------------------

% The Abstract Page
\addtotoc{Abstract}  % Add the "Abstract" page entry to the Contents

\abstract{Il contesto \'e importante! Nondimeno, sono state condotte poche ricerche sull'utilizzo dell'informazione contestuale con le neural network. In larga parte, l'utilizzo di tale informazione è stato limitato alle recurrent neural networks (RNN). Gli attention model e le capsule network sono due esempi recenti di come si possa introdurre l'informazione contestuale in modelli diversi dalle RNN, tuttavia entrambi gli algoritmi sono stati sviluppati dopo l'inizio di questo lavoro.

In questa tesi, dimostriamo come l'informazione contestuale possa essere utilizzata in due modi fondamentalmente differenti: implicitamente ed esplicitamente. Nel progetto DeepScore, dove l'utilizzo del contesto \'e veramente importante per il riconoscimento di molti oggetti di piccole dimensioni, dimostriamo di poter raggiungere lo stato dell'arte tramite la progettazione di architetture convolutive, e allo stesso tempo distinguere, implicitamente e correttamente, tra oggetti virtualmente idntici, ma con diversa semantica a seconda di ciò che sta loro intorno. In parallelo, dimostriamo che, progettando algoritmi motivati dalle teorie dei grafi e dei giochi, che prendono in considerazione l'intera struttura del dataset, possiamo raggiungere lo stato-dell-arte in task differenti come il semi-supervised e il similarity learning.

Al meglio delle nostre conoscenze, siamo i primi a integrare moduli di basati sulla teoria dei grafi, sapientemente sviluppati per problemi di similarity learning e progettati per considerare l'informazione contestuale, non solo superando in accuratezza gli altri modelli, ma anche guadagnando un miglioramento in termini di velocità, utilizzando un minor numero di parametri.
}

\abstract{Context matters! Nevertheless, there has not been much research in exploiting contextual information in deep neural networks. For most part, the entire usage of contextual information has been limited to recurrent neural networks. Attention models and capsule networks are two recent ways of introducing contextual information in non-recurrent models, however both of these algorithms have been developed after this work has started.

In this thesis, we show that contextual information can be exploited in $2$ fundamentally different ways: implicitly and explicitly. In the DeepScore project, where the usage of context is very important for the recognition of many tiny objects, we show that by carefully crafting convolutional architectures, we can achieve state-of-the-art results, while also being able to implicitly correctly distinguish between objects which are virtually identical, but have different meanings based on their surrounding. In parallel, we show that by explicitly designing algorithms (motivated from graph theory and game theory) that take into considerations the entire structure of the dataset, we can achieve state-of-the-art results in different topics like semi-supervised learning and similarity learning. 

To the best of our knowledge, we are the first to integrate graph-theoretical modules, carefully crafted for the problem of similarity learning and that are designed to consider contextual information, not only outperforming the other models, but also gaining a speed improvement while using a smaller number of parameters.
}

\clearpage  % Abstract ended, start a new page
%% ----------------------------------------------------------------

\setstretch{1.3}  % Reset the line-spacing to 1.3 for body text (if it has changed)

% The Acknowledgements page, for thanking everyone
\acknowledgements{
\addtocontents{toc}{\vspace{1em}}  % Add a gap in the Contents, for aesthetics
 I would like to thank my family, for being the most important part of my life. My late father Sejdi, who always wanted me to get a Ph.D. degree but whom will never see me getting it, my mother Kymete, my brothers Mentor, Armend and Petrit and my sister Merita. I would like to thank my sisters-in-law Alida and Ibe, and my nephews and nieces (Dian, Rina, Lejla, Anna, Samuel and the little Joel). I love you all!

I would like to give my deepest gratitude to my supervisor, professor Marcello Pelillo for pushing me to start a Ph.D. degree in the first place, and for advising me during the entire course of it. I believe that his invaluable advice has helped me become a better researcher, and see things in a different way. I would like to thank my friends and colleagues at Ca' Foscari University of Venice for helping me during the Ph.D. Thanks Ale, Marco, Leulee, Joshua, Yonatan, Mauro, Mara, Martina, Alvise and Stefano. Special thanks to Seba, whom was my primary collaborator during the entire Ph.D., and with whom I spent endless time (be it on Venice or Munich) discussing, coding, debugging and writing papers during the deadline sessions. I would like to thank Nicolla Miotello for being always helpful in all the administrative issues I had during the doctorate. I would like to thank the external reviewers, professor Marco Gori and professor Friedhelm Schwenker for giving valuable feedback on improving the thesis.

I spent a great year at Zurich University of Applied Sciences, being advised from professor Thilo Stadelmann. Thanks Thilo for having me there, and for helping me not only in the projects I was working on, but for giving me unconditional support and for being a great co-supervisor during the entire duration of my Ph.D. Thanks Thilo for volunteering to read an advanced draft of the thesis, and for giving me detailed feedback on how to improve it. I would like to thank Lukas with whom I did most of the work in DeepScore project while at ZHAW. I was lucky to have Lukas as my primary collaborator at ZHAW, and I found working with him both rewarding and enjoyable. I will never forget the time spent with the other members of the group (Mohammad, Kathy, Ana, Jonas, Mario, Melanie, Frank, Kurt, Andy and Martin).

I consider the highlight of my Ph.D. the time I spent at the Technical University of Munich. I went there (together with Seba) to work with professor Laura Leal-Taix\'{e}, a young professor I had only met twice before. Little I knew that during the next $9$ months, Laura would became my awesome supervisor, my mentor, a great friend, and beat me in table tennis, kicker, singstar, uno, codenames and bowling (though I had the last laugh in the Game of Thrones prediction game, but she does not accept it)! I will always be \textbf{\#grateful} to Laura for the opportunity she gave me to spend that time in her lab, for helping me become a better researcher, and for giving me the best supervision a student can ask for. I would also like to thank Laura's minions for making me feel part of the group. Tim, Qunjie, Patrick, Guillem, Aljosa, Aysim and Sergio, thank you guys, you rock. Special thanks to Maxim, with whom I worked a lot in a project which is not part of the thesis, but which was the most enjoyable project I have ever worked. If this group doesn't make the next big thing in computer vision, we are done! :)

Last but not least, I would like to thank Jose Alvarez for mentoring me at NVIDIA Research at Santa Clara. Together with him (and in collaboration with Laura) we are exploring an exciting project that deals with the combination of semi-supervised learning and active learning. I would like to thank the other members of the group (Akshay, Francois, Jiwoong and Maying). Finally, I like to thank Zhiding and Anima who helped me in the project, here at Nvidia.

}
\clearpage  % End of the Acknowledgements
%% ----------------------------------------------------------------

\pagestyle{fancy}  %The page style headers have been "empty" all this time, now use the "fancy" headers as defined before to bring them back

%% ----------------------------------------------------------------
\lhead{\emph{Contents}}  % Set the left side page header to "Contents"
\tableofcontents  % Write out the Table of Contents

%% ----------------------------------------------------------------
\lhead{\emph{List of Figures}}  % Set the left side page header to "List if Figures"
\listoffigures  % Write out the List of Figures

%% ----------------------------------------------------------------
\lhead{\emph{List of Tables}}  % Set the left side page header to "List of Tables"
\listoftables  % Write out the List of Tables

%% ----------------------------------------------------------------
\setstretch{1.5}  % Set the line spacing to 1.5, this makes the following tables easier to read
\clearpage  % Start a new page

\mainmatter	  % Begin normal, numeric (1,2,3...) page numbering
\pagestyle{plain}  % Return the page headers back to the "fancy" style

\chapter{Introduction}

\section{Introduction}

Since the publication of the AlexNet architecture \cite{DBLP:conf/nips/KrizhevskySH12}, deep learning \cite{DBLP:journals/nature/LeCunBH15, DBLP:journals/nn/Schmidhuber15} has been at the forefront of developments of machine learning, computer vision and artificial intelligence. The most successful class of deep learning models are undoubtedly the Convolutional Neural Networks (CNNs) conceived by \cite{DBLP:journals/pr/FukushimaM82}, developed by \cite{DBLP:journals/neco/LeCunBDHHHJ89} and revived by \cite{DBLP:conf/nips/KrizhevskySH12}. CNN-based models have been responsible for the advancements in image classification \cite{DBLP:conf/cvpr/HeZRS16}, image segmentation \cite{Long2015}, image recognition \cite{DBLP:conf/nips/RenHGS15} and many other computer vision applications \cite{Detectron2018}. The advantage of CNNs compared to more traditional machine learning techniques (especially applied to the task of computer vision) is that they are designed to be very good at feature extraction specifically for spatially correlated information like pixels in natural images. Additionally, CNN models are designed in such a way as to optimize the feature extraction and the task at hand (for example classification) all-together in an end-to-end fashion.

This thesis is heavily based on CNNs, and each chapter of it involves novel extensions of CNNs for different tasks of computer vision (classification, segmentation, detection, recognition, similarity learning, retrieval and clustering). We show the shortcomings of current usage of CNNs, and improve over them by either incorporating special 'contextual' modules, or carefully designing CNNs to implicitly exploit the context for the task at hand.

\section{The importance of contextual information}

\subsection{Explicit context}

It has been widely known that the usage of contextual information for machine learning tasks like classification is very important \cite{DBLP:journals/pami/HummelZ83}. The decisions on the classification of objects should not be dependent only in the local features, but also on the global information of the dataset (the similarity between objects). Nevertheless, the majority of deep learning algorithms ignore context and process the data observations in isolation. For more than two decades, the only clear usage of context in neural networks has been limited to Recurrent Neural Networks (RNNs) \cite{DBLP:journals/cogsci/Elman90}, a type of neural networks which take into considerations previous (and with modifications, future) samples, making them theoretically very suitable for the processing of sequences. However, there was a misguided belief that RNNs are hard to be trained because of the vanishing gradient problem which has a mathematical nature \cite{Hochreiter:90}. Despite that the problem was partially solved \cite{hochreiter1997long} by designing sub-modules in RNN cells (called gates), the usage of RNNs has been mostly limited in problems where the nature of the data is not sequential.

There have been attempts at combining CNNs with RNNs \cite{KarpathyThesis}, however these attempts have happened mostly when the task at hand had as inputs both images and sequences (like language). In cases where the input was not sequential (like many computer vision applications) the entire context is typically provided by the \textit{average} operator in the loss function. Training samples do not interect with each other, and the resulting loss function is purely based on local information.

For completeness, it needs to be said that during the course of this doctorate, there have been parallel works in integrating non-recurrent based context-aware modules in CNNs. Three such attempts have been \textit{attention mechanisms} \cite{DBLP:conf/nips/VaswaniSPUJGKP17}, \textit{capsule networks} \cite{DBLP:conf/nips/SabourFH17} and graph neural networks \cite{DBLP:journals/corr/abs-1806-01261}. The work there has been done in parallel, and can be seen as complementary to this research. At the same time, it shows that researchers are giving more considerations to the usage of context. In this thesis, any time we exploit context by using a context mechanism, we call it \textit{explicit context}.

\subsection{Implicit context}

Initially, it is believed that regular neural networks consider each sample in isolation. This belief was challenged when researchers started using stochastic gradient descent instead of gradient descent \cite{DBLP:series/lncs/LeCunBOM12}. For example, shuffling the training set during each epoch results in a better training performance than setting the order of samples in a deterministic way, with the worst performance being achieved when the order of the samples is given by class (first all elements of the first class, then the elements of the second class and so on). Clearly, despite the neural network having no designed mechanism to consider the contextual information, the network still insists to do so. The only operation that considers more than a sample in isolation is that of the average (or sum) applied at the end of the final loss, but even in case of total stochastic gradient descent (when only one sample is given in each mini-batch), the network still performs better when there is a stochastic order of samples. 

Even more interesting is the behavior of networks in the task of object recognition. Despite that most object detection models do not have designed mechanisms for context, they are still able (up to some degree) to give different predictions for the same object. This was first observed in \cite{DBLP:journals/corr/abs-1808-03305} where the authors made many toy experiments by copy-pasting an object in different images, and looking for the network predictions. A fridge in the kitchen gets classified correctly as a fridge, but if you copy and paste it into the sky, it gets classified as an airplane or a bird. While this might look a simple exciting but not useful experiments, it has clear consequences and can be exploited in different fields. For example, in the field of active semi-supervised learning, a similar strategy has been used to find the most informative samples. In \cite{DBLP:conf/cvpr/WangYZ0L18} the authors copied and pasted detected objects in different images that contain other objects. If the prediction for the same object were the same, then those objects were given a pseudo-label. On the other hand, if the prediction of the objects did not match, then those objects were considered hard, and needed a human oracle. In each case, it is clear that the surrounding of the objects play an important part in the classification score of a detected bounding box. Despite that the networks have no context mechanism, the convolution and pooling operators find a way of learning about the context. In this thesis, we call this type of context as \textit{implicit context}.

During the course of the doctorate, I was involved in DeepScore project \cite{DBLP:conf/icpr/TuggenerESPS18, DBLP:conf/ismir/TuggenerESS18, DBLP:journals/corr/abs-1810-05423} with the goal of solving the problem of object detection of musical objects. The problem is challenging because the number of objects in each musical sheet is orders of magnitudes higher than the number of objects in natural images. Traditional object detectors simply do not work. And even more challenging is the fact that different types of symbols might have an identical appearance (e.g. augmentation dot and staccato). Even if we have a perfect detector that finds the correct bounding box, it can not classify correctly the object inside it (in this case it will classify each object either as staccato or as augmentation dot). Adding contextual blocks is a possibility, but they are both expensive and it is not clear how they can be used in this problem. We found out that the easiest solution would be to design a new detector, which is an one stage detector.  By taking into consideration some simple intuition, in a single pass, it will both find the bounding box that surrounds an object and classify the object. In this way, our new detector is able to leverage the context in order to do efficient musical symbol recognition.

\subsection{Contributions}

The main contributions of this thesis are the following:

\begin{description}
  \item $\bullet$ Guided by the belief that context is important, we use a graph theoretical inspired module (which considers the entire structure of the dataset) as a pre-processing step in the training of CNNs for image classification where there is a lack of labelled data. 
  
  \item $\bullet$ We show that the mentioned graph theoretical module is differentiable, and inspired from it, we design a novel loss function for the task of similarity learning (Siamese Neural Networks). We call this loss function "Group Loss" and show that it has better properties than traditional loss functions used in Siamese architectures \cite{bromley1994signature}, while also achieves significantly better results.
  
  \item $\bullet$ We create one of the largest datasets (called DeepScores \cite{DBLP:conf/icpr/TuggenerESPS18}), specially tailored for the task of optical music recognition (OMR). 
  
  \item $\bullet$ Knowing that the OMR problem is very different from the task of natural image recognition, we design and implement a new CNN-based module which we call "Deep Watershed Detector" that achieves state-of-the-art results on DeepScores and other musical datasets. In the OMR problem, the context is very important (the objects' class is dependent in the surroundings, and identically looking objects might have different classes), so the design of the network architecture and its loss functions is carefully tailored to incorporate the usage of context.
\end{description}

\section{Papers of the author}

This thesis is mostly based in the following papers done during the course of the doctorate. The first two papers contain the part about the explicit usage of contextual information and are the core of the thesis:

\begin{description}
  \item {\color{blue} \textbf{Ismail Elezi*}, Alessandro Torcinovich*, Sebastiano Vascon* and Marcello Pelillo; \textit{Transductive label augmentation for improved deep network learning \cite{DBLP:conf/icpr/EleziTVP18}}; In Proceedings of IAPR International Conference on Pattern Recognition (ICPR 2018)} which deals with label augmentation for convolutional neural networks, performed by designing a pipeline which combines Graph Transuction Game (GTG) \cite{DBLP:journals/neco/ErdemP12} with CNNs. An extended version of the paper (containing many more experiments and comparisons) is given in \textbf{Chapter 3} and we achieve state-of-the-art results in the task of semi-supervised deep learning in cases where there are only a few labelled examples. The work can be considered as a first step on combining GTG with CNNs in an end-to-end manner. 
  
    \item {\color{blue} \textbf{Ismail Elezi}, Sebastiano Vascon, Alessandro Torcinovich, Marcello Pelillo and Laura Leal-Taix\'e; \textit{The Group Loss for Deep Metric Learning \cite{DBLP:journals/corr/groupLoss}}; submitted to European Conference on Computer Vision (ECCV 2020)} which deals with combining a graph trasduction inspired module in convolutional neural networks in an end-to-end manner for the task of similarity learning. An extended version of the paper (containing extra robustness analysis, different backbones, further comparisons and extra implementation details) is given on \textbf{Chapter 4} and we achieve state-of-the-art results in the task of similarity learning (image retrieval). The work presented there is the most important contribution of the thesis.
\end{description}  

The following four papers contain the part about the implicit usage of contextual information, where the author contributed a significant part of the work:

\begin{description}  
  \item {\color{blue} Lukas Tuggener, \textbf{Ismail Elezi}, J\"urgen Schmidhuber, Marcello Pelillo and Thilo Stadelmann; \textit{DeepScores-a dataset for segmentation, detection and classification of tiny objects \cite{DBLP:conf/icpr/TuggenerESPS18}}; In Proceedings of IAPR International Conference on Pattern Recognition (ICPR 2018)} describes the process of creating one of the largest computer vision datasets, with focus on musical symbols. An extended version of the paper is given in \textbf{Chapter 5}.
  
  \item {\color{blue} Lukas Tuggener, \textbf{Ismail Elezi}, J\"urgen Schmidhuber, Thilo Stadelmann; \textit{Deep watershed detector for music object recognition \cite{DBLP:conf/ismir/TuggenerESS18}}; In Proceedings of Conference of the International Society for Music Information Retrieval (ISMIR 2018)} describes the development of a convolutional-based end-to-end model for the task of optical music recognition. The work is described in \textbf{Chapter 6} and is the core work of the chapter.
  
  \item {\color{blue} Thilo Stadelmann, Mohammadreza Amirian, Ismail Arabaci, Marek Arnold, Gilbert François Duivesteijn, \textbf{Ismail Elezi}, Melanie Geiger, Stefan L\"orwald, Benjamin Bruno Meier, Katharina Rombach, Lukas Tuggener; \textit{Deep Learning in the Wild \cite{DBLP:conf/annpr/StadelmannAAADE18}}; In Proceedings of IAPR TC3 Workshop on Artificial Neural Networks in Pattern Recognition (ANNPR 2018)} describes a collection of industrial projects where deep learning has been used with the focus on explaining the difficulties of using deep learning for real world applications. The author contributed to this paper on the section describing difficulties on the DeepScore project. The work is described on \textbf{Chapter 6}.
  
  \item {\color{blue} \textbf{Ismail Elezi*}, Lukas Tuggener*, Marcello Pelillo, Thilo Stadelmann; \textit{DeepScores and Deep Watershed Detection: current state and open issues \cite{DBLP:journals/corr/abs-1810-05423}}; in The International Workshop on Reading Music Systems (WoRMS 2018) (ISMIR affiliated)}, describes the improvement of both the DeepScores dataset and the Deep Watershed Detector. An extended version of the short paper is given in \textbf{Chapter 6} where among others, it does a comparison with state-of-the-art models, showing considerable improvement.
\end{description}

The following paper is thematically related to The Group Loss paper \cite{DBLP:journals/corr/groupLoss}, explicitly using context, however the context is provided via Recurrent Neural Networks. For this reason, the paper is given in \textbf{Appendix A}:

\begin{description}  
  \item {\color{blue} Benjamin Bruno Meier, \textbf{Ismail Elezi}, Mohammadreza Amirian, Oliver D\"urr and Thilo Stadelmann; \textit{Learning neural models for end-to-end clustering \cite{DBLP:conf/annpr/MeierEADS18}}; In Proceedings of IAPR TC3 Workshop on Artificial Neural Networks in Pattern Recognition (ANNPR 2018)} describes an end-to-end clustering framework using residual bi-directional long short term memory networks.
\end{description}

The following paper was published during the doctorate, and is an extension of the author's master thesis:

\begin{description}  
  \item {\color{blue} Marcello Pelillo, \textbf{Ismail Elezi} and Marco Fiorucci; \textit{Revealing structure in large graphs: Szemeredi's regularity lemma and its use in pattern recognition \cite{DBLP:journals/prl/PelilloEF17}}; Pattern Recognition Letters (PRL 2017)} describes the usage of the regularity lemma in the context of graph summarization.
\end{description}

The work is only loosely connected to the work done in the doctorate and so it has been omitted from this thesis.

The following paper was published after the thesis' submission, and is not part of the thesis:

\begin{description}  
  \item {\color{blue} Maxim Maximov*, \textbf{Ismail Elezi*} and and Laura Leal-Taix\'e; \textit{CIAGAN: Conditional identity anonymization generative adversarial networks \cite{DBLP:journals/corr/abs-2005-09544}}; IEEE/CVF Computer Vision and Pattern Recognition (CVPR 2020)} describes a novel algorithm for face and body anonymization.
\end{description}

The following paper was done after the thesis' submission, with the author having a secondary role. The work is not part of the thesis:

\begin{description}
\item {\color{blue} Jiwoong Choi, \textbf{Ismail Elezi}, Hyuk-Jae Lee, Clement Farabet, Jose Alvarez; \textit{Deep Active Learning for Object Detection with Mixture Density Networks \cite{jiwoong}}; submitted to
Advances in Neural Information Processing Systems (NeurIPS 2020)} proposes a novel active learning method for the task of object detection.
\end{description} 

\section{How to read this thesis}

The first two chapters of this thesis introduce the problem and give the minimal and necessary background information in order to be able to read the remaining part of this thesis. Then this thesis gets separated into two different branches, which are independent from each other. The first and most important branch is that of the usage of context given in an explicit manner, where the context is given via a graph theoretical module called Graph Transduction Game (GTG) \cite{DBLP:journals/neco/ErdemP12}. We show that using GTG, we can significantly improve the results of classifications from CNNs where there is a lack of labelled data. Later, we show that the same algorithm can be put as a building block on top of the neural network, and combined with cross-entropy we create a new loss function (called group loss) which outperforms state-of-the-art methods on a wide range of image retrieval datasets. Thematically related with this problem, we develop a new clustering algorithm, which shows promising results in relatively simple datasets. The work there is described in Appendix A.

The second part of the thesis is fundamentally different and deals with the implicit usage of context in deep neural networks. While in the first part, we needed to give context-specific blocks, here by carefully designing CNN architectures and loss functions, we build a new object detector called Deep Watershed Detector, that is able to detect and recognize tiny symbols for the task of optical music recognition.

\begin{figure}
  \centering
  \includegraphics[width=1.0\columnwidth]{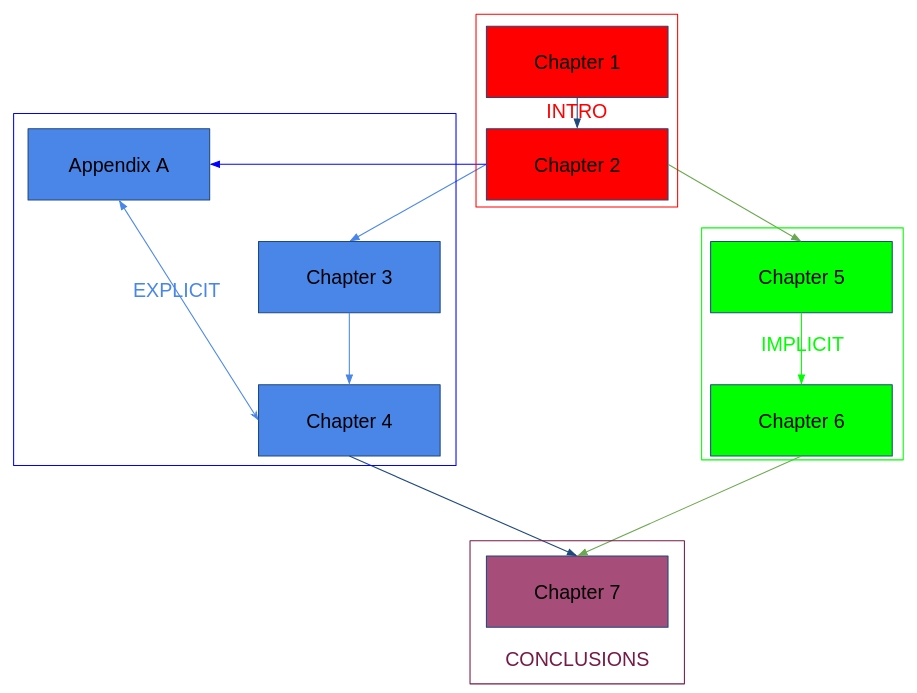}
  \caption{Thesis structure. The first two chapters provide the necessary information to read the remaining part of the thesis. Chapters 3 and 4 deal with explicit usage of the context using game and game theoretical models in deep learning, while chapters 5 and 6 deal with implicit usage of the context in convolutional neural networks for object recognition. Appendix A is related to Chapter 4 as they address similar problems, however they can be read separately considering that they use totally different ways of solving the problem. The part on explicit context is independent from the part of implicit context, and can be read independently.}   
  \label{fig:thesis}
\end{figure}

We conclude this thesis with Chapter 7, where we briefly summarize the work and show that the usage of context (be it implicit or explicit) is a very important step in building modern neural networks, and give directions to future research. A detailed graph of the structure of the thesis is given in Fig. \ref{fig:thesis}, as are given the dependencies of the chapters.

\chapter{Fundamentals of Deep Learning}

This chapter provides a brief description of machine learning and deep learning in order to make the thesis relatively self-sustainable. For a more thorough and slower-paced introduction we recommend the Deep Learning book \cite{DBLP:books/daglib/0040158}.

\section{Fundamentals of machine learning}

There are many problems (i.e image classification, speech recognition etc) where it is not clear how they can be solved via conventional computer programs. However, at the same time it is quite straightforward to collect a large number of examples, and to label them. In these cases, it can be both desirable and useful to use \emph{learning }in order to project some mappings between the input (data) and output (labels). 

There are several forms of learning, including supervised learning, unsupervised learning, semi-supervised learning and reinforcement learning. This thesis uses all forms of learning bar the last one.

\textit{Supervised learning} is the machine learning task of learning a function that maps an input to an output based on example input-output pairs. It infers a function from labeled training data consisting of a set of training examples. It is by far the most common type of learning in machine learning, and examples of it are image classification, image recognition, image segmentation, machine translation etc.

\textit{Unsupervised learning} is a type of learning that helps find previously unknown patterns in datasets without pre-existing labels. The most common examples of unsupervised learning are clustering, dimensionality reduction and image generation.

\textit{Semi-supervised learning} is the middle ground between supervised and unsupervised learning. In this type of learning, the majority of data do not have labels, but some of the data have labels, and the task of the learning is to propagate the labels from the labeled data to the unlabelled one. 

Considering that the majority of the work in machine learning, deep learning and this thesis is done in supervised learning, we give a more complete description of it.

\subsection{Supervised Learning}

Let $X$ be the data, $Y$ be the set of labels, and $D$ be the data distribution over $X \times Y$ that describes the data that we tend to observe. For every
sample $(x, y)$ from $D$, the variable $x$ is a typical input and $y$ is the corresponding (possibly noisy) desired output. The goal of supervised learning is to use a training set consisting of $n$ i.i.d. samples, $S = {(x_i, y_i)}_{i=1}^n \sim D^n$ in order to find a function $f : X \mapsto Y$ whose test error

\begin{equation}
Test_D(f) = \textbf{E}_{(x,y) \sim D} [L(f(x); y)] 
\end{equation}

is as low as possible. Here $L(z; y)$ is a loss function that measures the loss that we suffer whenever we predict $y$ as $z$. Once we find a function whose test error is small enough for our needs, the learning problem is solved.

Although it would be ideal to find the global minimizer of the test error

\begin{equation}
    f^* = argmin_f \: Test_D(f)
\end{equation}

doing so is fundamentally impossible. We can approximate the test error with the training error

\begin{equation}
Train_S(f) = \textbf{S}_{(x,y) \sim D} [L(f(x); y)] 
\end{equation}

(where we define $S$ as the uniform distribution over training cases counting duplicate cases multiple times) and find a function $f$ with a low training error. Given a model with large capacity, it is trivial to minimize the training error
by memorizing the training cases, which is very undesirable. Making sure that good performance on the training set translates into good performance on the test set is known as the generalization problem, which turns out to be conceptually easy to solve by restricting the allowable functions $f$ to a relatively small class of functions $F$:

\begin{equation}
    f^* = argmin_{f \in F} \: Train_S(f)
\end{equation}

Restricting $f$ to $F$ essentially solves the generalization problem, because it can be shown that when $log |F|$ is small relative to the size of the training set (so in particular, $|F|$ is finite) \cite{vapnik}, the training error is close to the test error for all functions $f \in F$ simultaneously. This lets us focus
on the algorithmic problem of minimizing the training error while being reasonably certain that the test error will be approximately minimized as well. Since the necessary size of the training set grows with $F$, we want $F$ to be as small as possible. At the same time, we want $F$ to be as large as possible to improve the performance of its best function. In practice, it is sensible to choose the largest possible $F$ that can be supported by the size of the training
set and the available computation. Unfortunately, there is no general recipe for choosing a good $F$ for a given machine learning problem. Effectively, it is
best to experiment with function classes that are similar to ones that are successful for related problems \cite{SutskeverThesis}.

\section{Fundamentals of neural networks}

The Feedforward Neural Networks are the most basic and widely used artificial neural networks. They consist of a number of layers of artificial units that are arranged into a layered configuration. Of particular interest are deep neural networks, which are believed to be capable of representing the highly complex functions that achieve high performance on difficult perceptual problems
such as vision, speech and language. 

\begin{figure}[ht!]
  \centering
\includegraphics[width=0.5\textwidth, trim={0cm 0cm 0cm 0cm},clip]{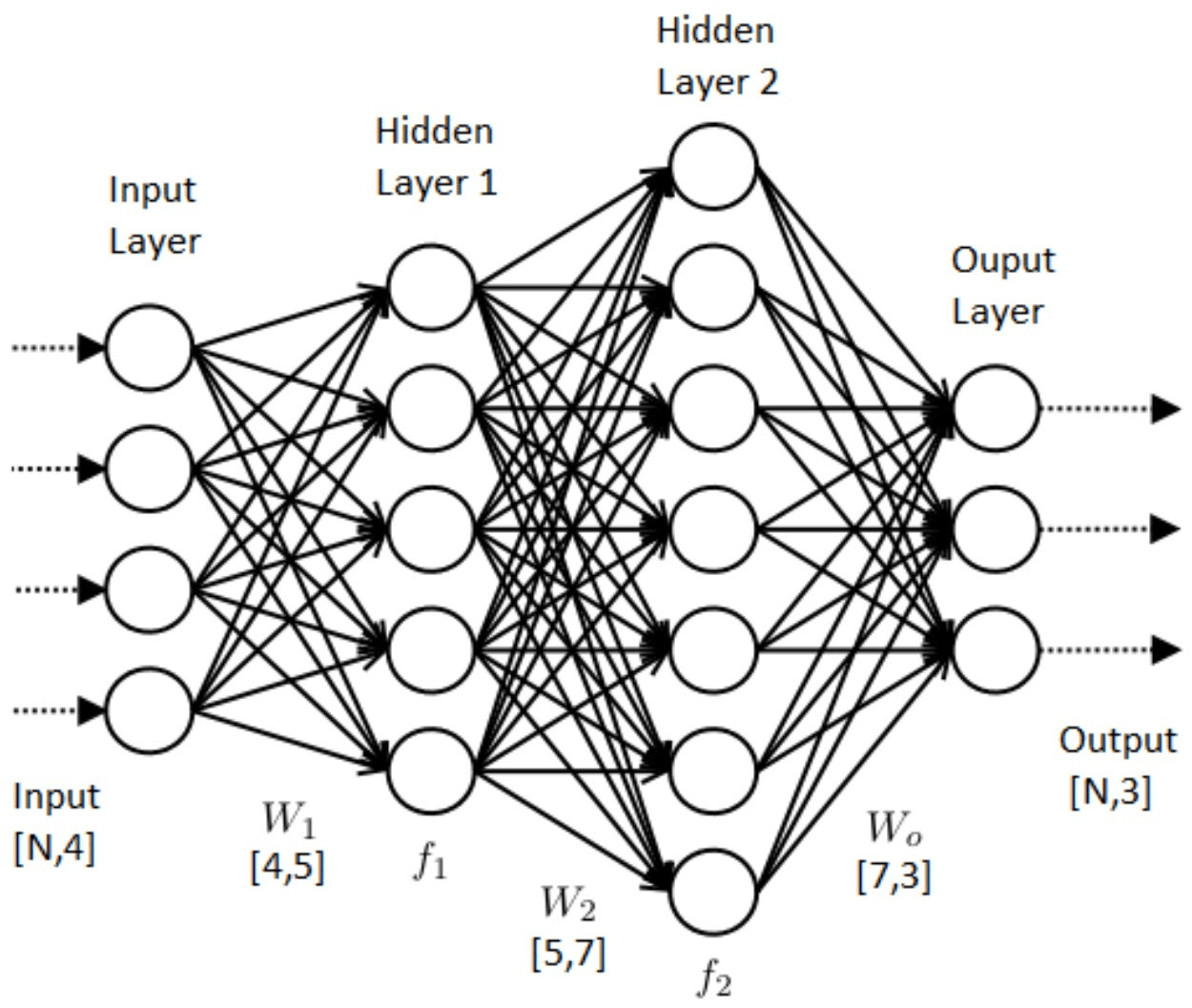} %_gtgsteps.eps}
  \caption{A fully connected feedforward neural network, containing 2 hidden layers, parametrized by 3 weight matrices.}
  \label{fig:fnn}
\end{figure}

A feedforward neural network with $n$ hidden layers is parametrized with $n+1$ weight matrices $(W_0, W_1, ..., W_{n})$ and $n+1$ vectors of biases $(b_0, b_1, ..., b_{n})$. Given an input $x$, the feedforward neural network computes the output $x$ given the following algorithm:

\begin{algorithmic}[H]
\STATE $z_0\gets x$
\FOR {$i$ from $1$ to $n+1$} 
        \STATE $x_i\gets W_{i-1}z_{i-1} + b_{i-1}$
        \STATE $z_i\gets act(x_i)$
\ENDFOR
\STATE $z\gets z_n$
\end{algorithmic}

where $act()$ represents a non-linear activation function. There are many possible activation functions, with the most popular ones coming from the family of rectified linear units (ReLU) \cite{DBLP:conf/icml/NairH10}.

\subsection{Backpropagation and Optimization}

The learning on deep neural networks typically consists of two procedures, being the computation of derivatives (gradients) and the adjustment of the weights based on the computed derivatives.

\begin{figure}[ht!]
  \centering
\includegraphics[width=0.9\textwidth, trim={0cm 0cm 0cm 0cm},clip]{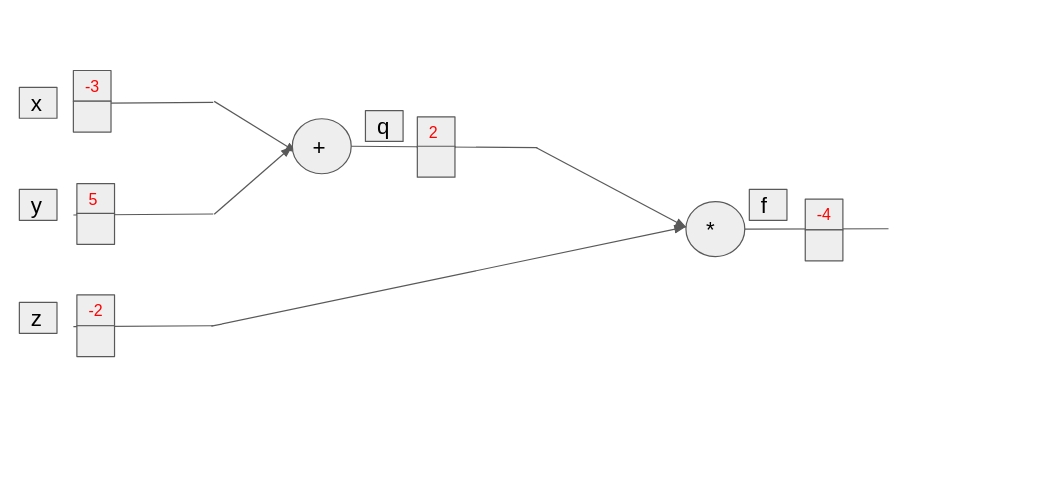} %_gtgsteps.eps}
  \caption{A computational graph representing a neural network during the forward pass. }
  \label{fig:forward-pass}
\end{figure}

Backpropagation \cite{Werbos:74, Rumelhart:1986:PDP:104279} typically implemented in modern deep learning libraries as the reverse mode of auto-differentiation \cite{Linnainmaa76} is the most used algorithm for the computation of derivatives in deep neural networks. While in essence it is a smart application of the chain rule of calculus, it has several interpretation, with perhaps the most intuitive one being the calculus graphs. Given a neural network represented by a computational graph (see Fig. \ref{fig:forward-pass}) first we do a forward pass followed by the computation of a loss function (eg. cross-entropy for classification or least-mean squared for regression).

\begin{figure}[ht!]
  \centering
\includegraphics[width=0.9\textwidth, trim={0cm 0cm 0cm 0cm},clip]{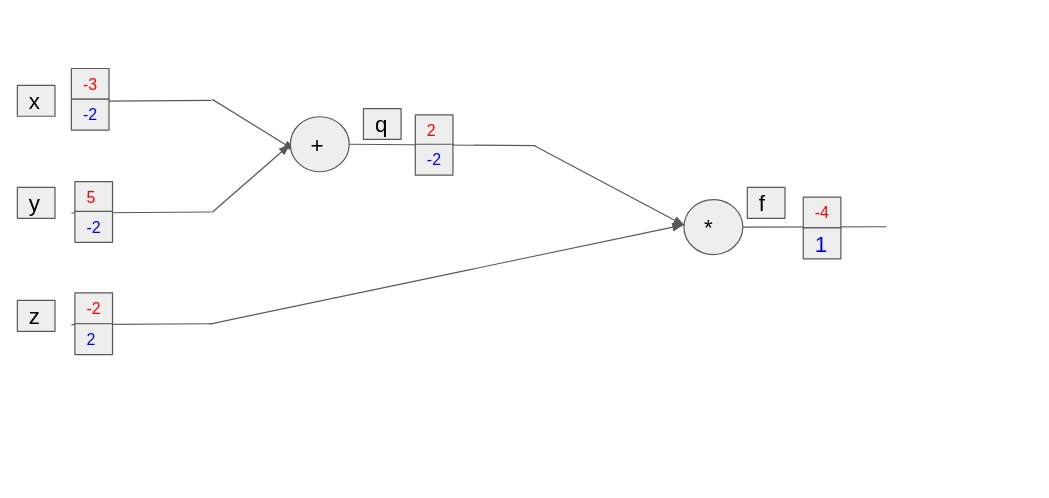} %_gtgsteps.eps}
  \caption{A computational graph representing a neural network during the backward pass. }
  \label{fig:backward-pass}
\end{figure}

After it, the derivatives are computed recursively (see Fig. \ref{fig:backward-pass}), where for each edge of the graph, the final derivative is the derivative of the edge, times the derivative of the nodes on the next layer which are connected to the edge. By storing the values of the derivatives in the graph (known as memoization), the derivatives do not need to be recomputed, making the computation of them linear in the number of edges.

After the computation of the derivatives, the weights of the network are adjusted. Most of the algorithms in neural networks use first-order optimizations, based on gradient descent \cite{Kiefer}. Given a function $F(\theta)$, gradient descent operates as follows:

\begin{algorithmic}[H]
\FOR {iterations} 
        \STATE $\theta_{t+1} \gets \theta_t - \alpha \nabla F(\theta_t)$
        \STATE $t\gets t+1$
\ENDFOR
\end{algorithmic}

where $\alpha$ is a hyperparameter representing the learning rate. On practice, there are a few considerations to be made. For each iteration, instead of using the entire dataset, only a small (randomly sampled) partition (called minibatch) of it is used. In these cases, the algorithm is called stochastic gradient descent (SGD). Perhaps surprisingly, in neural networks, SGD actually seems to outperform gradient descent, and recent studies have shown that using a small minibatch is actually desirable and has better generalization properties \cite{DBLP:conf/iclr/KeskarMNST17}. Additionally, the vanilla version of SGD is rarely used in practice. Instead, modifications of it are used being gradient descent with momentum \cite{lecun1998gradient}, accelerated gradient descent (Nesterov's momentum) \cite{Nesterov}, RMSProp \cite{rmsprop}, Adam \cite{DBLP:conf/iclr/Kingma14} etc, which typically reach higher generalization performance at a fraction of the computational cost.

\subsection{Convolutional Neural Networks (CNNs)}

Fully connected neural networks with hidden layers, a finite number of units and nonlinear activation functions have the ability of approximating any continuous function with arbitrary precision, making them universal approximators \cite{Hornik}. However, they are computationally costly, and have a large number of weights making them both non efficient and difficult to train while at the same time having poor generalization performances. 

Knowing that on rich-format data like images, speech and language there is structure, since the discovery of backpropagation, researchers have tried to exploit the structure of data in order to design more efficient types of feedforward neural networks. By far the most successful type of them have been the convolutional neural networks \cite{DBLP:journals/neco/LeCunBDHHHJ89} which are loosely inspired from visual cortex.

The core computational building block of a Convolutional Neural Network is the Convolutional Layer (or the CONV layer) which takes an input tensor and produces an output tensor by convolving the input with a set of filters. To make things more concrete, we will take an example with images. Suppose that our input is a color image $X$ (having $3$ channels) of size $224$ by $224$. Now consider a filter $w$ of size $3$ by $3$ by $3$ (see that the number of channels for the filter must be the same as the number of channels for the image, in this case $3$). We can convolve this filter by sliding it across all spatial positions of the input tensor and computing a dot product between a small chunk of $X$ and the filter $w$ at each position. The result will be an activation map, which in this case would have the dimensions $222$ by $222$. It is common to pad the images with zeros in order to not shrink the size of the images, in this case giving us an activation map of size $224$ by $224$. In a CONV layer, it is common to apply a set of filters (for example $128$) instead of applying a single filter. In this case, it will result with a feature map of size $224$ by $224$ by $128$. Intuitively, each filter has the capacity to “look for” certain local features in the input tensor and the parameters that make up the filters are trained with backpropagation and SGD.

\begin{figure}[ht!]
  \centering
\includegraphics[width=0.9\textwidth, trim={0cm 0cm 0cm 0cm},clip]{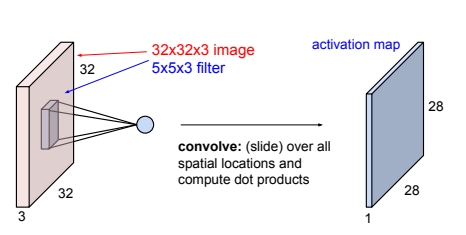} %_gtgsteps.eps}
  \caption{Illustration of convolving a $5$ by $5$ filter (which we will eventually learn) over a $3$ by $32$ by $3$ input array with stride 1 and with no input padding. The filters are always small spatially ($5$ vs. $32$), but always span the full depth of the input array ($3$). There are $28$ times $28$ unique positions for a $5$ by $5$ filter in a $32$ by $32$ input, so the convolution produces a $28$ by $28$ activation map, where each element
is the result of a dot product between the filter and the input. A convolutional layer has not just one but a set of different filters (e.g. $64$ of them), each applied in the same way and independently, resulting in their own activation maps. The activation maps are finally stacked together along depth to produce the output of the layer (e.g. $28$ by $28$ by $64$ array in this case). Figure reproduced from \cite{KarpathyThesis}.}
  \label{fig:conv}
\end{figure}

More generally, a convolutional layer for images (i.e. assuming input
tensors with three spatial dimensions):

\begin{description}
  \item $\bullet$ Accepts a tensor of size $W_1 \times H_1 \times D_1$
  \item $\bullet$ Requires $4$ hyperparameters: The number of filters $K$, their spatial extent $F$, the stride with which they are applied $S$, and the amount of zero padding on the borders of the input, $P$.
  \item $\bullet$ The convolutional layer produces an output volume of size $W_2 \times H_2 \times D_2$, where $W_2 = (W_1 - F + 2P)/S + 1$, $H_2=(H_1 - F + 2P)/S + 1$, and $D_2 = K$.
  \item $\bullet$ The number of parameters in each filter is $F \cdot F \cdot D_1$, for a total of $(F \cdot F \cdot D1) \cdot K$ weights and $K$ biases. In particular, note that the spatial extent of the filters is small in space $(F  \cdot F)$, but always goes through the full depth of the input tensor $(D_1)$.
  \item $\bullet$ In the output tensor, each d-th slice of the output (of size $W_2 \times H_2)$ is the result of performing a valid convolution of the d-th filter over the input tensor with a stride of S and then offsetting the result by d-th bias.
\end{description}

\textbf{Pooling layers.} In addition to convolutional layers, it is very common in CNNs to also have pooling layers that decrease the size of the representation with a fixed downsampling transformation (i.e. without any parameters). In particular, the pooling layers operate on each channel (activation map) independently and downsample them spatially. A commonly used setting is to use $2 \times 2$ filters with stride of $2$, where each filter computes the max operation (i.e. over $4$ numbers). The result is that an input tensor is downscaled exactly by a factor of $2$ in both width and height and the representation size is reduced by a factor of $4$, at the cost of losing some local spatial information. The most common types of pooling layers are \textit{max-pooling} where the highest value in the region is chosen, and \textit{average pooling} where the average value in a region is computed.

\textbf{CNNs.} CNNs are neural networks which contain (typically many) convolutional layers and a few pooling layers, followed by an output layer. Nowadays, it is common to have CNNs which contain tens to hundreds of convolutional layers (though researchers have trained CNNs which contain up to $10$ thousand layers) and millions to hundreds of billions of weights. During the last decade, CNNs have been at the forefront of not only deep learning, but artificial intelligence in general. Since the AlexNet architecture \cite{DBLP:conf/nips/KrizhevskySH12}, researchers have developed many efficient CNNs architectures. This thesis contain a heavy use of CNNs in all of the following chapters. On particular, we use ResNets \cite{DBLP:conf/cvpr/HeZRS16}, DenseNets \cite{DBLP:conf/cvpr/HuangLMW17} and GoogleNet \cite{DBLP:conf/cvpr/SzegedyLJSRAEVR15}.

\begin{figure}[ht!]
  \centering
\includegraphics[width=0.9\textwidth, trim={0cm 0cm 0cm 0cm},clip]{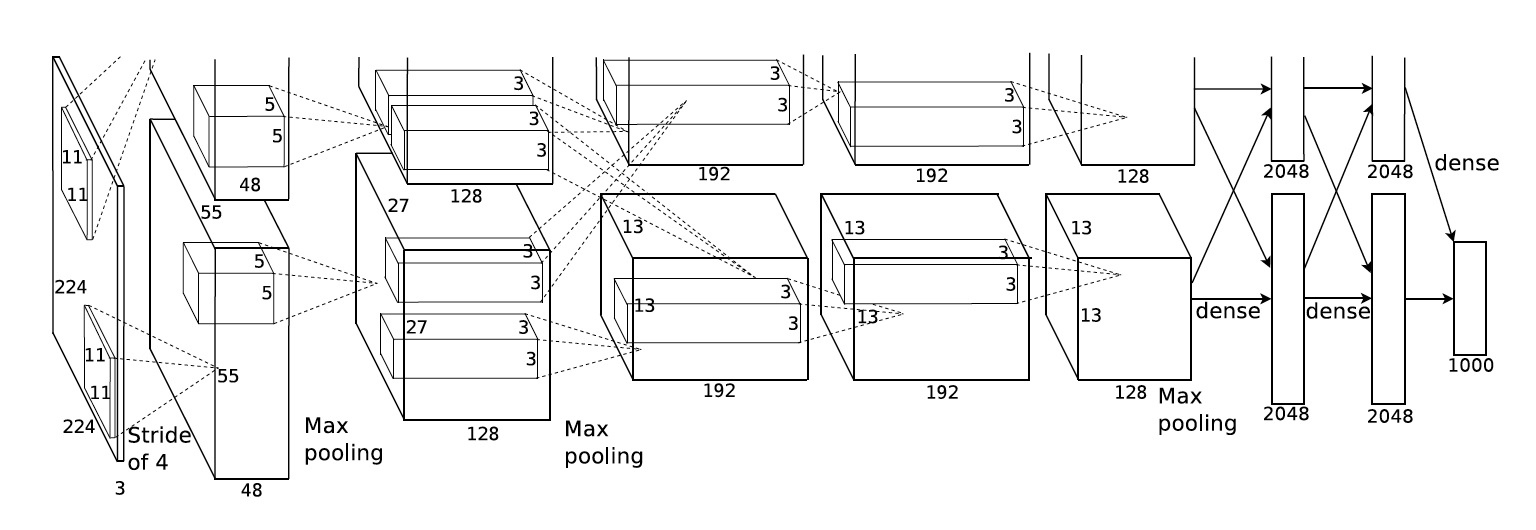} %_gtgsteps.eps}
  \caption{AlexNet \cite{DBLP:conf/nips/KrizhevskySH12} - the most famous CNN, which started the deep learning wave. }
  \label{fig:alexnet}
\end{figure}

\subsection{Recurrent Neural Networks (RNNs)}

There are many applications where the input and output are sequences. For example, in machine translation, it is desirable to not consider each word in isolation but to consider them as part of the sequences, and so instead of translating words, to translate the sequences. A recurrent neural network (RNN) is a connectivity pattern that processes a sequence of vectors $\{x_1, x_2,..., x_n\}$ using a recurrence formula of the form 
$h_t = f_\theta(h_{t-1}, x_t)$, where $f$ is a function and the same parameters $\theta$ are used at every time step, allowing us to process sequences with an arbitrary number of vectors. The hidden vector $h_t$ can be interpreted as a running summary of all vectors $x$ until that time step and the recurrence formula updates the summary based on the next vector. It is common to either use $h_0 = [0,...,0]$, or to treat $h_0$ as parameters and learn the starting hidden state.

\begin{figure}[ht!]
  \centering
\includegraphics[width=0.9\textwidth, trim={0cm 0cm 0cm 0cm},clip]{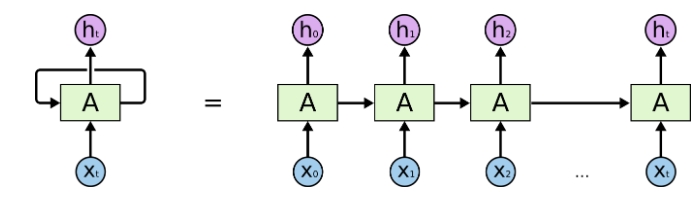} %_gtgsteps.eps}
  \caption{An unrolled recurrent neural network. Figure adapted from \cite{colah}.}
  \label{fig:rnn}
\end{figure}

Vanilla recurrent neural networks implement the following equation:

\begin{equation}
    h_t = tanh(W_{xh}x_t + W_{hh}h_{t-1})
\end{equation}

where $W_{xh}$ and $W_{hh}$ represent the transitional matrices from input to hidden state, and hidden state to hidden state respectively, $h$ represents the hidden state and the bias has been omitted for brevity. While theoretically RNNs are program approximators, in the form given above they tend to be very hard to train, with the gradients either vanishing or exploding \cite{Hochreiter:90, DBLP:journals/tnn/BengioSF94}.

\subsubsection{Long Short Term Memory Networks (LSTMs)}

In order to solve the above-mentioned problem, \cite{hochreiter1997long} modified the vanilla RNN to have extra gates which would allow the network to remember long-term dependencies, while at the same time to forget the irrelevant information. These networks which are called "Long Short Term Memory" networks have been widely used in machine translation, speech recognition and many other domains where long term dependencies are important. They can be implemented via the following equations:

\begin{equation}
    h_t = tanh(W_{hh}h_{t−1} + W_{hv}v_t + W_{hm}m_{t−1}) 
\end{equation}
\vspace{-1cm}
\begin{equation}
    i_t^g = sigmoid(W_{igh}h_t + W_{igv}v_t + W_{igm}m_{t-1} 
\end{equation}
\vspace{-1cm}
\begin{equation}
    i_t = tanh(W_{ih}h_t + W_{iv}v_t + W_{im}m_{t-1} 
\end{equation}
\vspace{-1cm}
\begin{equation}
    o_t = sigmoid(W_{oh}h_t + W_{ov}v_t + W_{om}m_{t-1} 
\end{equation}
\vspace{-1cm}
\begin{equation}
    f_t = sigmoid(b_f + W_{fh}h_t + W_{fv} v_t + W_{fm}m_{t-1})
\end{equation}
\vspace{-1cm}
\begin{equation}
    m_t = m_{t-1} \otimes f_t + i_t \otimes I_t^g
\end{equation}
\vspace{-1cm}
\begin{equation}
    m_t = m_{t} \otimes o_t
\end{equation}
\vspace{-1cm}
\begin{equation}
    o_t = g(W_{yh}h_t + W_{ym}m_t)
\end{equation}

where $\otimes$ represents the Haddamard product and $i$, $o$ and $f$ stand for input, output and forget gates. 

The only part of this thesis which depends on recurrent neural networks is Appendix A, the rest of the thesis can be read without any knowledge on RNNs.

\begin{figure}[ht!]
  \centering
\includegraphics[width=0.9\textwidth, trim={0cm 0cm 0cm 0cm},clip]{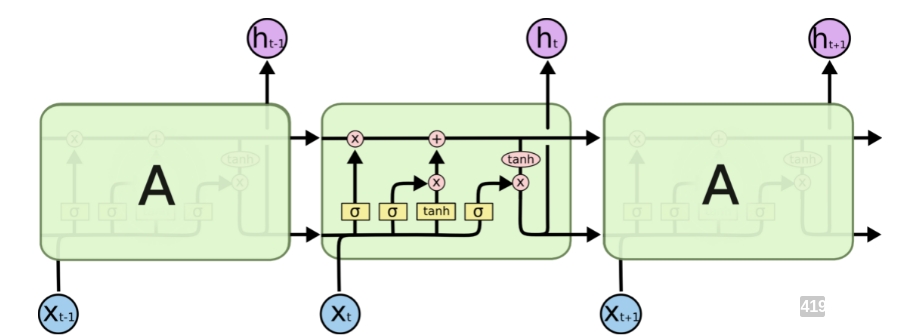} %_gtgsteps.eps}
  \caption{The LSTM module. Figure adapted from \cite{colah}.}
  \label{fig:lstm}
\end{figure}

\subsection{Regularization}

Deep neural networks have a massive number of parameters and so are prone to overfitting. In order to mitigate the problem, different types of regularization are used. Perhaps the most used form of regularization is the $l2$ regularization (at times wrongly called \textit{weight decay} \cite{DBLP:conf/iclr/LoshchilovH19}) where the large parameters are penalized. This can be achieved by augmenting the loss function with a new regularization term, as shown in the following equation:

\begin{equation}
    L(w) = L_0(w) + \lambda ||w||_2^2
\end{equation}

where $L_0(w)$ is the previous loss function and $\lambda$ is a hyperparameter. Consequently, the gradients become:

\begin{equation}
    \nabla_w L(w) = \nabla_w [L_0(w) + \lambda ||w||_2^2] = \nabla_w L_0(w) + 2 \lambda w
\end{equation}.

Another widely used deep learning-specific form of regularization is \textit{dropout} \cite{DBLP:journals/jmlr/SrivastavaHKSS14}. Dropout in forward pass simply drops units with probability $p$, making every unit less dependent in its neighbors. Additionally, by applying a different dropout mask (dropping different units) in each iteration, the resulted trained net can be considered as an ensemble. Dropout is widely used in fully connected neural networks, but is less used in CNNs. However, there exist usages of it in large CNNs, and with a slight modification, it can be used for probability calibration \cite{gal2016dropout}.

Another omnipresent form of regularization in neural networks is \textit{batch normalization} \cite{DBLP:conf/icml/IoffeS15}. When trained with batch normalization each feature map (layer) of a neural network is normalized using the mean and standard deviation. Then the features are scaled and shifted via $2$ learnable (by backpropagation) parameters $\gamma$ and $\beta$. Batch-normalization has shown to both improve the generalization performance and the speed of convergence for a neural network.

Finally, when working with images, it is extremely common to augment the training set by applying simple transformations to the images (horizontal and vertical shifting, rotation, random cropping etc). This type of regularization is called \textit{data augmentation} and is used in almost every computer vision application.

In this thesis, we have aggressively used all forms of regularization mentioned in this section, in many cases combining multiple forms of regularization (like batch normalization, data augmentation and l2 regularization).

\begin{figure}[ht!]
  \centering
\includegraphics[width=0.9\textwidth, trim={0cm 0cm 0cm 0cm},clip]{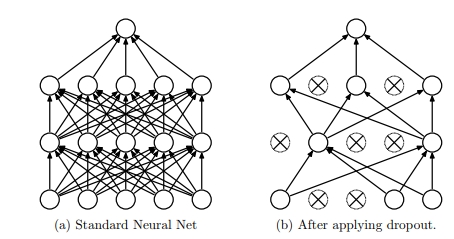} %_gtgsteps.eps}
  \caption{Dropout Neural Net Model. Left: A standard neural net with 2 hidden layers. Right:
An example of a thinned net produced by applying dropout to the network on the left.
Crossed units have been dropped. Figure adapted from \cite{DBLP:journals/jmlr/SrivastavaHKSS14}.}
  \label{fig:dropout}
\end{figure}

\subsection{Graphs in Neural Networks}

A large part of the thesis tries to combine game-theoretical approaches (or graph-theoretical inspired approaches) in the context of the neural networks. While our methods are not related to the other methods presented here, we need to acknowledge that the idea of using graph-like structures in neural networks is hardly new. For historical context, we describe a few methods in the following section.

\subsubsection{A General Framework for Adaptive Processing of Data Structures} The work of \cite{DBLP:journals/tnn/FrasconiGS98} is one of the first works that uses neural networks for arbitrary structured data. The work is mostly a theoretical work that gives directions on extending the concept of neural networks to other types of data, proposing a framework that attempts to unify adaptive models like artificial neural nets and belief nets for the problem of processing structured information. In particular, relations between data
variables are expressed by directed acyclic graphs, where both numerical and categorical values coexist. This is very different to most types of neural networks, that typically do not use categorical attributes. The general framework proposed in \cite{DBLP:journals/tnn/FrasconiGS98} can be regarded as an extension of both
recurrent neural networks and hidden Markov models to the
case of acyclic graphs. In particular, the authors study the supervised learning problem as the problem of learning transductions from an input structured space to an output structured space, where transductions are assumed to admit a recursive hidden statespace representation. The authors introduce a graphical formalism for representing this class of adaptive transductions by means of
recursive networks, i.e., cyclic graphs where nodes are labeled by
variables and edges are labeled by generalized delay elements, making possible to incorporate the symbolic and subsymbolic nature of data. 

\begin{figure}[ht!]
  \centering
\includegraphics[width=0.5\textwidth, trim={0cm 0cm 0cm 0cm},clip]{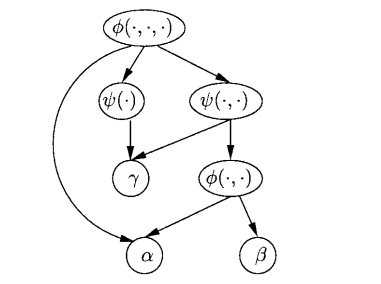} 
  \caption{A directed acyclic graph representing the logical term $\phi(\alpha,\psi(\gamma)), \psi(\gamma,\phi(\alpha,\beta))$ Figure reproduced from \cite{DBLP:journals/tnn/FrasconiGS98}.}
  \label{fig:frasconi}
\end{figure}

\subsubsection{The graph neural network model} An interesting work is that of \cite{DBLP:journals/tnn/ScarselliGTHM09} which explicitly uses graphs in the context of neural networks. Knowing that many underlying relationships among data in several
areas of science and engineering, e.g., computer vision, molecular chemistry, molecular biology, pattern recognition, and data mining, can be represented in terms of graphs, the authors propose a new neural network model, called graph neural network
(GNN) model, that extends existing neural network methods for
processing the data represented in graph domains. This GNN model, which can directly process most of the practically useful types of graphs, e.g., acyclic, cyclic, directed, and undirected, implements a function $\tau(G, n) \in R^m$  that maps a graph
and one of its nodes n into an m-dimensional Euclidean space.

The main strength of the work is that perhaps for the first time, the authors proposed a learning rule that can be combined with gradient-based methods. In particular, learning in GNNs consists of estimating the parameter $\omega$ 
such that $\varphi_\omega$ approximates the data in the learning data set:

\begin{equation}
L = \{(g_i, n_{i,j}, t_{i,j)}|g_i=(N_i,E_i) \in G;
     n_{i,j} \in N_i; t_{i,j} \in R^m, 1 < i < p, 1 < i < q_i\}
\end{equation}.

where $q_i$ is the number of supervised nodes in $g_i$. 

The learning task is posed as the minimization of a quadratic cost function:

\begin{equation}
e_w = \sum_{i=1}^p \sum_{j=1}^{q_i} (t_{i,j} - \varphi_w (g_i, n_{i,j})
\end{equation}.

which can be easily combined with gradient-based methods. Note, the method reminds to backpropagation-through-time (used in recurrent neural networks). The method showed success in various problems, including subgraph matching problem, inductive logic programming or web-page ranking. The method was introduced before the rise of the deep learning era, nevertheless showed that graphs can be combined with neural networks, reached good experimental results and might have served as an inspiration for the more recent methods \cite{DBLP:conf/iclr/KipfW17, DBLP:journals/corr/abs-1901-07984, DBLP:journals/corr/abs-1806-01261}.

\subsubsection{Semi-supervised classification with graph convolutional networks} Arguably the most famous method that combines graphs with CNNs is that of Kipf and Welling \cite{DBLP:conf/iclr/KipfW17}. The authors start from the framework of spectral graph convolutions \cite{DBLP:journals/corr/BrunaZSL13}, yet introduce simplifications that in many cases allow both for significantly faster training times and higher predictive accuracy, reaching state-of-the-art classification results on a number of benchmark graph datasets.

For this model, the goal is then to learn a function of signals/features on a graph $G=(V,E)$ which takes as input:

\begin{itemize}
    \item A feature description $x_i$ for every node $i$; summarized in a $N \times D$ feature matrix $X$ ($N$: number of nodes, $D$: number of input features).
    \item A representative description of the graph structure in matrix form; typically in the form of an adjacency matrix $A$ (or some function thereof).
\end{itemize}

and produces a node-level output $Z$ (an $N \times F$ feature matrix, where $F$ is the number of output features per node). Graph-level outputs can be modeled by introducing some form of pooling operation \cite{DBLP:conf/nips/DuvenaudMABHAA15}.

Every neural network layer can then be written as a non-linear function

\begin{equation}
H^(l+1) = f(H^l, A)
\end{equation}.

with $H(0)=X$ and $H(L)=Z$, $L$ being the number of layers. The specific
models then differ only in how $f(\cdot,\cdot)$ is chosen and parameterized.

As an example, they consider the following very simple form of a layer-wise propagation rule:

\begin{equation}
f(H^l,A) = \sigma(AH^lW^l)
\end{equation}.

where $W(l)$ is a weight matrix for the $l$-th neural network layer and $\sigma(\cdot)$ is a non-linear activation function like the ReLU.

But first, the authors address two limitations of this simple model: multiplication with $A$ means that, for every node, they sum up all the feature vectors of all neighboring nodes but not the node itself (unless there are self-loops in the graph). This can be fixed by enforcing self-loops in the graph by simply adding the identity matrix to $A$.

The second major limitation is that $A$ is typically not normalized and therefore the multiplication with $A$ will completely change the scale of the feature vectors. Normalizing $A$ such that all rows sum to one, i.e. $D^{-1}A$, where $D$ is the diagonal node degree matrix, gets rid of this problem. Multiplying with $D^{-1}A$ now corresponds to taking the average of neighboring node features. In practice, dynamics get more interesting when a symmetric normalization is used, i.e. $D^\frac{-1}{2}AD^\frac{-1}{2}$ (as this no longer amounts to mere averaging of neighboring nodes). By combining these two tricks, the authors reach the propagation rule defined as:

\begin{equation}
f(H^l,A) = \sigma(\hat{D}^{\frac{-1}{2}}\hat{A}\hat{D}^{\frac{-1}{2}}H^lW^l)
\end{equation}.

where $\hat{A}=A+I$, I is the identity matrix and $\hat{D}$ is the diagonal node degree matrix of $\hat{A}$.

\subsubsection{Discussion}

Graphs have been one of the main data-structures of computer science and machine learning, but they have hardly been used in deep learning. Despite that they were introduced in neural networks \cite{DBLP:journals/tnn/ScarselliGTHM09} before the latest wave of neural network research \cite{DBLP:conf/nips/KrizhevskySH12}, until recently they have found a limited usage in deep learning. The work of \cite{DBLP:conf/iclr/KipfW17} brought Graph CNNs at the front of the deep learning research, and since then many novel works and applications of graph CNNs followed. Recently, graphs in neural networks have been combined with message-passing techniques to solve many problems. The work of \cite{DBLP:conf/icml/GilmerSRVD17} focuses on using graph networks that can be useful in chemistry or drug discovery. The work of \cite{DBLP:conf/iccv/WangLSC019} proposes a novel attentive graph neural network for zero-shot video object segmentation. The recent work of \cite{DBLP:journals/corr/abs-1912-07515} achieves state-of-the-art multi-object tracking by leveraging the power of message-passing networks (implemented as graph CNNs). The reader is recommended to read the excellent survey of \cite{DBLP:journals/corr/abs-1806-01261} for more directions in the field.

The next two chapters of this thesis heavily use graphs in combinations with CNNs. In the next chapter we use a label-propagation method (based on graphs) \cite{DBLP:journals/neco/ErdemP12} to solve the problem of label-augmentation for CNNs. The fourth chapter is inspired from the same method to develop a brand novel loss function for Siamese networks. The work presented in this thesis was developed independently from the cited work, and is only loosely connected with them. Nevertheless, some similarities exist (especially with \cite{DBLP:conf/iclr/KipfW17}). The choice of our model was based on it working on probability space (which makes it very suitable to be combined with softmax cross-entropy in neural networks), and being very efficient and easy to work with a large number of samples (unlike \cite{DBLP:conf/iclr/KipfW17}). We describe the graph-transduction method \cite{DBLP:journals/neco/ErdemP12} in detail in the next chapter, use it to achieve label augmentation, and then in the following chapter use a graph-transduction based method for the problem of metric learning. We present in detail similarities and differences with the other graph methods in the conclusions.

\chapter{Transductive Label Augmentation for Improved Deep Network Learning}

\section{Disclaimer}

The work presented in this chapter is based on the following paper:

\begin{description}
  \item \textbf{Ismail Elezi}, Alessandro Torcinovich, Sebastiano Vascon and Marcello Pelillo; \textit{Transductive label augmentation for improved deep network learning \cite{DBLP:conf/icpr/EleziTVP18}}; In Proceedings of IAPR International Conference on Pattern Recognition (ICPR 2018)
\end{description} 

The contributions of the author are the following:

\begin{description}
  \item[$\bullet$ Coming] up with the pipeline of the algorithm.
  \item[$\bullet$ Writing] the vast majority of the code.
  \item[$\bullet$ Performing] the majority of the experiments.
  \item[$\bullet$ Writing] a considerable part of the paper.
\end{description}

\section{Introduction}

Deep neural networks (DNNs) have met with success multiple tasks, and testified a constantly increasing popularity, being able to deal with the vast heterogeneity of data and to provide state-of-the-art results across many different fields and domains \cite{DBLP:journals/nature/LeCunBH15,DBLP:journals/nn/Schmidhuber15}.
Convolutional Neural Networks (CNNs) \cite{DBLP:journals/pr/FukushimaM82,DBLP:journals/neco/LeCunBDHHHJ89} are one of the protagonists of this success. Starting from AlexNet \cite{DBLP:conf/nips/KrizhevskySH12}, until the most recent convolutional-based architectures \cite{DBLP:conf/cvpr/SzegedyLJSRAEVR15,DBLP:conf/cvpr/HeZRS16,DBLP:conf/cvpr/HuangLMW17} CNNs have proved to be especially useful in the field of computer vision, improving the classification accuracy in many datasets \cite{DBLP:conf/cvpr/DengDSLL009, Krizhevsky2009}.

However, a common caveat of large CNNs is that they require a lot of training data in order to work well. In the presence of classification tasks on small datasets, typically those networks are \emph{pre-trained} in a very large dataset like ImageNet \cite{DBLP:conf/cvpr/DengDSLL009}, and then \emph{finetuned} on the dataset the problem is set on. The idea is that the pre-trained network has stored a decent amount of information regarding features which are common to the majority of images, and in many cases this knowledge can be transferred to different datasets or to solve different problems (image segmentation, localization, detection, etc.). This technique is referred as \emph{transfer learning} \cite{DBLP:conf/nips/YosinskiCBL14} and has been an important ingredient in the success and popularization of CNNs. Another important technique -- very often paired with the previous one -- is \emph{data augmentation}, through which small transformations are directly applied on the images. A nice characteristic of data augmentation is its agnosticism toward algorithms and datasets. \cite{DBLP:conf/cvpr/CiresanMS12} used this technique to achieve state-of-the-art results in MNIST dataset \cite{Lecun1998}, while \cite{DBLP:conf/nips/KrizhevskySH12} used the method almost without any changes to improve the accuracy of their CNN in the ImageNet dataset \cite{DBLP:conf/cvpr/DengDSLL009}. Since then, data augmentation has been used in virtually every implementation of CNNs in the field of computer vision.

Despite the practicality of the above-mentioned techniques, when the number of images per class is extremely small, the performances of CNNs rapidly degrade and leave much to be desired. The high availability of unlabeled data only solves  half of the problem, since the manual labeling process is usually costly, tedious and prone to human error. Under these assumptions, we propose a new method to perform an automatic labeling, called \emph{transductive label augmentation}. Starting from a very small labeled dataset, we set an automatic label propagation procedure, that relies on graph transduction techniques, to label a large unlabeled set of data. This method takes advantage of second-order similarity information among the data objects, a source of information which is not directly exploited by traditional techniques. To assess our statements, we perform a series of experiments with different CNN architectures and datasets, comparing the results with a first-order ``label propagator'' in addition to competing label propagation techniques and purely deep learning based methods.

In summary, our contributions of this chapter are as follows: a) by using graph transductive approaches, we propose and develop the aforementioned label augmentation method and use it to improve the accuracy of state-of-the-art CNNs in datasets where the number of labels is limited; b) by gradually increasing the number of labeled objects, we give detailed results in four standard computer vision datasets and compare the results with the results of CNNs; c) we replace our transductive algorithm with linear support vector machines (SVM) \cite{DBLP:journals/ml/CortesV95} to perform label augmentation and compare the results; d) we replace our method with other label propagation techniques and compare the results; e) we compare our method with other deep learning methods; f) we give directions for future work and how the method can be used on other domains. 

\begin{figure*}[ht!]
 \centering
 \includegraphics[width=\textwidth]{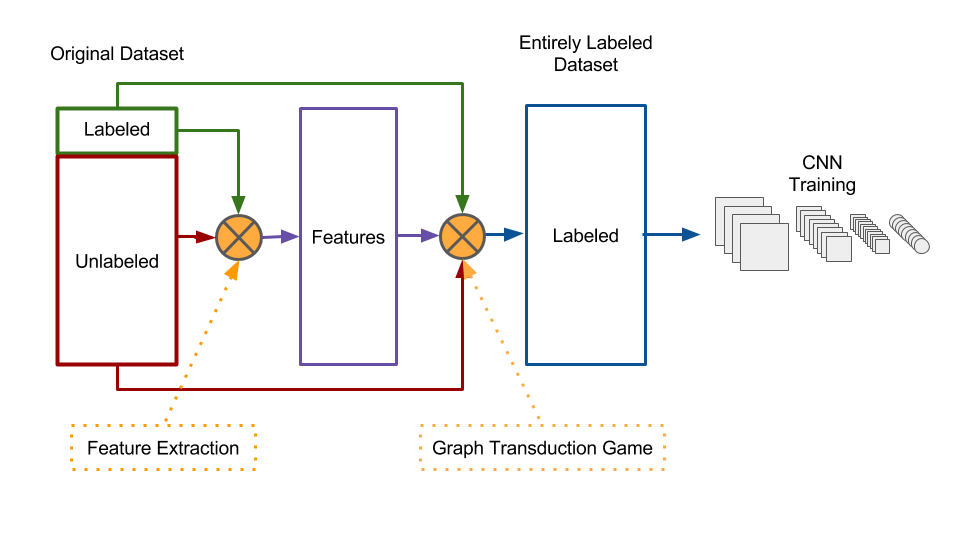}
 \caption{The pipeline of our method. The dataset consists of labeled and unlabeled images. First, we extract features from the images, and then we feed the features (and the labels of the labeled images) to graph transduction games. For the unlabeled images, we use a uniform probability distribution as 'soft-labeling'. The final result is that the unlabeled points get labeled, thus the entire dataset can be used to train a convolutional neural network.}
\end{figure*}

\section{Related Work}

Semi-supervised label propagation has a long history of usage in the field of machine learning. Starting from an initial large dataset, with a small portion of labeled observations the traditional way of using semi-supervised learning is to train a classifier only in the labeled part, and then use the classifier to predict labels for the unlabeled part. The labels predicted in this way are called \emph{pseudo-labels}. The classifier is then trained in the entire dataset, considering the pseudo-labels as if they were real labels.

Different methods with the same intent have been previously proposed. In deep learning in particular, there have been devised algorithms to use data with a small number of labeled observations. \cite{Lee_pseudo-label:the} trained the network jointly in both the labeled and unlabeled points. The final loss function is a weighted loss of both labeled and unlabeled points, where in the case of the unlabeled points, the pseudo-label is determined by the highest score proposed by the model. \cite{DBLP:conf/cvpr/HausserMC17} optimized a CNN on such a way as to produce embeddings that have high similarities for the observations that belong to the same class. \cite{DBLP:conf/nips/KingmaMRW14} used a totally different approach, developing a generative model that allows for effective generalization from small labeled datasets to large unlabeled ones. Recently, new methods have been developed \cite{DBLP:conf/iclr/LaineA17, DBLP:conf/iclr/TarvainenV17, DBLP:journals/pami/MiyatoMKI19}. The reader is encouraged to read \cite{DBLP:conf/nips/OliverORCG18} for a realistic evaluation and comparison of the most common forms of deep learning-based semi-supervised learning.

In all the mentioned methods, the way how the unlabeled data has been used can be considered as an intrinsic property of their engineered neural networks. Our choice of CNNs as the algorithm used for the experiments was motivated because CNNs are state-of-the-art models in computer vision, but the approach is more general than that. The method presented in this chapter does not even require a neural network and in principle, non-feature based observations (i.e graphs) can be considered, as long as a similarity measure can be derived for them. At the same time, the method shows good results in relatively complex image datasets, improving over the results of state-of-the-art CNNs.

\subsection{Graph Transduction Game}\label{sec:gtg}
Graph Transduction (GT) is a subfamily of semi-supervised learning that aims to classify unlabeled objects starting from a small set of labeled ones. In particular, in GT the data is modeled as a graph whose vertices are the objects in a dataset. The provided label information is then propagated all over the unlabeled objects through the edges, weighted according to the consistency of object pairs. The reader is encouraged to refer to \cite{ZhuSemiSupervised} for a detailed description of algorithms and applications on graph transduction.

In \cite{ZhuSemiSupervised}, GT takes in input $W$ along with initial probability distributions for every objects -- one-hot labels for $(f_i, y_i) \in L$, soft labels for $f_i \in U$ -- and iteratively applies a function $P: \Delta^m \rightarrow \Delta^m$ where $\Delta^m$ is the standard simplex. At each iteration, if the distributions of labeled objects have changed, they are reset. Once the algorithm reaches the convergence, the resulting final probabilities give a labeling over the entire set of objects.

In this chapter, we follow the approach proposed in \cite{DBLP:journals/neco/ErdemP12} called Graph Transduction Game (GTG), where the authors interpret the graph transduction task as a non-cooperative multiplayer game. The same methodology has been successfully applied in different context, e.g. bioinformatics \cite{DBLP:journals/prl/Vascon18} and matrix factorization \cite{DBLP:conf/icpr/TripodiVP16}.

In GTG the transduction task is modeled as a non-cooperative game in which the players represent the observations and the pure strategies represent the possible labels. The players (observations) play a game in which they choose strategies (labels) such that their payoff is maximized by progressively modifying their preferences over the labels proportionally to the similarities among the players and their own preferences.
During every round of the game, each player updates its probability of choosing a particular strategy according to the received payoff. The more similar the players are, the more they will affect each other in choosing the same strategy. The game is performed until a point of convergence is reached, the so-called \emph{Nash equilibrium} \cite{Nash1951}. At equilibrium condition all the players have chosen the strategy that provides them the highest payoff. At this point, no one has any incentive to deviate from their choices, thus a consistent 
labeling \cite{miller1991copositive} is reached.

More formally, we define a set of \emph{players} $\mathcal{I} = \{1, \dots, n\}$ and a set of \emph{pure strategies} $S = \{1, \dots, m\}$ shared among all players (here $n$ is the number of the observations in the mini-batch and $m$ the number of labels). The set of players is divided into \emph{labeled} (denoted by $L$, in this work we also refer to them as \textit{anchors}) and \emph{unlabeled} (denoted by $U$), with $\mathcal{I} = L \cup U$ being their union set. Each player $i$ is associated with a \emph{mixed strategy} $x_i$, which is a probability distribution over $S$, modeling the uncertainty in picking one label over another. Each mixed strategy lies in the $m$-dimensional \emph{standard simplex} $\Delta^m$ defined as $\Delta^m = \left\lbrace x_i \in \mathbb{R}^m | \ \sum_{h = 1}^{m} x_{ih} = 1, x_{ih} \geq 0 \right\rbrace$. The union of all the mixed strategies of the players composes a \emph{mixed strategy profile} $x \in \Delta^{n \times m}$ which corresponds to a particular step of the game. The matrix $x$ evolves at each iteration of the game. We denote with $x(t)$ the situation of the game at the $t$-th iteration, while with $x_{ih}(t)$ we point to the probability of picking the $h$-strategy adopted by player $i$ at time $t$. In the following, we denote the \textit{pure strategy} $h$ for player $i$ with the mixed strategy $e_i^{(h)}$, a vector of size $m$ with $1$ at position $h$ and $0$ elsewhere (one-hot labeling).

\vspace{-0.3cm}
\paragraph{Game initialization}
The starting point of the game ($t = 0$) is encoded into the initial mixed strategy profile $x(0)$. Prior knowledge on the strategies of the players can be injected in each mixed strategy, drifting the starting point. The mixed strategies of each labeled players $x_i(0) \in L$ are simply set to their one-hot labeling. This choice, along with an appropriate update rule, ensures that the labeled players never change their strategy during the process. As for the unlabeled players, their strategies can be set to either some prior distribution (coming, for example, from a neural network) or to an uniform distribution, i.e. $x_{ih}(0) = 1 / m, \ \forall h \in S$.
\vspace{-0.3cm}
\paragraph{Payoff definition}
The game updates are driven by the player choices of a strategy towards another, which in turn is based on their mixed strategies and pairwise similarities among the players. To quantify the best choices, a tuple of \emph{payoff functions} $u = (u_1, \dots, u_n)$ s.t. $u: \Delta^{n \times m} \rightarrow \mathbb{R}_{\ge 0}^n$ is defined. Each payoff function $u_i$ quantifies the gain that player $i$ obtains given the actual configuration of the mixed strategy profile. It is worth stressing the fact that the payoff functions take into account the mixed strategy of every player, fitting gracefully within the context of this chapter. 
Let $(e_i^{(h)}, x_{-i})$ define a mixed strategy profile where all players $j \in \mathcal{I} \setminus \{i\}$ play their mixed strategy $x_j$ while player $i$ plays the mixed strategy $e_i^{(h)}$, instead. Then: 
\vspace{-0.1cm}
\begin{equation}
u_i(e_i^{(h)}, x_{-i}) = \sum_{j \in U}{(A_{ij} x_j)_h} + \sum_{k = 1}^m{\sum_{j \in L_k}{A_{ij}(h, k)}} \label{eqn:payoff_pure}
\end{equation}
\\
\vspace{-0.5cm}
\begin{equation} %\\
u_i(x) = \sum_{j \in U}{x_i^T A_{ij} x_j}+\sum_{k = 1}^m{\sum_{j \in L_k}{x_i^T (A_{ij})_k}} \label{eqn:average_payoff}
\end{equation}

where $A_{ij} \in \mathbb{R}^{m \times m}$ is the \emph{partial payoff matrix} between the pair of players $(i, j)$. In particular, $A_{ij} = \omega_{ij} \cdot I_m$ with $\omega_{ij}$ being the similarity of players $i$ and $j$ while $I_m$ is an identity matrix of size $m \times m$. The Equation \ref{eqn:payoff_pure} quantifies the payoff obtained by player $i$ when it plays the pure strategy $h$, while Equation \ref{eqn:average_payoff} compute the overall payoff of player $i$ considering the entire strategy profile $x$. 

\paragraph{Iterative procedure}
The goal of GTG is to reach an equilibrium condition in which the players are satisfied with their chosen strategy and have no incentives to change them. This condition is known as Nash Equilibrium \cite{Nash1951}, and %in the context of this work 
corresponds to the so-called \emph{consistent labeling}. 

The evolution of the game towards an equilibrium point is computed through a dynamical system, namely the \emph{Replicator Dynamics} (RD) \cite{smith1982evolution, weibull1997evolutionary}. In our case, we adopted the discrete version of the dynamics:
\begin{equation}
    x_{ih}(t + 1) = x_{ih}(t)\frac{u_i(e_i^{(h)}, x_{-i}(t))}{u_i(x(t))}
    \label{eq:RD}
\end{equation}
where $t$ defines the current iteration of the process.
The dynamics are typically run until two consecutive steps do not differ significantly or a maximum number of iterations is reached. 
Finally, it is worth mentioning that Equation \ref{eq:RD} can be written in a more compact way allowing a fast GPU implementation:
\begin{equation}
    x_i(t + 1) = \frac{x_i(t) \odot (W x(t))_i}{x_i(t)(W x(t))_i^T}
\end{equation}
where $\odot$ represents the Hadamard (element-wise) product.

\begin{figure}[t!]
 \centering
 \includegraphics[width=\textwidth]{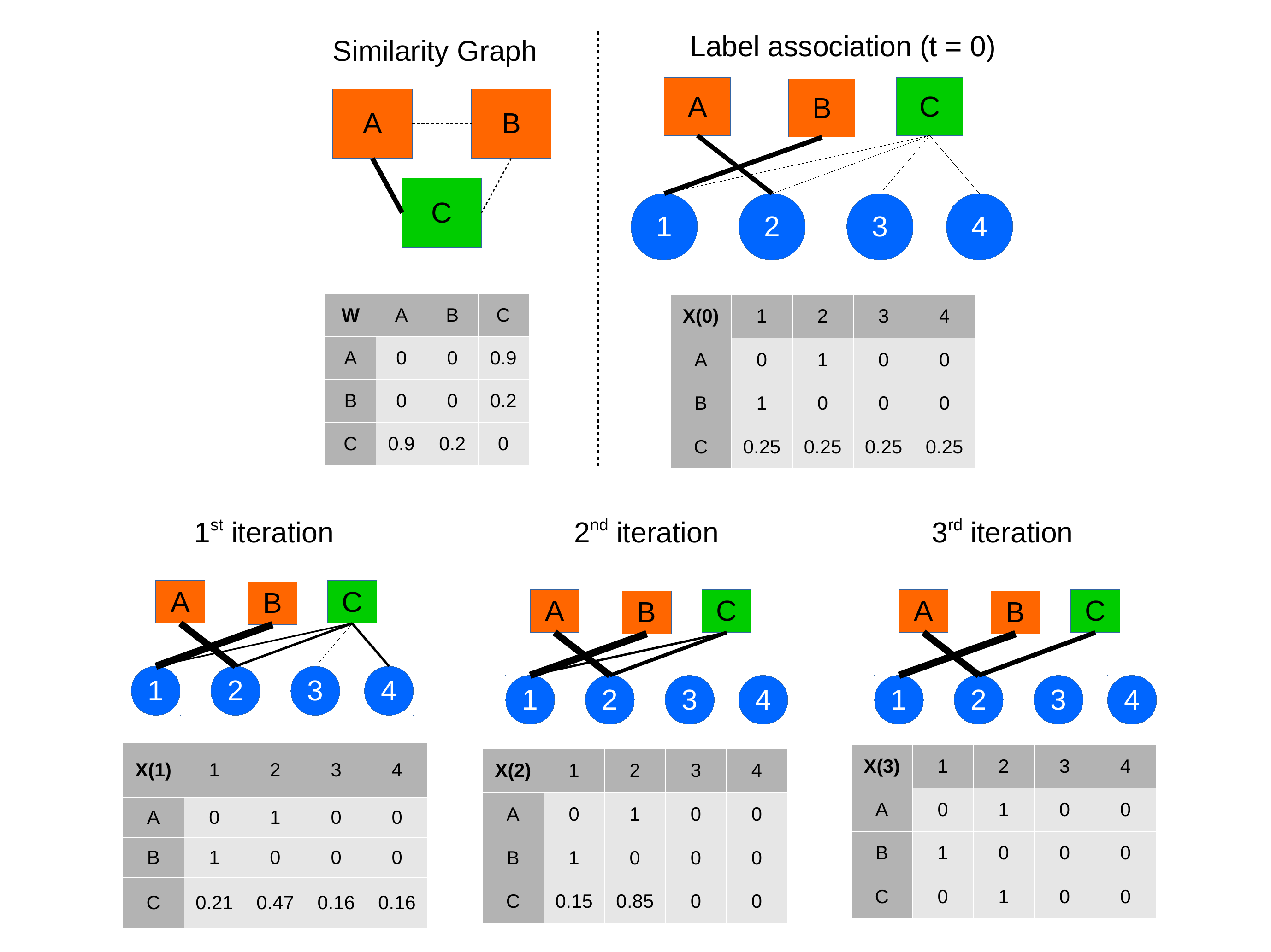}
 \caption{The dynamics of the GTG. The algorithm takes in input similarities between objects and hard/soft labelings of the object themselves. After three iterations, the algorithm has converged, generating a pseudo-label with 100\% confidence.}
\end{figure}

\section{Label Generation}
The previously explained framework can be applied to a dataset with many unlabeled objects to perform an automatic labeling and thus increase the availability of training objects. In this chapter we deal with datasets for image classification, but our approach can be applied in other domains too.

\textbf{Preliminary step}: both the labeled and unlabeled sets can be refined to obtain more informative feature vectors. In this chapter, we used fc7 features of CNNs trained on ImageNet, but in principle, any type of features can be considered. Our particular choice was motivated because fc7 features work significantly better than traditional computer vision features (SIFT \cite{DBLP:journals/ijcv/Lowe04} and its variations). While this might seem counter-intuitive (using pre-trained CNNs on ImageNet, while we are solving the problem of limited labeled data), we need to consider that our datasets are different from ImageNet (they come from different distributions), and by using some other dataset to pre-train our networks, we are not going against the spirit of our idea.

\textbf{Step 1}: the objects are assigned to initial probability distributions, needed to start the GTG. The labeled ones use their respective one-hot label representations, while the unlabeled ones can be set to a uniform distribution among all the labels. In presence of previous possessed information, some labels can be directly excluded in order to start from a multi-peaked distribution, which if chosen wisely, can improve the final results.

\textbf{Step 2}: the extracted features are used to compute the similarity matrix $W$. The literature \cite{DBLP:conf/nips/Zelnik-ManorP04} presents multiple methods to obtain a $W$ matrix 
and extra care should be taken when performing this step, since an incorrect choice in its computation can determine a failure in the transductive labeling.

\textbf{Step 3}: once $W$ is computed, graph transduction game can be played (up to convergence) among the objects to obtain the final probabilities which determine the label for the unlabeled objects.

The resulting labeled dataset can then be used to train a classification model. This is very convenient for several reasons: 1) CNNs are fully parametric models, so we do not need to store the training set in memory like in the case of graph transduction. In some aspect, the CNN is approximating in a parametric way the GTG algorithm; 2) the inference stage on CNNs is extremely fast (real-time); 3) CNN features can be used for other problems, like image segmentation, detection and classification, something that we cannot do with graph-transduction or with classical machine learning methods (like SVM). In the next section we will report the results obtained from state-of-the-art CNNs, and compare those results with the same CNNs trained only on the labeled part of the dataset.

\section{Experiments}

\begin{table}[]
\centering
\begin{tabular}{|c|c|l|l|l|l|l|}
\hline
\multirow{2}{*}{\begin{tabular}[c]{@{}c@{}}Accuracy\\ 2\% labelled\end{tabular}} & \multicolumn{2}{c|}{Caltech}      & \multicolumn{2}{c|}{Indoors} & \multicolumn{2}{c|}{Scenenet} \\ \cline{2-7} 
                                                                                  & RN18                      & DN121 & RN18         & DN121         & RN18          & DN121         \\ \hline
		   GTG + CNN & \textbf{0.529} & \textbf{0.588} & \textbf{0.478} & \textbf{0.506} & \textbf{0.46} &	\textbf{0.455} \\
		   LS + CNN & 			0.459 & 	     0.517 &      	  0.434 & 		   0.486 &		  	0.359 & 		 0.435 \\
		   LH + CNN & 			0.393 & 	     0.463 &      	  0.372 & 		   0.438 &		  	0.312 & 		 0.319 \\
		   LP + CNN & 			0.397 & 	     0.462 &      	  0.373 & 		   0.425 &		  	0.293 & 		 0.357 \\
           CNN &			0.193 & 		 0.216 & 		  0.315 & 		0.302 & 			0.08 & 		 0.18 \\ \hline
\multirow{2}{*}{\begin{tabular}[c]{@{}c@{}}F score\\ 2\% labeled\end{tabular}}   & \multicolumn{2}{c|}{Caltech}      & \multicolumn{2}{c|}{Indoors} & \multicolumn{2}{c|}{Scenenet} \\ \cline{2-7} 
                                                                                  & \multicolumn{1}{l|}{RN18} & DN121 & RN18         & DN121         & RN18          & DN121         \\ \hline
		   GTG + CNN & \textbf{0.471} & \textbf{0.534} & \textbf{0.336} & \textbf{0.393} & \textbf{0.439} &	\textbf{0.435} \\
		   LS + CNN & 			0.392 & 	     0.462 &      	  0.352 & 		   0.403 &		  	0.342 & 		 0.417 \\
		   LH + CNN & 			0.367 & 	     0.446 &      	  0.262 & 		   0.335 &		  	0.331 & 		 0.342 \\
		   LP + CNN & 			0.321 & 	     0.381 &      	  0.29 & 		   0.111 &		  	0.278 & 		 0.344 \\
           CNN &			0.091 & 		 0.108 & 		  0.151 & 		0.131 & 			0.076 & 		 0.18 \\ \hline
\end{tabular}
\caption{The results of our algorithm, compared with the results of Label Spreading (LS), Label Harmonic (LH), Label Propagation (LP) and CNN, when only 2\% of the dataset is labeled. We see that in all three datasets and two different neural networks, our approach gives significantly better results than the competing approaches.} 
\label{2percent}
\vspace{1.5mm}
\end{table}

\begin{table}[]
\centering
\begin{tabular}{|c|c|l|l|l|l|l|}
\hline
\multirow{2}{*}{\begin{tabular}[c]{@{}c@{}}Accuracy\\ 5\% labelled\end{tabular}} & \multicolumn{2}{c|}{Caltech}      & \multicolumn{2}{c|}{Indoors} & \multicolumn{2}{c|}{Scenenet} \\ \cline{2-7} 
                                                                                  & RN18                      & DN121 & RN18         & DN121         & RN18          & DN121         \\ \hline
		   GTG + CNN & \textbf{0.667} & \textbf{0.71} & \textbf{0.552} & \textbf{0.585} & \textbf{0.628} &	\textbf{0.626} \\
		   LS + CNN & 			0.589 & 	     0.647 &      	  0.496 & 		   0.561 &		  	0.523 & 		 0.4562 \\
		   LH + CNN & 			0.589 & 	     0.665 &      	  0.527 & 		   0.555 &		  	0.549 & 		 0.588 \\
		   LP + CNN & 			0.532 & 	     0.60 &      	  0.454 & 		   0.502 &		  	0.442 & 		 0.513 \\
           CNN &			0.44 & 		 0.526 & 		  0.425 & 		0.438 & 			0.381 & 		 0.456 \\ \hline
\multirow{2}{*}{\begin{tabular}[c]{@{}c@{}}F score\\ 5\% labeled\end{tabular}}   & \multicolumn{2}{c|}{Caltech}      & \multicolumn{2}{c|}{Indoors} & \multicolumn{2}{c|}{Scenenet} \\ \cline{2-7} 
                                                                                  & \multicolumn{1}{l|}{RN18} & DN121 & RN18         & DN121         & RN18          & DN121         \\ \hline
		   GTG + CNN & \textbf{0.624} & \textbf{0.674} & \textbf{0.44} & 0.476 & \textbf{0.606} &	\textbf{0.62} \\
		   LS + CNN & 			0.544 & 	     0.601 &      	  0.428 & 		   \textbf{0.503} &		  	0.511 & 		 0.557 \\
		   LH + CNN & 			0.542 & 	     0.636 &      	  0.444 & 		   0.482 &		  	0.531 & 		 0.574 \\
		   LP + CNN & 			0.477 & 	     0.551 &      	  0.394 & 		   0.432 &		  	0.43 & 		 0.506 \\
           CNN &			0.37 & 		 0.467 & 		  0.279 & 		0.291 & 			0.376 & 		 0.448 \\ \hline
\end{tabular}
\caption{The results of our algorithm, compared with the results of Label Spreading (LS), Label Harmonic (LH), Label Propagation (LP) and CNN, when only 5\% of the dataset is labeled. We see that in all three datasets and two different neural networks, our approach gives significantly better results than the competing approaches.} 
\label{5percent}
\vspace{1.5mm}
\end{table}

\begin{table}[]
\centering
\begin{tabular}{|c|c|l|l|l|l|l|}
\hline
\multirow{2}{*}{\begin{tabular}[c]{@{}c@{}}Accuracy\\ 10\% labelled\end{tabular}} & \multicolumn{2}{c|}{Caltech}      & \multicolumn{2}{c|}{Indoors} & \multicolumn{2}{c|}{Scenenet} \\ \cline{2-7} 
                                                                                  & RN18                      & DN121 & RN18         & DN121         & RN18          & DN121         \\ \hline
		   GTG + CNN & \textbf{0.714} & \textbf{0.746} & \textbf{0.577} & \textbf{0.628} & \textbf{0.675} &	\textbf{0.681} \\
		   LS + CNN & 			0.636 & 	     0.702 &      	  0.541 & 		   0.592 &		  	0.631 & 		 0.608 \\
		   LH + CNN & 			0.646 & 	     0.716 &      	  0.548 & 		   0.595 &		  	0.578 & 		 0.643 \\
		   LP + CNN & 			0.594 & 	     0.672 &      	  0.49 & 		   0.553 &		  	0.499 & 		 0.565 \\
           CNN &			0.599 & 		 0.655 & 		  0.527 & 		0.563 & 			0.544 & 		 0.599 \\ \hline
\multirow{2}{*}{\begin{tabular}[c]{@{}c@{}}F score\\ 10\% labeled\end{tabular}}   & \multicolumn{2}{c|}{Caltech}      & \multicolumn{2}{c|}{Indoors} & \multicolumn{2}{c|}{Scenenet} \\ \cline{2-7} 
                                                                                  & \multicolumn{1}{l|}{RN18} & DN121 & RN18         & DN121         & RN18          & DN121         \\ \hline
		   GTG + CNN & \textbf{0.681} & \textbf{0.717} & \textbf{0.49} & \textbf{0.558} & \textbf{0.646} &	\textbf{0.665} \\
		   LS + CNN & 			0.601 & 	     0.675 &      	  0.465 & 		   0.549 &		  	0.62 & 		 0.601\\
		   LH + CNN & 			0.607 & 	     0.689 &      	  0.466 & 		   0.523 &		  	0.568 & 		 0.634 \\
		   LP + CNN & 			0.545 & 	     0.635 &      	  0.411 & 		   0.48 &		  	0.488 & 		 0.554 \\
           CNN &			0.554 & 		 0.615 & 		  0.414 & 		0.466 & 			0.538 & 		 0.589 \\ \hline
\end{tabular}
\caption{The results of our algorithm, compared with the results of Label Spreading (LS), Label Harmonic (LH), Label Propagation (LP) and CNN, when only 10\% of the dataset is labeled. We see that in all three datasets and two different neural networks, our approach gives significantly better results than the competing approaches.} 
\label{10percent}
\vspace{1.5mm}
\end{table}

In order to assess the quality of the algorithm, we used it to automatically label three known realistic datasets, namely \emph{Caltech-256} \cite{caltech}, \emph{Indoor Scene Recognition} \cite{DBLP:conf/cvpr/QuattoniT09} and \emph{SceneNet-100} \cite{DBLP:conf/eccv/KadarB14}. \emph{Caltech-256} contains $30607$ images belonging to $256$ different categories and it is used for object recognition tasks. \emph{Indoor Scene Recognition} is a dataset containing $15620$ images of different common places (restaurants, bedrooms, etc.), divided in $67$ categories and, as the name says, it is used for scene recognition. \emph{SceneNet-100} database is a publicly available online ontology for scene understanding that organizes scene categories according to their perceptual relationships. The dataset contains $10000$ real-world images, separated into $100$ different classes.

Each dataset was split in a training (70\%) and a testing (30\%) set. In addition, we further randomly split the training set in a small labeled part and a large unlabeled one, according to three different percentages for labeled objects (2\%, 5\%, 10\%). For feature representation, we used two models belonging to state-of-the-art CNN families of architectures, ResNet and DenseNet. In particular we used the smallest models offered in PyTorch library, the choice motivated by the fact that our datasets are relatively small, and so models with smaller number of parameters are expected to work better. The features were combined to generate the similarity matrix $W$. The matrix for GTG model was initialized as described in the previous section. We ran the GTG algorithm up to convergence, with the pseudo-labels being computed by doing an $argmax$ over the final probability vectors.

%\input{results_figure_relative}
%%%%%%%%%%%%%%%%%%%%%%%%%%%%%%
\begin{figure*}[]
\centering
\begin{subfigure}{\includegraphics[width=.4\textwidth]{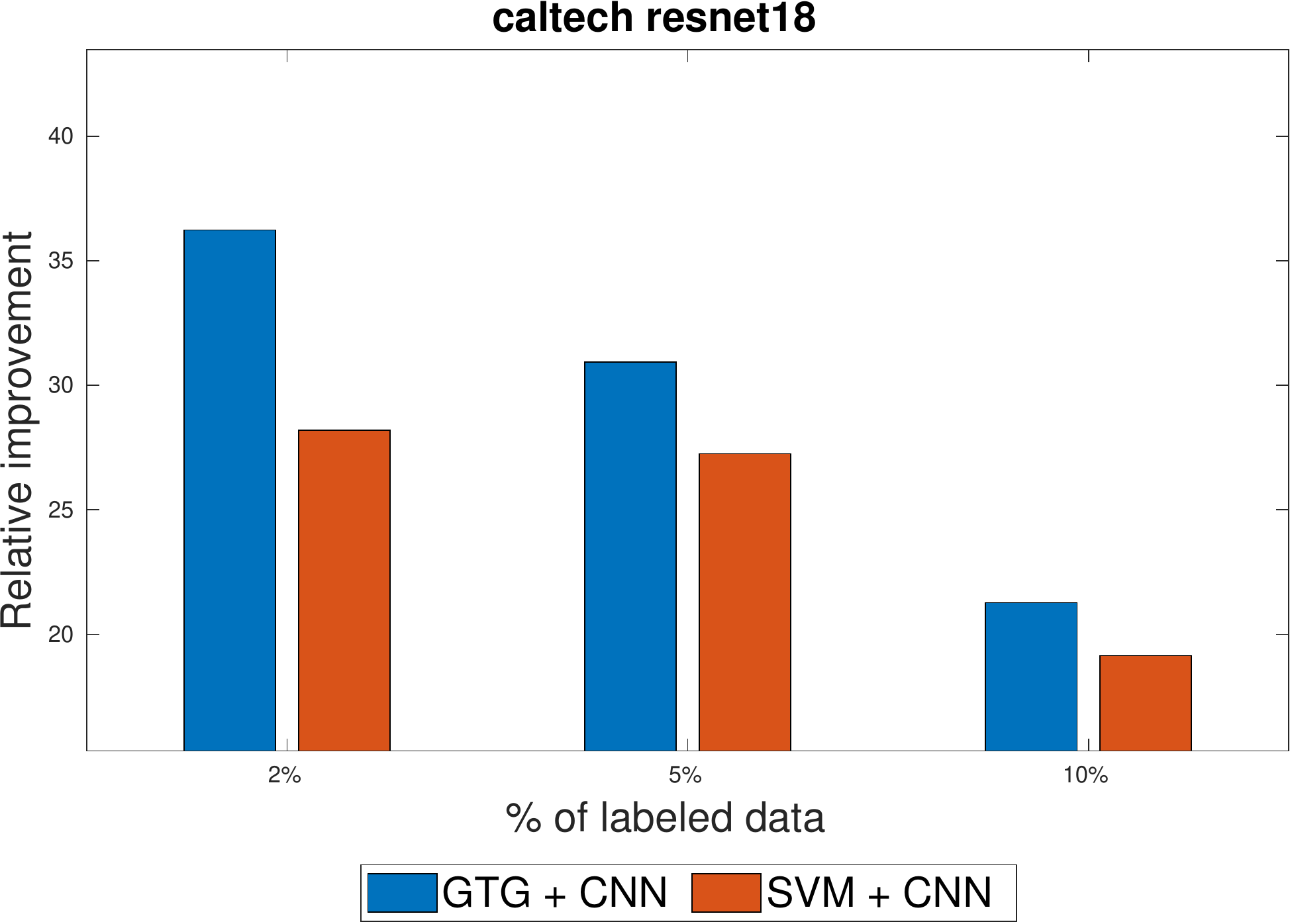}} \end{subfigure}
\begin{subfigure}{\includegraphics[width = .4\textwidth]{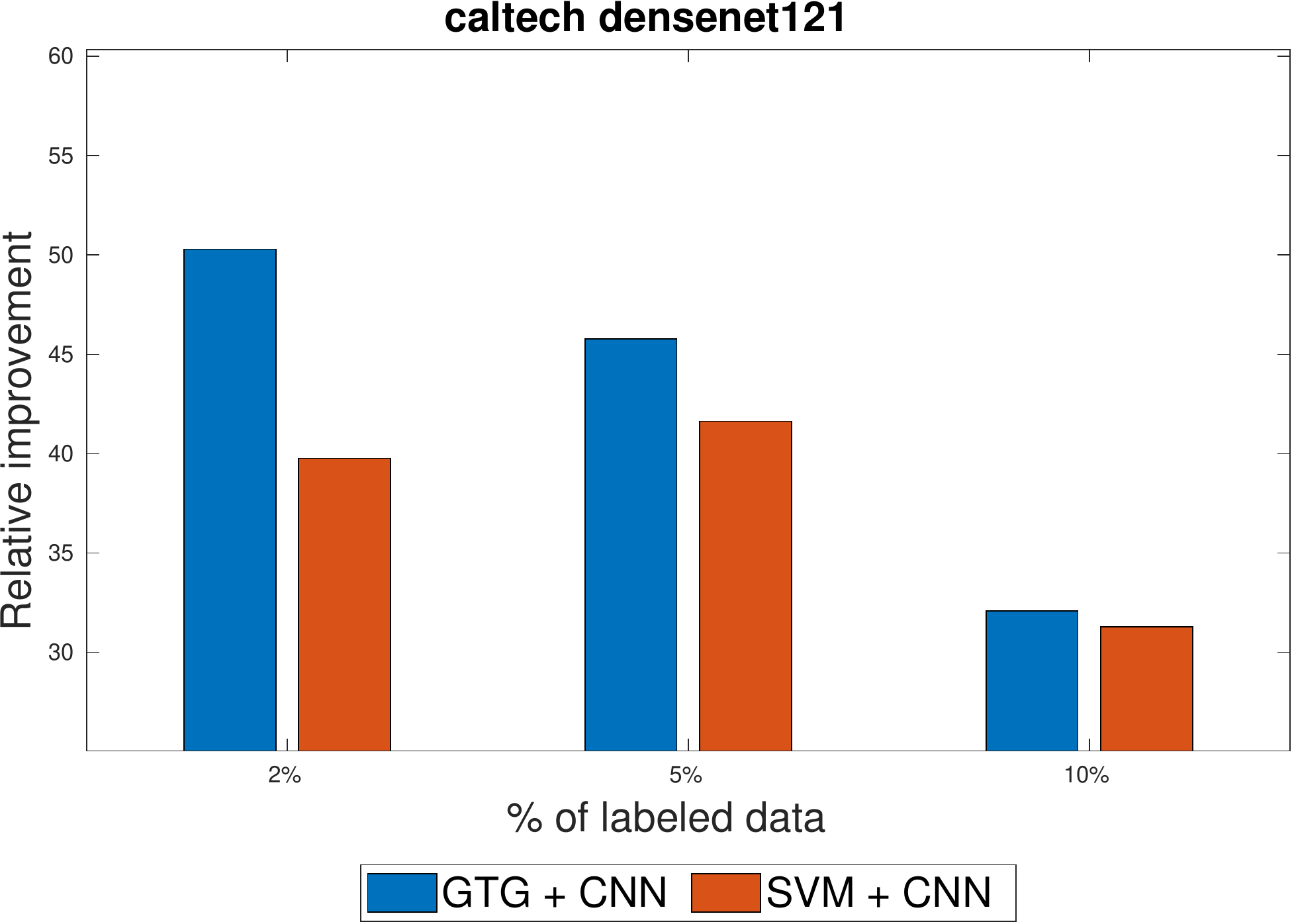}} \end{subfigure}
\begin{subfigure}{\includegraphics[width = .4\textwidth]{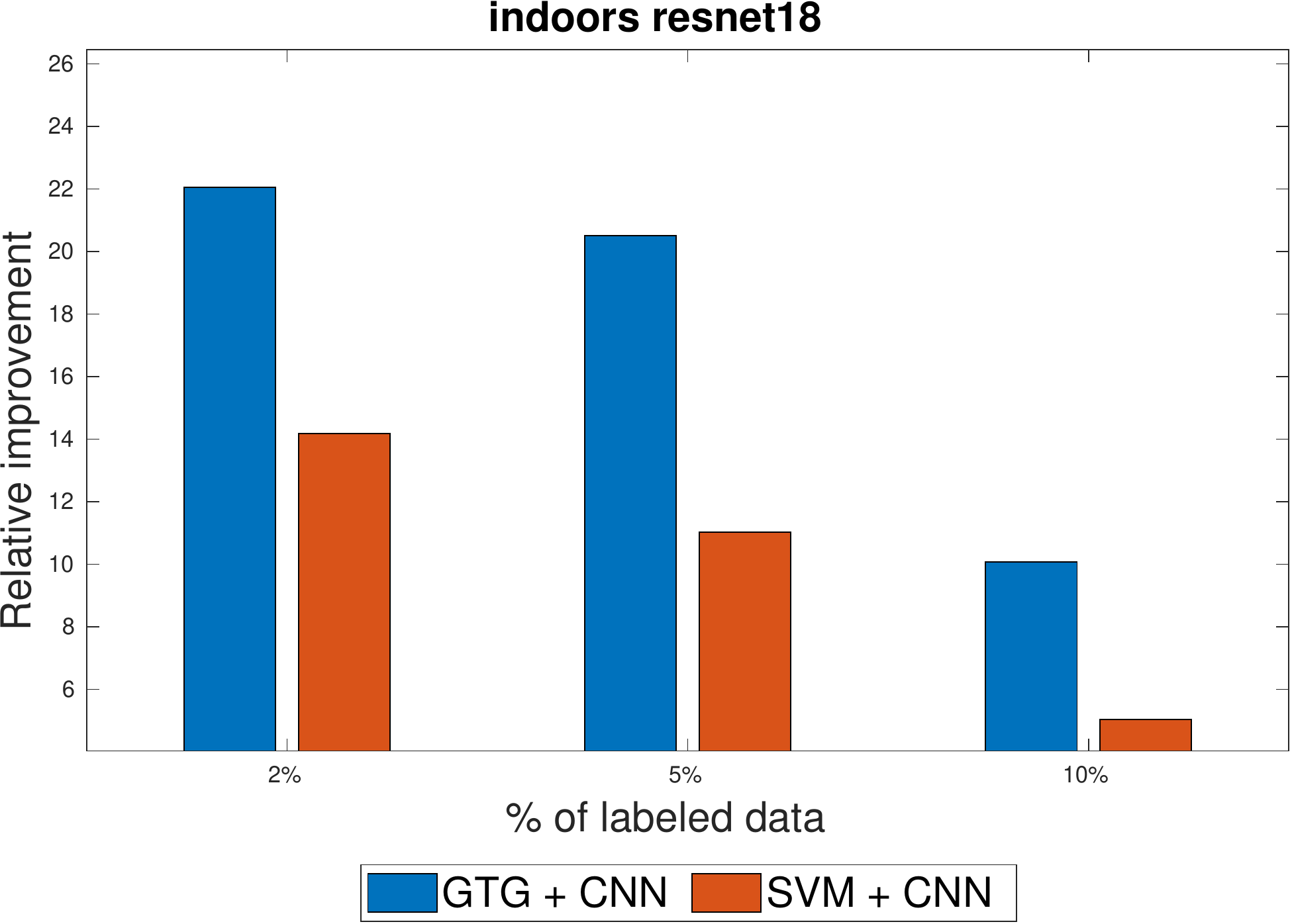}} \end{subfigure}
\begin{subfigure}{\includegraphics[width = .4\textwidth]{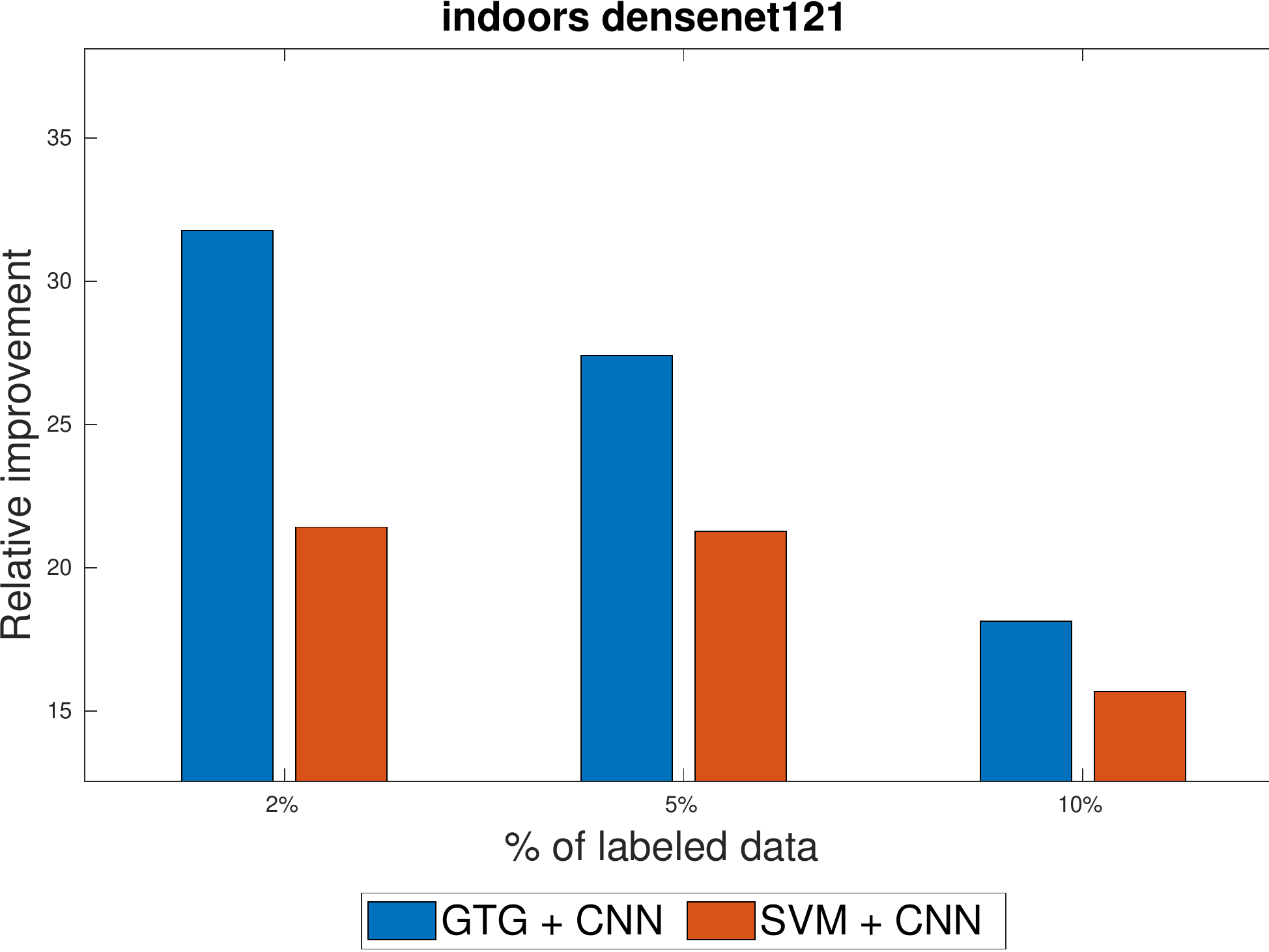}} \end{subfigure}
\begin{subfigure}{\includegraphics[width = .4\textwidth]{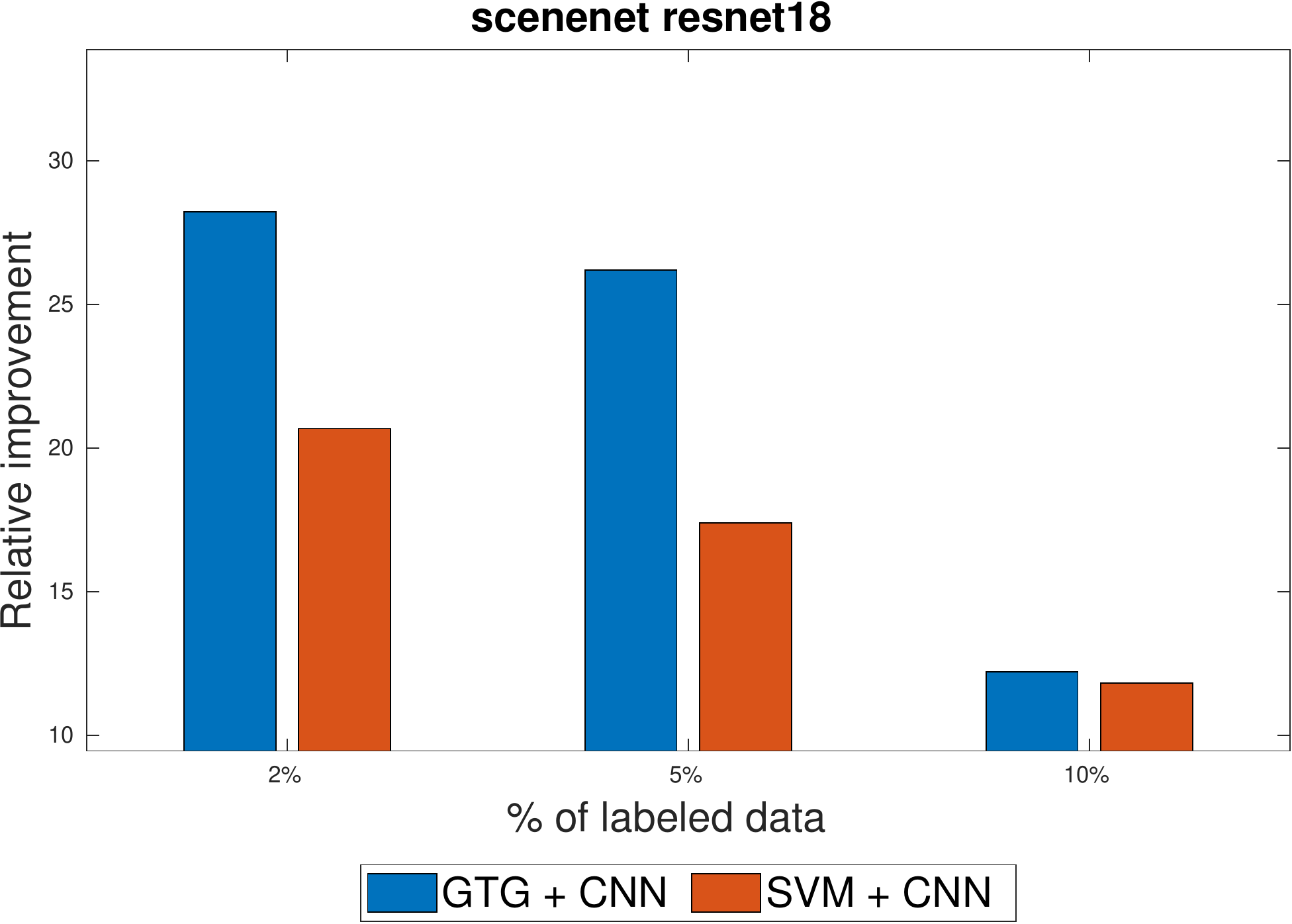}} \end{subfigure}
\begin{subfigure}{\includegraphics[width = .4\textwidth]{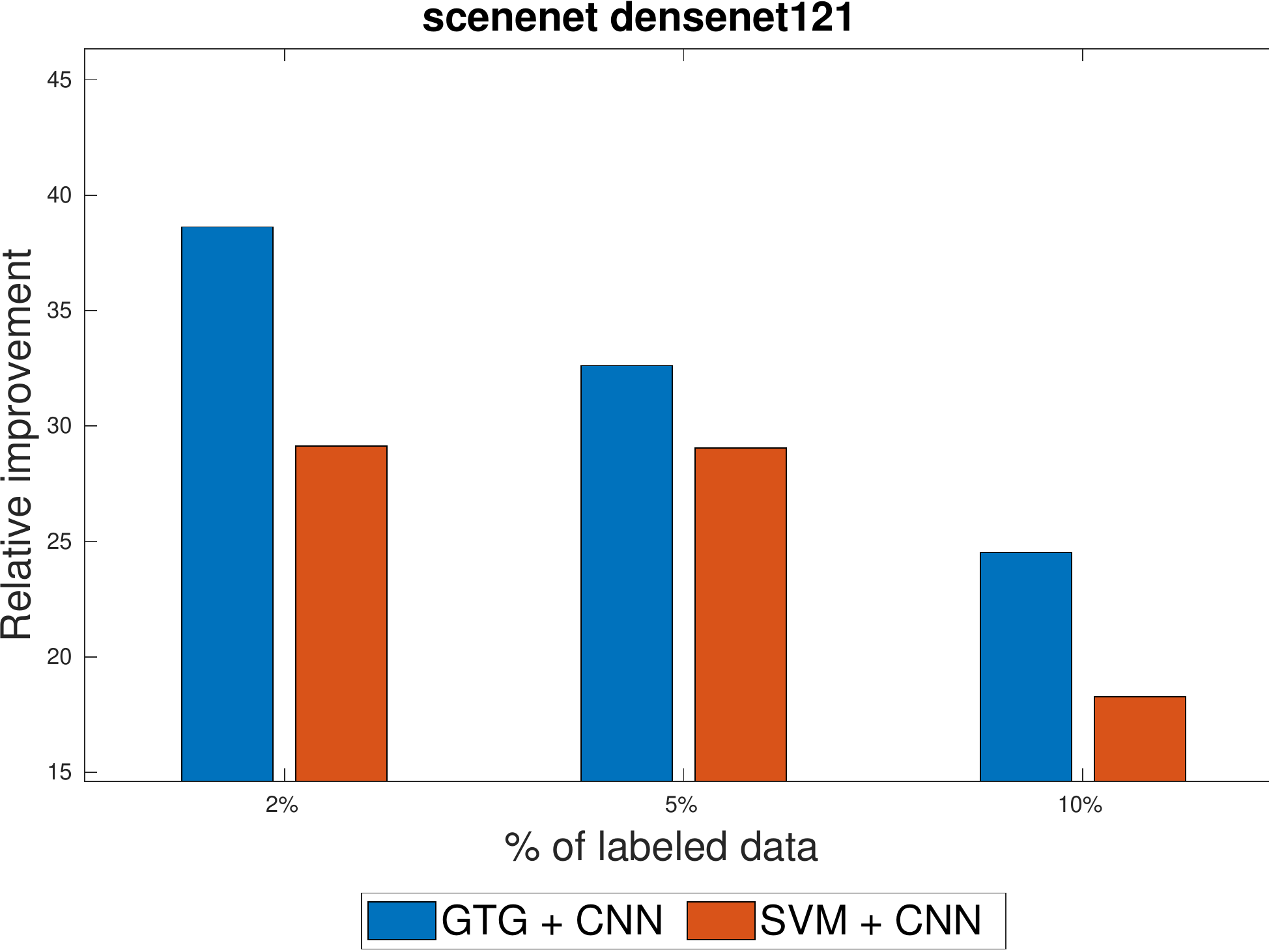}} \end{subfigure}
\caption{Results obtained on different datasets and CNNs. Here the relative improvements with respect to the CNN accuracy is reported. As can be seen, the biggest advantage of our method compared to the other two approaches, is when the number of labeled points is extremely small (2\%). When the number of labeled points increases, the difference on accuracy becomes smaller, but nevertheless our approach continues being significantly better than CNN, and in most cases, it gives better results than the alternative approach.}
\label{fig:results}
\end{figure*}
%%%%%%%%%%%%%%%%%%%%%%%%%%%%%%
We then trained \emph{ResNet18} (RN18) and \emph{DenseNet121} (DN121) in the entire dataset, by not having a distinction between labels and pseudo-labels, using Adam optimizer \cite{DBLP:conf/iclr/Kingma14} with $3 * 10^{-4}$ learning rate. We think that the results reported in this section are conservative, and can be improved with a more careful training of the networks, and by doing an exhaustive search over the space of hyper-parameters.

For comparison, we performed an alternative approach, by replacing GTG with a first-order information algorithm, namely linear SVM, in addition to other well-known label-propagation algorithms (namely label propagation \cite{zhu2002learning}, label spreading \cite{DBLP:conf/nips/ZhouBLWS03}, label harmonic \cite{DBLP:conf/icml/ZhuGL03}. While we experimented also with kernel SVM, we saw that its results are significantly worse than those of linear SVM, most likely because the features were generated from a CNN and so they are already quite good, having transformed the feature space in order to solve the classification problem linearly. 

On Table \ref{2percent} we give the results of the accuracy and F score on the testing set, in all three datasets, while the number of labels is only 2\% for each of the datasets ($400$ observations for Caltech-256, $200$ observations for Indoor, and $140$ observations for Scenenet). In all three datasets, and both CNNs, our results are significantly better than those of CNNs trained only in the labeled data, or the results of the competing second-order label augmenters. Table \ref{5percent} and Table \ref{10percent} give the results of the accuracy and F score while the number of labeled images is 5\%, respectively 10\%. It can be seen that with the number of labeled points increasing, the performance boost of our model becomes smaller, but our performance still gives better (or equal) results to the alternative approaches in all cases, and it gives significantly better results than CNN in all cases.

Figure \ref{fig:results} shows the results of our approach compared with the linear SVM and with the results of CNN. We plotted the relative improvement of our model and the alternative approach over CNN. When the number of labels is very small (2\%), in all three datasets we have significantly better improvements compared with the alternative approach. Increasing the number of labels to 5\% and 10\%, this trend persists. In all cases, our method gives significant improvements compared to CNN trained on only the labeled part of the dataset, with the most interesting case (only 2\% of labeled observations), our model gives 36.24\% relative improvement over CNN for \emph{ResNet18} and 50.29\% relative improvement for \emph{DenseNet121}.

\subsection{Comparison with Deep Learning models}

We also compared our approach with purely deep-learning models. Specifically, we compared our model with $\Pi$-model \cite{DBLP:conf/iclr/LaineA17}, Mean-Teacher \cite{DBLP:conf/iclr/TarvainenV17}, Pseudo-Label \cite{Lee_pseudo-label:the} and VAT \cite{DBLP:journals/pami/MiyatoMKI19}, methods which have been intensively evaluated in \cite{DBLP:conf/nips/OliverORCG18}, from whom we also got the code. We were particularly interested in the case where the number of labels was extremely small (from $1$ to $25$ for class). As in \cite{DBLP:conf/nips/OliverORCG18}, we used the \textit{CIFAR-10} \cite{Krizhevsky2009} dataset, and we used the exact pre-processing and training scheme as given on the paper. Table \ref{tab:cifar10} shows the results of our method in comparison with other deep learning based, label propagation based and transfer learning methods. We see that our method significantly outperforms all the other methods, with the difference becoming smaller while the number of labelled points increases. However, deep learning methods perform better when the number of labels is higher than $500$, suggesting that a common ground can be reached with our method being used when the number of labels is extremely small, while deep learning methods should be used when the number of labels increases.

We also did the same experiment as \cite{DBLP:conf/nips/OliverORCG18}, where the network instead of being pre-trained on \textit{ImageNet} dataset, it was pre-trained in a subset of \textit{ImageNet}, which does not contain any similar classes to the classes of \textit{CIFAR-10}. The  omitted classes can be found in Appendix F of \cite{DBLP:conf/nips/OliverORCG18}. We see in Table \ref{tab:cifar10-2} that the performances of all algorithms bar Label Harmonic suffer for a few percentage points, with our method still being by far the best method where the number of labelled points is $10$, $50$ and $100$, but it gets massively outperformed from Pseudo-Label \cite{Lee_pseudo-label:the} method where the number of labelled points is $250$. Nevertheless, the extremely good performance of our model where the number of labelled points is extremelly small shows a relative robustness over the choice of the dataset the network has been pre-trained.

\begin{table}[]
\centering
\resizebox{0.8\textwidth}{!}{%
\begin{tabular}{@{}llccccccccccc@{}}
\toprule
\textbf{Method} & \textbf{10} & \textbf{50} & \textbf{100} & \textbf{250} \\ \midrule
Pi Model & 0.1 & 0.1 & 0.362 & 0.741\\
Mean Teacher & 0.1 & 0.1 & 0.266 & 0.734 \\
Pseudo-Labels & 0.13 & 0.335 & 0.603 & 0.778\\
VAT & 0.119 & 0.227 & 0.321 & 0.705\\
Transfer Learning & 0.289 & 0.518 & 0.605 & 0.711 \\
LS + CNN & 0.534 & 0.666 & 0.714 & 0.758 \\
HF + CNN & 0.113 & 0.232 & 0.352 & 0.659 \\
\midrule
\textbf{GTG + CNN} & \textbf{0.575} & \textbf{0.733} & \textbf{0.764} & \textbf{0.791} \\ \bottomrule
\end{tabular}%
}
\caption{The results of our method in \textit{CIFAR-10} dataset, compared with the results of other deep learning approaches, where the network has been pre-trained in \textit{ImageNet} dataset. $10$, $50$, $100$, and $250$ represent the total number of labeled points in the dataset.}
\label{tab:cifar10}
\end{table}

\begin{table}[]
\centering
\resizebox{0.8\textwidth}{!}{%
\begin{tabular}{@{}llccccccccccc@{}}
\toprule
\textbf{Method} & \textbf{10} & \textbf{50} & \textbf{100} & \textbf{250} \\ \midrule
Pi Model & 0.1 & 0.1 & 0.11 & 0.681\\
Mean Teacher & 0.1 & 0.1 & 0.12 & 0.695 \\
Pseudo-Labels & 0.132 & 0.236 & 0.498 & \textbf{0.764}\\
VAT & 0.118 & 0.193 & 0.323 & 0.641\\
Transfer Learning & 0.271 & 0.506 & 0.589 & 0.689 \\
LS + CNN & 0.484 & 0.621 & 0.684 & 0.703 \\
HF + CNN & 0.113 & 0.309 & 0.556 & 0.642 \\
\midrule
\textbf{GTG + CNN} & \textbf{0.514} & \textbf{0.656} & \textbf{0.713} & 0.719 \\ \bottomrule
\end{tabular}%
}
\caption{The results of our method in \textit{CIFAR-10} dataset, compared with the results of other deep learning approaches, where the network has been pre-trained in a subset of \textit{ImageNet} dataset, which does not contain any class which is similar to classes of \textit{CIFAR-10}. $10$, $50$, $100$, and $250$ represent the total number of labeled points in the dataset.}
\label{tab:cifar10-2}
\end{table}

\section{Conclusions and Future Work}

In this section, we proposed and developed a game-theoretic model which can be used as a semi-supervised learning algorithm in order to label the unlabeled observations and so augment datasets. Different types of algorithms (including state-of-the-art CNNs) can then be trained on the extended dataset, where the ``pseudo-labels'' can be treated as normal labels.

Our method massively outperforms the other non deep learning based methods, and in the cases of datasets which have only a few labelled points, it massively outperforms deep learning methods where the semi-supervision is intrinsic part of the model itself. Additionally, we offer a different perspective, developing a model which is algorithm-agnostic, and which doesn't even need the data to be on feature-based format, while also being competitive with state-of-the-art methods.

Part of the future work will consist on tailoring our model specifically towards convolutional neural networks and to study if it complements the other deep-learning  semi-supervised models. Additionally, we are working on making the model end-to-end, where the GTG algorithm will be part of the neural network, instead of being used as a pre-processing sete. Finally, we believe that the true potential of the model can be unleashed when the data is in some non-traditional format. In particular, we plan to use our model in the fields of bio-informatics and natural language processing, where non-conventional learning algorithms need to be developed. A direct extension of this work is to embed into the model the similarity between classes which has been proven to significantly boost the performances of learning algorithms.

\chapter{The Group Loss for Deep Metric Embedding}

\section{Disclaimer}

The work presented in this chapter is based on the following paper:

\begin{description}
  \item \textbf{Ismail Elezi}, Sebastiano Vascon, Alessandro Torcinovich, Marcello Pelillo and Laura Leal-Taix\'e; \textit{The Group Loss for Deep Metric Learning \cite{DBLP:journals/corr/groupLoss}}; submitted to European Conference on Computer Vision (ECCV 2020)
\end{description} 

The contributions of the author are the following:

\begin{description}
  \item[$\bullet$ Coming] up with the pipeline and the modifications of the algorithm.
  \item[$\bullet$ Writing] the code.
  \item[$\bullet$ Performing] all the experiments.
  \item[$\bullet$ Writing] the majority of the paper.
\end{description}

\section{Introduction}
Measuring object similarity is at the core of many important machine learning problems like clustering and object retrieval. 
For visual tasks, this means learning a distance function over images. With the rise of deep neural networks, the focus has rather shifted towards learning a feature embedding that is easily separable using a simple distance function, such as the Euclidean distance. 
In essence, objects of the same class (similar) should be close by in the learned manifold, while objects of a different class (dissimilar) should be far away.

Historically, the best performing approaches get deep feature embeddings from the so-called siamese networks \cite{bromley1994signature}, which are typically trained using the contrastive loss \cite{bromley1994signature} or the triplet loss \cite{DBLP:conf/nips/SchultzJ03,DBLP:journals/jmlr/WeinbergerS09}. 
A clear drawback of these losses is that they only consider pairs or triplets of data points, missing key information about the relationships between all members of the mini-batch. On a mini-batch of size $n$, despite that the number of pairwise relations between samples is $\mathcal{O}(n^2)$, contrastive loss uses only $\mathcal{O}(n/2)$ pairwise relations, while triplet loss uses $\mathcal{O}(2n/3)$ relations.
Additionally, these methods consider only the relations between objects of the same class (positives) and objects of other classes (negatives), without making any distinction that negatives belong to different classes.
This leads to not taking into consideration the global structure of the embedding space, and consequently results in lower clustering and retrieval performance. 
To compensate for that, researchers rely on other tricks to train neural networks for deep metric learning: intelligent sampling \cite{DBLP:conf/iccv/ManmathaWSK17}, multi-task learning \cite{DBLP:conf/cvpr/ZhangZLZ16} or hard-negative mining \cite{DBLP:conf/cvpr/SchroffKP15}. 
Recently, researchers have been increasingly working towards exploiting in a principled way the global structure of the embedding space \cite{DBLP:journals/corr/abs-1906-07589,DBLP:conf/cvpr/Cakir0XKS19,DBLP:conf/cvpr/0003CBS18,DBLP:conf/cvpr/WangHKHGR19}, typically by designing ranking loss functions instead of following the classic triplet formulations.

% MAIN idea of gruop loss
In a similar spirit, we propose {\it Group Loss}, a novel loss function for deep metric learning that considers the similarity between all samples in a mini-batch. To create the mini-batch, we sample from a fixed number of classes, with samples coming from a class forming a \textit{group}. Thus, each mini-batch consists of several randomly chosen groups, and each group has a fixed number of samples. An iterative, fully-differentiable label propagation algorithm is then used to build feature embeddings which are similar for samples belonging to the same group, and dissimilar otherwise. 

At the core of our method lies an iterative process called {replicator dynamics} \cite{weibull1997evolutionary,DBLP:journals/neco/ErdemP12}, that refines the local information, given by the softmax layer of a neural network, with the global information of the mini-batch given by the similarity between embeddings. 
The driving rationale is that the more similar two samples are, the more they affect each other in choosing their final label and tend to be grouped together in the same group, while dissimilar samples do not affect each other on their choices.
Neural networks optimized with the Group Loss learn to provide similar features for samples belonging to the same class, making clustering and image retrieval easier.

Our \textbf{contribution} in this work is four-fold: 
\begin{itemize}
    \item We propose a novel loss function to train neural networks for deep metric embedding that takes into account the local information of the samples, as well as their similarity.
    \item We propose a differentiable label-propagation iterative model to embed the similarity computation within backpropagation, allowing end-to-end training with our new loss function.
    \item We perform a comprehensive robustness analysis showing the stability of our module with respect to the choice of hyperparameters. 
    \item We show state-of-the-art qualitative and quantitative results in several standard clustering and retrieval datasets. 
\end{itemize}

\begin{figure*}[ht!]
\includegraphics[width=\textwidth, trim={0 4.9cm 0.2cm 0},clip]{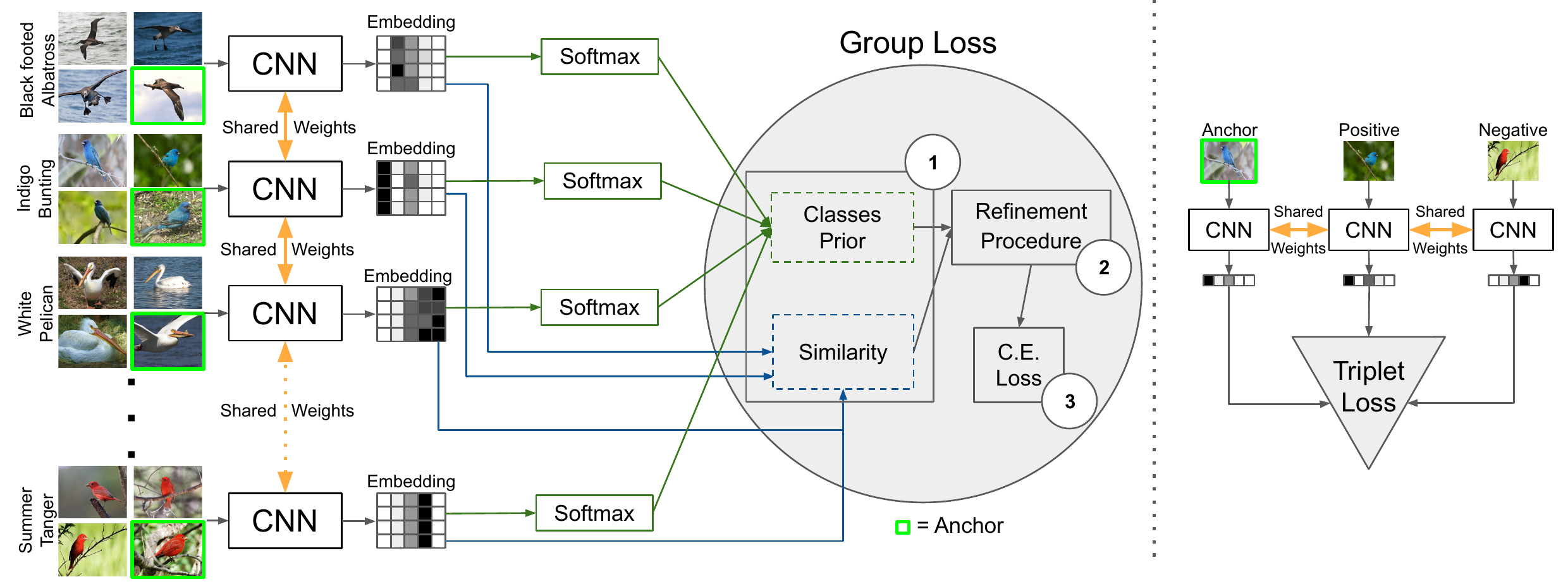} %architecture.jpg}
\centering
\caption{A comparison between a neural model trained with the Group Loss (left) and the triplet loss (right). Given a mini-batch of images belonging to different classes, their embeddings are computed through a convolutional neural network. Such embeddings are then used to generate a similarity matrix that is fed to the Group Loss along with prior distributions of the images on the possible classes. The green contours around some mini-batch images refer to \emph{anchors}. It is worth noting that, differently from the triplet loss, the Group Loss considers multiple classes and the pairwise relations between all the samples. Numbers from \raisebox{.4pt}{\textcircled{\raisebox{-1pt} {1}}} to \raisebox{.4pt}{\textcircled{\raisebox{-1pt} {3}}} refer to the Group Loss steps, see Sec \ref{sec:gl_overview} for the details.
} 
\vspace{-0.5cm}
\label{fig:pipeline}
%left, lower, right, upper {0 5.1cm 0.6cm 0}
\end{figure*}

\section{Related Work}

\noindent\textbf{Classical metric learning losses.} The first attempt at using a neural network for feature embedding was done in the seminal work of Siamese Networks \cite{bromley1994signature}. A cost function called \textit{contrastive loss} was designed in such a way as to minimize the distance between pairs of images belonging to the same cluster, and maximize the distance between pairs of images coming from different clusters. In \cite{DBLP:conf/cvpr/ChopraHL05}, researchers used the principle to successfully address the problem of face verification. %\noindent\textbf{Triplet loss.}
Another line of research on convex approaches for metric learning led to the triplet loss \cite{DBLP:conf/nips/SchultzJ03,DBLP:journals/jmlr/WeinbergerS09}, which was later combined with the expressive power of neural networks \cite{DBLP:conf/cvpr/SchroffKP15}. The main difference from the original Siamese network is that the loss is computed using triplets (an anchor, a positive and a negative data point). 
The loss is defined to make the distance between features of the anchor and the positive sample smaller than the distance between the anchor and the negative sample. The approach was so successful in the field of face recognition and clustering, that soon many works followed. The majority of works on the Siamese architecture consist of finding better cost functions, resulting in better performances on clustering and retrieval. In \cite{DBLP:conf/nips/Sohn16}, the authors generalized the concept of triplet by allowing a joint comparison among $N - 1$ negative examples instead of just one. 
\cite{DBLP:conf/cvpr/SongXJS16} designed an algorithm for taking advantage of the mini-batches during the training process by lifting the vector of pairwise distances within the batch to the matrix of pairwise distances, thus enabling the algorithm to learn feature embedding by optimizing a novel structured prediction objective on the lifted problem. The work was later extended in \cite{DBLP:conf/cvpr/SongJR017}, proposing a new metric learning scheme based on structured prediction that is designed to optimize a clustering quality metric, i.e., the normalized mutual information \cite{DBLP:journals/corr/abs-1110-2515}. Better results were achieved on \cite{DBLP:conf/iccv/WangZWLL17}, where the authors proposed a novel angular loss, which takes angle relationship into account. A very different problem formulation was given by \cite{DBLP:conf/icml/LawUZ17}, where the authors used a spectral clustering-inspired approach to achieve deep embedding. A recent work presents several extensions of the triplet loss that reduce the bias in triplet selection by adaptively correcting the distribution shift on the selected triplets \cite{DBLP:conf/eccv/YuLGDT18}. 

\noindent\textbf{Sampling and ensemble methods.} Knowing that the number of possible triplets is extremely large even for moderately-sized datasets, and having found that the majority of triplets are not informative \cite{DBLP:conf/cvpr/SchroffKP15}, researchers also investigated sampling. In the original triplet loss paper \cite{DBLP:conf/cvpr/SchroffKP15}, it was found that using semi-hard negative mining, the network can be trained to a good performance, but the training is computationally inefficient. 
The work of \cite{DBLP:conf/iccv/ManmathaWSK17} found out that while the majority of research is focused on designing new loss functions, selecting training examples plays an equally important role. The authors proposed a distance-weighted sampling procedure, which selects more informative and stable examples than traditional approaches, achieving excellent results in the process. A similar work was that of \cite{DBLP:conf/eccv/GeHDS18} where the authors proposed a hierarchical version of triplet loss that learns the sampling all-together with the feature embedding. 
The majority of recent works has been focused on complementary research directions such as intelligent sampling \cite{DBLP:conf/iccv/ManmathaWSK17,DBLP:conf/eccv/GeHDS18,DDBLP:conf/cvpr/Duan2019,DDBLP:conf/cvpr/Wand2019,DDBLP:conf/cvpr/Xu2019} or ensemble methods \cite{DBLP:conf/eccv/XuanSP18,DDBLP:conf/cvpr/Sanakoyeu2019,DBLP:conf/eccv/KimGCLK18,DBLP:conf/iccv/OpitzWPB17,DBLP:conf/iccv/YuanYZ17}. As we will show in the experimental section, these can be combined with our novel loss.

\noindent\textbf{Other related problems.} In order to have a focused and concise paper, we mostly discuss methods which tackle image ranking/clustering in standard datasets. Nevertheless, we acknowledge related research on specific applications such as person re-identification or landmark recognition, where researchers are also gravitating towards considering the global structure of the mini-batch. In \cite{DBLP:conf/cvpr/0003CBS18} the authors propose a new hashing method for learning binary embeddings of data by optimizing Average Precision metric. In \cite{DBLP:journals/corr/abs-1906-07589,DBLP:conf/cvpr/0003LS18} authors study novel metric learning functions for local descriptor matching on landmark datasets. \cite{DBLP:conf/cvpr/Cakir0XKS19} designs a novel ranking loss function for the purpose of few-shot learning. Similar works that focus on the global structure have shown impressive results in the field of person re-identification \cite{DBLP:conf/cvpr/ZhaoXC19,DBLP:journals/corr/abs-1904-11397}.

\noindent\textbf{Classification-based losses.} The authors of \cite{DBLP:conf/iccv/Movshovitz-Attias17} proposed to optimize the triplet loss on a different space of triplets than the original samples, consisting of an anchor data point and similar and dissimilar learned proxy data points. These proxies approximate the original data points so that a triplet loss over the proxies is a tight upper bound of the original loss. The final formulation of the loss is shown to be similar to that of softmax cross-entropy loss, challenging the long-hold belief that classification losses are not suitable for the task of metric learning. Recently, the work of \cite{DBLP:journals/corr/abs-1811-12649} showed that a carefully tuned normalized softmax cross-entropy loss function combined with a balanced sampling strategy can achieve competitive results. A similar line of research is that of \cite{DBLP:conf/aaai/ZhengJSZWH19}, where the authors use a combination of normalized-scale layers and Gram-Schmidt optimization to achieve efficient usage of the softmax cross-entropy loss for metric learning. 
The work of \cite{DBLP:journals/corr/abs-1909-05235} goes a step further by taking into consideration the similarity between classes. Furthermore, the authors use multiple centers for class, allowing them to reach state-of-the-art results, at a cost of significantly increasing the number of parameters of the model. In contrast, we propose a novel loss that achieves state-of-the-art results without increasing the number of parameters of the model.

\section{Group Loss}

Most loss functions used for deep metric learning \cite{DBLP:conf/cvpr/SchroffKP15,DBLP:conf/cvpr/SongXJS16,DBLP:conf/nips/Sohn16,DBLP:conf/cvpr/SongJR017,DBLP:conf/iccv/WangZWLL17,DDBLP:conf/cvpr/Wand2019,DBLP:conf/cvpr/WangHKHGR19,DBLP:conf/icml/LawUZ17,DBLP:conf/eccv/GeHDS18,DBLP:conf/iccv/ManmathaWSK17} do not use a classification loss function, e.g., cross-entropy, but rather a loss function based on embedding distances. 
The rationale behind it, is that what matters for a classification network is that the output is correct, which does not necessarily mean that the embeddings of samples belonging to the same class are similar. 
Since each sample is classified independently, it is entirely possible that two images of the same class have two distant embeddings that both allow for a correct classification. 
We argue that a classification loss can still be used for deep metric learning if the decisions do not happen independently for each sample, but rather jointly for a whole {\it group}, i.e., the set of images of the same class in a mini-batch. In this way, the method pushes for images belonging to the same class to have similar embeddings.
%

%\lau{proposed:}
Towards this end, we propose {\it Group Loss}, an iterative procedure that uses the global information of the mini-batch to refine the local information provided by the softmax layer of a neural network.
This iterative procedure categorizes samples into different \textit{groups}, and enforces consistent labelling among the samples of a group.
While softmax cross-entropy loss judges each sample in isolation, the Group Loss allows us to judge the overall class separation for {\it all} samples.
%jointly takes into consideration all the samples in a mini-batch. 
%
In section \ref{refine}, we show the differences between the softmax cross-entropy loss and Group Loss, and highlight the mathematical properties of our new loss.

\subsection{Overview of Group Loss}\label{sec:gl_overview}

Given a mini-batch $\mathcal{B}$ consisting of $n$ images, consider the problem of assigning a class label $\lambda \in \Lambda = \{1, \dots, m\}$ to each image in $\mathcal{B}$. 
In the remainder of the manuscript, $X=(x_{i\lambda})$ represents a $n \times m$ (non-negative) matrix of image-label soft assignments. In other words, each row of $X$ represents a probability distribution over the label set $\Lambda$ ($\sum_{\lambda} x_{i\lambda}=1 \mbox{ for all } i=1\dots n$). 

The proposed model
consists of the following steps (see also Fig. \ref{fig:pipeline} and Algorithm \ref{algo}):

\begin{enumerate}%[label=\protect\circled{\arabic*}]
    \item \textbf{Initialization}: Initialize $X$, the image-label assignment using the softmax outputs of the neural network. Compute the $n \times n$ pairwise similarity matrix $W$ using the neural network embedding.%\SV{~(see Fig.\ref{fig:pipeline}.1)}.
    \item \textbf{Refinement}: Iteratively, refine $X$ considering the similarities between all the mini-batch images, as encoded in $W$, as well as their labeling preferences.%\SV{~(see Fig.\ref{fig:pipeline}.2)}.
    \item \textbf{Loss computation}: Compute the cross-entropy loss of the refined probabilities and update the weights of the neural network using backpropagation.%\SV{~(see Fig.\ref{fig:pipeline}.3)}.
\end{enumerate}

We now provide a more detailed description of the three steps of our method.

\subsection{Initialization}

\noindent{\bf Image-label assignment matrix.} The initial assignment matrix denoted $X(0)$, comes from the softmax output of the neural network. 
We can replace some of the initial assignments in matrix $X$ with one-hot labelings of those samples. We call these randomly chosen samples {\it anchors}, as their assignments do not change during the iterative refine process and consequently do not directly affect the loss function. However, by using their correct label instead of the predicted label (coming from the softmax output of the NN), they guide the remaining samples towards their correct label.

\noindent{\bf Similarity matrix.} A measure of similarity is computed among all pairs of embeddings (computed via a CNN) in $\mathcal{B}$ to generate a similarity matrix $W \in \mathbb{R}^{n \times n}$. 
In this work, we compute the similarity measure using the Pearson's correlation coefficient \cite{DBLP:journals/rsl/Pearson95}:
\begin{equation}\label{eq:pearson}
    \omega(i, j) = \frac{\mathrm{Cov}[\phi(I_i), \phi(I_j)]}{\sqrt{\mathrm{Var}[\phi(I_i)]\mathrm{Var}[\phi(I_j)]}} %\delta(i \ne j)
\end{equation} 
for $i\neq j$, and set $\omega(i, i)$ to $0$.
The choice of this measure over other options such as cosine layer, Gaussian kernels, or learned similarities, is motivated by the observation that the correlation coefficient uses data standardization, thus providing invariance to scaling and translation -- unlike the cosine similarity, which is invariant to scaling only -- and it does not require additional hyperparameters, unlike Gaussian kernels \cite{DBLP:conf/icpr/EleziTVP18}. 
The fact that a measure of the linear relationship among features provides a good similarity measure can be explained by the fact that the computed features are actually a highly non-linear function of the inputs. Thus, the linear correlation among the embeddings actually captures a non-linear relationship among the original images.

\subsection{Refinement}
In this core step of the proposed algorithm, the initial assignment matrix $X(0)$ is refined in an iterative manner, taking into account the similarity information provided by matrix $W$. $X$ is updated in accordance with the \emph{smoothness assumption}, which prescribes that similar objects should share the same label.

To this end, let us define the \emph{support} matrix $\Pi=(\pi_{i\lambda}) \in R^{n \times m}$ as
\begin{equation}
    \Pi = W X \label{eqn:pi}
\end{equation}
whose $(i, \lambda)$-component
\begin{equation}
    \pi_{i\lambda} = \sum_{j=1}^{n}w_{ij}x_{j\lambda}
\end{equation}
represents the \emph{support} that the current mini-batch gives to the hypothesis that the $i$-th image in $\mathcal{B}$ belongs to class $\lambda$. Intuitively, in obedience to the smoothness principle, $\pi_{i\lambda}$  is expected to be high if images similar to $i$ are likely to belong to class $\lambda$.

\begin{figure}
  \centering
  \includegraphics[width=1in]{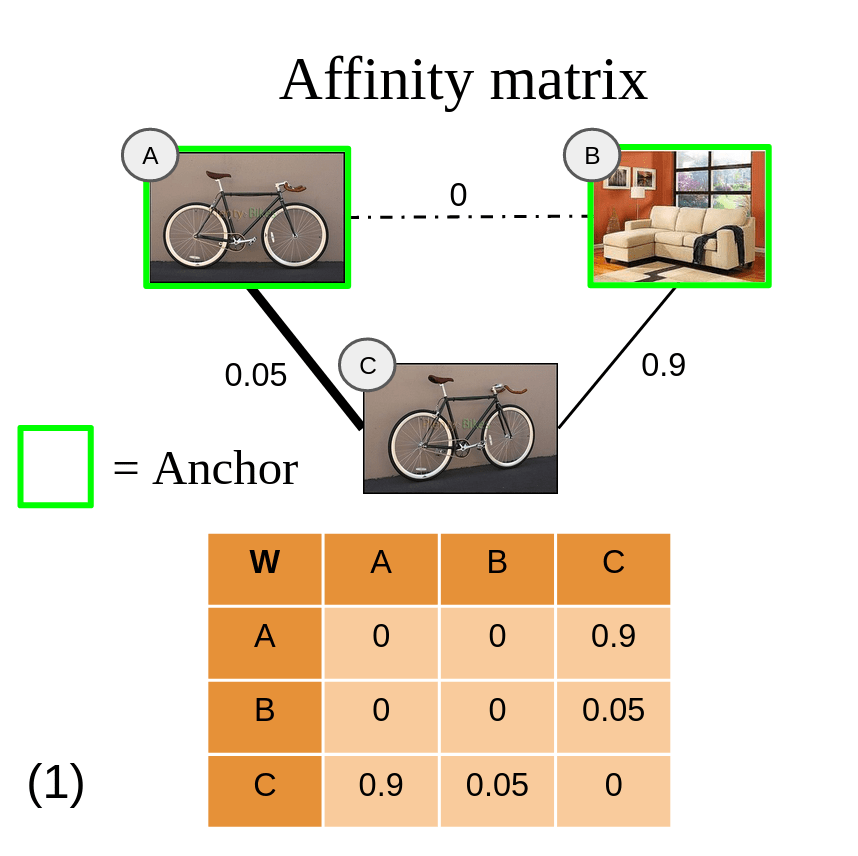}
  \includegraphics[width=0.19\textwidth]{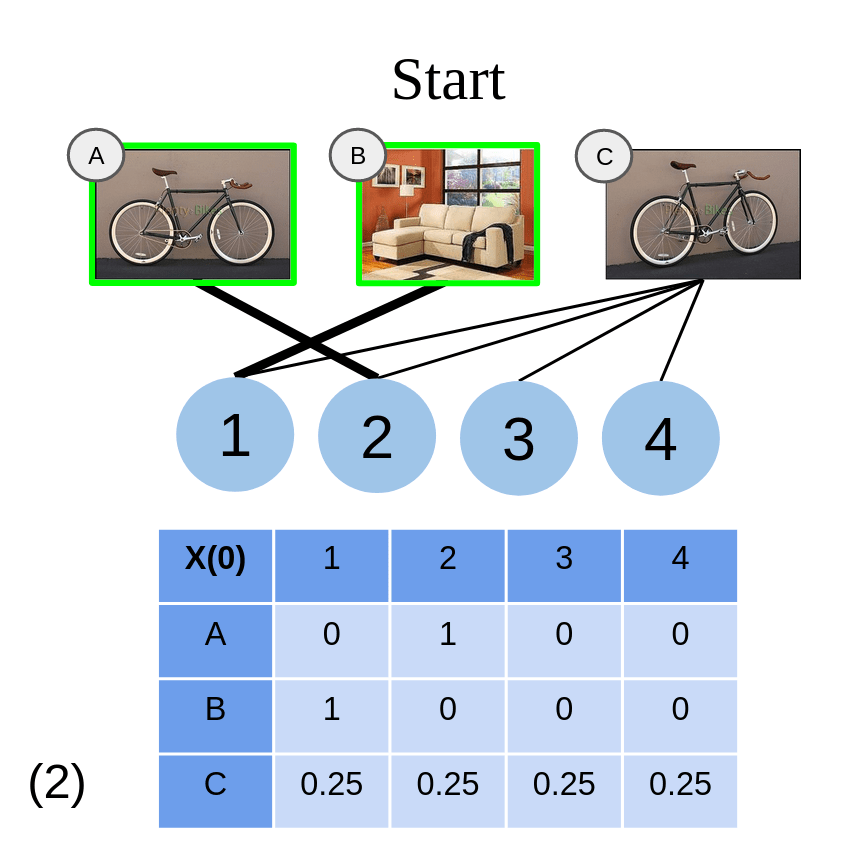}
  \includegraphics[width=0.19\textwidth]{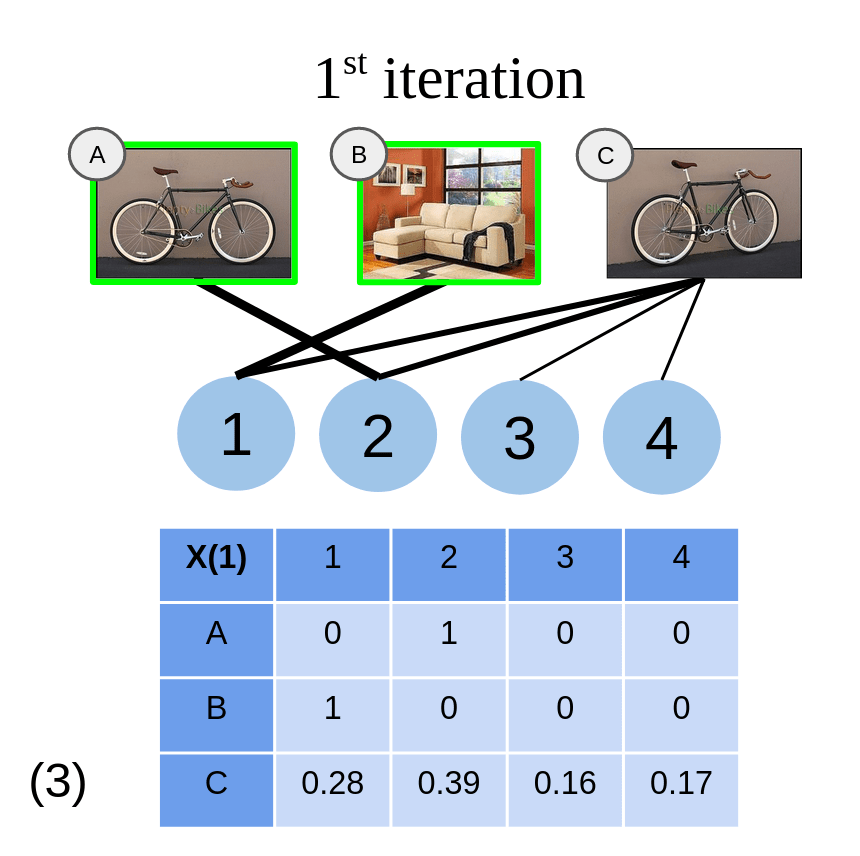}
  \includegraphics[width=0.19\textwidth]{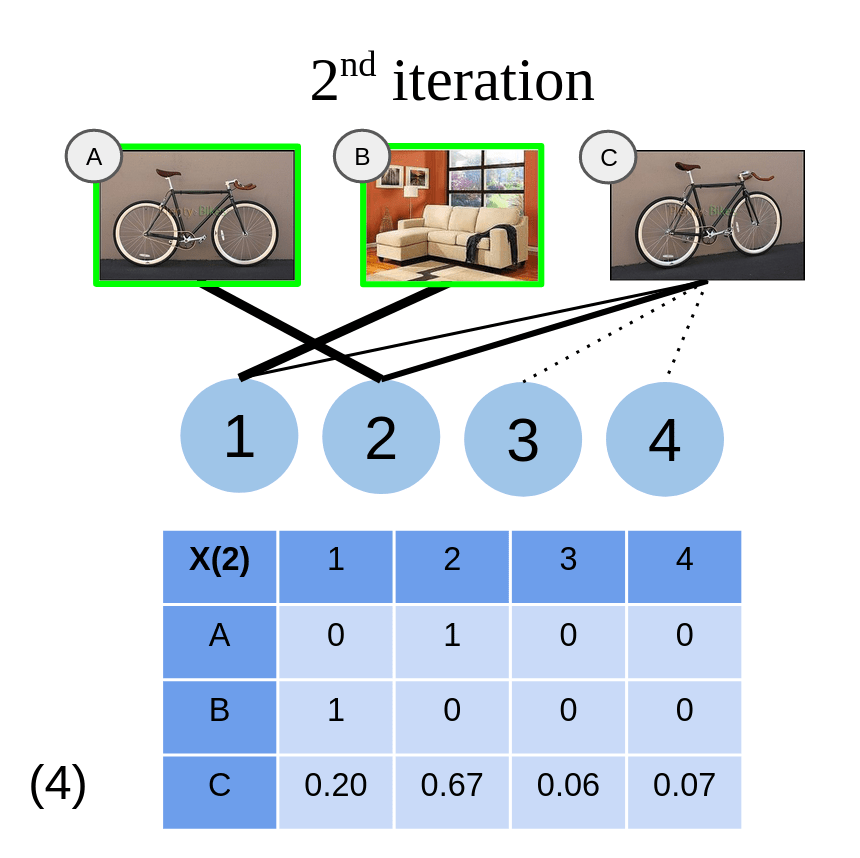}
  \includegraphics[width=0.19\textwidth]{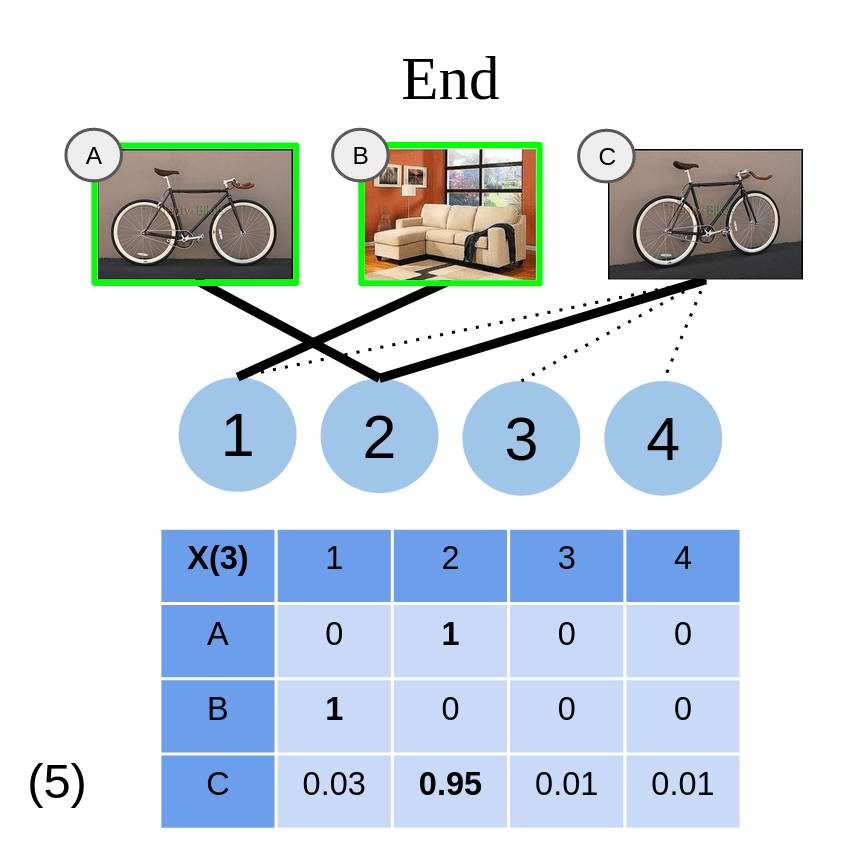}
  \caption{A  toy example of  the  refinement  procedure, where the goal is to classify sample C based on the similarity with samples A and B. From left to right: (1) The Affinity matrix used to update the soft assignments. (2) The initial labeling of the matrix. (3-4) The process iteratively refines the soft assignment of the unlabeled sample C. (5) At the end of the process, sample C gets the same label of A, (A, C) being more similar than (B, C).}
  \label{fig:gtg_prior}
  %\vspace{-0.3cm}
\end{figure}

Given the initial assignment matrix $X(0)$, our algorithm refines it using the following update rule:
\begin{equation}
x_{i\lambda}(t+1) = \frac{x_{i\lambda}(t)\pi_{i\lambda}(t)}{\sum_{\mu=1}^m{x_{i\mu}(t)\pi_{i\mu}(t)}}
\label{eqn:rd}
\end{equation}
where the denominator represents a normalization factor which guarantees that the rows of the updated matrix sum up to one. 
This is known as multi-population replicator dynamics in evolutionary game theory \cite{weibull1997evolutionary} and is equivalent to nonlinear relaxation labeling processes \cite{RosHumZuc76,DBLP:journals/jmiv/Pelillo97}.

In matrix notation, the update rule (\ref{eqn:rd}) can be written as:
\begin{equation}
X(t+1) = Q^{-1}(t) \left[ X(t) \odot \Pi(t) \right] \label{eq:rd_mat}
\end{equation}
where 
\begin{equation}
Q(t) = \mbox{diag}(\left[X(t) \odot \Pi(t)\right] \mathbb{1} )
\end{equation}
and $\mathbb{1}$ is the all-one $m$-dimensional vector.
$\Pi(t) = W X(t)$ as defined in (\ref{eqn:pi}), and $\odot$ denotes the Hadamard (element-wise) matrix product.
In other words, the diagonal elements of $Q(t)$ represent the normalization factors in (\ref{eqn:rd}), which can also be interpreted as the average support that object $i$ obtains from the current mini-batch at iteration $t$.
Intuitively, the motivation behind our update rule is that at each step of the refinement process, for each image $i$, a label $\lambda$ will increase its probability $x_{i\lambda}$ if and only if its support $\pi_{i\lambda}$ is higher than the average support among all the competing label hypothesis $Q_{ii}$.\footnote{This can be motivated by a Darwinian survival-of-the-fittest selection principle, see e.g. \cite{weibull1997evolutionary}.}

Thanks to the Baum-Eagon inequality \cite{DBLP:journals/jmiv/Pelillo97}, it is easy to show that the dynamical system defined by (\ref{eqn:rd}) has very nice convergence properties. In particular, it strictly increases at each step the following functional:

\begin{equation}
F(X) = \sum_{i = 1}^{n}\sum_{j = 1}^{n}\sum_{\lambda = 1}^{m} w_{ij}  x_{i\lambda} x_{j\lambda} \label{eqn:avgconsistency}
\end{equation}
which represents a measure of ``consistency'' of the assignment matrix $X$, in accordance to the smoothness assumption ($F$ rewards assignments where highly similar objects are likely to be assigned the same label).
In other words:
\begin{equation}
F(X(t+1)) \geq F(X(t))
\end{equation}
with equality if and only if $X(t)$ is a stationary point.
Hence, our update rule (\ref{eqn:rd}) is, in fact, an algorithm for maximizing the functional $F$ over the space of row-stochastic matrices.
Note, that this contrasts with classical gradient methods, for which an increase in the objective function is guaranteed only when infinitesimal steps are taken, and determining the optimal step size entails computing higher-order derivatives. Here, instead, the step size is implicit and yet, at each step, the value of the functional increases. 

\label{refine}
% \SV{why we have formulated }
\subsection{Loss computation}
Once the labeling assignments converge (or in practice, a maximum number of iterations is reached), we apply the cross-entropy loss to quantify the classification error and backpropagate the gradients. Recall, the refinement procedure is optimized via \textit{replicator dynamics}, as shown in the previous section. By studying Equation (\ref{eq:rd_mat}), it is straightforward to see that it is composed of fully differentiable operations (matrix-vector and scalar products), and so it can be easily integrated within backpropagation. 
Although the refining procedure has no parameters to be learned, its gradients can be backpropagated to the previous layers of the neural network, producing, in turn, better embeddings for similarity computation.
%making, in turn, the network produce more optimal embeddings for the future labelings of the samples, and their similarities.

\subsection{Summary of the Group Loss} 
In this section, we proposed the Group Loss function for deep metric learning. 
During training, the Group Loss works by grouping together similar samples based on both the similarity between the samples in the mini-batch and the local information of the samples. The similarity between samples is computed by the correlation between the embeddings obtained from a CNN, while the local information is computed with a softmax layer on the same CNN embeddings. Using an iterative procedure, we combine both sources of information and effectively bring together embeddings of samples that belong to the same class.

During inference, we simply forward pass the images through the neural network
to compute their embeddings, which are directly used for image retrieval within a nearest neighbor search scheme.
The iterative procedure is not used during inference, thus making the feature extraction as fast as that of any other competing method. 

\SetKwInput{KwInput}{Input}

\begin{algorithm}[t] 
\caption{The Group Loss}
\label{algo}
\DontPrintSemicolon 
  \KwInput{input : Set of pre-processed images in the mini-batch $\mathcal{B}$, set of labels $y$, 
%   number of images per mini-batch $b$, number of classes $m$,
  neural network $\phi$ with learnable parameters $\theta$, similarity function $\omega$, number of iterations $T$}  
  
  1) Compute feature embeddings $\phi(\mathcal{B}, \theta)$ via the forward pass

  2) Compute the similarity matrix $W = [\omega(i, j)]_{ij}$
  
  3) Initialize the matrix of priors $X(0)$ from the softmax layer
  
  4) \For{t = 0, \dots, T-1}
  {
    $Q(t) = \mbox{diag}(\left[X(t) \odot \Pi(t)\right] \mathbb{1} )$\\
    $X(t+1) = Q^{-1}(t) \left[ X(t) \odot \Pi(t) \right]$
  }
      
  5) Compute the cross-entropy $J(X(T), y)$
  
  6) Compute the derivatives $\partial J / \partial \theta$ via backpropagation, and update the weights $\theta$
\end{algorithm}

\subsection{Alternative loss formulation}
In the main paper, we formulated the loss as an iterative dynamical system, followed by the cross-entropy loss function. In this way, we encourage the network to predict the same label for samples coming from the same class. 
One might argue that this is not necessarily the best loss for metric learning, in the end, we are interested in bringing similar samples closer together in the embedding space, without the need of having them classified correctly.
Even though several works have shown that a classification loss can be used for metric learning \cite{DBLP:conf/iccv/Movshovitz-Attias17,DBLP:journals/corr/abs-1811-12649,DBLP:journals/corr/abs-1909-05235}, we test whether this is also the best formulation for our loss function.

We therefore experiment with a different loss function which encourages the network to produce similar label distributions (soft labels) for the samples coming from the same class.
We first define Kullback-Leibler divergence for two distributions $P$ and $Q$ as:

\begin{equation}
    D_{KL}(P||Q) = \sum_{x \in X} P(x) log \frac{P(x)}{Q(x)}.
\end{equation}

We then minimize the divergence between the predicted probability (after the iterative procedure) of samples coming from the same class. Unfortunately, this loss formulation results in lower performances on both \textit{CUB-200-2011} \cite{WahCUB_200_2011} ($3pp$) and \textit{Cars 196} \cite{KrauseStarkDengFei-Fei_3DRR2013} ($1.5pp$). Thus, we report the experiments in only with the original loss formulation.

\subsection{Dealing with negative similarities}

Equation (4) in the paper assumes that the matrix of similarity is non-negative.
However, for similarity computation, we use a correlation metric (see Equation (1)) which produces values in the range $[-1, 1]$.
In similar situations, different authors propose different methods to deal with the negative outputs.
The most common approach is to shift the matrix of similarity towards the positive regime by subtracting the biggest negative value from every entry in the matrix \cite{DBLP:journals/neco/ErdemP12}.
Nonetheless, this shift has a side effect: If a sample of class $k_1$ has very low similarities to the elements of a large group of samples of class $k_2$, these similarity values (which after being shifted are all positive) will be summed up. If the cardinality of class $k_2$ is very large, then summing up all these small values lead to a large value, and consequently affect the solution of the algorithm.
What we want instead, is to ignore these negative similarities, hence we propose {\it clamping}. More concretely, we use a \textit{ReLU} activation function over the output of Equation (1).

We compare the results of shifting vs clamping. On the \textit{CARS 196} dataset, we do not see a significant difference between the two approaches.
However, on the \textit{CUBS-200-2011} dataset, the Recall@1 metric is $51$ with shifting, much below the $64.3$ obtained when using clamping.
We investigate the matrix of similarities for the two datasets, and we see that the number of entries with negative values for the \textit{CUBS-200-2011} dataset is higher than for the \textit{CARS 196} dataset. This explains the difference in behavior, and also verifies our hypothesis that clamping is a better strategy to use within {\it Group Loss}.

\subsection{Temperature scaling}
We mentioned that as input to the Group Loss (step 3 of the algorithm) we initialize the matrix of priors $X(0)$ from the softmax layer of the neural network. 
Following the works of \cite{DBLP:conf/icml/GuoPSW17,DBLP:conf/nips/BerthelotCGPOR19,DBLP:journals/corr/abs-1811-12649}, we apply a sharpening function to reduce the entropy of the softmax distribution. We use the common approach of adjusting the {\it temperature} of this categorical distribution, known as temperature scaling. 
Intuitively, this procedure calibrates our network and in turn, provides more informative prior to the dynamical system. Additionally, this calibration allows the dynamical system to be more effective in adjusting the predictions, i.e, it is easier to change the probability of a class if its initial value is $0.6$ rather than $0.95$. The function is implemented using the following equation:

\begin{equation}
    T_{softmax}(z_i) = \frac{e^{z_i/T}}{\sum_i{e^{z_i/T}}},
\end{equation}

which can be efficiently implemented by simply dividing the prediction logits by a constant $T$.

\section{Experiments}
In this section, we compare the Group Loss with state-of-the-art deep metric learning models on both image retrieval and clustering tasks. Our method achieves state-of-the-art results in three public benchmark datasets. 

\subsection{Implementation details}
We use the PyTorch \cite{Paszke17} library for the implementation of the Group Loss. We choose GoogleNet \cite{DBLP:conf/cvpr/SzegedyLJSRAEVR15} with batch-normalization \cite{DBLP:conf/icml/IoffeS15} as the backbone feature extraction network. We pretrain the network on \textit{ILSVRC 2012-CLS} dataset \cite{DBLP:journals/corr/RussakovskyDSKSMHKKBBF14}. For pre-processing, in order to get a fair comparison, we follow the implementation details of \cite{DBLP:conf/cvpr/SongJR017}. The inputs are resized to $256 \times 256$ pixels, and then randomly cropped to $227 \times 227$. Like other methods except for \cite{DBLP:conf/nips/Sohn16}, we use only a center crop during testing time. We train all networks in the classification task for $10$ epochs. We then train the network in the Group Loss task for $60$ epochs using Adam optimizer \cite{DBLP:conf/iclr/Kingma14} with learning rate $0.0002$ set for all networks and all datasets. After $30$ epochs, we lower the learning rate by multiplying it by $0.1$. We find the hyperparameters using random search \cite{DBLP:journals/jmlr/BergstraB12}. We use small mini-batches of size $30 - 100$. As sampling strategy, on each mini-batch, we first randomly sample a fixed number of classes, and then for each of the chosen classes, we sample a fixed number of samples. For the weight decay ($L2$-regularization) parameter, we search over the interval $[0.1, 10^{-16}]$, while for learning rate we search over the interval $[0.1, 10^{-5}]$, choosing $0.0002$ as the learning rate for all networks and all datasets.

\subsection{Benchmark datasets}
We perform experiments on $3$ publicly available datasets, evaluating our algorithm on both clustering and retrieval metrics. For training and testing, we follow the conventional splitting procedure \cite{DBLP:conf/cvpr/SongXJS16}.
\vspace{0.3cm}

\textbf{CUB-200-2011} \cite{WahCUB_200_2011} is a dataset containing $200$ species of birds with $11,788$ images, where the first $100$ species ($5,864$ images) are used for training and the remaining $100$ species ($5,924$ images) are used for testing.

\vspace{0.3cm}

\textbf{Cars 196} \cite{KrauseStarkDengFei-Fei_3DRR2013} dataset is composed of $16,185$ images belonging to $196$ classes. We use the first $98$ classes ($8,054$ images) for training and the other $98$ classes ($8,131$ images) for testing.
\vspace{0.3cm}

\textbf{Stanford Online Products} dataset, as introduced in \cite{DBLP:conf/cvpr/SongXJS16}, contains $22,634$ classes with $120,053$ product images in total, where $11,318$ classes ($59,551$ images) are used for training and the remaining $11,316$ classes ($60,502$ images) are used for testing.

\subsection{Evaluation metrics}
Based on the experimental protocol detailed above, we evaluate retrieval performance and clustering quality on data
from unseen classes of the $3$ aforementioned datasets. For the retrieval task, we calculate the percentage of the testing examples whose $K$ nearest neighbors contain at least one example of the same class. This quantity is also known as Recall@K \cite{DBLP:journals/pami/JegouDS11} and is the most used metric for image retrieval evaluation. 

Similar to all other approaches, we perform clustering using K-means algorithm \cite{MacQueen} on the embedded features. Like in other works, we evaluate the clustering quality using the Normalized Mutual Information measure (NMI) \cite{DBLP:journals/corr/abs-1110-2515}. The choice of NMI measure is motivated by the fact that it is invariant to label permutation, a desirable property for cluster evaluation.

\subsection{Main Results}

We now show the results of our model and comparison to state-of-the-art methods. 
Our main comparison is with other loss functions, e.g., triplet loss. %,  without including any other tricks such as sampling or ensembles.
 To compare with perpendicular research on intelligent sampling strategies or ensembles, and show the power of the Group Loss, we propose a simple ensemble version of our method. Our ensemble network is built by training $l$ independent neural networks with the same hyperparameter configuration. During inference, their embeddings are concatenated. Note, that this type of ensemble is much simpler than the works of \cite{DBLP:conf/iccv/YuanYZ17,DBLP:conf/eccv/XuanSP18,DBLP:conf/eccv/KimGCLK18,DBLP:journals/corr/abs-1801-04815,DDBLP:conf/cvpr/Sanakoyeu2019}, and is given only to show that, when optimized for performance, our method can be extended to ensembles giving higher clustering and retrieval performance than other methods in the literature. 
%
%Nevertheless, we consider the main focus of comparison should be the other loss functions that do not use advanced sampling or ensemble methods. 
Finally, in the interest of space, we only present results for Inception network \cite{DBLP:conf/cvpr/SzegedyLJSRAEVR15}, as this is the most popular backbone for the metric learning task, which enables fair comparison among methods. In supplementary material, we present results for other backbones, and include a discussion about the methods that work by increasing the number of parameters (capacity of the network) \cite{DBLP:journals/corr/abs-1909-05235}, or use more expressive network architectures.

\noindent{\bf Loss comparison.} In Table \ref{tab:res_loss} we present the results of our method and compare them with the results of other approaches. On the \textit{CUB-200-2011} dataset, we outperform the other approaches by a large margin, with the second-best model (Classification \cite{DBLP:journals/corr/abs-1811-12649}) having circa $5$ percentage points($pp$) lower absolute accuracy in Recall@1 metric. On the NMI metric, our method achieves a score of $67.9$ which is  $1.7pp$ higher than the second-best method. Similarly, on \textit{Cars 196}, our method achieves best results on Recall@1, with Classification \cite{DBLP:journals/corr/abs-1811-12649} coming second with a $2pp$ lower score. %Proxy-NCA \cite{DBLP:conf/iccv/Movshovitz-Attias17} is second on the NMI metric, with a $6pp$ lower score.
On \textit{Stanford Online Products}, our method reaches the best results on the Recall@1 metric, around $1.5pp$ higher than Classification \cite{DBLP:journals/corr/abs-1811-12649} and Proxy-NCA \cite{DBLP:conf/iccv/Movshovitz-Attias17}. On the same dataset, when evaluated on the NMI score, our loss outperforms any other method, be those methods that exploit advanced sampling, or ensemble methods.

\begin{table}[t]
\centering
    \resizebox{\textwidth}{!}{%
    \begin{tabular}{@{}l|ccccc|ccccc|cccc@{}}
\toprule

\textbf{} & \multicolumn{5}{c}{CUB-200-2011} & \multicolumn{5}{c}{CARS 196} & \multicolumn{4}{c}{Stanford Online Products}\\ \hline
\textbf{Loss} &  \textbf{R@1} & \textbf{R@2} & \textbf{R@4} & \textbf{R@8} & \textbf{NMI} &  \textbf{R@1} & \textbf{R@2} & \textbf{R@4} & \textbf{R@8} & \textbf{NMI} & \textbf{R@1} & \textbf{R@10} & \textbf{R@100} & \textbf{NMI} \\ \midrule
Triplet \cite{DBLP:conf/cvpr/SchroffKP15} & 42.5 & 55 & 66.4 & 77.2 & 55.3 & 51.5 & 63.8 & 73.5 & 82.4 & 53.4 & 66.7 & 82.4 & 91.9 & 89.5 \\

Lifted Structure \cite{DBLP:conf/cvpr/SongXJS16} & 43.5 & 56.5 & 68.5 & 79.6 & 56.5 &  53.0 & 65.7 & 76.0 & 84.3 & 56.9 & 62.5 & 80.8 & 91.9 & 88.7 \\

Npairs \cite{DBLP:conf/nips/Sohn16} &  51.9 & 64.3 & 74.9 & 83.2 & 60.2 &  68.9 & 78.9 & 85.8 & 90.9 & 62.7 &  66.4 & 82.9 & 92.1 & 87.9  \\

Facility Location \cite{DBLP:conf/cvpr/SongJR017} &  48.1 & 61.4 & 71.8 & 81.9 & 59.2 &  58.1 & 70.6 & 80.3 & 87.8 & 59.0 & 67.0 & 83.7 & 93.2 & 89.5 \\

Angular Loss \cite{DBLP:conf/iccv/WangZWLL17} &  54.7 & 66.3 & 76 & 83.9 & 61.1 &  71.4 & 81.4 & 87.5 & 92.1 & 63.2 &  70.9 & 85.0 & 93.5 & 88.6\\

Proxy-NCA \cite{DBLP:conf/iccv/Movshovitz-Attias17} &  49.2 & 61.9 & 67.9 & 72.4 & 59.5 &  73.2 & 82.4 & 86.4 & 88.7 & 64.9 &  73.7 & - & - & 90.6 \\

Deep Spectral \cite{DBLP:conf/icml/LawUZ17} &  53.2 & 66.1 & 76.7 & 85.2 & 59.2 &  73.1 & 82.2 & 89.0 & 93.0 & 64.3 &  67.6 & 83.7 & 93.3& 89.4 \\ 

Classification \cite{DBLP:journals/corr/abs-1811-12649} &  59.6 & 72 & 81.2 & 88.4 & 66.2 &  81.7 & 88.9 & 93.4 & 96 & 70.5 &  73.8 & \textbf{88.1} & \textbf{95} & 89.8 \\

Bias Triplet \cite{DBLP:conf/eccv/YuLGDT18} & 46.6 & 58.6 & 70.0 & - &  - &  79.2 & 86.7 & 91.4  & - & - & 63.0 & 79.8 & 90.7 &- \\ \hline

\textbf{Ours} &  \textbf{64.3} & \textbf{75.8} & \textbf{84.1} & \textbf{90.5} & \textbf{67.9} &  \textbf{83.7} & \textbf{89.9} & \textbf{93.7} & \textbf{96.3} & \textbf{70.7} &  \textbf{75.1} & 87.5 & 94.2 &\textbf{90.8}\\ \bottomrule 
\end{tabular}}
\caption{Retrieval and Clustering performance on \textit{CUB-200-2011}, \textit{CARS 196} and \textit{Stanford Online Products} datasets. Bold indicates best results. }
\label{tab:res_loss}
\end{table}

\vspace{0.5cm}

\noindent{\bf Loss with ensembles.} In Table \ref{tab:loss_ensembles} we present the results of our ensemble, and compare them with the results of other ensemble and sampling approaches. Our ensemble method (using $5$ neural networks) is the highest performing model in \textit{CUB-200-2011}, outperforming the second-best method (Divide and Conquer \cite{DDBLP:conf/cvpr/Sanakoyeu2019}) by $1pp$ in Recall@1 and by $0.4pp$ in NMI. In \textit{Cars 196} our method outperforms the second best method (ABE 8 \cite{DBLP:conf/eccv/KimGCLK18}) by $2.8pp$ in Recall@1. The second best method in NMI metric is the ensemble version of RLL \cite{DBLP:conf/cvpr/WangHKHGR19} which gets outperformed by $2.4pp$ from the Group Loss. In \textit{Stanford Online Products}, our ensemble reaches the third-highest result on the Recall@1 metric (after RLL \cite{DBLP:conf/cvpr/WangHKHGR19} and GPW \cite{DDBLP:conf/cvpr/Wand2019}) while increasing the gap with the other methods in NMI metric.

\begin{table}[t]
\centering
    \resizebox{\textwidth}{!}{%
    \begin{tabular}{@{}l|ccccc|ccccc|cccc@{}}
\toprule

\textbf{} & \multicolumn{5}{c}{CUB-200-2011} & \multicolumn{5}{c}{CARS 196} & \multicolumn{4}{c}{Stanford Online Products}\\ \hline
\textbf{Loss+Sampling} & \textbf{R@1} & \textbf{R@2} & \textbf{R@4} & \textbf{R@8} & \textbf{NMI} &  \textbf{R@1} & \textbf{R@2} & \textbf{R@4} & \textbf{R@8} & \textbf{NMI} & \textbf{R@1} & \textbf{R@10} & \textbf{R@100} & \textbf{NMI} \\ \midrule

Samp. Matt.  \cite{DBLP:conf/iccv/ManmathaWSK17} &  63.6 & 74.4 & 83.1 & 90.0 & 69.0 &  79.6 & 86.5 & 91.9 & 95.1 & 69.1 &  72.7 & 86.2 & 93.8 & 90.7 \\
Hier. triplet  \cite{DBLP:conf/eccv/GeHDS18} & 57.1 & 68.8 & 78.7 & 86.5 & - &  81.4 & 88.0 & 92.7 & 95.7 &  - & 74.8 & 88.3 & 94.8 & - \\
DAMLRRM  \cite{DDBLP:conf/cvpr/Xu2019} &  55.1 & 66.5 & 76.8 & 85.3 & 61.7 &  73.5 & 82.6 & 89.1 & 93.5 & 64.2 &  69.7 & 85.2 & 93.2 & 88.2\\
DE-DSP  \cite{DDBLP:conf/cvpr/Duan2019} &  53.6 & 65.5 & 76.9 & 61.7 & - &  72.9 & 81.6 & 88.8 & - & 64.4 & 68.9 & 84.0 & 92.6 & 89.2 \\
RLL 1 \cite{DBLP:conf/cvpr/WangHKHGR19} & 57.4 & 69.7 & 79.2 & 86.9 & 63.6 & 74 & 83.6 & 90.1 & 94.1 &   65.4 &   76.1 & 89.1 & 95.4 & 89.7 \\
GPW  \cite{DDBLP:conf/cvpr/Wand2019} & 65.7 & 77.0 & 86.3 & 91.2 & - &  84.1 & 90.4 & 94.0 & 96.5 & - & 78.2 & 90.5 & 96.0 & - \\ \hline \hline

\textbf{Teacher-Student} &  \\ \hline
RKD \cite{DBLP:conf/cvpr/Park2019} &  61.4 & 73.0 & 81.9 & 89.0 & - &  82.3 & 89.8 & 94.2 & 96.6 & - & 75.1 & 88.3 & 95.2  & -\\ \hline \hline

\textbf{Loss+Ensembles} &  \\ \hline
BIER 6  \cite{DBLP:conf/iccv/OpitzWPB17} & 55.3 & 67.2 & 76.9 & 85.1 &  - & 75.0 & 83.9 & 90.3 & 94.3 &  - &  72.7 & 86.5 & 94.0  & -\\
HDC 3  \cite{DBLP:conf/iccv/YuanYZ17} &  54.6 & 66.8 & 77.6 & 85.9 & - &  78.0 & 85.8 & 91.1 & 95.1 & - & 70.1 & 84.9 & 93.2 & - \\
%DRE 48  \cite{DBLP:conf/eccv/XuanSP18} &58.9 & 69.6 & 78.4 & 85.6 &  62.1 &  84.2 & 89.4 & 93.2 & 95.5 &  71 & - & - & - & -  \\
ABE 2  \cite{DBLP:conf/eccv/KimGCLK18} &  55.7 & 67.9 & 78.3 & 85.5 & - & 76.8 & 84.9 & 90.2 & 94.0 & - & 75.4 & 88.0 & 94.7  & -  \\
ABE 8  \cite{DBLP:conf/eccv/KimGCLK18} &  60.6 & 71.5 & 79.8 & 87.4 & - & 85.2 & 90.5 & 94.0 & 96.1 &  - & 76.3 & 88.4 &  94.8 & - \\
A-BIER 6 \cite{DBLP:journals/corr/abs-1801-04815} & 57.5 & 68.7 & 78.3 & 86.2 & - &  82.0 & 89.0 & 93.2 & 96.1 &  - & 74.2 & 86.9 & 94.0 & - \\
D and C 8  \cite{DDBLP:conf/cvpr/Sanakoyeu2019} & 65.9 & 76.6 & 84.4 & 90.6 & 69.6 & 84.6 & 90.7 & 94.1 & 96.5 &  70.3 &   75.9   & 88.4 & 94.9 &90.2 \\ 
RLL 3  \cite{DBLP:conf/cvpr/WangHKHGR19} & 61.3 & 72.7 & 82.7 & 89.4 & 66.1 & 82.1 & 89.3 & 93.7 & 96.7 & 71.8 &  \textbf{79.8} & \textbf{91.3} & \textbf{96.3} & 90.4\\
\hline
\textbf{Ours 2-ensemble} & 65.8 & 76.7 & 85.2 & 91.2 & 68.5 &  86.2 & 91.6 & 95.0 & 97.1 & \textbf{91.1} & 75.9 & 88.0 & 94.5 & 72.6\\
\textbf{Ours 5-ensemble} &  \textbf{66.9} & \textbf{77.1} & \textbf{85.4} & \textbf{91.5} & \textbf{70.0} &  \textbf{88.0} & \textbf{92.5} & \textbf{95.7} & \textbf{97.5} & \textbf{74.2} &  76.3 & 88.3 & 94.6 & \textbf{91.1} \\ \bottomrule
\end{tabular}}
\caption{Retrieval and Clustering performance of our ensemble compared with other ensemble and sampling methods. Bold indicates best results. }
\label{tab:loss_ensembles}
\end{table}

\vspace{0.8cm}

\subsubsection{Qualitative results}
\begin{figure*}[ht!]
\centering
\begin{minipage}{.333\textwidth}
  \flushleft %\centering
\includegraphics[width=\textwidth, trim={0.5cm 4cm 12cm 0.1cm},clip]{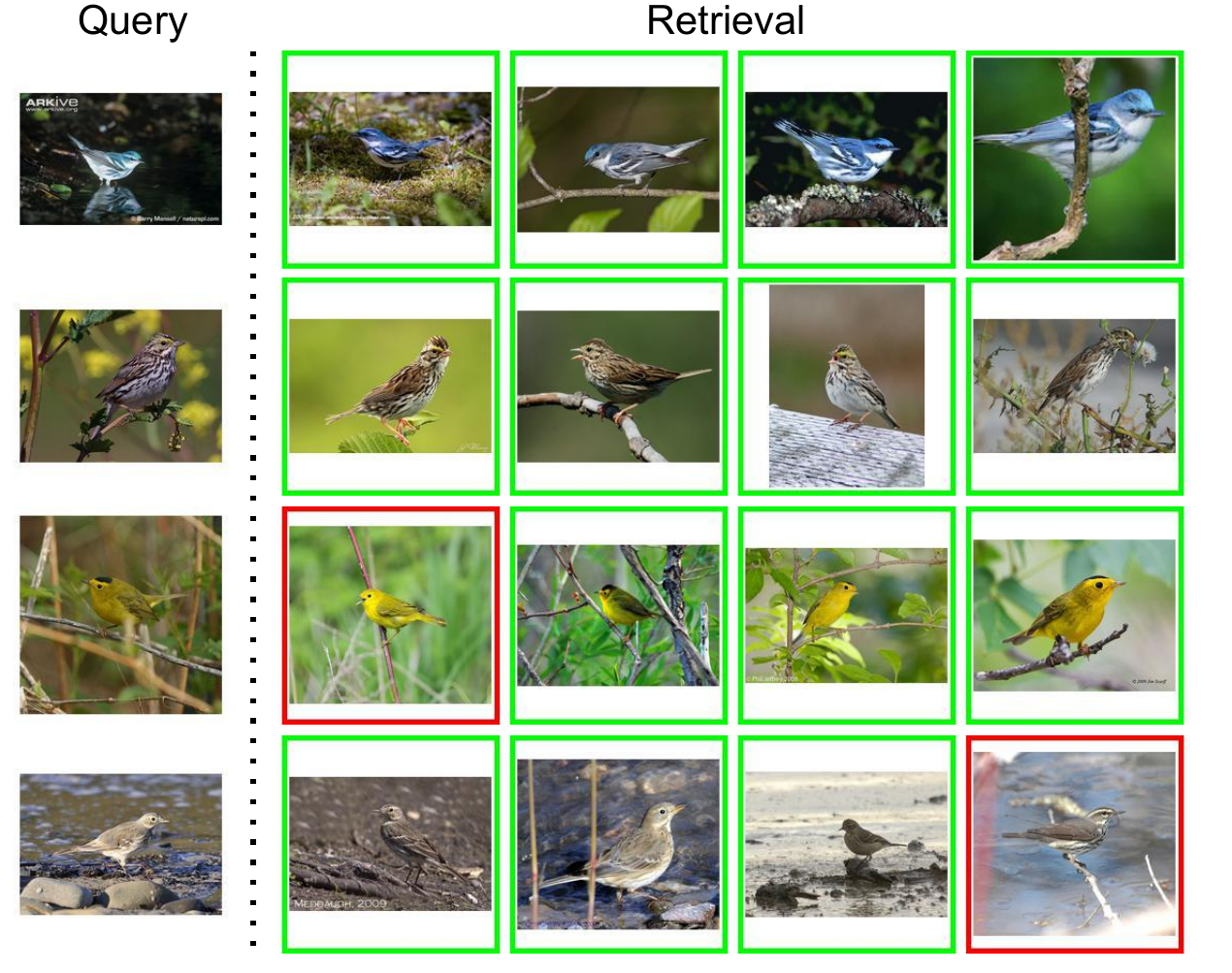}
\end{minipage}% \\
\begin{minipage}{.333\textwidth}
  \centering
\includegraphics[width=\textwidth, trim={0.5cm 4cm 12cm 0.1cm},clip]{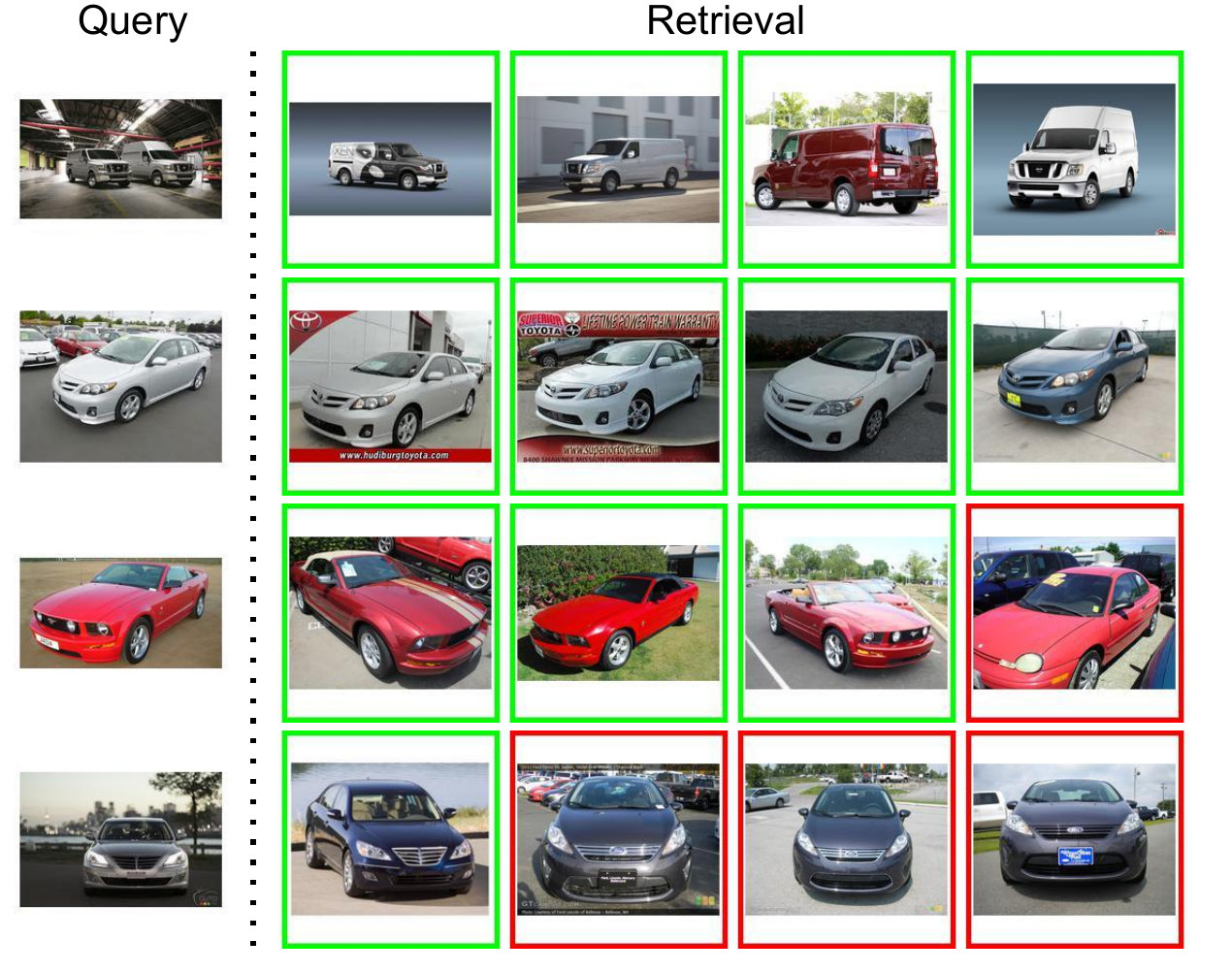}
\end{minipage}% \\
\begin{minipage}{.333\textwidth}
  \flushright %\centering
\includegraphics[width=\textwidth, trim={0.5cm 4cm 12cm 0.1cm},clip]{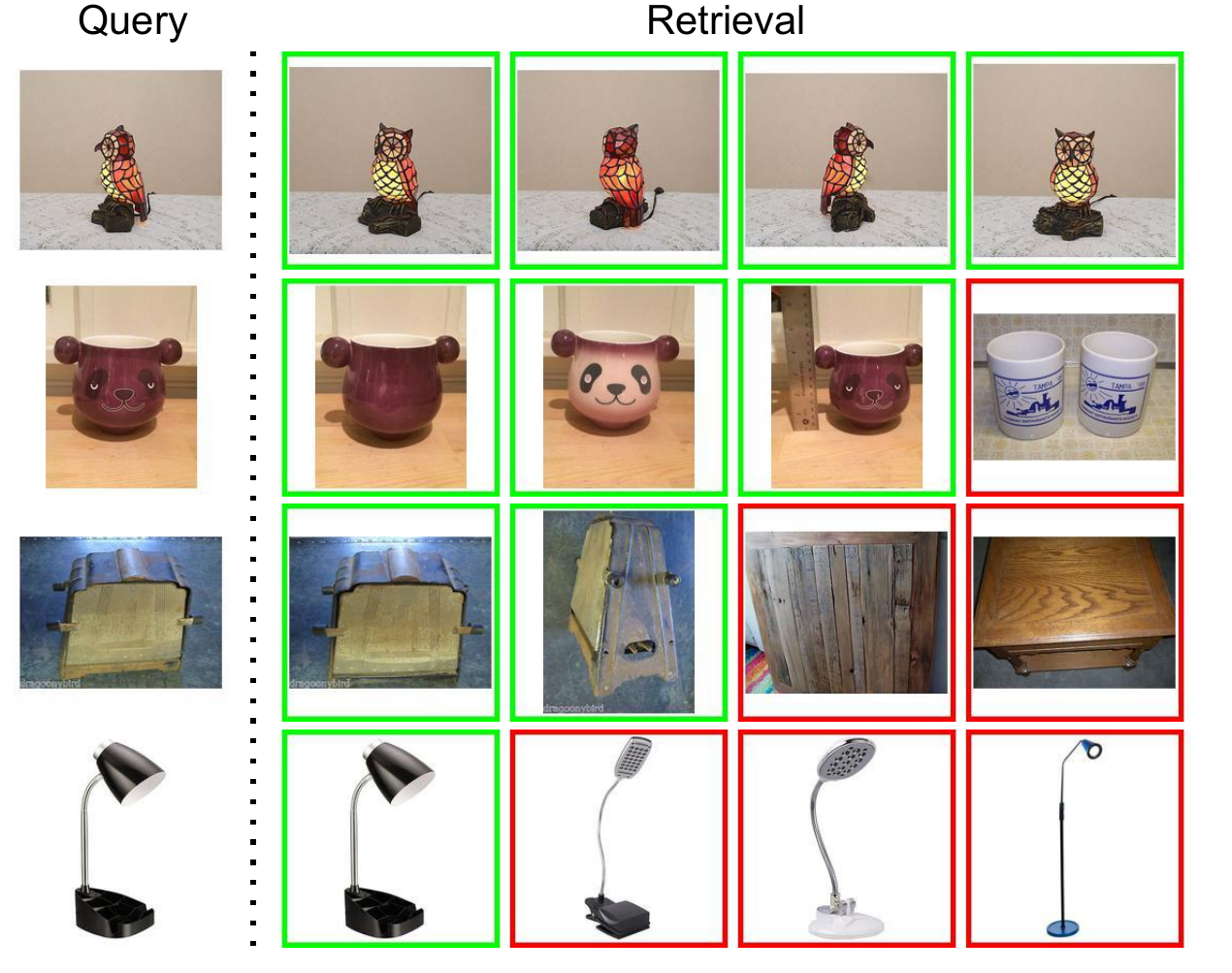}
\end{minipage}

\caption{Retrieval results on a set of images from the
\textit{CUB-200-2011} (left), \textit{Cars 196} (middle), and \textit{Stanford Online Products} (right) datasets using our Group Loss model. Left column contains query images. The results are ranked by distance. Green square indicates that the retrieved image is from the same class as query image, while the red box indicate that the retrieved image is from a different class.}  
\label{fig:retrieval}
\end{figure*}

In Fig. \ref{fig:retrieval} we present qualitative results on the retrieval task in all three datasets. In all cases, the query image is given on the left, with the four nearest neighbors given on the right. 
Green boxes indicate the cases where the retrieved image is of the same class as the query image, and red boxes indicate a different class. 
As we can see, our model is able to perform well even in cases where the images suffer from occlusion and rotation. On the \textit{Cars 196} dataset, we see a successful retrieval even when the query image is taken indoors and the retrieved image outdoors, and vice-versa. % the query might be indoor with the retrieved image outdoors, and vice versa. 
The first example of \textit{Cars 196} dataset is of particular interest. Despite the fact that the query image contains $2$ cars, all four nearest neighbors which have been retrieved have the same class as the query image, showing the robustness of the algorithm to uncommon input image configurations. We provide the results of t-SNE \cite{DBLP:journals/ml/MaatenH12} projection in the supplementary material.

\subsection{Robustness analysis}

\begin{figure}[t!]
\centering
\begin{minipage}[t!]{.49\textwidth}
  \centering
\includegraphics[width=\textwidth, trim={0cm 0cm 0cm 0cm},clip]{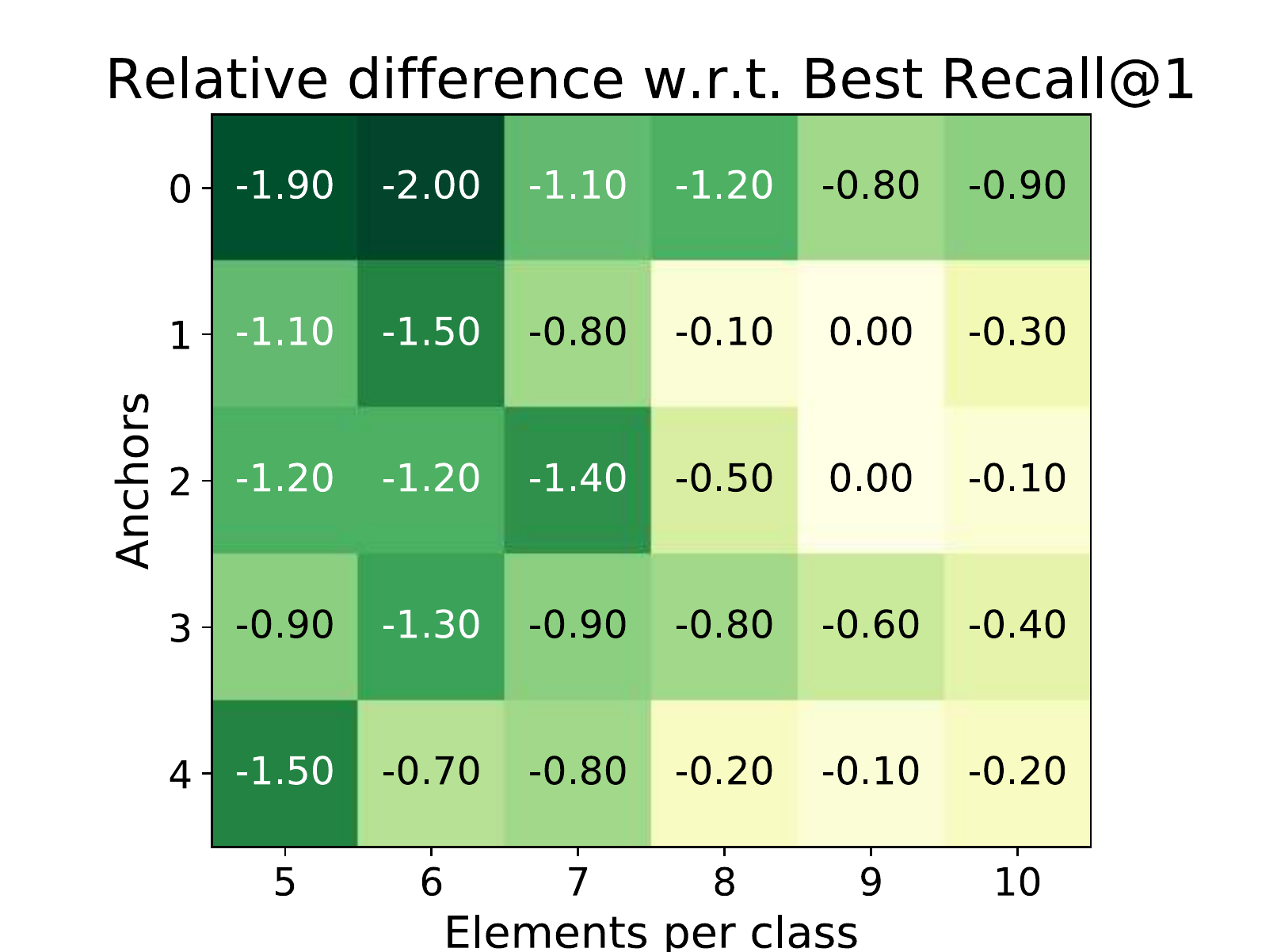} %ablation_labeled_double_loss.eps}
  %\caption{\SV{Positive/negative effect on the Recall@1 by changing the number of anchors and the number of samples per class.}}
  \caption{The effect of the number of anchors and the number of samples per class.}
  \label{fig:abl_labeledpoints}
\end{minipage}
\hfill %
\begin{minipage}[t!]{.49\textwidth}
  \centering
\includegraphics[width=\textwidth, trim={0cm 0cm 0cm 0cm},clip]{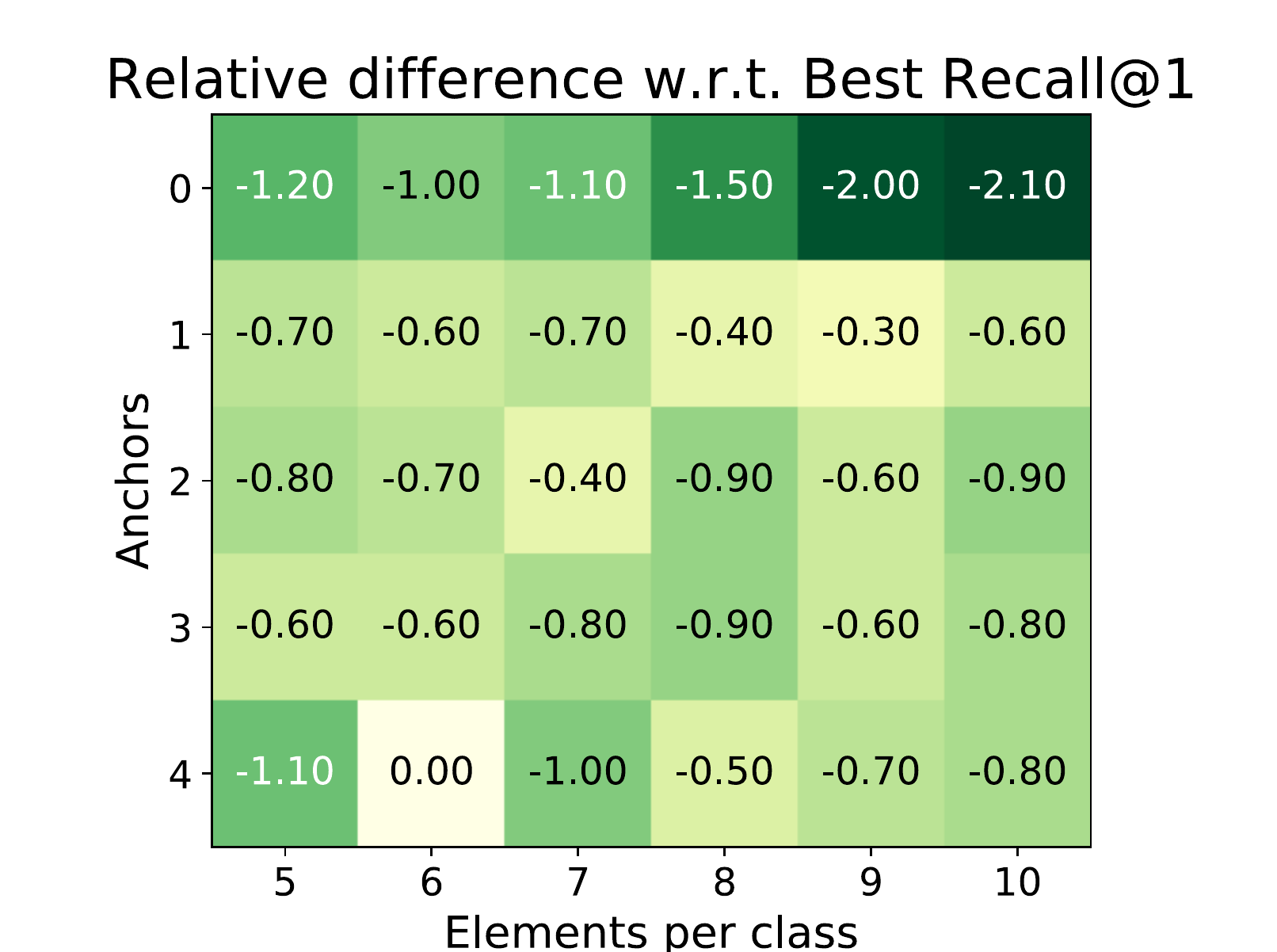} %ablation_classperbatch_withdouble.eps}
  \caption{The effect of the number of anchors and the number of samples per class.}
  \label{fig:cars_anchors}
\end{minipage}
\hfill %
\end{figure}

\begin{figure}[t!]
\centering
\begin{minipage}[t!]{.49\textwidth}
  \centering
\includegraphics[width=\textwidth, trim={0cm 0cm 0cm 0cm},clip]{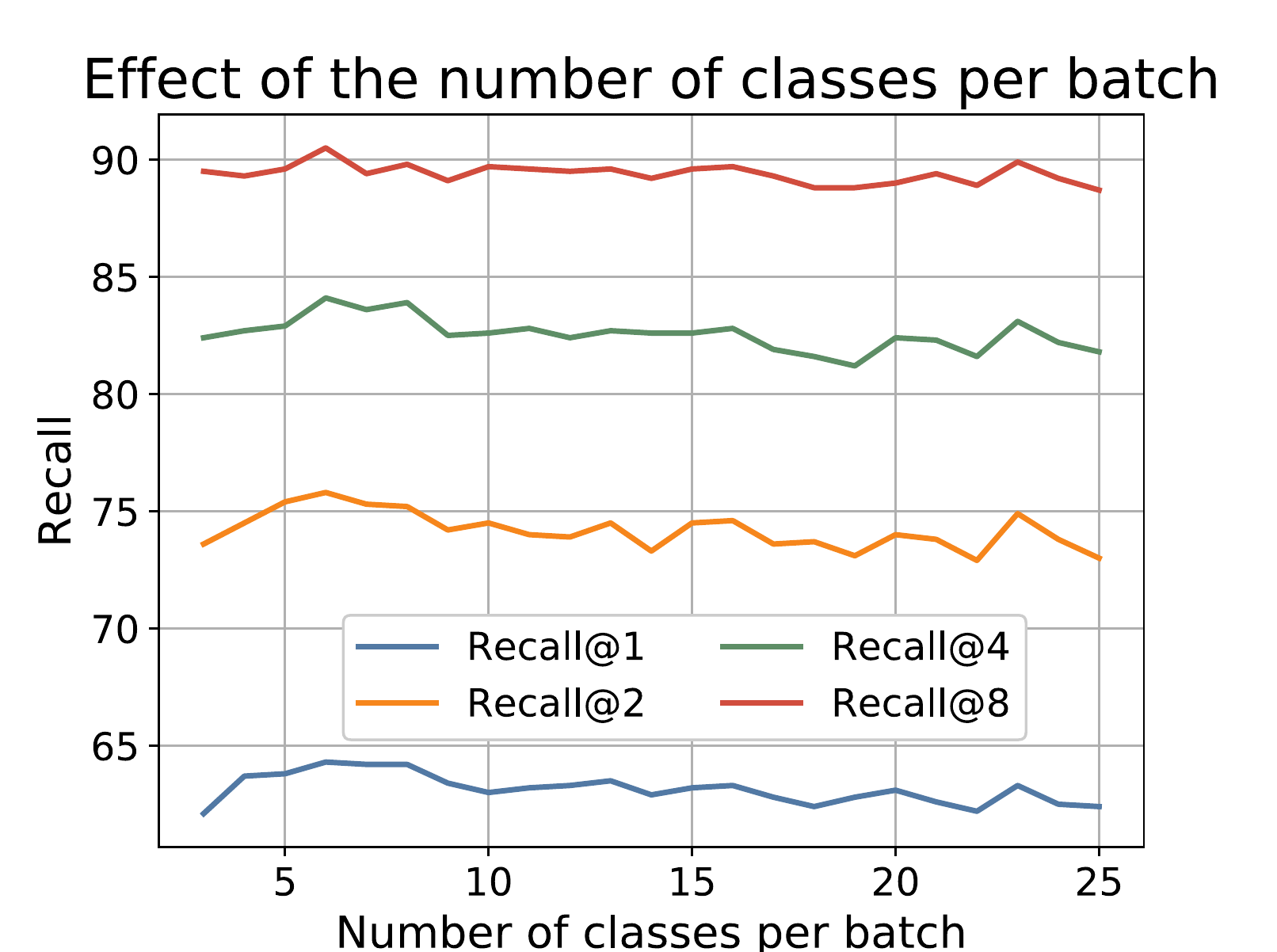} %ablation_classperbatch_withdouble.eps}
  \caption{The effect of the number of classes per mini-batch.}
  \label{fig:abl_classes}
\end{minipage}
\hfill %
\begin{minipage}[t!]{.49\textwidth}
  \centering
\includegraphics[width=\textwidth, trim={0cm 0.4cm 0cm 0cm},clip]{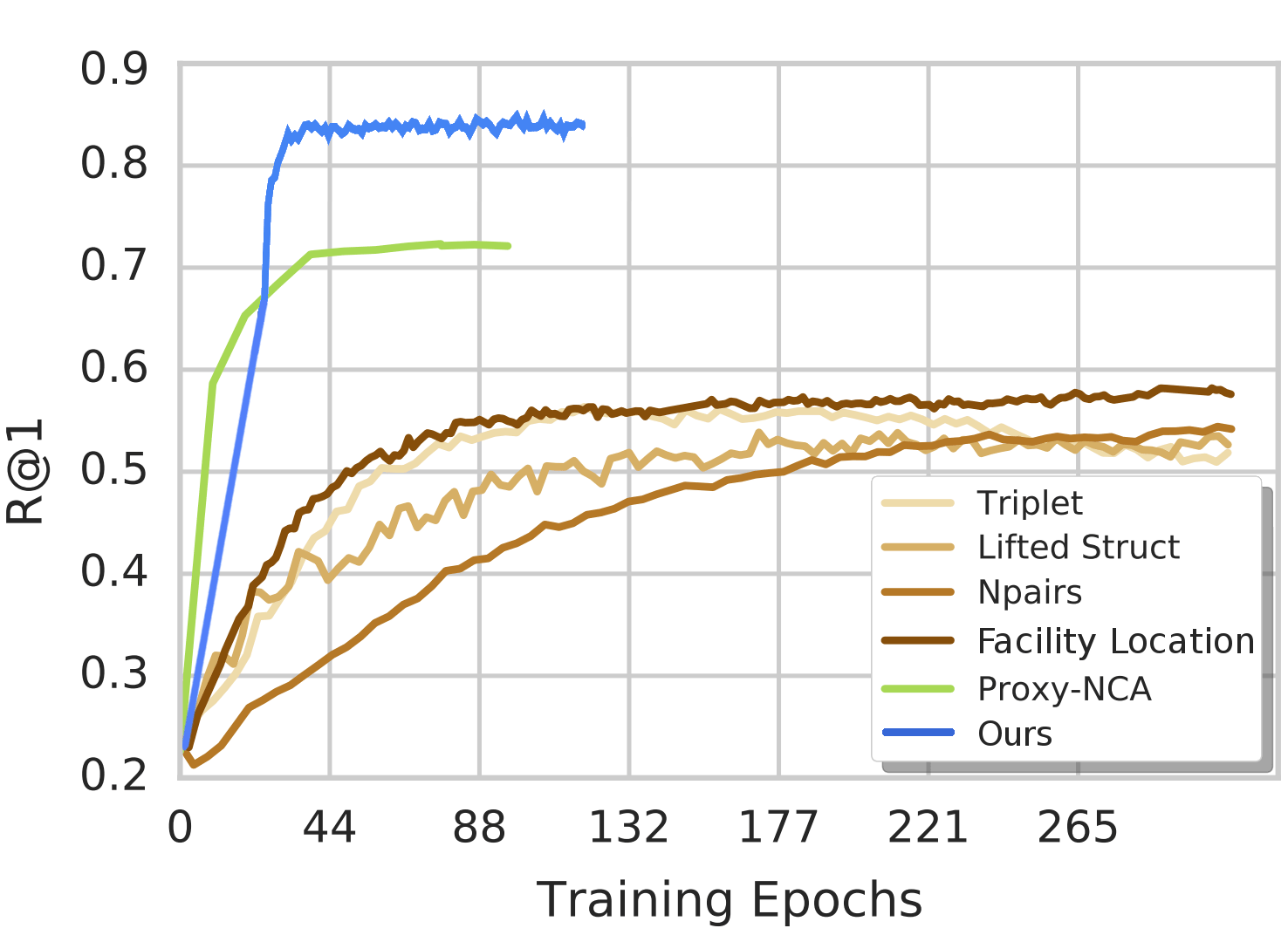} %ablation_classperbatch_withdouble.eps}
  \caption{Recall@1 as a function of training epochs on Cars196 dataset. Figure adapted from \cite{DBLP:conf/iccv/Movshovitz-Attias17}.}
  \label{fig:abl_convergence}
\end{minipage}
\hfill %
\end{figure}

\textbf{Number of anchors.} 
In Fig.~\ref{fig:abl_labeledpoints}, we show the effect of the number of anchors with respect to the number of samples per class. We do the analysis on \textit{CUB-200-2011} dataset and give a similar analysis for \textit{CARS} dataset in the supplementary material. The results reported are the percentage point differences in terms of Recall@1 with respect to the best performing set of parameters (see $Recall@1=64.3$ in Tab.~\ref{tab:res_loss}). 
The number of anchors ranges from 0 to 4, while the number of samples per class varies from 5 to 10. 
It is worth noting that our best setting considers 1 or 2 anchors over 9 samples. Moreover, even when we do not use any anchor, the difference in Recall@1 is no more than $2pp$.

We report the same analysis for the \textit{Cars 196} \cite{KrauseStarkDengFei-Fei_3DRR2013} dataset, leading us to the same conclusions. We increase the number of elements per class from $5$ to $10$, and in each case, we vary the number of anchors from $0$ to $4$. We show the results in Fig. \ref{fig:cars_anchors}.
Note, the results decrease mainly when we do not have any labeled sample, i.e., when we use zero anchors.
The method shows the same robustness as on the \textit{CUB-200-2011} \cite{WahCUB_200_2011} dataset, with the best result being only $2.1$ percentage points better at the Recall@1 metric than the worst result.

\textbf{Number of classes per mini-batch.}
In Fig.~\ref{fig:abl_classes}, we present the change in Recall@1 on the \textit{CUB-200-2011} dataset if we increase the number of classes we sample at each iteration. The best results are reached when the number of classes is not too large. This is a welcome property, as we are able to train on small mini-batches, known to achieve better generalization performance \cite{DBLP:conf/iclr/KeskarMNST17}.

%\vspace{0.2cm}
\textbf{Convergence rate.} In Fig.~\ref{fig:abl_convergence}, we present the convergence rate of the model on the \textit{Cars 196} dataset. Within the first $30$ epochs, our model achieves state-of-the-art results, making our model significantly faster than other approaches. Note, that other models, with the exception of Proxy-NCA \cite{DBLP:conf/iccv/Movshovitz-Attias17}, need hundreds of epochs to converge. 
Additionally, we compare the training time with Proxy-NCA \cite{DBLP:conf/iccv/Movshovitz-Attias17}. On a single Volta V100 GPU, the average running time of our method per epoch is $23.59$ seconds on \textit{CUB-200-2011} and $39.35$ seconds on \textit{Cars 196}, compared to $27.43$ and $42.56$ of Proxy-NCA \cite{DBLP:conf/iccv/Movshovitz-Attias17}.
Hence, our method is faster than one of the fastest methods in the literature. Note, the inference time of every method is the same because the network is used only for feature embedding extraction during inference.

\begin{figure*}[t]
\centering
\begin{minipage}[t]{.49\textwidth}
  \centering
\includegraphics[width=\textwidth, trim={0cm 0cm 0cm 0cm},clip]{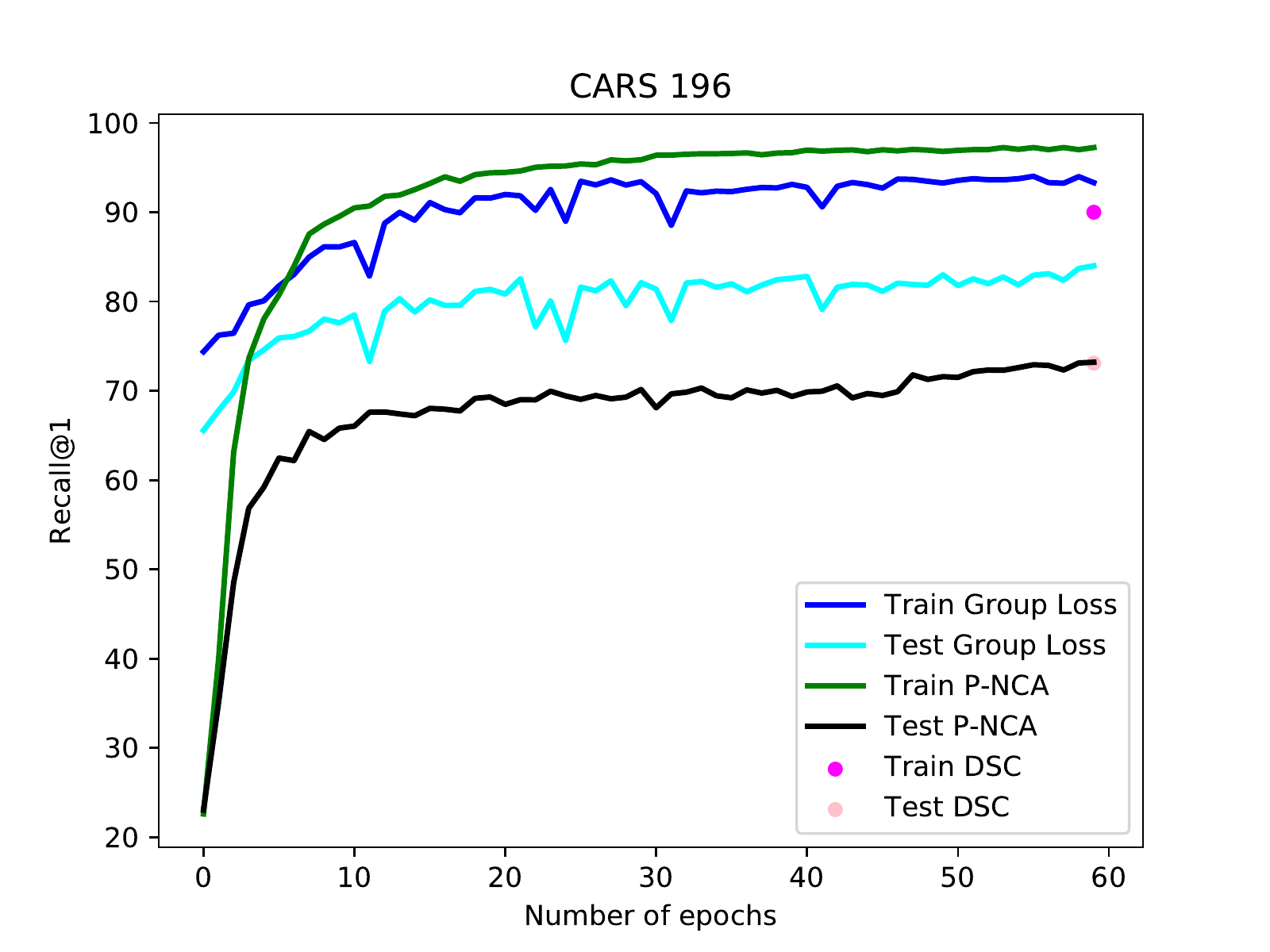} %ablation_labeled_double_loss.eps}
  \caption{Training vs testing Recall@1 curves on \textit{Cars 196} dataset.}
  \label{fig:regularization_cars}
\end{minipage}
\hfill %
\begin{minipage}[t]{.49\textwidth}
  \centering
\includegraphics[width=\textwidth, trim={0cm 0cm 0cm 0cm},clip]{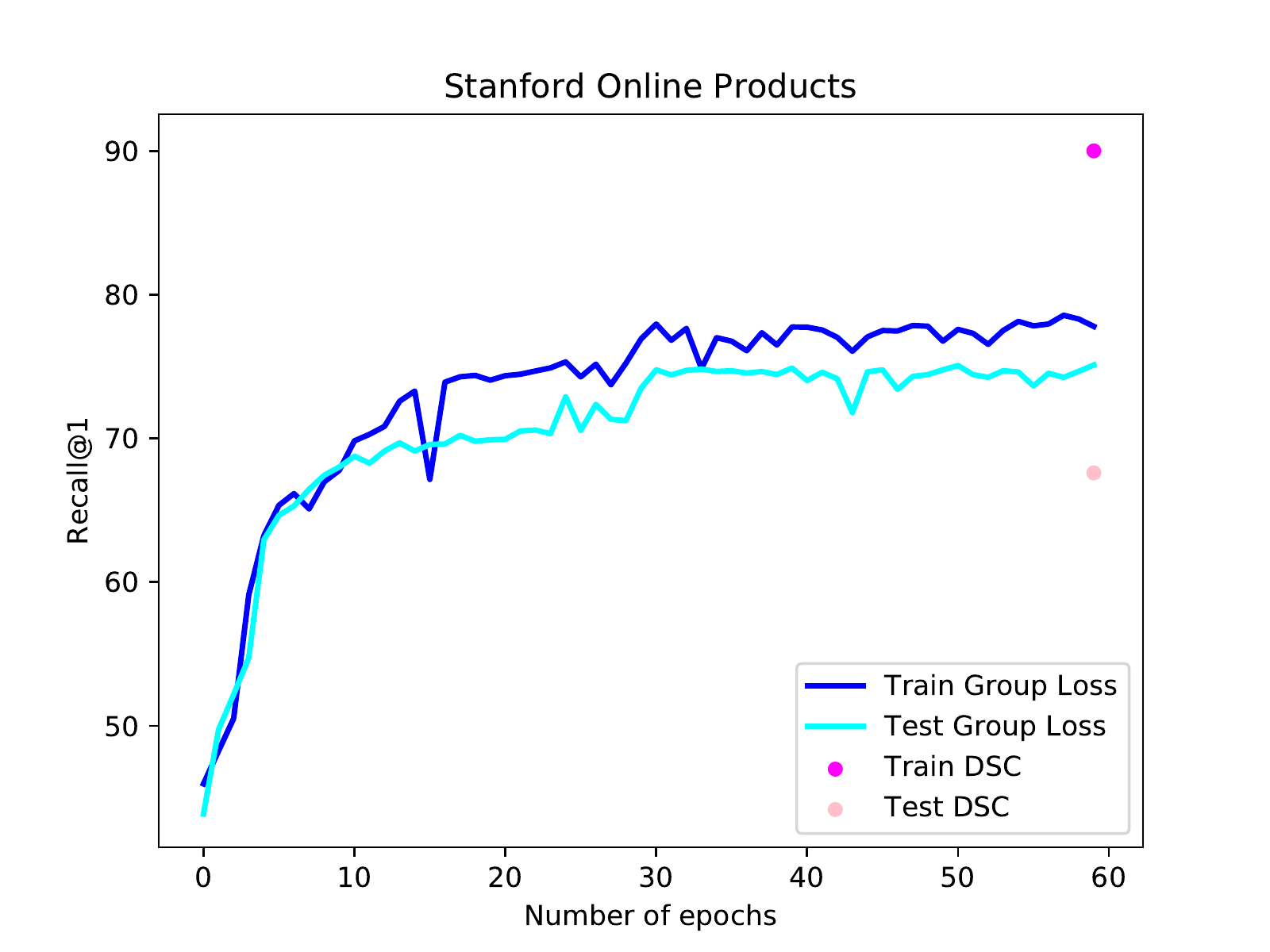} %ablation_classperbatch_withdouble.eps}
  \caption{Training vs testing Recall@1 curves on \textit{Stanford Online Products} dataset.}
  \label{fig:regularization_sop}
\end{minipage}
\hfill %
\end{figure*}

\textbf{Implicit regularization and less overfitting.} In Figures \ref{fig:regularization_cars} and \ref{fig:regularization_sop}, we compare the results of training vs. testing on \textit{Cars 196} \cite{KrauseStarkDengFei-Fei_3DRR2013} and \textit{Stanford Online Products} \cite{DBLP:conf/cvpr/SongXJS16} datasets.
We see that the difference between Recall@1 at train and test time is small, especially on \textit{Stanford Online Products} dataset.
On \textit{Cars 196} the best results we get for the training set are circa $93\%$ in the Recall@1 measure, only $9$ percentage points ($pp$) better than what we reach in the testing set.
From the works we compared the results with, the only one which reports the results on the training set is \textit{Deep Spectral Clustering Learning} \cite{DBLP:conf/icml/LawUZ17}. They reported results of over $90\%$ in all metrics for all three datasets (for the training sets), much above the test set accuracy which lies at $73.1\%$ on \textit{Cars 196} and $67.6\%$ on \textit{Stanford Online Products} dataset.
This clearly shows that our method is much less prone to overfitting.

We further implement the P-NCA \cite{DBLP:conf/iccv/Movshovitz-Attias17} loss function and perform a similar experiment, in order to be able to compare training and test accuracies directly with our method. In Figure \ref{fig:regularization_cars}, we show the training and testing curves of P-NCA on the \textit{Cars 196} \cite{KrauseStarkDengFei-Fei_3DRR2013} dataset. We see that while in the training set, P-NCA reaches results of $3pp$ higher than our method, in the testing set, our method outperforms P-NCA by around $10pp$. Unfortunately, we were unable to reproduce the results of the paper \cite{DBLP:conf/iccv/Movshovitz-Attias17} on \textit{Stanford Online Products} dataset.

Furthermore, even when we turn off $L2$-regularization, the generalization performance of our method does not drop at all.
Our intuition is that by taking into account the structure of the entire manifold of the dataset, our method introduces a form of regularization.
We can clearly see a smaller gap between training and test results when compared to competing methods, indicating less overfitting.
We plan to further investigate this phenomenon in future work.

\subsection{Other backbones}

In the previous section, we perform all experiments using a GoogleNet backbone with batch normalization. This choice is motivated by the fact that most methods use this backbone, making comparisons fair. 
In this section, we explore the performance of our method for other backbone architectures, to show the generality of our proposed loss formulation. 
We choose to train a few networks from Densenet family \cite{DBLP:conf/cvpr/HuangLMW17}. Densenets are a modern CNN architecture which show similar classification accuracy to GoogleNet in most tasks (so they are a similarly strong classification baseline \footnote{The classification accuracy of different backbones can be found in the following link: \url{https://pytorch.org/docs/stable/torchvision/models.html}. BN-Inception's top 1/top 5 error is 7.8\%/25.2\%, very similar to those of Densenet121 (7.8\%/25.4\%).}). 
Furthermore, by training multiple networks of the same family, we can study the effect of the capacity of the network, i.e., how much can we gain from using a larger network? 
Finally, we are interested in studying if the choice of hyperparameters can be transferred from one backbone to another.

We present the results of our method using Densenet backbones in Tab. \ref{tab:densenets}. 
We use the same hyperparameters as the ones used for the GoogleNet experiments, reaching state-of-the-art results on both \textit{CARS 196} \cite{KrauseStarkDengFei-Fei_3DRR2013} and \textit{Stanford Online Products} \cite{DBLP:conf/cvpr/SongXJS16} datasets, even compared to ensemble and sampling methods. 
The results in \textit{Stanford Online Products} \cite{DBLP:conf/cvpr/SongXJS16} are particularly impressive considering that this is the first time any method in the literature has broken the $80$ point barrier in Recall@1 metric. 
We also reach state-of-the-art results on the \textit{CUB-200-2011} \cite{WahCUB_200_2011} dataset when we consider only methods that do not use ensembles (with the Group Loss ensemble reaching the highest results in this dataset). 
We observe a clear trend when increasing the number of parameters (weights), with the best results on both \textit{CARS 196} \cite{KrauseStarkDengFei-Fei_3DRR2013} and \textit{Stanford Online Products} \cite{DBLP:conf/cvpr/SongXJS16} datasets being achieved by the largest network, Densenet161 (whom has a lower number of convolutional layers than Densenet169 and Densenet201, but it has a higher number of weights/parameters).

Finally, we study the effects of hyperparameter optimization. Despite that the networks reached state-of-the-art results even without any hyperparameter tuning, we expect a minimum amount of hyperparameters tuning to help. 
To this end, we used random search \cite{DBLP:journals/jmlr/BergstraB12} to optimize the hyperparameters of our best network on the \textit{CARS 196} \cite{KrauseStarkDengFei-Fei_3DRR2013} dataset. We reach a $90.7$ score ($2pp$ higher score than the network with default hyperparameters) in Recall@1, and $77.6$ score ($3pp$ higher score than the network with default hyperparameters) in NMI metric, showing that individual hyperparameter optimization can boost the performance. 
The score of $90.7$ in Recall@1 is not only by far the highest score ever achieved, but also the first time any method has broken the $90$ point barrier in Recall@1 metric when evaluated on the \textit{CARS 196} \cite{KrauseStarkDengFei-Fei_3DRR2013} dataset.

\begin{table}[]
\begin{tabular}{|l|l|l|l|l|l|l|l|l|l|}
\hline
\multicolumn{1}{|c|}{\multirow{2}{*}{Model}} & \multicolumn{3}{c|}{CUB}                                                          & \multicolumn{3}{c|}{CARS}                                                         & \multicolumn{3}{c|}{SOP}                                                          \\ \cline{2-10} 
\multicolumn{1}{|c|}{}                       & \multicolumn{1}{c|}{Params} & \multicolumn{1}{c|}{R@1} & \multicolumn{1}{c|}{NMI} & \multicolumn{1}{c|}{Params} & \multicolumn{1}{c|}{R@1} & \multicolumn{1}{c|}{NMI} & \multicolumn{1}{c|}{Params} & \multicolumn{1}{c|}{R@1} & \multicolumn{1}{c|}{NMI} \\ \hline
GL Densenet121                 & 7056356      &   \textbf{65.5}  & 69.4 &  7054306    & 88.1  & 74.2  &  18554806   &   78.2 &   91.5     \\
GL Densenet161                 &  26692900     &  64.7  &  68.7 &  26688482    & \textbf{88.7}  & 74.6 & 51473462    & \textbf{80.3} &  \textbf{92.3}        \\
GL Densenet169                 &  12650980     &  65.4   & \textbf{69.5} &  12647650    & 88.4 & 75.2 & 31328950    &    79.4 &  92.0    \\
GL Densenet201                 &  18285028     &  63.7  & 68.4  &  18281186    &  88.6 & \textbf{75.8}  & 39834806    &    79.8   &  92.1     \\ \hline
GL Inception v2                 &   10845216    &  64.3  & 67.9  &  10846240    & 83.7   & 70.7 & 16589856    & 75.1   & 90.8       \\ \hline \hline
SofTriple 10  \cite{DBLP:journals/corr/abs-1909-05235}             & 11307040      &  65.4  & 69.3  &   11296800   & 84.5  & 70.1 &  68743200   & 78.3  & 92        \\ \hline
\end{tabular}
\caption{The results of Group Loss in Densenet backbones and comparisons with SoftTriple loss \cite{DBLP:journals/corr/abs-1909-05235}}
\label{tab:densenets}
\end{table}

\subsection{Comparisons with SoftTriple loss \cite{DBLP:journals/corr/abs-1909-05235}}

A recent paper (SoftTriple loss \cite{DBLP:journals/corr/abs-1909-05235}, ICCV 2019) explores another type of classification loss for the problem of metric learning. The main difference between our method and \cite{DBLP:journals/corr/abs-1909-05235} is that our method checks the similarity between samples, and then refines the predicted probabilities (via a dynamical system) based on that information. \cite{DBLP:journals/corr/abs-1909-05235} instead deals with the intra-class variability, but does not explicitly take into account the similarity between the samples in the mini-batch. 
They propose to add a new layer with $10$ units per class. 

We compare the results of \cite{DBLP:journals/corr/abs-1909-05235} with our method in Tab. \ref{tab:densenets}. SoftTriple loss \cite{DBLP:journals/corr/abs-1909-05235} reaches a higher result than our method in all three datasets in Recall@1 metric, and higher results than the Group Loss on the \textit{CUB-200-2011} and \textit{Stanford Online Products} datasets in NMI metric. However, this comes at a cost of significantly increasing the number of parameters. 
On the \textit{Stanford Online Products} dataset in particular, the number of parameters of \cite{DBLP:journals/corr/abs-1909-05235} is $68.7$ million. In comparison, we (and the other methods we compare the results with in the main paper) use only $16.6$ million parameters. 
In effect, their increase in performance comes at the cost of using a neural network which is 4 times larger as ours, making results not directly comparable. 
Furthermore, using multiple centres is crucial for the performance of \cite{DBLP:journals/corr/abs-1909-05235}. Fig. 4 of the work \cite{DBLP:journals/corr/abs-1909-05235} shows that when they use only $1$ centre per class, the performance drops by $3pp$, effectively making \cite{DBLP:journals/corr/abs-1909-05235} perform worse than the Group Loss by $2pp$. 

We further used the official code implementation to train their network using only one center on the  \textit{CARS 196} \cite{KrauseStarkDengFei-Fei_3DRR2013} dataset, reaching $83.1$ score in Recall@1, and $70.1$ score in NMI metric, with each score being $0.6pp$ lower than the score of The Group Loss. 
Essentially, when using the same backbone, SoftTriple loss \cite{DBLP:journals/corr/abs-1909-05235} reaches lower results than our method. 

As we have shown in the previous section, increasing the number of parameters improves the performances of the network, but it is not a property of the loss function. In fact, a similarly sized network to theirs (Densenet 169) consistently outperforms SoftTriple loss, as can be seen in Tab. \ref{tab:densenets}. 
For this reason, we keep this comparison in the supplementary material, while we leave for the main paper the comparisons with more than $20$ methods that use the same backbone.

\section{t-SNE on CUB-200-2011 dataset}

\begin{figure*}[ht]
\centering
\includegraphics[width=0.99\textwidth, trim={1cm 3cm 0.5cm 5cm},clip]{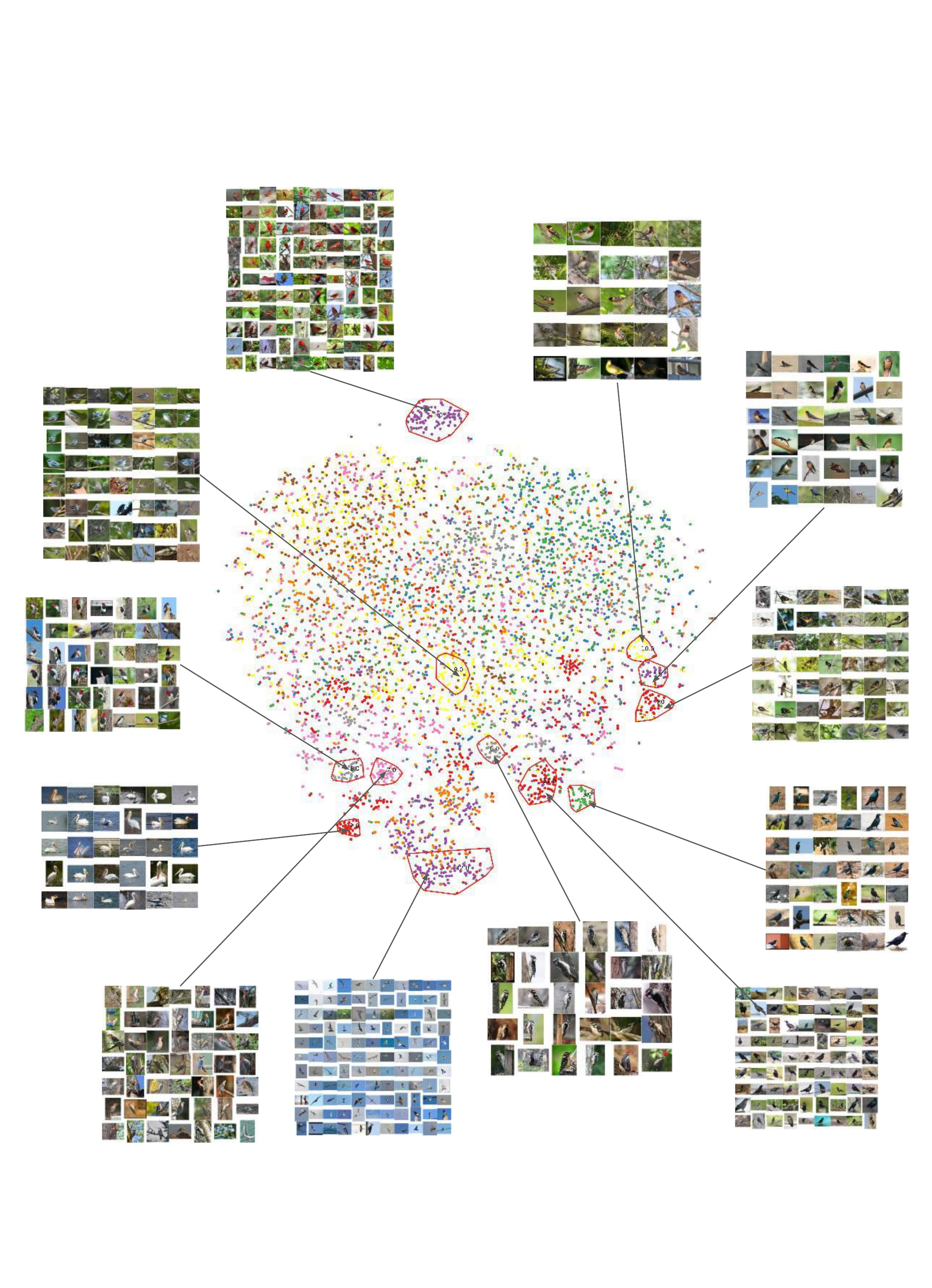}
  \caption{t-SNE \cite{DBLP:journals/ml/MaatenH12} visualization of our embedding
on the CUB-200-2011 \cite{WahCUB_200_2011} dataset, with some clusters highlighted. Best viewed on a monitor
when zoomed in.}
  \label{fig:t-sne}
\end{figure*}

Fig. \ref{fig:t-sne} visualizes the t-distributed stochastic neighbor embedding (t-SNE) \cite{DBLP:journals/ml/MaatenH12} of the embedding vectors obtained by our method on the \textit{CUB-200-2011} \cite{WahCUB_200_2011} dataset. 
The plot is best viewed on a high-resolution monitor when zoomed in. We highlight several representative groups by enlarging the corresponding regions in the corners. Despite the large pose and appearance variation, our method efficiently generates a compact feature mapping that preserves semantic similarity.

\section{Conclusions and Future Work}
\vspace{0.3cm}

In this work, we proposed the Group Loss, a new loss function for deep metric learning that goes beyond triplets. 
By considering the content of a mini-batch, it promotes embedding similarity across all samples of
the same class, while enforcing dissimilarity for elements of different classes.
This is achieved with a fully-differentiable layer that is used to train a convolutional network in an end-to-end fashion.
We show that our model outperforms state-of-the-art methods on several datasets, and at the same time shows fast convergence.

In our work, we did not consider any advanced and intelligent sampling strategy. Instead, we randomly sample objects from a few classes at each iteration. Sampling has shown to have a very important role in feature embedding \cite{DBLP:conf/iccv/ManmathaWSK17}, therefore, we will explore in future work sampling techniques which can be suitable for our module. Additionally, we are going to investigate the applicability of Group Loss to other problems, such as person re-identification, landmark matching and deep semi-supervised learning.

\chapter{DeepScores - a Dataset for Segmentation, Detection and Classification of Tiny Objects}

\section{Disclaimer}

The work presented in this chapter is based on the following paper:

\begin{description}  
  \item Lukas Tuggener, \textbf{Ismail Elezi}, Jurgen Schmidhuber, Marcello Pelillo and Thilo Stadelmann; \textit{DeepScores-a dataset for segmentation, detection and classification of tiny objects \cite{DBLP:conf/icpr/TuggenerESPS18}}; In Proceedings of IAPR International Conference on Pattern Recognition (ICPR 2018)
\end{description}

The contributions of the author are the following:

\begin{description}
  \item[$\bullet$ Building] the ground truth for the segmentation task.
  \item[$\bullet$ Experimenting] with SOTA detectors in the dataset.
  \item[$\bullet$ Writing] a part of the paper.
\end{description}

\section{Introduction}

Increased availability of data and computational power has been often followed by progress in computer vision and machine learning. The recent rise of deep learning in computer vision for instance has been promoted by availability of large image datasets \cite{DBLP:conf/cvpr/DengDSLL009} and increased computational power provided by GPUs \cite{gpu2004, Raina2009}.

Optical music recognition (OMR) \cite{Rebelo2012} is a classical and challenging area of computer vision that aims at converting scans of written music to machine-readable form, much like optical character recognition (OCR) \cite{mori1999} does it for printed text. To the best of our knowledge, there are no OMR systems yet that fully leverage the power of deep learning. We conjecture that this is caused in part by the lack of publicly available datasets of written music, big enough to train deep neural networks. The \emph{DeepScores} dataset has been collected with OMR in mind, but addresses important aspects of next generation computer vision research that pertain to the size and number of objects per image.

Although there is already a number of clean, large datasets available to the computer vision community \cite{DBLP:conf/cvpr/DengDSLL009, DBLP:journals/ijcv/EveringhamGWWZ10, DBLP:conf/eccv/LinMBHPRDZ14}, those datasets are similar to each other in the sense that for each image there are a few large objects of interest. Object detection approaches that have shown state-of-the-art performance under these circumstances, such as Faster R-CNN \cite{DBLP:conf/nips/RenHGS15}, SSD \cite{DBLP:conf/eccv/LiuAESRFB16} and YOLO \cite{Redmon2016}, demonstrate very poor off-the-shelf performances when applied to environments with large input images containing multiple small objects (see Section \ref{sec_baselines}). 

Sheets of written music, on the other hand, usually have dozens to hundreds of small salient objects. The class distribution of musical symbols is strongly skewed and the symbols have a large variability in size. Additionally, the OMR problem is very different from modern OCR \cite{Goodfellow2013, Lee16}: while in classical OCR, the text is basically a 1D signal (symbols to be recognized are organized in lines of fixed height, in which they extend from left to right or vice versa), musical notation can additionally be stacked arbitrarily also on the vertical axis, thus becoming a 2D signal. This superposition property would exponentially increase the number of symbols to be recognized, if approached the usual way (which is intractable from a computational as well as from a classification point of view). It also makes segmentation very hard and does not imply a natural ordering of the symbols as for example in the SVHN dataset \cite{Netzer2011}.

In this work, we present the \emph{DeepScores} dataset with the following contributions: a) a curated and publicly available collection of hundreds of thousands of musical scores, containing tens of millions of objects to construct a high quality dataset of written music; b) available ground truth for the tasks of object detection, semantic segmentation, and classification; c) comprehensive comparisons with other computer vision datasets (see Section \ref{sec_comparison}) and a quantitative and qualitative analysis of \emph{DeepScores} (see Section \ref{sec_statistics}); d) computation of an object classification baseline (see Section \ref{sec_baselines}) together with an outlook on how to facilitate next generation computer vision research using \emph{DeepScores} (see Section \ref{sec_conclusions}).

\begin{figure}[t]
\centering
\includegraphics[width=0.8\textwidth]{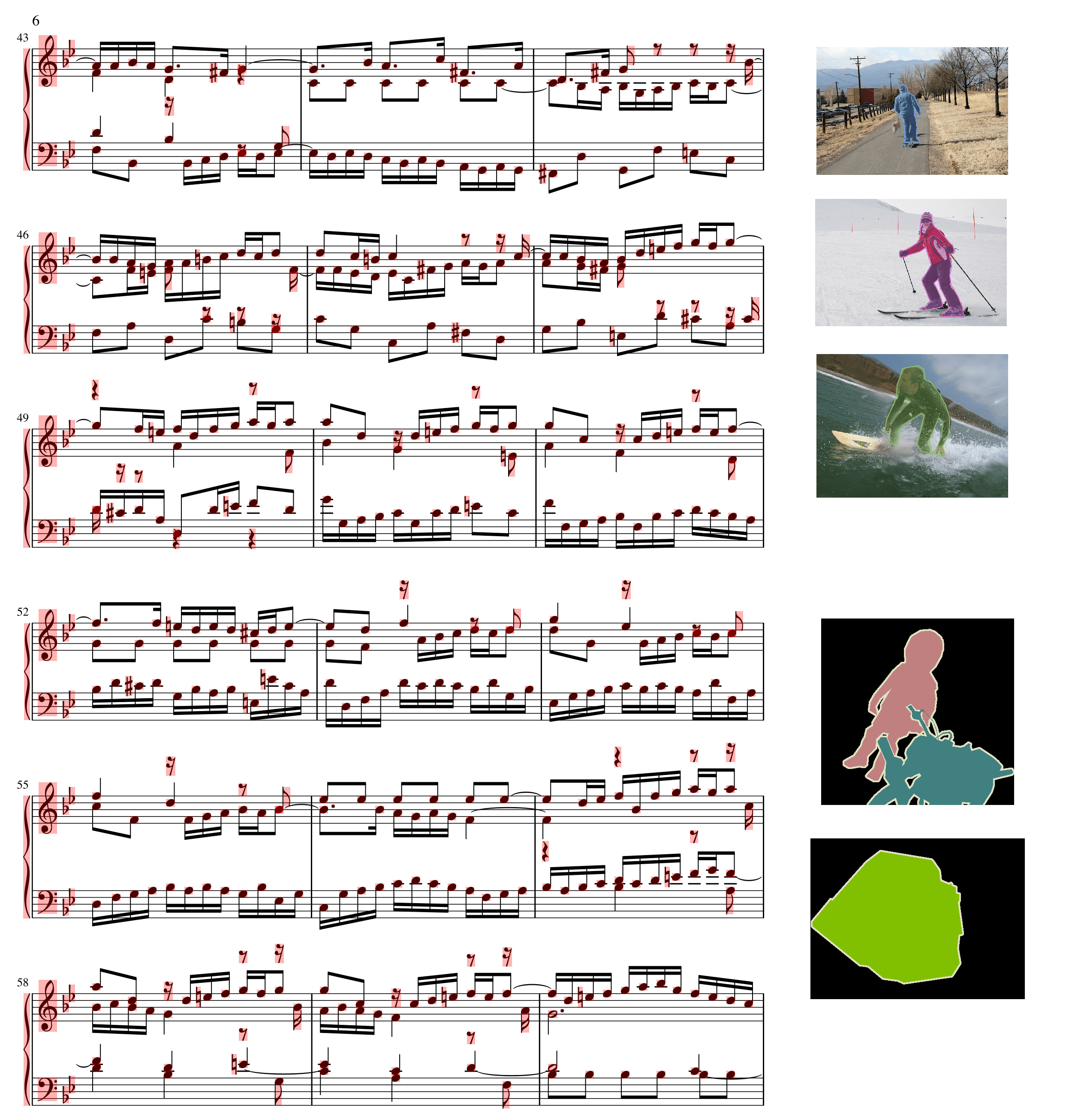}
\caption{A typical image and ground truth from the \emph{DeepScores} dataset (left), next to examples from the MS-COCO (3 images, top right) and PASCAL VOC (2 images, bottom right) datasets. Even though the music page is rendered at a much higher resolution, the objects are still smaller; the size ratio between the images is realistic despite all images being downscaled.}
\label{fi:1}
\end{figure}

\section{\emph{DeepScores} in the context of other datasets}
\label{sec_comparison}

\emph{DeepScores} is a high quality dataset consisting of pages of written music, rendered at $400$ dots per inch (dpi). It has $300'000$ full pages as images, containing tens of millions of objects, separated in $118$ classes. The aim of the dataset is to facilitate general research on small object recognition, with direct applicability to the recognition of musical symbols. We provide the dataset with three different kinds of ground truths (in the order of progressively increasing task complexity): object classification, semantic segmentation, and object detection.

\begin{figure}
\begin{center}

\includegraphics[width=.75\linewidth]{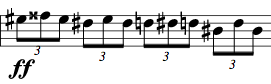}

\includegraphics[width=.75\linewidth]{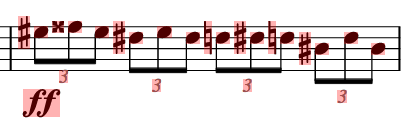}

\includegraphics[width=.75\linewidth]{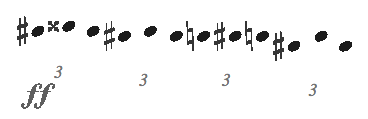}

\includegraphics[width=.5\linewidth]{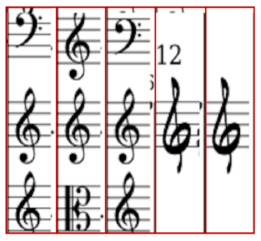}

\end{center}  
\caption{Examples for the different flavors of ground truth available in \emph{DeepScores}. From top to bottom: Snippet of an input image; Bounding boxes over single objects from previous snippet for object detection; Color-based pixel level labels (the differences can be hard to see, but there is a distinct color per symbol class) for semantic segmentation; Patches centered around specific symbols for object classification.}
\label{fi:2}  
\end{figure}

\textbf{Object classification} in the context of computer vision is the procedure of labeling an image with a single label. Its recent history is closely linked to the success of deep convolutional learning models \cite{LeCun1990}, leading to superhuman performance \cite{DBLP:conf/cvpr/CiresanMS12} and subsequent ImageNet object classification breakthroughs \cite{DBLP:conf/nips/KrizhevskySH12}. Shortly afterwards, similar systems achieved human-level accuracy also on ImageNet \cite{Simonyan2014, DBLP:conf/cvpr/SzegedyLJSRAEVR15, DBLP:conf/cvpr/HeZRS16}. Generally speaking, the ImageNet dataset \cite{DBLP:conf/cvpr/DengDSLL009} was a key ingredient to the success of image classification algorithms.

In \emph{DeepScores}, we provide data for the classification task even though classifying musical symbols in isolation is not a challenging problem compared to classifying ImageNet images. But providing the dataset for classification, in addition to a neural network implementation that achieves high accuracy (see Section \ref{sec_baselines}), might help to address the other two tasks. In fact, the first step in many computer vision models is to use a deep convolutional neural network pre-trained on ImageNet, and alter it for the task of image segmentation or image detection \cite{Long2015, DBLP:conf/nips/RenHGS15}. We expect that the same technique can be used when it comes to detecting very small objects.

\textbf{Semantic segmentation} is the task of labeling each pixel of the image with one of the possible classes. State-of-the-art models are typically based on fully convolutional architectures \cite{Ciresan2012isbi, Long2015}. The task arguably is a significantly more difficult problem than image classification, with the recent success being largely attributed to the release of high quality datasets like PASCAL VOC \cite{DBLP:journals/ijcv/EveringhamGWWZ10} and MS-COCO \cite{DBLP:conf/eccv/LinMBHPRDZ14}. 

In \emph{DeepScores}, we provide ground truth for each pixel in all the images, having roughly $10^{12}$ labeled pixels in the dataset. In the next section, we compare these figures with existing datasets. 

\textbf{Object detection} is the by far most interesting and challenging task: to classify all the objects in the image, and at the same time to find their precise position in the image. State-of-the-art algorithms are pipeline convolutional models, typically having combined cost functions for detection and classification \cite{DBLP:conf/nips/RenHGS15, DBLP:conf/cvpr/RedmonF17, DBLP:conf/eccv/LiuAESRFB16}. The task can be combined with segmentation, which means that the algorithm is required to provide masks (instead of bounding boxes) for each of the objects in the image \cite{DBLP:conf/iccv/HeGDG17}. It differs from mere segmentation in the fact that the result shows which pixels together form an object. Similar to the case of semantic segmentation above, the PASCAL VOC and especially MS-COCO datasets have played an important part on the recent success of object detection algorithms.

In \emph{DeepScores}, we provide bounding boxes and labels for each of the musical symbols in the dataset. With around $80$ million objects, this makes our dataset the largest one released so far, and highly challenging: the above-mentioned algorithms did not work well on our dataset in preliminary comprehensive experiments. We attribute this to the fact that most of the models used for object detection are fitted to datasets which have few but large objects. On the contrary, our dataset contains a lot of very small objects, which means that new models might need to be created in order to deal with it.

\subsection{Comparisons with computer vision datasets}
\label{comparison}

Compared with some of the most used datasets in the field of computer vision, \emph{DeepScores} has by far the largest number of objects, in addition of having the highest resolution. In particular, images of \emph{DeepScores} have a resolution of $1'894$ x $2'668$ pixels, which is at least four times higher than the resolutions of datasets we compare with. Table \ref{num_of_classes} contains quantitative comparisons of \emph{DeepScores} with other datasets, while the following paragraphs bring in also qualitative aspects.

\textbf{SVHN}, the street view house numbers dataset \cite{Netzer2011}, contains $600'000$ labeled digits cropped from street view images. Compared to \emph{DeepScores}, the number of objects in SVHN is two orders of magnitude lower, and the number of objects per image is two to three orders of magnitude lower.

\textbf{ImageNet} \cite{DBLP:conf/cvpr/DengDSLL009} contains a large number of images and (as a competition) different tracks (classification, detection and segmentation) that together have proven to be a solid foundation for many computer vision projects. However, the objects in ImageNet are quite large, while the number of objects per image is very small. Unlike ImageNet, \emph{DeepScores} tries to address this issue by going to the other extreme, providing a very large number of very small objects, with images having significantly higher resolution than all the other mentioned datasets.

\textbf{PASCAL VOC} \cite{DBLP:journals/ijcv/EveringhamGWWZ10} is a dataset which has been assembled mostly for the tasks of detection and segmentation. Compared to ImageNet, the dataset has slightly more objects per image, but the number of images is comparatively small: our dataset is one order of magnitude bigger in the number of images, and three orders of magnitude bigger in the number of objects.

\textbf{MS-COCO} \cite{DBLP:conf/eccv/LinMBHPRDZ14} is a large upgrade over PASCAL VOC on both the number of images and number of objects per image. With more than $300$K images containing more than $3$ millions of objects, the dataset is very useful for various tasks in computer vision. However, like ImageNet, the number of objects per image is still more than one order of magnitude lower than in our dataset, while the objects are relatively large. 

\paragraph{Other datasets\\}

A number of other datasets have been released during the years, which have helped the progress of the field, and some of them have been used for different competitions. \textbf{MNIST} \cite{Lecun1998} is the first ``large'' dataset in the fields of machine learning and computer vision. It has tens of thousands of $28$x$28$ pixels grayscale images, each containing a handwritten digit. The dataset is a solved classification problem and during the last decade has been used mostly for prototyping new models. Nowadays, this is changing, with more challenging datasets like \textbf{CIFAR-10}/\textbf{CIFAR-100} \cite{Krizhevsky2009} being preferred. Similar to MNIST, those datasets contain an object per image ($32$x$32$ color pixels), which do not make them ideal for more challenging problems like detection and segmentation. 

\textbf{Caltech-101}/\textbf{Caltech-256} \cite{caltech} are more interesting datasets considering that both the resolution and the number of images are larger. Still, the images contain only a single object, making them only useful for the process of image classification. \textbf{SUN} \cite{Xiao2010} is a scene understanding dataset, containing over $100$k images, each labeled with a single class. 

The online and offline Chinese handwriting databases, \textbf{CASIA-OLHWDB} and \textbf{CASIA-HWDB} \cite{Liu2011}, were produced by $1'020$ writers using a digital pen on paper, such that both online and offline data were obtained. The samples include both isolated characters and handwritten texts (continuous scripts). Both datasets have millions of samples, separated into $7'356$ classes, making them far more interesting and challenging than digit datasets.

The German traffic sign recognition benchmark \textbf{(GTSRB)} is a multi-category classification competition held at IJCNN 2011 \cite{Stallkamp2011}. The corresponding dataset comprises a comprehensive collection of more than $50'000$ lifelike traffic sign images, reflecting the strong variations in visual appearance of signs due to distance, illumination, weather conditions, partial occlusions, and rotations. The dataset has $43$ classes with unbalanced class frequencies.

\begin{table}[t]	
\centering
\begin{tabular}{ |p{3cm}||p{2.0cm}|p{2.0cm}|p{2.cm}|p{2.cm}| }
 \hline
 \bf{Dataset} & \bf{\#classes} & \bf{\#images} & \bf{\#objects} & \bf{\#pixels}\\
 \hline
 \hline
 MNIST & 10 & 70k & 70k & 55m\\ 
 \hline
 CIFAR-10 & 10 & 60k & 60k & 61m\\
 \hline
 CIFAR-100 & 100 & 60k & 60k & 61m\\
 \hline
 Caltech-101 & 101 & 9k & 9k & 700m\\
 \hline
 Caltech-256 & 256  & 31k & 31k & 2b\\
 \hline
 SUN & 397 & 17k & 17k & 6b\\
 \hline
 PASCAL VOC & 21 & 10k & 30k & 2.5b\\
 \hline
 MS COCO & 91 & 330k &3.5m & 100b\\
 \hline
 ImageNet & 200 & \textbf{500k} & 600k & 125b\\
 \hline
 SVHN & 10 & 200k & 630k & 4b\\
 \hline
 CASIA online & \textbf{7356} &  5090 & 1.35& nn\\
 \hline
 CASIA offline & \textbf{7356} & 5090 & 1.35m & nn\\
 \hline
 GTSRB & 43 & 50k & 50k & nn\\
 \hline
 \emph{DeepScores} & 118 & 300k & \textbf{80m} & \textbf{1.5t}\\
 \hline
\end{tabular}
\caption{Information about the number of classes, images and objects for some of the most common used datasets in computer vision. The number of pixels is estimated due to most datasets not having fixed image sizes. We used the SUN 2012 object detection specifications for SUN, and the statistics of ILSVRC 2014 \cite{DBLP:conf/cvpr/DengDSLL009} detection task for ImageNet.}
\label{num_of_classes}
\end{table}

\subsection{Comparisons with OMR datasets}

A number of OMR datasets have been released in the past with a specific focus on the computer music community. \emph{DeepScores} will be of use both for general computer vision as well as to the OMR community (compare Section \ref{sec_baselines}).

\paragraph{Handwritten scores\\}

The Handwritten Online Musical Symbols dataset \textbf{HOMS} \cite{Zaragoza2014} is a reference corpus with around $15'000$ samples for research on the recognition of online handwritten music notation. For each sample, the individual strokes that the musician wrote on a Samsung Tablet using a stylus were recorded and can be used in online and offline scenarios.

The \textbf{CVC-MUSCIMA} database \cite{Fornes2012} contains handwritten music images, which have been specially designed for writer identification and staff removal tasks. The database contains $1'000$ music sheets written by $50$ different musicians with characteristic handwriting styles. 

\textbf{MUSICMA++} \cite{Hajic2017} is a dataset of handwritten music for musical symbol detection that is based on the MUSCIMA dataset. It contains $91'255$ written symbols, consisting of both notation primitives and higher-level notation objects, such as key signatures or time signatures. There are $23'352$ notes in the dataset, of which $21'356$ have a full notehead, $1'648$ have an empty notehead, and $348$ are grace notes.  

The \textbf{Capitan Collection} \cite{Zaragoza2016} is a corpus collected via an electronic pen while tracing isolated music symbols from early manuscripts. The dataset contains information on both the sequence followed by the pen (capitan stroke) as well as the patch of the source under the tracing itself (capitan score). In total, the dataset contains $10'230$ samples unevenly spread over $30$ classes.

Further OMR datasets of printed scores are reviewed by the \textbf{OMR-Datasets} project\footnote{See https://apacha.github.io/OMR-Datasets/.}. \emph{DeepScores} is by far larger than any of these or the above-mentioned dataset, containing more images and musical symbols than all the other datasets combined. In addition, \emph{DeepScores} contains only real-world scores (i.e., symbols in context as they appear in real written music), while the other datasets are either synthetic or reduced (containing only symbols in isolation or just a line per image). The sheer scale of \emph{DeepScores} makes it highly usable for the modern deep learning algorithms. While convolutional neural networks have been used before for OMR \cite{vanderWel17}, \emph{DeepScores} for the first time enables the training of very large and deep models. 

\section{The \emph{DeepScores} dataset}
\label{sec_statistics}

\begin{figure*}[h!]
  \includegraphics[width=\textwidth]{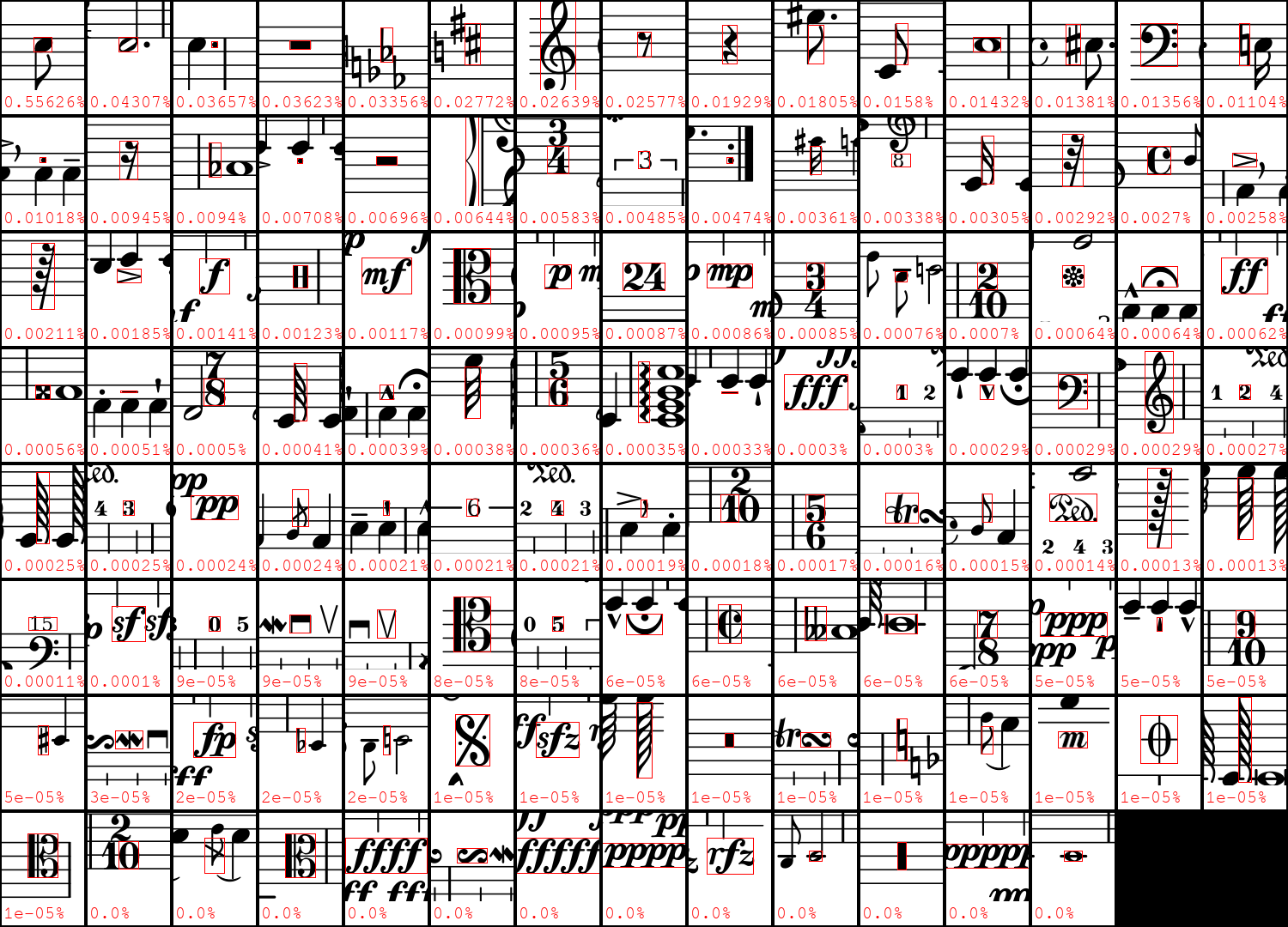}
  \caption{Examples for each of the $118$ classes present in \emph{DeepScores}, ordered by their frequency of occurrence. Even though the resolution is reduced for this plot, some of the larger symbols like \texttt{brace} (row 2, column 6) or \texttt{gClef} (row 1, column 7) are only shown partially to keep a fixed size for each patch. The symbols' full resolution in the dataset is such that the inter-line distance between two staff lines amounts to $20$ pixels.}
  \label{fi:3} 
\end{figure*}

\subsection{Quantitative properties}
\emph{DeepScores} contains around $300'000$ pages of digitally rendered music scores and has ground truth for $118$ different symbol classes. The number of labeled music symbol instances is roughly $80$ million ($4$-$5$ orders of magnitudes higher than in the other music datasets; when speaking of symbols, we mean labeled musical symbols that are to be recognized as objects in the task at hand). The number of symbols on one page can vary from as low as $4$ to as high as $7'664$ symbols. On average, a sheet (i.e., an image) contains around $243$ symbols. Table \ref{statistics} gives the mean, standard deviation, median, maximum and minimum number of symbols per page in the second column. 

\begin{table}[t] 
\centering
\begin{tabular}{ |l||r|r| }
 \hline
 \bf{Statistic} & \bf{Symbols per sheet} & \bf{Symbols per class} \\
 \hline
 \hline
 Mean & 243 & 650k\\
 \hline
 Std. dev. & 203 & 4m\\
 \hline
 Maximum & 7'664 & 44m\\
 \hline
 Minimum & 4 & 18\\
 \hline
 Median & 212 & 20k\\
 \hline
\end{tabular}
\caption{Statistical measures for the occurrences of symbols per musical sheet and per class (rounded).}
\label{statistics}
\end{table}

\begin{figure}[h]
\centering
\includegraphics[width=0.5\textwidth]{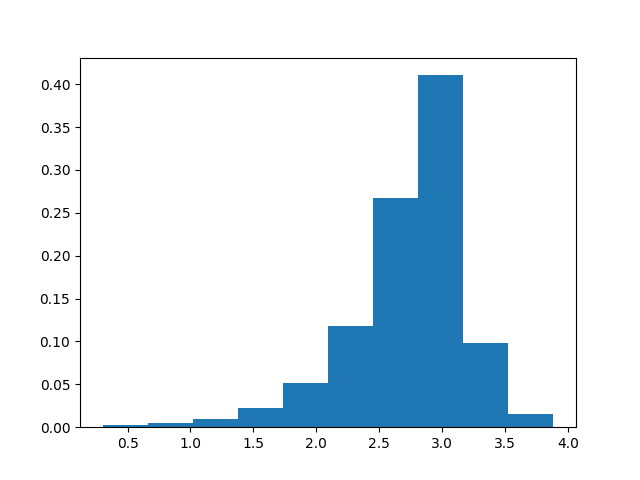}
\caption{Histogram for the distribution of symbols over all images (logarithmic scale on abscissa, ordinate weighted to give unit area). The majority of images contain from $100$ to $1000$ objects.}
\label{fi:4}
\end{figure}

Another interesting aspect of \emph{DeepScores} is the class distribution (see Figure \ref{fi:4}). Obviously, some classes contain more symbols than other classes (see also Table \ref{statistics}, column 3). It can be seen that the average number of elements per class is $600k$ but the standard deviation is $4m$, illustrating that the distribution of symbols per class is very skewed.

Figure \ref{fi:3} visualizes the symbol classes together with their occurrence probability. The most common class is \texttt{noteheadBlack}, which provides slightly more than half of the symbols in the dataset. The top $10$ classes are responsible for $86$\% of the musical symbols found.

\subsection{Flavors of ground truth}

In order for \emph{DeepScores} to be useful for as many applications as possible, we offer ground truth for three different tasks. For object classification, there are up to $5'000$ labeled image patches per class. This means we do not provide each of the $80$m symbols as a single patch for classification purposes, but constrain the dataset for this simpler task to a random subset of reasonable size (see Section \ref{sec_baselines}). The patches have a size of $45$ x $170$ and contain the full original context of the symbol (i.e., they are cropped out of real world musical scores). Each patch is centered around the symbol's bounding box (see Figure \ref{fi:2}). 

For object detection, there is an accompanying XML file for each image in \emph{DeepScores}. The XML file has an \texttt{object} node for each symbol instance present on the page, which contains class and bounding box coordinates. 

For semantic segmentation, there is an accompanying PNG file for each image. This PNG has identical size as the initial image, but each pixel has been recolored to represent the symbol class it is part of. As in Figure \ref{fi:2}, the background is white, with the published images using grayscale colors from $0$ to $118$ for ease of use in the softmax layer of potential models.

\subsection{Dataset construction}

\emph{DeepScores} is constructed by synthesizing from a large collection of written music in a digital format: crowd-sourced MusicXML files publicly available from MuseScore\footnote{https://musescore.com} and used by permission. The rendering of MuscXML with accompanying ground truth for the three flavors of granularity is done by a custom software using the SVG back-end of the open-source music engraving software LilyPond. The rendered SVG files not only contain all the musical symbols, but also additional tags that allow for identifying what musical symbol each SVG path belongs to. 

\begin{figure}[t]
\centering
\includegraphics[width=0.9\textwidth]{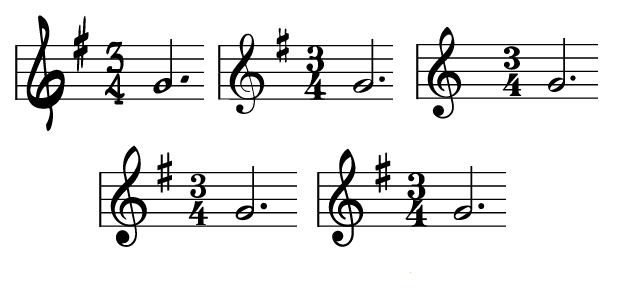}
\caption{The same patch, rendered using five different fonts.}
\label{fi:7}
\end{figure}

To achieve a realistic variety in the data even though all images are digitally rendered and therefore have perfect image quality, five different music fonts have been used for rendering (see Figure \ref{fi:7}). Python scripts finally extract the three types of ground truth from this basis of SVG data and save the images as PNG using the CairoSVG library.  

A key feature of a dataset is the definition of the classes to be included. Due to their compositional nature, there are many ways to define classes of music symbols: is it for example a ``c'' note with duration $8$ (\texttt{noteheadBlack}) or is it a black notehead (\texttt{noteheadBlack}) and a flag (\texttt{flag8thUp} or \texttt{flag8thDown})? Adding to this complexity, there is a huge number of special and thus infrequent symbols in music notation. The selected set is the result of many discussions with music experts and contains the most important symbols. We decided to use atomic symbol parts as classes which makes it possible for everyone to define composite symbols in an application-dependent way. 

\section{Anticipated use and impact}
\label{sec_baselines}

\subsection{Unique challenges}

\begin{figure}[h]
\centering
\includegraphics[width=0.4\textwidth]{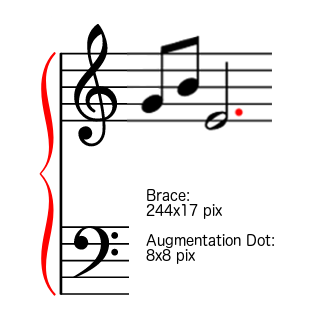}
\caption{The possible size difference of objects in music notation, illustrated by \texttt{brace} and \texttt{augmentationDot}.}
\label{fi:5}
\end{figure}

\begin{figure}[h]
\begin{center}
\includegraphics[width=0.7\linewidth]{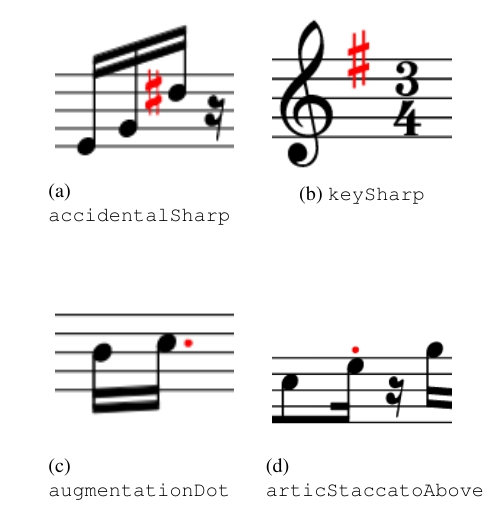}
\end{center}
\caption{Examples for the importance of context for classifying musical symbols: in both rows, the class of otherwise similar looking objects changes depending on the surrounding objects.}
\label{fi:6}
\end{figure}

One of the key challenges this dataset poses upon modeling approaches is the sheer amount of objects on a single image. Two other properties of music notation impose challenges: First, there is a big variability in object size as can be seen for example in Figure \ref{fi:5}. Second, music notation has the special feature that context matters: two objects having identical appearance can belong to a different class depending on the local surroundings (see Figure \ref{fi:6}). To our knowledge there is no other freely available large scale dataset that shares this trait. 

Moreover, datasets like ImageNet are close to being perfectly balanced, with the number of images/objects per class being a constant. This clearly isn't the case with the \emph{DeepScores} dataset, where the most common class contains more than half of the symbols in the dataset, and the top $10$\% of classes contain more than $85$\% of the symbols in the entire dataset. This extremely skewed distribution resembles many real-world cases for example in anomaly detection and industrial quality control.

\subsection{Towards next-generation computer vision}

Classifying the musical symbols in \emph{DeepScores} should nevertheless not be a problem: all these symbols have very clear black and white borders, their shape has limited variability and they are rendered at very high resolution (see Figure \ref{fi:3}). Due to these reasons we assumed that classification on \emph{DeepScores} should be a relatively easy task, given CNNs usually deal well with these kinds of objects. To support this assumption, we fitted a simple residual-CNN \cite{DBLP:conf/cvpr/HeZRS16} with $25$ convolutional layers and about $8$ million trainable parameters. Using the Adam optimizer \cite{DBLP:conf/iclr/Kingma14} with the hyper-parameters proposed by the authors, we reached an accuracy of over $0.98$ in just ten epochs. This shows that classification will indeed not be a big issue and CNNs are able to deal with labels that not only depend on an object but also its surroundings.

Detection, however, is more challenging: we evaluated SSD's and YOLO's fitness for the detection task on \emph{DeepScores} and applied Faster R-CNN - with very little success. We conjecture that one of the main problems is that these region proposal-based systems seem to become computationally overwhelmed for this type of data, due to the sheer number of proposals necessary to find the many small objects.

Both observations - easy classification but challenging detection - lie at the heart of what we think makes \emph{DeepScores} very useful: it offers the challenging scenario of many tiny objects that cannot be approached using current datasets (see Section \ref{sec_comparison}). On the other hand, \emph{DeepScores} is probably the easiest scenario of that kind, because classifying single musical objects is relatively easy and the dataset contains a vast amount of training data. \emph{DeepScores} thus is a prime candidate to develop next generation computer vision methods that scale to many tiny objects on large images: many real-world problems deal with high-resolution images, with images containing hundreds objects and with images containing very small objects in them. This might be automated driving and other robotics use cases, medical applications with full-resolution imaging techniques as data sources, or surveillance tasks e.g. in sports arenas and other public places.

Finally, \emph{DeepScores} will be a valuable source for pre-training models: 
transfer learning has been one of the most important ingredients in the advancement of computer vision. The first step in many computer vision models \cite{Long2015,DBLP:conf/nips/RenHGS15} is to use a deep convolutional neural network pre-trained on ImageNet, and alter it for the task of image segmentation or object detection, or use it on considerably smaller, task-dependent final training sets. \emph{DeepScores} will be of value specifically in the area of OMR, but more generally to allow the development of algorithms that focus on the fine-grained structure of smaller objects while simultaneously being able to scale to many objects of that nature. 

\section{Conclusions}
\label{sec_conclusions}

We have presented the conception and creation of \emph{DeepScores} - the largest publicly and freely available dataset for computer vision applications in terms of image size and contained objects. Compared to other well-known datasets, \emph{DeepScores} has large images (more than four times larger than the average) containing many (one to two orders of magnitude more) very small (down to a few pixels, but varying by several orders of magnitude) objects that change their class belonging depending on the visual context. The dataset is made up of sheets of written music, synthesized from the largest public corpus of MusicXML. It comprises ground truth for the tasks of object classification, semantic segmentation and object detection. 

We have argued that the unique properties of \emph{DeepScores} make the dataset suitable for use in the development of general next generation computer vision methods that are able to work on large images with tiny objects. This ability is crucial for real-world applications like robotics, automated driving, medical image analysis or surveillance, besides OMR. We have motivated that object classification is relatively easy on \emph{DeepScores}, making it therefore the potentially cheapest way to work on a challenging detection task. We thus expect impact on general object detection algorithms. 

A weakness of the \emph{DeepScores} dataset is that all the data is digitally rendered. Linear models (or piecewise linear models like neural networks) have been shown to not generalize well when the distribution of the real-world data is far from the distribution of the dataset the model has been trained on \cite{DBLP:conf/cvpr/TorralbaE11, Szegedy2013}. Future work \emph{on} the dataset will include developing and publishing scripts to perturb the data in order to make it look more like real (scanned) written music, and evaluation of the transfer performance of models trained on \emph{DeepScores}.

Future work \emph{with} the dataset will - besides the general impact predicted above - directly impact OMR: the full potential of deep neural networks is still to be realized on musical scores. 

\chapter{Deep Watershed Detector for Music Object Recognition}

\section{Disclaimer}

The work presented in this chapter is based on the following papers:

\begin{description}  
  \item Lukas Tuggener, \textbf{Ismail Elezi}, J\"urgen Schmidhuber, Thilo Stadelmann; \textit{Deep watershed detector for music object recognition \cite{DBLP:conf/ismir/TuggenerESS18}}; In Proceedings of Conference of the International Society for Music Information Retrieval (ISMIR 2018)
  
  \item Thilo Stadelmann, Mohammadreza Amirian, Ismail Arabaci, Marek Arnold, Gilbert François Duivesteijn, \textbf{Ismail Elezi}, Melanie Geiger, Stefan L\"orwald, Benjamin Bruno Meier, Katharina Rombach, Lukas Tuggener; \textit{Deep Learning in the Wild \cite{DBLP:conf/annpr/StadelmannAAADE18}}; In Proceedings of IAPR TC3 Workshop on Artificial Neural Networks in Pattern Recognition (ANNPR 2018)
  
  \item \textbf{Ismail Elezi}, Lukas Tuggener, Marcello Pelillo, Thilo Stadelmann; \textit{DeepScores and Deep Watershed Detection: current state and open issues \cite{DBLP:journals/corr/abs-1810-05423}}; in in The International Workshop on Reading Music Systems (WoRMS 2018) (ISMIR affiliated).
\end{description}

The contributions of the author are the following:

\begin{description}
  \item[$\bullet$ Writing] code for the dataset preparation.
  \item[$\bullet$ Writing] a part of the paper. \cite{DBLP:conf/ismir/TuggenerESS18}.
  \item[$\bullet$ Writing] a part of the paper \cite{DBLP:conf/annpr/StadelmannAAADE18}.
  \item[$\bullet$ Writing] a considerable part of the paper \cite{DBLP:journals/corr/abs-1810-05423}.
  \item[$\bullet$ Performing] many experiments in all the mentioned datasets. 
  \item[$\bullet$ Improving] the results of DWD (in mAP score) by more than 100\%.
  \item[$\bullet$ Leading] the work on the creation of the scanned dataset.
  \item[$\bullet$ Doing] preliminary investigations on domain transfer.
  
\end{description}

\section{Introduction and Problem Statement}\label{sec:introduction}
The goal of Optical Music Recognition (OMR) is to transform images of printed or handwritten music scores into machine readable form, thereby understanding the semantic meaning of music notation \cite{omr}. It is an important and actively researched area within the music information retrieval community. 
The two main challenges of OMR are: first the accurate detection and classification of music objects in digital images; and second, the reconstruction of valid music in some digital format. This work is focusing solely on the first task, meaning that we recover position and class  (based on the shape only) of every object without inferring any higher level information.

Recent progress in computer vision \cite{Detectron2018} thanks to the adaptation of convolutional neural networks (CNNs) \cite{DBLP:journals/pr/FukushimaM82, DBLP:journals/neco/LeCunBDHHHJ89} provide a solid foundation for the assumption that OMR systems can be drastically improved by using CNNs as well. Initial results of applying deep learning \cite{DBLP:journals/nn/Schmidhuber15} to heavily restricted settings such as staffline removal \cite{DBLP:journals/eswa/GallegoC17}, symbol classification \cite{DBLP:conf/icmla/PachaE17} or end-to-end OMR for monophonic scores \cite{DBLP:conf/ismir/Calvo-ZaragozaV17}, support such expectations. 

In this work, we introduce a novel general object detection method called Deep Watershed Detector (DWD) motivated by the following two hypotheses: a) deep learning can be used to overcome the classical OMR approach of having hand-crafted pipelines of many preprocessing steps \cite{DBLP:journals/ijdar/RebeloCC10} by being able to operate in a fully data-driven fashion; b) deep learning can cope with larger, more complex inputs than simple glyphs, thereby learning to recognize musical symbols in their context. This will disambiguate meanings (e.g., between staccato and augmentation dots) and allow the system to directly detect a complex alphabet. 

DWD operates on full pages of music scores in one pass without any preprocessing besides interline normalization and detects handwritten and digitally rendered music symbols without any restriction on the alphabet of symbols to be detected. We further show that it learns meaningful representation of music notation and achieves state-of-the art detection rates on common symbols. 

\begin{figure*}
\centering
\includegraphics[scale=0.2]{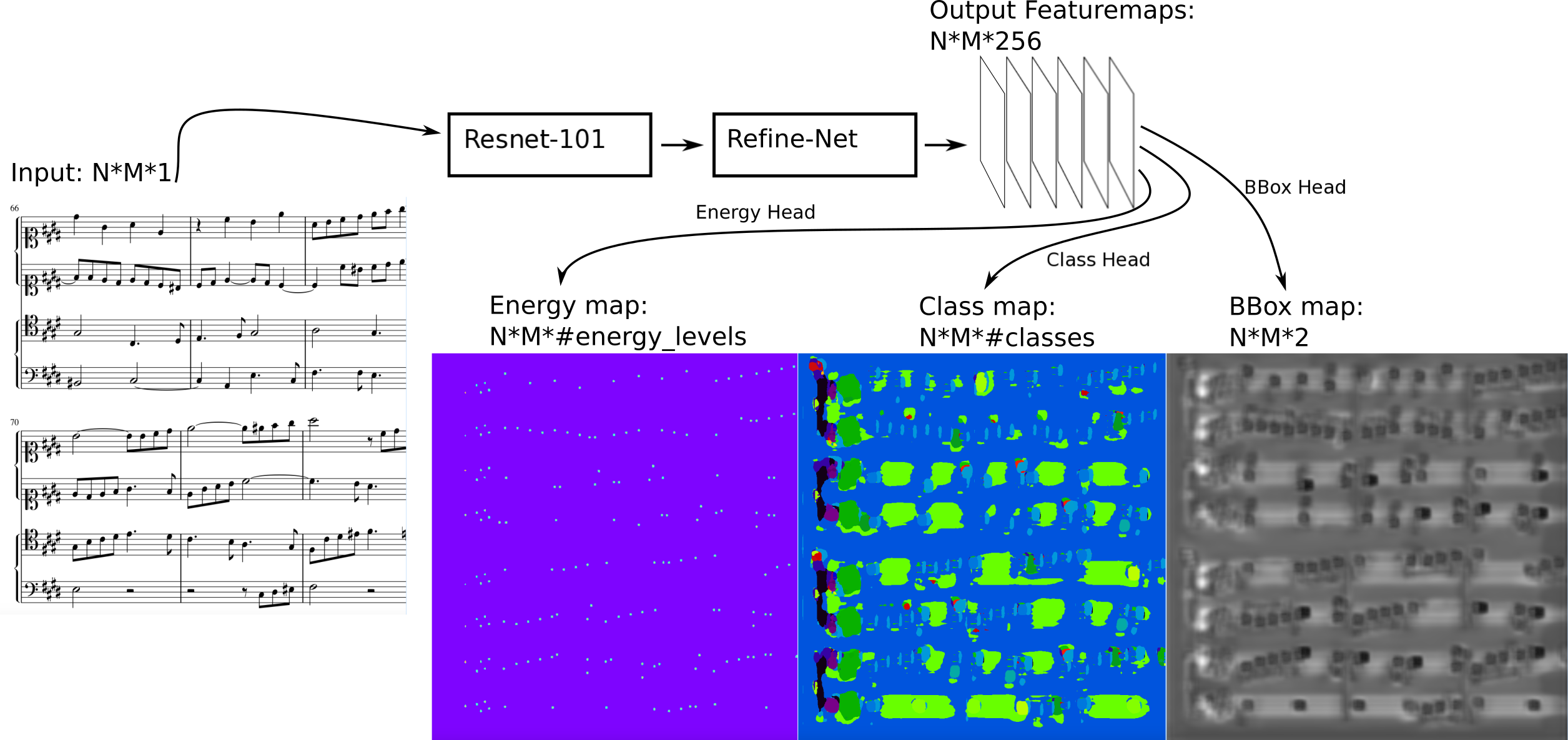}
\caption{Schematic of the Deep Watershed Detector model with three distinct output heads. $N$ and $M$ are the height and width of the input image, $\mathrm{\#classes}$ denotes the number of symbols and $\mathrm{\#energy\_levels}$ is a hyperparameter of the system.} 
\label{fig:network}
\end{figure*}

\section{Related Work}\label{sec:related}
The visual detection and recognition of objects is one of the most central problems in the field of computer vision. With the recent developments of CNNs, many competing CNN-based approaches have been proposed to solve the problem. R-CNNs \cite{DBLP:conf/cvpr/GirshickDDM14}, and in particular their successors \cite{DBLP:conf/nips/RenHGS15}, are generally considered to be state-of-the-art models in object recognition, and many developed recognition systems are based on R-CNN. On the other hand, researchers have also proposed models which are tailored towards computational efficiency instead of detection accuracy. YOLO systems \cite{DBLP:conf/cvpr/RedmonF17} and Single-Shot Detectors \cite{DBLP:conf/eccv/LiuAESRFB16} while slightly compromising on accuracy, are significantly faster than R-CNN models, and can even achieve super real-time performance.

A common aspect of the above-mentioned methods is that they are specifically developed to work on cases where the images are relatively small, and where images contain a small number of relatively large objects \cite{DBLP:journals/ijcv/EveringhamGWWZ10, DBLP:conf/eccv/LinMBHPRDZ14}. On the contrary, musical sheets usually have high-resolution, and contain a very large number of very small objects, making the mentioned methods not suitable for the task.

The watershed transform is a well understood method that has been applied to segmentation for decades \cite{beucher1992watershed}. 
Bai and Urtasun \cite{DBLP:conf/cvpr/BaiU17} were first to propose combining the strengths of deep learning with the power of this classical method. They proposed to directly learn the energy (in our application the distance to an object center) for the watershed transform such that all dividing ridges are at the same height. As a consequence, the components can be extracted by a cut at a single energy level without leading to over-segmentation. The model has been shown to achieve state of the art performance on object segmentation.

For the most part, OMR detectors have been rule-based systems working well only within a hard set of constraints \cite{DBLP:journals/ijdar/RebeloCC10}. Typically, they require domain knowledge, and work well only on simple typeset music scores with a known music font, and a relatively small number of classes \cite{DBLP:journals/ejasp/RossantB07}. When faced with low-quality images, complex or even handwritten scores \cite{DBLP:conf/icfhr/BaroRF16}, the performance of these models quickly degrades, to some degree because errors propagate from one step to another \cite{DBLP:conf/icmla/PachaE17}. Additionally, it is not clear what to do when the classes change, and in many cases, this requires building the new model from scratch.

In response to the above mentioned issues some deep learning based, data driven approaches have been developed. Hajic and Pecina \cite{DBLP:journals/corr/abs-1708-01806} proposed an adaptation of Faster R-CNN with a custom region proposal mechanism based on the morphological skeleton to accurately detect noteheads, while Choi et al. \cite{DBLP:conf/icdar/ChoiCRZ17} were able to detect accidentals in dense piano scores
with high accuracy, given previously detected noteheads, that are being used as input-features to the network. A big limitation of both approaches is that the experiments have been done only on a tiny vocabulary of the musical symbols, and therefore their scalability remains an open question.

To our knowledge, the best results so far has been reported in the work of Pacha and Choi \cite{DBLP:conf/iwdas/Pacha2018} where they explored many models on the \emph{MUSCIMA++ }\cite{DBLP:conf/icdar/HajicP17b} dataset of handwritten music notation. They got the best results with a Faster R-CNN model, achieving an impressive score on the standard mAP metric. A serious limitation of that work is that the system was not designed in an end-to-end fashion and needs heavy pre- and post-processing. In particular, they cropped the images in a context-sensitive way, by cutting images first vertically and then horizontally, such that each image contains exactly one staff and has a width-to-height-ratio
of no more than $2:$1, with about $15\%$ horizontal overlap to adjacent slices. In practice, this means that all objects significantly exceeding the size of such a cropped region will neither  appear in the training nor testing data, as only annotations that have an intersection-over-area of $0.8$ or higher between the object and the cropped region are considered part of the ground truth. Furthermore, all the intermediate results must be combined to one concise final prediction, which is a non-trivial task.

\section{Deep Watershed Detection}\label{sec:methods}

\begin{figure}
\centering

\includegraphics[width=0.7\textwidth]{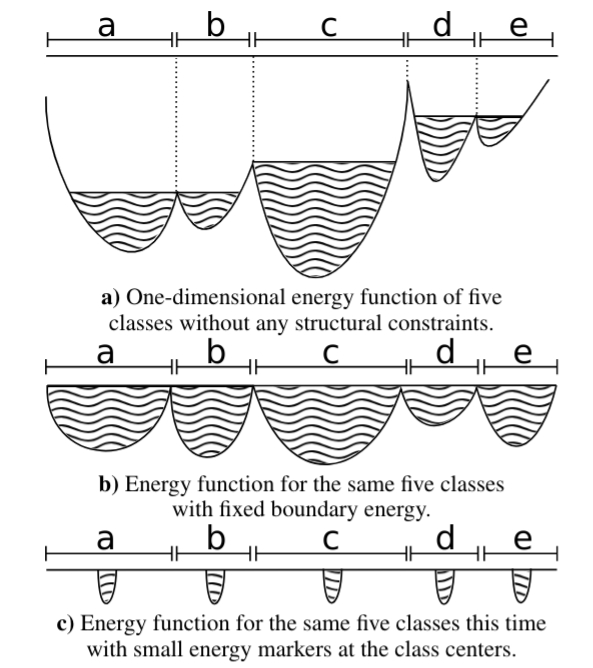}
\caption{Illustration of the watershed transform applied to different one-dimensional functions.}
\label{fig:watershed_energy}
\end{figure}

In this section we present the Deep Watershed Detector (DWD) as a novel object detection system, built on the idea of the deep watershed transform \cite{DBLP:conf/cvpr/BaiU17}. The watershed transform \cite{beucher1992watershed} is a mathematically well understood method with a simple core idea that can be applied to any topological surface. The algorithm starts filling up the surface from all the local minima, with all the resulting basins corresponding to connected regions. When applied to image gradients, the basins correspond to homogeneous regions of said image (see Fig. \ref{fig:watershed_energy}a). One key drawback of the watershed transform is its tendency to over segment. This issue can be addressed by using the deep watershed transform. It combines the classical method with deep learning by training a deep neural network to create an energy surface based on an input image. This has the advantage that one can design the energy surface to have certain properties. When designed in a way that all segmentation boundaries have energy zero, the watershed transform is reduced to a simple cutoff at a fixed energy level (see Fig. \ref{fig:watershed_energy}b). An objectness energy of this fashion has been used by Bai and Urtasun for instance segmentation \cite{DBLP:conf/cvpr/BaiU17}. Since we want to do object detection, we further simplify the desired energy surface to having small conical energy peaks of radius $n$ pixels at the center of each object and be zero everywhere else (see Fig. \ref{fig:watershed_energy}c).

More formally, we define our energy surface (or: energy map) $M^{e}$ as follows:

\begin{equation}\label{energy_map}
M^{e}_{(i,j)} = max \begin{cases}
    \ \underset{c \in C}{\mathrm{argmax}}[E_{max}\cdot(1-\frac{\sqrt{(i-c_i)^2+(j-c_j)^2}}{r})]\\
    0 
    \end{cases}
\end{equation}

where $M^{e}_{(i,j)}$ is the value of $M^e$ at position $(i,j)$, $C$ is the set of all object centers and $c_i, c_j$ are the center coordinates of a given center $c$. $E_{max}$ corresponds to the maximum energy and $r$ is the radius of the center marking.

At first glance this definition might lead to the misinterpretation that object centers that are closer together than $r$ cannot be disambiguated using the watershed transform on $M^e$. This is not the case since we can cut the energy map at any given energy level between $1$ and $E_{max}$. However, using this method it is not possible to detect multiple bounding boxes that share the exact same center.

\subsection{Retrieving Object Centers}
After computing an estimate $\hat M^e$ of the energy map, we retrieve the coordinates of detected objects by the following steps:
\begin{enumerate}
\item Cut the energy map at a certain fixed energy level and then binarize the result.
\item Label the resulting connected components, using the two-pass algorithm \cite{Wu:2009:OTC:1529867.1529869}. Every component receives a label $l$ in $1...n$, for every component $o^l$ we define $O^l_{ind}$ as the set of all tuples $(i,j)$ for which the pixel with coordinates $j$ and $i$ is part of $o^l.$ %by thilo: what is n - number of classes? Why O_ind (what's the intuition behind the name)?
\item The center $\hat c^l$ of any component $o^l$ is given by its center of gravity: 
\begin{equation}\label{center}
\hat c^l = o^l_{center} = \vert O^l_{ind} \vert^{-1}  \cdot \sum_{(i,j) \in O^l_{ind}}{(i,j)}
\end{equation}
\end{enumerate}
We use these component centers $\hat c$ as estimates for the object centers $c$.

\subsection{Object Class and Bounding Box}
In order to recover bounding boxes we do not only need the object centers, but also the object classes and bounding box dimensions. To achieve this we output two additional maps $M^c$ and $M^b$ as predictions of our network. $M^c$ is defined as:

\begin{equation}\label{Mc}
M^{c}_{(i,j)} = \begin{cases}
    \Lambda_{(i,j)},& \text{if } M^{e}_{(i,j)} > 0\\
    \Lambda_{background} ,  & \text{otherwise}
\end{cases}
\end{equation}

where $\Lambda_{backgroud}$ is the class label indicating background and $\Lambda_{(i,j)}$ is the class label associated with the center $c$ that is closest to $(i,j)$. We define our estimate for the class of component $o^l$ by a majority vote of all values $\hat M^{c}_{(i,j)}$ for all $(i,j) \in O^l_{ind}$, where $\hat M^c$ is the estimate of $M^c$. Finally, we define the bounding box map $M^{b}$ as follows:
\begin{equation}\label{bbox_map}
M^{b}_{(i,j)} = \begin{cases}
    (y^{l}, x^{l}),& \text{if } M^{e}_{(i,j)} > 0\\
    (0,0) ,  & \text{otherwise}
\end{cases}
\end{equation}
where $y^{l}$ and $x^{l}$ are the width and height of the bounding box for component $o^l$. Based on this we define our bounding box estimation as the average of all estimations for label $l$: %by thilo: what is l: a class (and n then the number of classes)? Or is n the number of objects on a given image?
\begin{equation}\label{bbox_estimation}
(\hat y^l, \hat x^l) = \vert O^l_{ind} \vert^{-1} \cdot\sum_{(i,j) \in O^l_{ind}}{\hat M^b_{(i,j)}}
\end{equation}

\begin{figure*}
\centering

\includegraphics[scale=0.35]{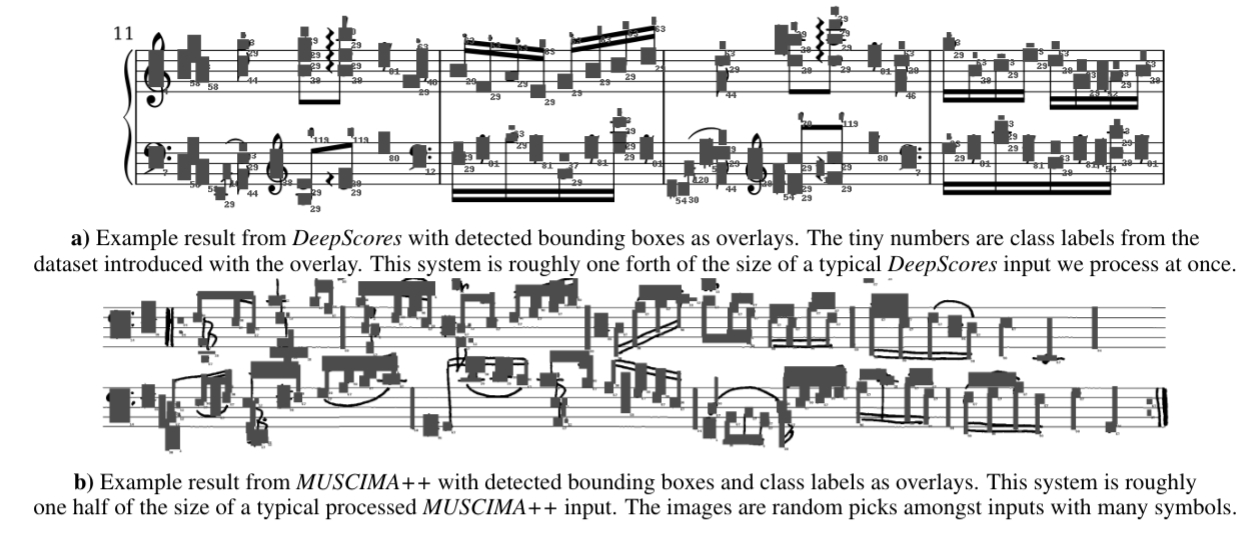}
\caption{Detection results for  \emph{DeepScores} and \emph{MUSCIMA++} examples, drawn on crops from corresponding input images.}
\label{fig:ex_detection}
\end{figure*}

\subsection{Network Architecture and Losses}\label{subsec:architecture}
As mentioned above we use a deep neural network to predict the dense output maps $M^e$, $M^c$ and $M^b$ (see Fig. \ref{fig:network}). The base neural network for this prediction can be any fully convolutional network with the same input and output dimensions. We use a ResNet-101 \cite{DBLP:conf/cvpr/HeZRS16} (a special case of a Highway Net \cite{greff2015nips}) in conjunction with the elaborate RefineNet \cite{DBLP:conf/cvpr/LinMSR17} upsampling architecture. For the estimators defined above it is crucial to have the highest spacial prediction resolution possible. Our network has three output layers, all of which are an $1$ by $1$ convolution applied to the last feature map of the RefineNet.

\subsubsection{Energy prediction} We predict a quantized and one-hot encoded version of $M^e$, called $M^{eo}$, by applying a 1 by 1 convolution of depth $E_{max}$ to the last feature map of the base network. The loss of the prediction $\hat M^{eo}$, $loss^e$, is defined as the cross-entropy between $M^{eo}$ and $\hat M^{eo}$.

\subsubsection{Class prediction} We again use the corresponding one-hot encoded version $M^{co}$ and predict it using an $1$ by $1$ convolution, with the depth equal to the number of classes, on the last feature map of the base network. The cross-entropy $loss^c$ is calculated between $M^{co}$ and $\hat M^{co}$. Since it is not the goal of this prediction to distinguish between foreground and background, all the loss stemming from locations with $M^e=0$ will get masked out.

\subsubsection{Bounding box prediction} $M^b$ is predicted in its initial form using an $1$ by $1$ convolution of depth $2$ on the last feature map of the base network. The bounding box loss $loss^b$ is the mean-squared difference between $M^b$ and $\hat M^b$. For $loss^b$, the components stemming from background locations will be masked out analogous to $loss^c$.

\subsubsection{Combined prediction}
We want to jointly train in all tasks, therefore we define a total loss $loss^{tot}$ as: \begin{equation}\label{loss_tot}
loss^{tot} = w_1 * \frac{loss^e}{v^e} + w_2 * \frac{loss^c}{v^c} + w_3 * \frac{loss^b}{v^b}
\end{equation}
where the $v^.$ are running means of the corresponding losses and the scalars $w_.$ are hyper-parameters of the DWD network. We purposefully use very short extraction heads of one convolutional layer; by doing so we force the base network to do all three tasks simultaneously. We expect this leads to the base network learning a meaningful representation of music notation, from which it can extract the solutions of the three above defined tasks. %by thilo: what is the "detection head" (I have a solid idea, but never heared the term before)?

\section{Experiments and Results}\label{sec:experiments-dwd}

\subsection{Used Datasets}\label{subsec:datasets}

For our experiments we use two datasets: \emph{DeepScores} \cite{DBLP:conf/icpr/TuggenerESPS18} and \emph{MUSCIMA++} \cite{DBLP:conf/icdar/HajicP17b}.

\emph{DeepScores} is currently the largest publicly available dataset of musical sheets with ground truth for various machine learning tasks, consisting of high-quality pages of written music, rendered at $400$ dots per inch. The dataset has $300,000$ full pages as images, containing tens of millions of objects, separated in $123$ classes. We randomly split the set into training and testing, using $200k$ images for training and $50k$ images each for testing and validation. The dataset being so large allows efficient training of large convolutional neural networks, in addition to being suitable for transfer learning \cite{DBLP:conf/nips/YosinskiCBL14}.

\emph{MUSCIMA++} is a dataset of handwritten music notation for musical symbol detection. It contains $91,255$ symbols spread into $140$ pages, consisting of both notation primitives and higher-level notation objects, such as key signatures or time signatures. It features 105 object classes. There are $23,352$ notes in the dataset, of which $21,356$ have a full notehead, $1,648$ have an empty notehead, and $348$ are grace notes. We randomly split the dataset into training, validation, and testing, with the training set consisting of $110$ pages, while validation and testing each consists of $15$ pages.

\subsection{Network Training and Experimental Setup}
We pre-train our network in two stages in order to achieve reasonable results. First we train the ResNet on music symbol classification using the \emph{DeepScores} classification dataset \cite{DBLP:conf/icpr/TuggenerESPS18}. Then, we train the ResNet and RefineNet jointly on semantic segmentation data also available from \emph{DeepScores}. After this pre-training stage we are able to use the network on the tasks defined above in Sec. \ref{subsec:architecture}.

Since music notation is composed of hierarchically organized sub-symbols, there does not exist a canonical way to define a set of atomic symbols to be detected (e.g., individual numbers in time signatures vs. complete time signatures). We address this issue using a fully data-driven approach by detecting atomic classes as they are provided by the two datasets.

We rescale every input image to the desired interline value. We use $10$ pixels for \emph{DeepScores} and $20$ pixels for \emph{MUSCIMA++}. Other than that we apply no preprocessing. We do not define a subset of target objects for our experiments, but attempt to detect all classes for which there is ground truth available. We always feed single images to the network, i.e. we only use batch size = $1$. During training we crop the full page input (and the ground truth) to patches of $960$ by $960$ pixels using randomized coordinates. This serves two purposes: it saves GPU memory and performs efficient data augmentation. This way the network never sees the exact same input twice, even if we train for many epochs. For all of the results described below we train individually on $loss^e$, $loss^c$ and $loss^b$ and then refine the training using $loss^{tot}$. It turns out that the prediction of $M^e$ is the most fragile to effects introduced by training on the other losses, therefore we retrain on $loss^e$ again after training on the individual losses in the order defined above, before moving on to $loss^{tot}$. All the training is done using the RMSProp optimizer \cite{rmsprop} with a learning rate of $0.001$ and a decay rate of $0.995$.

\begin{table}
 \begin{center}
\begin{footnotesize}
\begin{tabular}{|rl|rl|}
\toprule
             Class &    AP@$\frac{1}{2}$ & Class &      AP@$\frac{1}{4}$ \\
\midrule
          rest16th &  0.8773 &            tuplet6 &  0.9252 \\
     noteheadBlack &  0.8619 &           keySharp &  0.9240 \\
          keySharp &  0.8185 &           rest16th &  0.9233 \\
           tuplet6 &  0.8028 &      noteheadBlack &  0.9200 \\
       restQuarter &  0.7942 &    accidentalSharp &  0.8897 \\
           rest8th &  0.7803 &           rest32nd &  0.8658 \\
      noteheadHalf &  0.7474 &       noteheadHalf &  0.8593 \\
         flag8thUp &  0.7325 &            rest8th &  0.8544 \\
       flag8thDown &  0.6634 &        restQuarter &  0.8462 \\
   accidentalSharp &  0.6626 &  accidentalNatural &  0.8417 \\
 accidentalNatural &  0.6559 &          flag8thUp &  0.8279 \\
           tuplet3 &  0.6298 &            keyFlat &  0.8134 \\
     noteheadWhole &  0.6265 &        flag8thDown &  0.7917 \\
         dynamicMF &  0.5563 &            tuplet3 &  0.7601 \\
          rest32nd &  0.5420 &      noteheadWhole &  0.7523 \\
        flag16thUp &  0.5320 &              fClef &  0.7184 \\
         restWhole &  0.5180 &          restWhole &  0.7183 \\
          timeSig8 &  0.5180 &       dynamicPiano &  0.7069 \\
    accidentalFlat &  0.4949 &     accidentalFlat &  0.6759 \\
           keyFlat &  0.4685 &         flag16thUp &  0.6621 \\
\bottomrule
\end{tabular}
\end{footnotesize}
\end{center}
 \caption{AP with overlap $0.5$ and overlap $0.25$ for the twenty best detected classes of the \emph{DeepScores} dataset.}
 \label{tab:ap_deepscores}
\end{table}

\begin{table}
 \begin{center}
\begin{footnotesize}
\begin{tabular}{|rl|rl|}
\toprule
          Class &  AP@$\frac{1}{2}$ & Class &  AP@$\frac{1}{4}$ \\
\midrule
      half-rest &  0.8981 &  whole-rest &  0.9762 \\
           flat &  0.8752 &     ledger-line &  0.9163 \\
        natural &  0.8531 &       half-rest &  0.8981 \\
     whole-rest &  0.8226 &            flat &  0.8752 \\
  notehead-full &  0.8044 &         natural &  0.8711 \\
          sharp &  0.8033 &            stem &  0.8377 \\
 notehead-empty &  0.7475 &    staccato-dot &  0.8302 \\
           stem &  0.7426 &   notehead-full &  0.8298 \\
   quarter-rest &  0.6699 &           sharp &  0.8121 \\
       8th-rest &  0.6432 &          tenuto &  0.7903 \\
         f-clef &  0.6395 &  notehead-empty &  0.7475 \\
      numeral-4 &  0.6391 &    duration-dot &  0.7285 \\
       letter-c &  0.6313 &       numeral-4 &  0.7158 \\
       letter-c &  0.6313 &        8th-flag &  0.7055 \\
       8th-flag &  0.6051 &    quarter-rest &  0.6849 \\
           slur &  0.5699 &        letter-c &  0.6643 \\
           beam &  0.5188 &        letter-c &  0.6643 \\
 time-signature &  0.4940 &        8th-rest &  0.6432 \\
   staccato-dot &  0.4793 &            beam &  0.6412 \\
       letter-o &  0.4793 &          f-clef &  0.6395 \\
\bottomrule
\end{tabular}
\end{footnotesize}
\end{center}
 \caption{AP with overlap $0.5$ and overlap $0.25$ for the twenty best detected classes from \emph{MUSCIMA++}.}
 \label{tab:ap_muscima}
\end{table}

Since our design is invariant to how many objects are present on the input (as long as their centers do not overlap) and we want to obtain bounding boxes for full pages at once, we feed whole pages to the network at inference time. The maximum input size is only bounded by the memory of the GPU. For typical pieces of sheet music this is not an issue, but pieces that use very small interline values (e.g. pieces written for conductors) result in very large inputs due to the interline normalization. At about $10.5$ million pixels even a Tesla P40 with $24$ gigabytes runs out of memory. 

\subsection{Initial Results}\label{subsec:results}

Table \ref{tab:ap_deepscores} shows the average precision (AP) for the twenty best detected classes with an overlap of the detected bounding box and ground truth of $50\%$ and $25\%$, respectively. We observe that in both cases there are common symbol classes that get detected very well, but there is also a steep fall off. The detection rate outside the top twenty continues to drop and is almost zero for most of the rare classes. We further observe that there is a significant performance gain for the lower overlap threshold, indicating that the bounding-box regression is not very accurate. 

Fig. \ref{fig:ex_detection} shows an example detection for qualitative analysis. It confirms the conclusions drawn above. The rarest symbol present, an arpeggio, is not detected at all, while the bounding boxes are sometimes inaccurate, especially for large objects (note that stems, bar-lines and beams are not part of the \emph{DeepScores} alphabet and hence do not constitute missed detections). On the other hand, staccato dots are detected very well. This is surprising since they are typically hard to detect due to their small size and the context-dependent interpretation of the symbol shape (compare the dots in dotted notes or F-clefs). We attribute this to the opportunity of detecting objects in context, enabled by training on larger parts of full raw pages of sheet music in contrast to the classical processing of tiny, pre-processed image patches or glyphs. 

The results for the experiments on \emph{MUSCIMA++} in Tab. \ref{tab:ap_muscima} and Fig. \ref{fig:ex_detection}b show a very similar outcome. This is intriguing because it suggests that the difficulty in detecting digitally rendered and handwritten scores might be smaller than anticipated. We attribute this to the fully data-driven approach enabled by deep learning instead of hand-crafted rules for handling individual symbols. It is worth noting that ledger-lines are detected with very high performance (see AP@$\frac{1}{4}$). This explains the relatively poor detection of note-heads on \emph{MUSCIMA++}, since they tend to overlap.

\begin{figure}
\begin{center}
\includegraphics[scale=0.2]{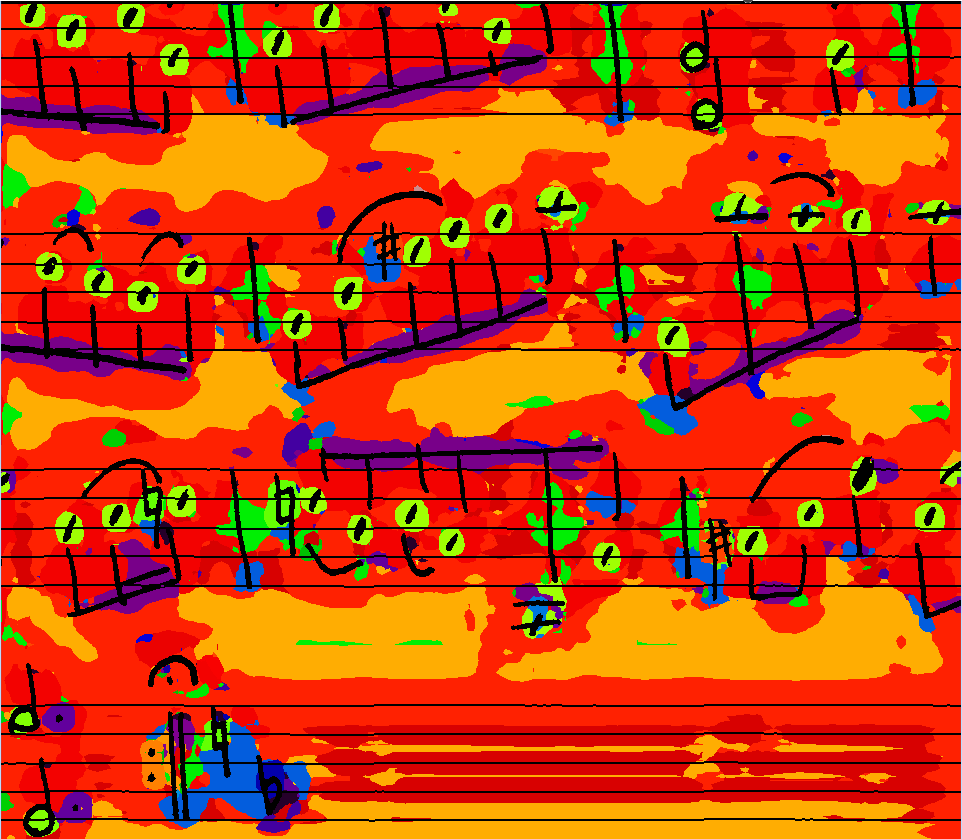}
\caption{Estimate of a class map $\hat M^c$ for every input pixel with the corresponding \emph{MUSCIMA++} input overlayed. }
\label{fig:ex_overlayed}
\end{center}
\end{figure}

Fig. \ref{fig:ex_overlayed} shows an estimate for a class map with its corresponding input overlayed. Each color corresponds to one class. This figure proofs that the network is learning a sensible representation of music notation: even though it is only trained to mark the centers of each object with the correct colors, it learns a primitive segmentation mask. This is best illustrated by the (purple) segmentation of the beams.

\section{Deep Watershed Detector in the Wild}

We highlight four typical issues when applying deep learning techniques to practical OMR: (a) the absence of a comprehensive dataset; (b) the extreme class imbalance present in written music with respect to symbols; (c) the issues of state-of-the-art object detectors with music notation (many tiny and compound symbols on large images); and (d) the transfer from synthetic data to real world examples. 

\subsection{Dealing with imbalanced data} While typical academic training datasets are nicely balanced \cite{DBLP:conf/cvpr/DengDSLL009,DBLP:journals/ijcv/EveringhamGWWZ10}, this is rarely the case in datasets sourced from real world tasks. Music notation (and therefore \emph{DeepScores}) shows an extreme class imbalance (see Figure \ref{fi:3}). For example, the most common class (note head black) contains more than $55$\% of the symbols in the entire dataset, and the top $10$ classes contain more than $85$\% of the symbols. At the other extreme, there is a class which is present only once in the entire dataset, making its detection by pattern recognition methods nearly impossible (a ``black swan'' is no pattern). However, symbols that are rare are often of high importance in the specific pieces of music where they appear, so simply ignoring the rare symbols in the training data is not an option. A common way to address such  imbalance is the use of a weighted loss function. 

This is not enough in our case: first, the imbalance is so extreme that naively reweighing loss components leads to numerical instability; second, the signal of these rare symbols is so sparse that it will get lost in the noise of the stochastic gradient descent method \cite{DBLP:conf/ismir/TuggenerESS18}, as many symbols will only be present in a tiny fraction of the mini batches. Our current answer to this problem is \emph{data synthesis} \cite{ng2018yearning}, using a three-fold approach to synthesize image patches with rare symbols: (a) we locate rare symbols which are present at least $300$ times in the dataset, and crop the parts containing those symbols including their local context (other symbols, staff lines etc.); (b) for rarer symbols, we locate a semantically similar but more common symbol in the dataset (based on some expert-devised notion of symbol similarity), replace this common symbol with the rare symbol and add the resulting page to the dataset. This way, synthesized sheets still have semantic sense, and the network can learn from syntactically correct context symbols. We then crop patches around the rare symbols similar to the previous approach; (c) for rare symbols without similar common symbols, we automatically ``compose''  music containing those symbols. 

Then, during training, we augment each input page in a mini batch with $12$ randomly selected synthesized crops of rare symbols (of size $130 \times 80$ pixels) by putting them in the margins at the top of the page. This way, the neural network (on expectation) does not need to wait for more than $10$ iterations to see every class which is present in the dataset. Preliminary results show improvement, though more investigation is needed: overfitting on extreme rare symbols is still likely, and questions remain regarding how to integrate the concept of patches (in the margins) with the idea of a full page classifier that considers all context.

\subsection{Generalizing to real-world data} The basic assumption in machine learning for training and test data to stem from the same distribution is often violated in field applications. In the present case, domain adaptation is crucial: our training set consists of synthetic sheets created by LilyPond scripts \cite{DBLP:conf/icpr/TuggenerESPS18}, while the final product will work on scans or photographs of printed sheet music. These test pictures can have a wide variety of impairments, such as bad printer quality, torn or stained paper etc. While some work has been published on the topic of \emph{domain transfer} \cite{DBLP:conf/iccv/GebruHF17}, the results are non-satisfactory. The core idea to address this problem here is transfer learning \cite{DBLP:conf/nips/YosinskiCBL14}: the neural network shall learn the core task of the full complexity of music notation from the synthetic dataset (symbols in context due to full page input), and use a much smaller dataset to adapt to the real world distributions of lighting, printing and defect.

\begin{figure}[t]
  \centering
  \includegraphics[width=0.9\textwidth]{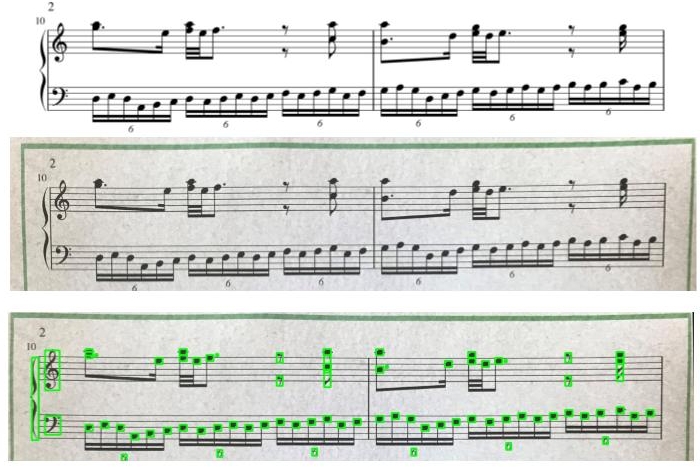}
  \caption{Top: part of a synthesized image from \emph{DeepScores}; middle: the same part, printed on old paper and photographed using a cell phone; bottom: the same image, automatically retrofitted (based on the dark green lines) to the original image coordinates for ground truth matching (ground truth overlayed in neon green boxes).}
  \label{fig:realworld}
\end{figure}

We construct this post-training dataset by carefully choosing several hundred representative musical sheets, printing them with different types of printers on different types of paper, and finally scanning or photographing them. We then use the \texttt{BFMatcher} function from OpenCV to align these images with the original musical sheets to use all the ground truth annotation of the original musical sheet for the real-world images (see Figure \ref{fig:realworld}). This way, we get annotated real-looking images ``for free'' that have much closer statistics to real-world images than images from \emph{DeepScores}. With careful tuning of the hyperparameters (especially the regularization coefficient), we get promising - but not perfect - results during the inference stage.

\vspace{1cm}

\section{Improvements on the dataset and the detector}

\subsection{Shortcomings of the initial release}
At its initial release, \emph{DeepScores} had two main weaknesses: first, it was fully geared towards our application in conjunction with Audiveris; many common symbols that were not interesting in that context have been omitted, which severely limited the usability of \emph{DeepScores} in other contexts. Second, \emph{DeepScores} consist only of synthetically rendered music sheets, since labelling hundreds of thousands of music sheets by hand is prohibitively expensive. However, the common use case for OMR is scans or even photos of music sheets. This discrepancy can lead to severe performance drops between model training and actual use.

\subsection{Enhanced character set}
In an effort to make \emph{DeepScores} more universally usable we created a new version---called \emph{DeepScores-extended}---containing annotations for a far greater number of symbols. According to our knowledge and discussions with other members of the community, no crucial symbols are missing from the \emph{DeepScores-extended} annotations. The full list of supported symbols is available online\footnote{$tuggeluk.github.io/deepscores\_syms\_list$}.

\subsection{Richer musical information}
\begin{figure}[ht]
	\begin{center}
    \includegraphics[width=0.8\textwidth]{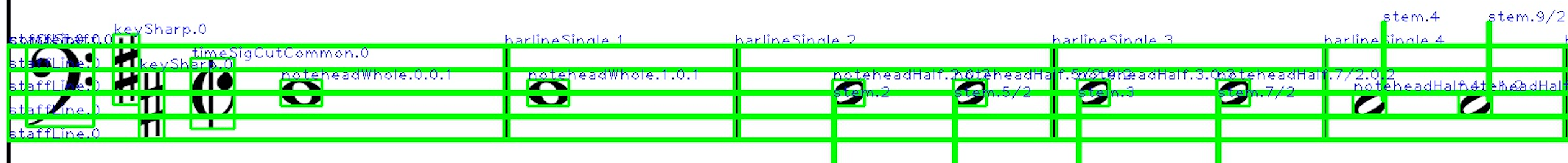}
    \caption{Small piece of music notation with DeepScores-extended annotations overlayed. The naming is either classname.onset or classname.onset.relativecoordinate.duration, depending on availability.}
    \label{fig:annotations_extended}
    \end{center}
    \vspace{-0.8cm}
\end{figure}

While the interest of the authors lies in the detection of musical symbols, this task is not the full problem of OMR. The reconstruction of semantically valid music from detected symbols is at least as challenging as the detection. To enable research focused on reconstructing higher-level information, we have added additional information to the \emph{DeepScores} annotations. Every labeled object now has an \emph{onset} tag that tells the start beat of the the given object. All noteheads additionally have their relative position on the staff as well as their duration in their annotation (see Figure \ref{fig:annotations_extended}).
%Todo by thilo: are the updates available online? please state so and give a link (probably earlier, together with footnote 1)

\subsection{Planned improvements}
\begin{figure}[ht]
	\begin{center}
    \includegraphics[width=0.9\textwidth]{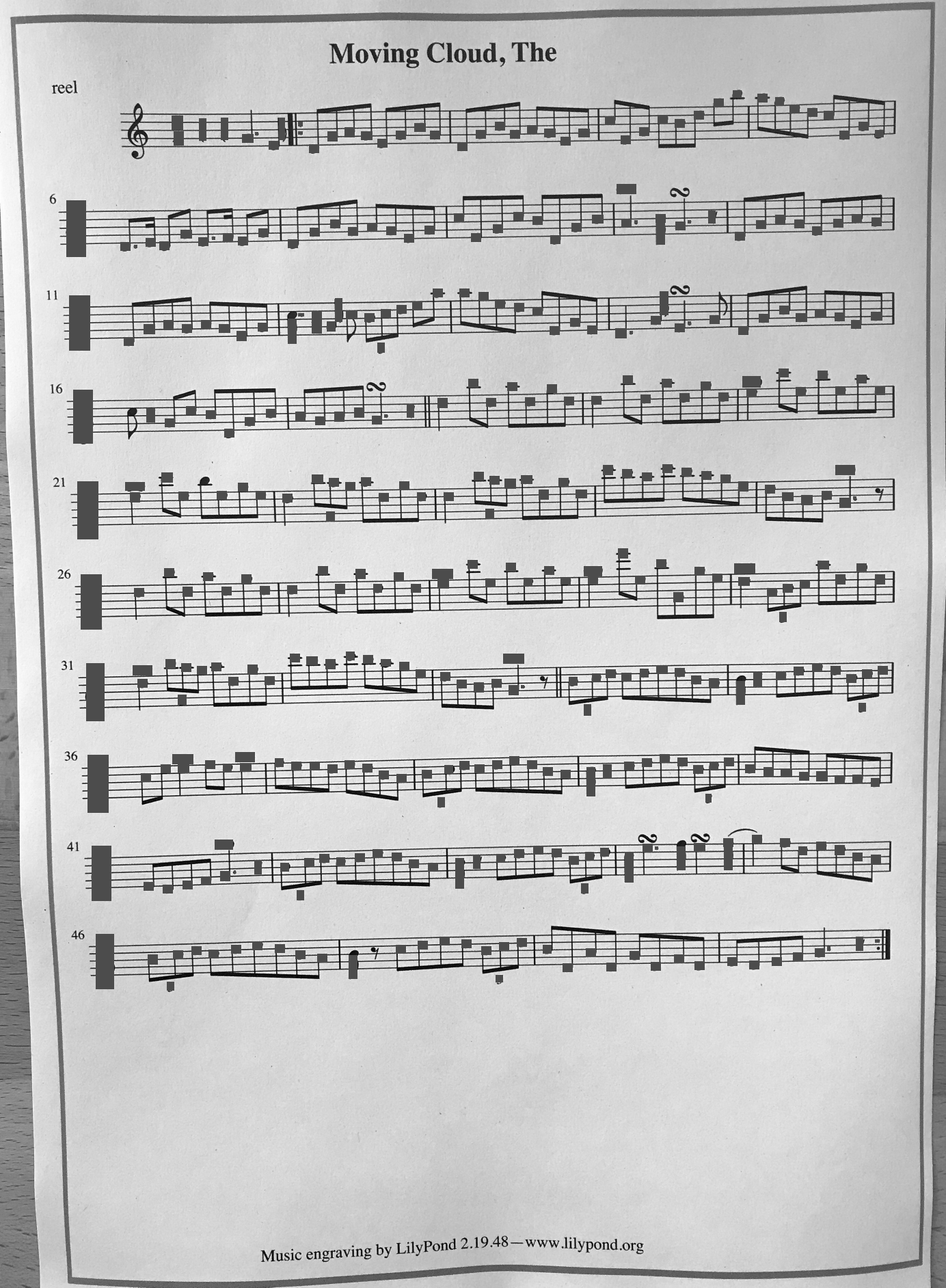}
    \caption{Preliminary results of our model (grey boxes) on a photo of a printed sheet. While not perfect (for example, our model misses the clef in the first row), they already look promising.}
    \label{fig:data_scanned}
    \end{center}
    \vspace{-0.8cm}
\end{figure}

A drawback of the \emph{DeepScores} dataset is that it is synthetic. We are currently working on a much smaller dataset, meant for transfer-learning, that consists of pages originally taken from \emph{DeepScores} that are printed and then digitized again. Then, through a global centering and orientation alignment of the scan, the original annotations are made valid again for the scanned version. We use different printers, scanners, cell-phone cameras, and paper qualities to make the noise introduced by this process resemble the real world use case as much as possible. Naively training a Deep Watershed Detector on this new dataset, we observed that the detector was unable to find anything on the testing set despite that the loss function converged. This led us to believe that severe overfitting is going on, and we were able to get promising results by simply adding l2-regularization and performing more careful training (see Figure \ref{fig:data_scanned} for a qualitative result of the detector on the new dataset).

\section{Further Research on Deep Watershed Detection}
\subsection{Augmenting inputs}

\emph{DeepScores}, unlike many academic datasets, is extremely unbalanced. In fact, the most common class (notehead black) contains more symbols than the rest of the classes combined, while the top $10$ classes contain more than $85$\% of the symbols. However, some of the rare symbols are important and simply dismissing them might lead to semantic problems during the reconstruction of valid music in some digital format. Initially, we tried to solve the problem by using a weighted loss function which penalizes more severely the mistakes on the rare symbols, but to no avail. In \cite{DBLP:conf/ismir/TuggenerESS18} we conjecture that the inbalance is so extreme that simply weighting the loss function leads to numerical instability, while at the same time the signal from these rare symbols is so sparse that it will get lost in the noise of stochastic gradient descent during the training: many symbols will be present only in a tiny fraction of mini batches. Both of these problems do not get solved by a weighted loss function. 

\begin{figure}[t]
	\begin{center}
    \includegraphics[width=0.9\textwidth]{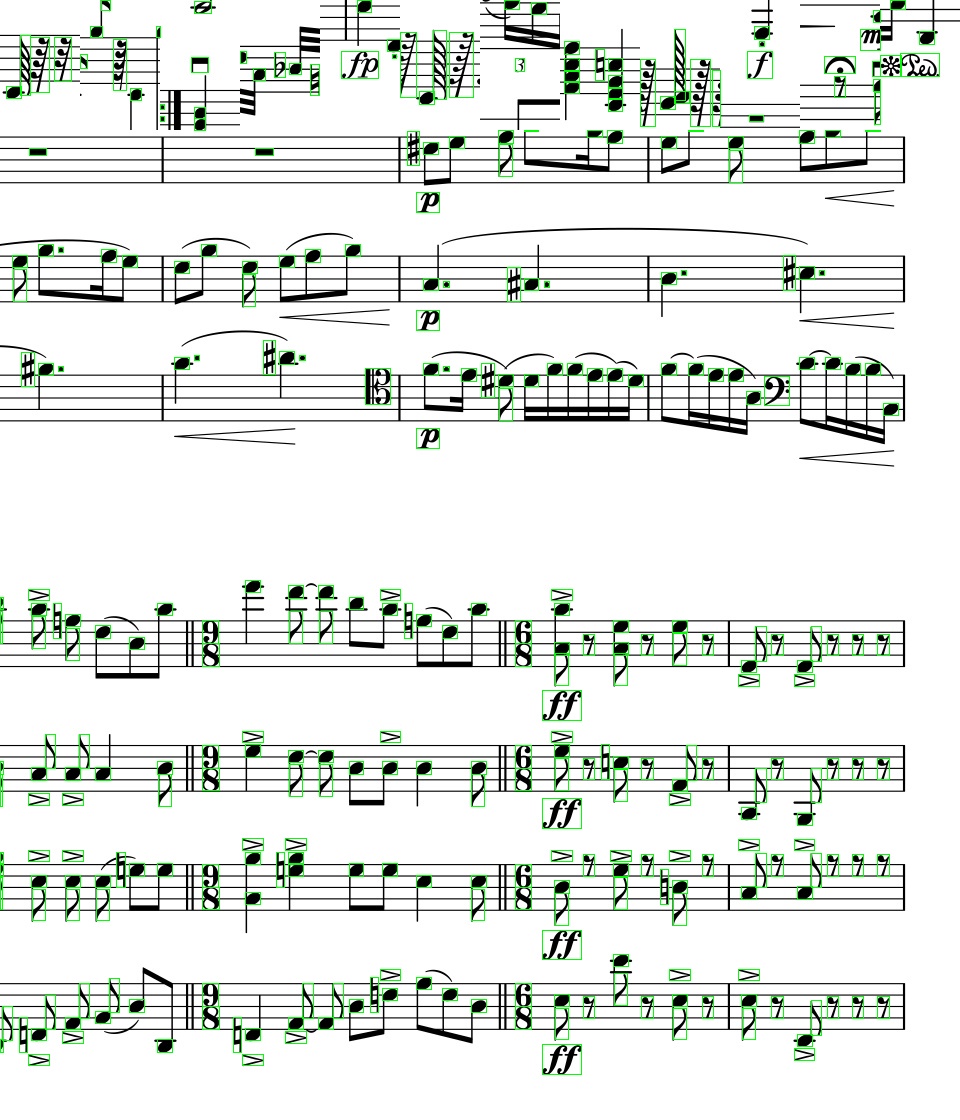}
    \caption{A musical score where $12$ small images have been augmented at the top of $7$ regular staves. The bounding boxes are marked in green.}
    \label{fig:inp_aug}
    \end{center}
    \vspace{-0.8cm}
\end{figure}

Our current answer to this problem is oversampling rare classes by data synthesis, where we locate rare symbols in the dataset, and during training, we append these symbols at the top of the musical sheets (see Figure \ref{fig:inp_aug}). More specifically, we augment each input page in a mini-batch with with $12$ randomly selected synthesized crops of rare symbols (of size $130 \times 80$ pixels) by putting them in the margins at the top of the page. Directions on the choice of the creation of augmented symbols are given on \cite{DBLP:conf/annpr/StadelmannAAADE18}. This way, the neural network (on expectation) does not need to wait for more than $10$ iterations to see every class which is present in the dataset. At the same time, we have been experimenting with pre-training the net with fully synthetic scores where the classes are fully balanced and then retraining it on the full \emph{DeepScores} dataset. The two approaches are complementary and preliminary results show improvement, though more investigation is needed: overfitting on extremely rare symbols is still likely, and questions remain regarding how to integrate the concept of patches (in the margins) with the idea of a full page classifier that considers all context.

%by thilo: "of $7$ regular staves" - is this addition of mine correct? otherwise please change or remove.

\subsection{Cached bounding boxes}
The biggest problem of the Deep Watershed Detector (DWD) on a fundamental level is that the bounding box regression is inaccurate. This is possibly due to the fact that convolutional networks produce smooth outputs, but the bounding box map can be very non-smooth. This "smoothing-bias" creates an averaging over all bounding boxes and leads to an overestimation of small bounding boxes and an underestimating of large ones. We currently address this issue by using cached bounding boxes per class as a prediction, being quite accurate for most classes but completely unusable for others. This is a not a satisfactory solution and has to be improved. We are considering multiple approaches including different bounding box encoding in the output layer or usage of the DWD localization as an object proposal system in an R-CNN style detection scheme.

\section{Final results}

After we implemented the mentioned improvement in our detector, we compared our results with state-of-the-art OMR detectors. We compared the results of our detector with the results reported on \cite{DBLP:journals/as/Pacha18}, where the authors reported numbers of Faster R-CNN \cite{DBLP:conf/nips/RenHGS15}, RetinaNet \cite{DBLP:conf/iccv/LinGGHD17} and a custom net based on U-Net \cite{DBLP:conf/miccai/RonnebergerFB15} on \emph{DeepScores} and \emph{MUSCIMA++} datasets. Additionally, we reported the results of our detector in the scanned version of \emph{DeepScores}. All results are given on MS-COCO mAP \cite{DBLP:conf/eccv/LinMBHPRDZ14}, where the mAP score is computed at [0.5, 0.55, ..., 0.95] and then averaged.

\begin{table}[h]
\centering
\resizebox{0.9\textwidth}{!}{%
\begin{tabular}{@{}llccccccccccc@{}}
\toprule
\textbf{map \%} & \textbf{DeepScores (synthetic)} & \textbf{Musicma++ (handwritten)} & \textbf{DeepScores (scans)} \\ \midrule
Faster R-CNN & 19.6 & 3.9 & -\\
RetinaNet & 9.8 & 7.7 &  - \\
U-Net & 24.8 & 16.6 & - \\
\midrule
\textbf{DWDNet} & \textbf{41.4} & \textbf{19.9} & 47.3 \\ \bottomrule
\end{tabular}%
}
\caption{Results of our detector in \emph{DeepScores}, \emph{Musicma++} and \emph{DeepScores-scans} and the comparison with Faster R-CNN, RetinaNet and U-Net}
\label{tab:res_dwd_final}
\end{table}

As can be seen in Table \ref{tab:res_dwd_final}, our method massively outperforms the other $3$ detectors. In case of \emph{DeepScores} our method reaches almost twice as high score as the next best method. Similarly, in \emph{MUSCIMA++} our method outperforms the other three method by a large margin.

Being a single-stage detector, our method is as fast as RetinaNet, and around an order of magnitude faster than Faster R-CNN detector. The U-Net detector presented in \cite{DBLP:journals/as/Pacha18} is inspired from U-Net segmentation network \cite{DBLP:conf/miccai/RonnebergerFB15}, with the caveat being that the segmentation happens sequentially for every class. This makes the method at the very best case slow (around two orders of magnitude slower than our method and RetinaNet), if not totally unscalable when the number of classes increase. So, not only our detector reaches the best results in mAP score, but it is also as fast as the fastest competitive detector.

\section{Conclusions and Future Work}\label{sec:conc}
\begin{figure}
\begin{center}
\includegraphics[scale=0.2]{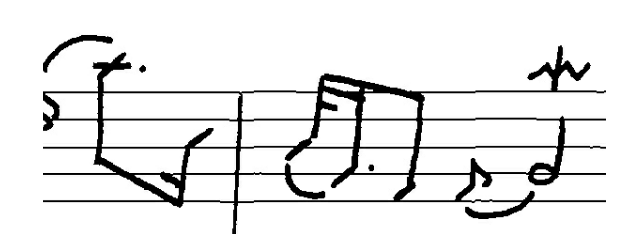}
\caption{Typical input snippet used by Pacha et al. \cite{DBLP:conf/iwdas/Pacha2018}}
\label{fig:snippet_apacha}
\end{center}
\end{figure}

We have presented a novel method for object detection that is specifically tailored to detect many tiny objects on large inputs. We have shown that it is able to detect common symbols of music notation with high precision, both in digitally rendered music as well as in handwritten music, without a drop in performance when moving to the "more complicate" handwritten input. This suggests that deep learning based approaches are able to deal with handwritten sheets just as well as with digitally rendered ones, additionally to their benefit of recognizing objects in their context and with minimal preprocessing as compared to classical OMR pipelines. Pacha et al.\cite{DBLP:conf/iwdas/Pacha2018} show that higher detection rates, especially for uncommon symbols, are possible when using R-CNN on small snippets (cp. Fig. \ref{fig:snippet_apacha}). Despite their higher scores, it is unclear how recognition performance is affected when results of overlapping and potentially disagreeing snippets are aggregated to full page results. A big advantage of our end-to-end system is the complete avoidance of error propagation in longer recognition pipeline of independent components like classifiers, aggregators, etc \cite{lecun1998gradient}. Moreover, our full-page end-to-end approach has the advantages of speed (compared to a sliding window patch classifier), change of domain (we use the same architecture for both the digital and handwritten datasets) and is easily integrated into complete OMR frameworks.

\begin{figure}
\begin{center}
\includegraphics[scale=0.2]{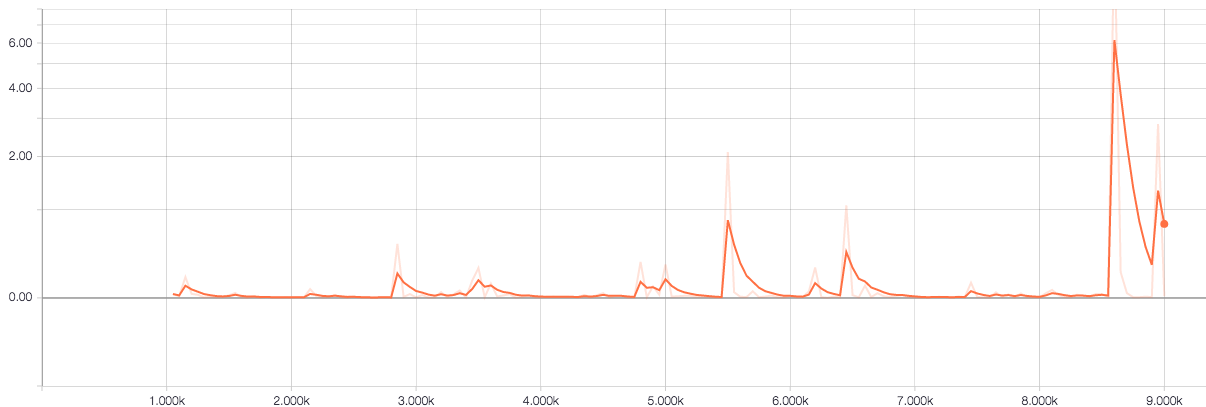}
\caption{Evolution of $loss^b$ (on the ordinate) of a sufficiently trained network, when training for another 8000 iterations (on the abscissa).}
\label{fig:loss_evolution}
\end{center}
\end{figure}

Arguably the biggest problem we faced is that symbol classes in the dataset are heavily unbalanced. Considering that originally we did not do any class-balancing, this imbalance had its effect in training. We observed that in cases where the symbol is common, we get a very high average precision, but it quickly drops when symbols become less common. Furthermore, it is interesting to observe that the neural network actually forgets about the existence of these rarer symbols: Fig. \ref{fig:loss_evolution} depicts the evolution of $loss^b$ of a network that is already trained and gets further trained for another $8,000$ iterations. When faced with an image containing rare symbols, the initial loss is larger than the loss on more common images. But to our surprise, later during the training process, the loss actually increases when the net encounters rare symbols again, giving the impression that the network is actually treating these symbols as outliers and ignoring them. 

We solved the problem by a combination of better training (weighted loss functions, l2 regularization and better hyperparamter search) and a series of data augmentations. In the end this resulted with our detector reaching almost twice as high results in mAP metric, while at the same time being as fast as one-stage competing detectors.

As future work, we plan to investigate the ability of our method beyond OMR on natural images. Initially we will approach canonical datasets like \emph{PASCAL VOC} \cite{DBLP:journals/ijcv/EveringhamGWWZ10} and \emph{MS-COCO} \cite{DBLP:conf/eccv/LinMBHPRDZ14} that have been at the front-line of object recognition tasks. However, images in those datasets are not exactly natural, and for the most part they are simplistic (small images, containing a few large objects). Recently, researchers have been investigating the ability of state-of-the-art recognition systems on more challenging natural datasets, like DOTA \cite{DBLP:journals/corr/abs-1711-10398}, and unsurprisingly, the results leave much to be desired. The DOTA dataset shares a lot of similarities with musical datasets, with images being high resolution and containing hundreds of small objects, making it a suitable benchmark for our DWD method to recognize tiny objects.

\chapter{Discussion and Conclusions}

In this thesis, we studied the effect of contextual information in deep neural networks. Loosely speaking, the contextual information can be given to a CNN either explicitly (by incorporating special building blocks that take into consideration the structure of the dataset) or implicitly (by carefully constraining the CNN to take into consideration the surrounding objects. We investigated both approaches, finding out that in either case, giving contextual information helps neural networks in many different tasks like classification, recognition or similarity learning.

\section{Implicit Context}

There has been a long-standing belief in the community of machine learning that CNNs make local decisions if they are not augmented by building blocks that take into consideration the global information. This belief was challenged by \cite{DBLP:journals/corr/abs-1808-03305} where the authors found out that for the tasks of object detection and recognition, not only that context matters a lot, but CNNs with large receptive fields use context in an implicit manner.

\begin{figure}[ht!]
  \centering
\includegraphics[width=0.99\textwidth]{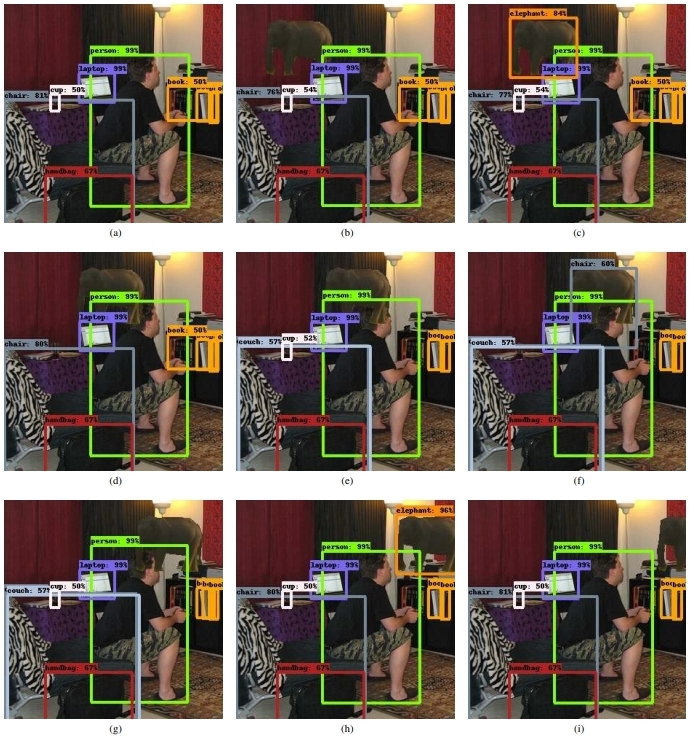}
%\vspace{-0.5cm}
\caption{Detecting an elephant in a room. A state-of-the-art object detector detects multiple images in a living-room (a). A transplanted object (elephant) can remain undetected in many situations and arbitrary locations (b,d,e,g,i). It can assume incorrect identities such as a chair (f). The object has a non-local effect, causing other objects to disappear (cup, d,f, book, e-i ) or switch identity (chair switches to couch in e). It is recommended to view this image in color online. Figure taken from \cite{DBLP:journals/corr/abs-1808-03305}} 
\label{fig:elephant}
\end{figure}

\begin{figure}
\centering
\subfigure[Augmentation dot\label{fig:augmentation_dot}]{\includegraphics[width=0.35\textwidth]{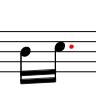}}
\subfigure[Stacatto\label{fig:stacatto}]{\includegraphics[width=0.4\textwidth]{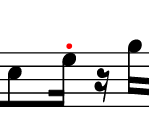}}
\caption{Two symbols which look the same, but have totally different meanings (and so classes). By carefully designing our CNN to implicitly consider contextual information, we are able to distinguish between "Augmentation dot" and "Stacatto".}
\end{figure}

Fig. \ref{fig:elephant} shows how the surroundings of an image directly effect the results of a state-of-the-art object detector, despite that there is no explicit way of looking for the context in the image. Motivated by similar beliefs, we designed and developed a state-of-the-art object detector for the task of optical music recognition, called DWDNet \cite{DBLP:conf/ismir/TuggenerESS18}. DWDNet is an one-stage object detector, meaning that it finds all objects in an image in one go (unlike two-stage detectors which initially find interesting regions and then classify those regions by taking into consideration only the pixels of that region), and by doing so, it is forced to consider the contextual information. By carefully training it and improving over the dataset it is trained on, we managed to not only outperform the current best object detectors, but also to be able to correctly distinguish between symbols which look visually identical, but have totally different meaning, as shown in Fig. \ref{fig:augmentation_dot} and Fig. \ref{fig:stacatto}.

\section{Explicit Context}

Even more interesting to us is the explicit usage of context, which has been a relatively non-studied topic until recently. We used the graph theoretical well-known algorithm called "Graph Transduction Game (GTG)" to initially do a label augmentation of the dataset, which would allow us to train CNNs even in cases where there is a sparsity of labelled data. In extreme cases - where the number of labelled data is only $10$ to $250$, we were able to massively outperform the previously best deep  semi-supervised learning models. Later on, inspired from GTG we designed a differentiable module which we called "The Group Loss", put it on top of CNN, allowing us to get state-of-the-art results in the task of similarity learning, clustering and image retrieval.

Among the possible graph-based label propagation algorithms \cite{zhu2002learning,DBLP:conf/icml/ZhuGL03,DBLP:conf/nips/ZhouBLWS03,zhou2004workshop}, we choose Graph Transduction Game (GTG) \cite{DBLP:journals/neco/ErdemP12}. The motivations that drive our choice are the following:

\begin{itemize}
  \item GTG takes into account the similarity between objects and the relation between all objects in the mini-batch, ensuring that similar objects belong to the same group and dissimilar objects belong to different groups;
  \item it is differentiable, hence perfectly fit for an end-to-end learning;
  \item it can be implemented in a vectorized format as product of matrices, making it computationally efficient; 
  \item it allows the injection of prior knowledge, which we can get for free from the same neural model used to compute the embedding;
  \item it can be applied in small mini-batches (of size $30$-$100$), unlike Deep Spectral Embedding \cite{DBLP:conf/icml/LawUZ17} which uses large mini-batches in the size of thousands. Training with small batches was empirically showed to achieve higher generalization performance \cite{DBLP:conf/iclr/KeskarMNST17}, thus making the usage of small mini-batches very desirable.
  \item by working in the standard simplex, it outputs probability distributions instead of hard cluster assignments, thus allowing the usage of the cross-entropy loss function and establishing a natural link between supervised learning (classes) and unsupervised learning (clusters).
\end{itemize}

Nevertheless, on chapter 3 the model could have been replaced with any of the other models \cite{zhu2002learning,DBLP:conf/icml/ZhuGL03,DBLP:conf/nips/ZhouBLWS03,zhou2004workshop}. In fact, we keep the same framework but replace the GTG with other models in order to show the superior performance of GTG. On the other hand, the work on chapter 4 is not GTG-based, but more GTG-inspired and has evolved on its own thing. The combination of an iterative-procedure that works in standard simplex (probability-space) with the softmax-layer of neural network is very strong, and makes it very natural. While we do not see a way of using some other propagation model \cite{zhu2002learning,DBLP:conf/icml/ZhuGL03,DBLP:conf/nips/ZhouBLWS03,zhou2004workshop} for the task of similarity learning, it needs to be said that in principle, a graph neural network based method could be used for the same task. In particular, our work can be extended in something that looks like \cite{DBLP:conf/iclr/KipfW17} where the propagation rule is learned (instead of using replicator dynamics). However, for practical purposes, this is extremely hard to be achieved because it is difficult to make graph neural network models work with mini-batches. Not surprisingly, the biggest success of \cite{DBLP:conf/iclr/KipfW17} has come in citation network datasets when the number of samples is not too high, and the samples are represented by low-dimensional features. Training such networks in high-dimensional datasets (for example large-image datasets) comes with extreme memory requirements, and with the current technology is not easily achievable. Mini-batch stochastic gradient descent can potentially alleviate this issue. The procedure of generating mini-batches, however, should take into account the
number of layers in the Graph Convolutional Network model, as the $K$th-order neighborhood for a Graph Convolutional Network with $K$ layers has to be stored in memory for an exact procedure. For very large and densely connected graph datasets are needed. So, while extending our framework to a purely Graph Convolutional Network model is desirable and would allow to learn the propagation rule (instead of pre-determining it like in The Group Loss \cite{DBLP:journals/corr/groupLoss} work) doing so is not straight-forward. Not surprisingly, there are not many clear successes of Graph Convolutional Networks in the field of computer vision, especially when dealing with large datasets. Nevertheless, this remains as something to be done as part of future work.

\section{Discussion}

Regardless if the contextual information is provided implicitly in neural networks, or contextual modules are added in neural networks, it is important to carefully consider and involve context when the network is making decisions. While for many decades, the only way to provide context in a network was assumed to be via recurrent connections, recently the investigators have developed new methods of doing so \cite{DBLP:journals/corr/abs-1808-03305, DBLP:conf/nips/SabourFH17, DBLP:conf/nips/VaswaniSPUJGKP17}. This thesis is a step in the same direction, where we investigated ways of both implicitly and explicitly using context to help convolutional neural networks make informed and global decisions. We found that in both cases, the context plays a very important role and modern neural networks suited to complicated problems need to carefully consider ways of involving context.

\appendix
\chapter{Learning Neural Models for End-to-End Clustering}

\section{Disclaimer}

The work presented in this chapter is based on the following paper:

\begin{description}  
  \item Benjamin Bruno Meier, \textbf{Ismail Elezi}, Mohammadreza Amirian, Oliver D\"urr and Thilo Stadelmann; \textit{Learning neural models for end-to-end clustering \cite{DBLP:conf/annpr/MeierEADS18}}; In Proceedings of IAPR TC3 Workshop on Artificial Neural Networks in Pattern Recognition (ANNPR 2018)
\end{description}

The contributions of the author are the following:

\begin{description}
  \item[$\bullet$ Doing] some limited experiments.
  \item[$\bullet$ Writing] a part of the paper.
\end{description}

\section{Introduction}
\begin{figure}[t]
  \centering
  \includegraphics[width=1.0\columnwidth]{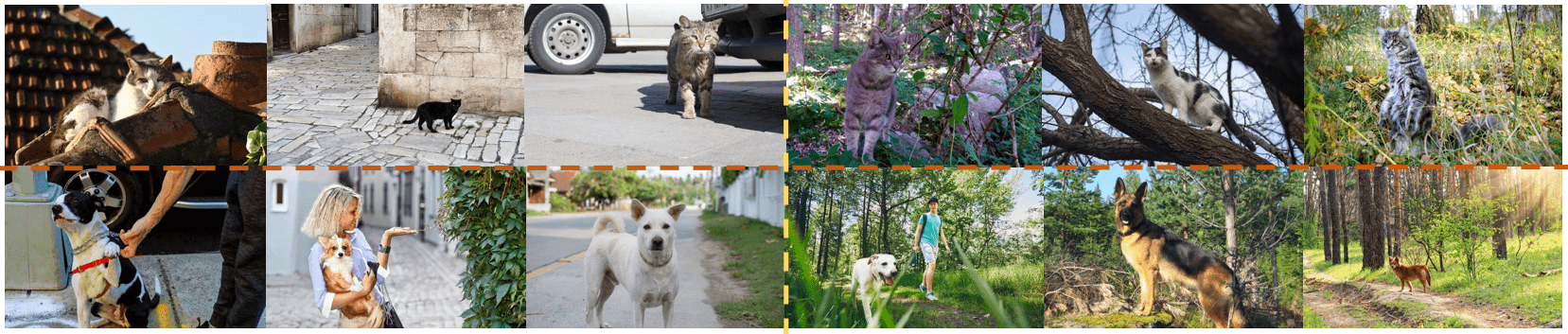}
  \caption{Images of cats (top) and dogs (bottom) in urban (left) and natural (right) environments.}   
  \label{fig:cats_dogs}
\end{figure}

Consider the illustrative task of grouping images of cats and dogs by \emph{perceived} similarity: depending on the intention of the user behind the task, the similarity could be defined by animal type (foreground object), environmental nativeness (background landscape, cp. Fig. \ref{fig:cats_dogs}) etc. This is characteristic of clustering perceptual, high-dimensional data like images \cite{kampffmeyer2017} or sound \cite{lukic2017speaker}: a user typically has some similarity criterion in mind when thinking about naturally arising groups (e.g., pictures by holiday destination, or persons appearing; songs by mood, or use of solo instrument). As defining such a similarity for every case is difficult, it is desirable to learn it. At the same time, the learned model will in many cases not be a classifier---the task will not be solved by classification---since the number and specific type of groups present at application time are not known in advance (e.g., speakers in TV recordings; persons in front of a surveillance camera; object types in the picture gallery of a large web shop). 

Grouping objects with machine learning is usually approached with clustering algorithms \cite{kaufman1990finding}. Typical ones like K-means \cite{MacQueen}, EM \cite{jin2011expectation}, hierarchical clustering \cite{murtagh1983survey} with chosen distance measure, or DBSCAN \cite{ester1996density} each have a specific inductive bias towards certain similarity structures present in the data (e.g., K-means: Euclidean distance from a central point; DBSCAN: common point density). Hence, to be applicable to above-mentioned tasks, they need high-level features that already encode the aspired similarity measure. This may be solved by learning salient embeddings \cite{mikolov2013efficient} with a deep metric learning approach \cite{hoffer2015deep}, followed by an off-line clustering phase using one of the above-mentioned algorithm.

However, it is desirable to combine these distinct phases (learning salient features, and subsequent clustering) into an end-to-end approach that can be trained globally \cite{lecun1998gradient}: it has the advantage of each phase being perfectly adjusted to the other by optimizing a global criterion, and removes the need of manually fitting parts of the pipeline. Numerous examples have demonstrated the success of neural networks for end-to-end approaches on such diverse tasks as speech recognition \cite{amodei2016deep}, robot control \cite{levine2016end}, scene text recognition \cite{shi2017end}, interactive segmentation \cite{DBLP:journals/pami/MequanintAP19}, image retrieval \cite{DBLP:journals/corr/groupLoss, DBLP:conf/icpr/ZemeneAP16, DBLP:journals/corr/abs-1808-05075}, person re-identification \cite{DBLP:journals/corr/abs-1904-11397} or music transcription \cite{sigtia2016end}.

In this work, we present a conceptually novel approach that we call \emph{``learning to cluster''} in the above-mentioned sense of grouping high-dimensional data by some perceptually motivated similarity criterion. For this purpose, we define a novel neural network architecture with the following properties: (a) during training, it receives pairs of similar or dissimilar examples to learn the intended similarity function implicitly or explicitly; (b) during application, it is able to group objects of groups never encountered before; (c) it is trained end-to-end in a supervised way to produce a tailor-made clustering model and (d) is applied like a clustering algorithm to find both the number of clusters as well as the cluster membership of test-time objects in a fully probabilistic way.

\begin{figure}[t!]
  \centering
  \includegraphics[width=0.85\columnwidth]{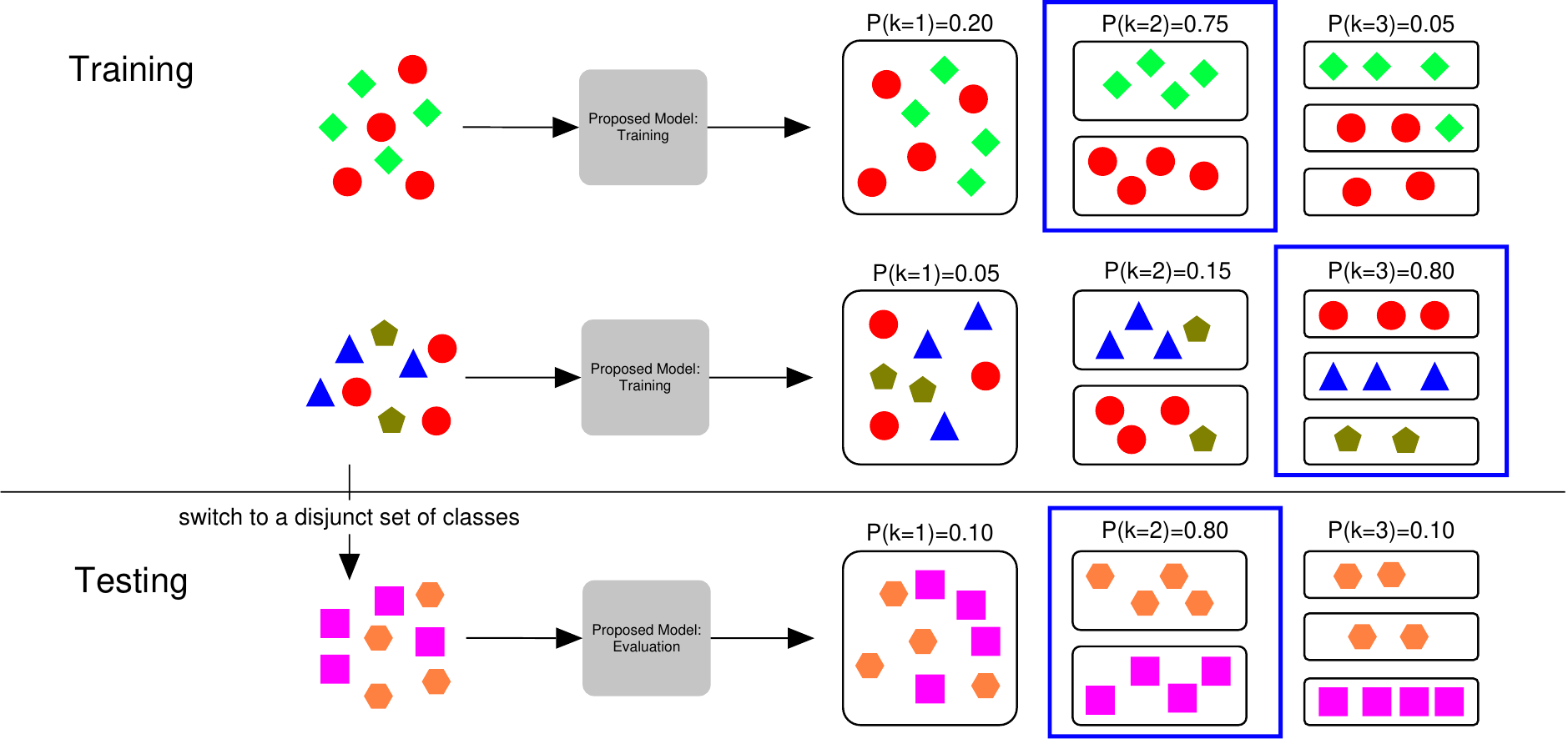}
  \caption{Training vs. testing: cluster types encountered during application/inference are never seen in training. Exemplary outputs (right-hand side) contain a partition for each $k$ ($1$--$3$ here) and a corresponding probability (best highlighted blue).}
  \label{fig:training_vs_evaluation}
\end{figure}

Our approach builds upon ideas from \emph{deep metric embedding}, namely to learn an embedding of the data into a representational space that allows for specific perceptual similarity evaluation via simple distance computation on feature vectors. However, it goes beyond this by adding the actual clustering step---grouping by similarity---directly to the same model, making it trainable end-to-end. Our approach is also different from \emph{semi-supervised clustering} \cite{Basu02semi-supervisedclustering}, which uses labels for some of the data points in the inference phase to guide the creation of groups. In contrast, our method uses absolutely no labels during inference, and moreover doesn't expect to have seen any of the groups it encounters during inference already during training (cp. Fig. \ref{fig:training_vs_evaluation}). Its training stage may be compared to creating K-means, DBSCAN etc. in the first place: it creates a specific clustering model, applicable to data with certain similarity structure, and once created/trained, the model performs ``unsupervised learning'' in the sense of finding groups. Finally, our approach differs from traditional cluster \emph{analysis} \cite{kaufman1990finding} in how the clustering algorithm is applied: instead of looking for patterns in the data in an unbiased and exploratory way, as is typically the case in unsupervised learning, our approach is geared towards the use case where users know perceptually what they are looking for, and can make this explicit using examples.  We then learn appropriate features and the similarity function simultaneously, taking full advantage of end-to-end learning. 

Our main contribution in this work is the creation of a neural network architecture that learns to \emph{group} data, i.e., that outputs the same ``label'' for ``similar'' objects regardless of (a) it has ever seen this group before; (b) regardless of the actual value of the label (it is hence not a ``class''); and (c) regardless of the number of groups it will encounter during a single application run, up to a predefined maximum. This is novel in its concept and generality (i.e., learn to cluster previously unseen groups end-to-end for arbitrary, high-dimensional input without any optimization on test data). Due to this novelty in approach, we focus here on the general idea and experimental demonstration of the principal workings, and leave comprehensive hyperparameter studies and optimizations for future work. In Sec. \ref{sec:related_work}, we compare our approach to related work, before presenting the model and training procedure in detail in Sec. \ref{sec:our_model}. We evaluate our approach on different datasets in Sec \ref{sec:experiments}, showing promising performance and a high degree of generality for data types ranging from 2D points to audio snippets and images, and discuss these results with conclusions for future work in Sec. \ref{sec:conclusions}.

\section{Related Work}
\label{sec:related_work}

Learning to cluster based on neural networks has been approached mostly as a supervised learning problem to extract embeddings for a subsequent off-line clustering phase. The core of all deep metric embedding models is the choice of the loss function. Motivated by the fact that the softmax-cross entropy loss function has been designed as a classification loss and is not suitable for the clustering problem per se, \emph{Chopra et al.} \cite{DBLP:conf/cvpr/ChopraHL05} developed a ``Siamese'' architecture, where the loss function is optimized in a way to generate similar features for objects belonging to the same class, and dissimilar features for objects belonging to different classes. A closely related loss function called ``triplet loss'' has been used by \emph{Schroff et al.} \cite{DBLP:conf/cvpr/SchroffKP15} to get state-of-the-art accuracy in face detection. The main difference from the Siamese architecture is that in the latter case, the network sees same and different class objects with every example. It is then optimized to jointly learn their feature representation. A problem of both approaches is that they are typically difficult to train compared to a standard cross entropy loss.

\emph{Song et al.} \cite{DBLP:conf/cvpr/SongXJS16} developed an algorithm for taking full advantage of all the information available in training batches. They later refined the work \cite{DBLP:conf/cvpr/SongJR017} by proposing a new metric learning scheme based on structured prediction, which is designed to optimize a clustering quality metric (normalized mutual information \cite{DBLP:journals/corr/abs-1110-2515}). Even better results were achieved by \emph{Wong et al.} \cite{DBLP:conf/iccv/WangZWLL17}, where the authors proposed a novel angular loss, and achieved state-of-the-art results on the challenging real-world datasets \emph{Stanford Cars} \cite{KrauseStarkDengFei-Fei_3DRR2013} and \emph{Caltech Birds} \cite{DBLP:journals/ijcv/BransonHWPB14}. On the other hand, \emph{Lukic et al.} \cite{lukic2017speaker} showed that for certain problems, a carefully chosen deep neural network can simply be trained with softmax-cross entropy loss and still achieve state-of-the-art performance in challenging problems like speaker clustering. Alternatively, \emph{Wu et al.} \cite{DBLP:conf/iccv/ManmathaWSK17} showed that state-of-the-art results can be achieved simply by using a traditional margin loss function and being careful on how sampling is performed during the creation of mini-batches.

On the other hand, attempts have been made recently that are more similar to ours in spirit, using deep neural networks only and performing clustering end-to-end \cite{DBLP:journals/corr/abs-1801-07648}. They are trained in a fully unsupervised fashion, hence solve a different task then the one we motivated above (that is inspired by speaker- or image clustering based on some human notion of similarity). Perhaps first to group objects together in an unsupervised deep learning based manner where \emph{Le et al.} \cite{DBLP:conf/icml/LeRMDCCDN12}, detecting high-level concepts like cats or humans. \emph{Xie et al.} \cite{xie2016unsupervised} used an autoencoder architecture to do clustering, but experimental evaluated it only simplistic datasets like \emph{MNIST}. CNN-based approaches followed, e.g. by \emph{Yang et al.} \cite{DBLP:conf/cvpr/YangPB16}, where clustering and feature representation are optimized together. \emph{Greff et al.} \cite{DBLP:conf/nips/GreffSS17} performed perceptual grouping (of pixels within an image into the objects constituting the complete image, hence a different task than ours) fully  unsupervised using a neural expectation maximization algorithm. Our work differs from above-mentioned works in several respects: it has no assumption on the type of data, and solves the different task of grouping whole input objects.

\section{A model for end-to-end clustering of arbitrary data}
\label{sec:our_model}

\begin{figure}[t!]
  \centering
  \includegraphics[width=1\columnwidth]{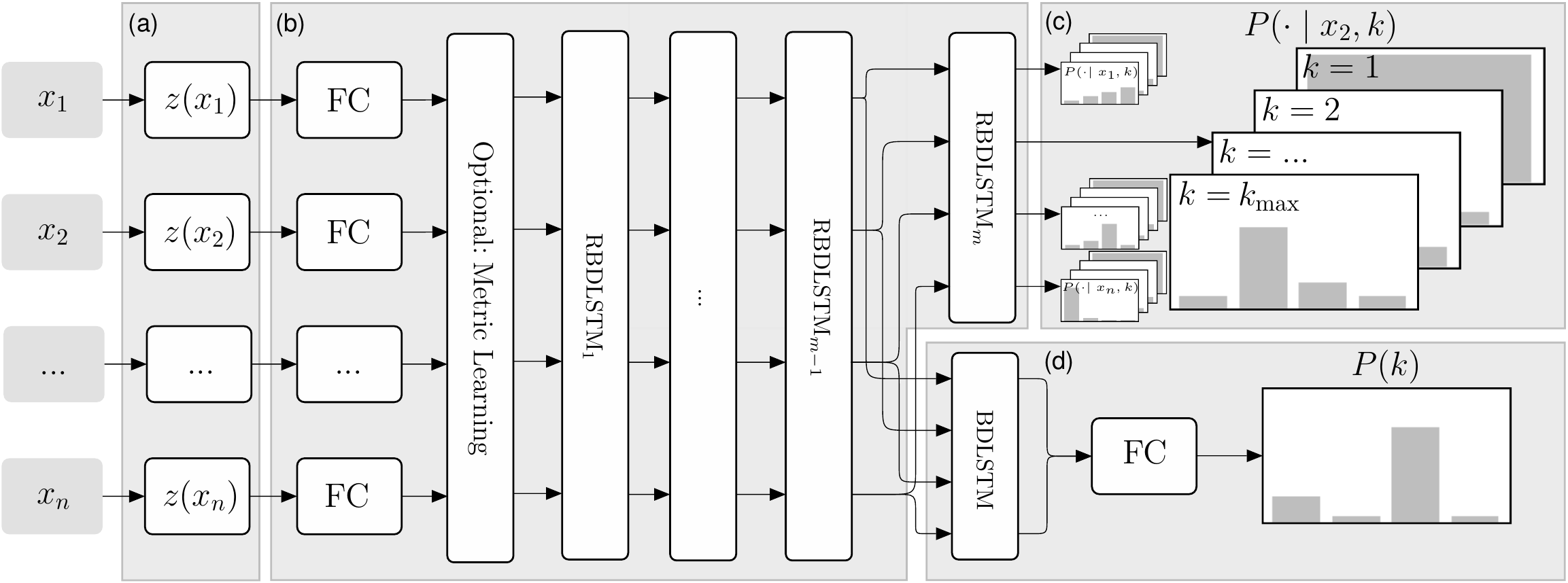}
  \caption{Our complete model, consisting of \mbox{(a)} the embedding network, \mbox{(b)} clustering network (including an optional metric learning part, see Sec. \ref{sec:metric_learning}), \mbox{(c)} cluster-assignment network and \mbox{(d)} cluster-count estimating network.}   
  \label{fig:clustering}
\end{figure}

Our method learns to cluster end-to-end purely ab initio, without the need to explicitly specify a notion of similarity, only providing the information whether two examples belong together. It uses as input $n \ge 2$ examples $x_i$, where $n$  may be different during training and application and constitutes the number of objects that can be clustered at a time, i.e. the maximum number of objects in a partition. The network's output is two-fold: a probability distribution $P(k)$ over the cluster count $1 \leq k \leq k_\mathrm{max}$; and probability distributions $P(\cdot\mid x_i,k)$ over all possible cluster indexes for each input example $x_i$ and for each $k$. 

\subsection{Network architecture}
The network architecture (see Fig. \ref{fig:clustering}) allows the flexible use of different input types, e.g. images, audio or 2D points. An input $x_i$ is first processed by an embedding network (a) that produces a lower-dimensional representation $z_i=z(x_i)$. The dimension of $z_i$ may vary depending on the data type. For example, 2D points do not require any embedding network. A fully connected layer (FC) with $\mathrm{LeakyReLU}$ activation at the beginning of the clustering network (b) is then used to bring all embeddings to the same size. This approach allows to use the identical subnetworks (b)--(d) and only change the subnet (a) for any data type. The goal of the subnet (b) is to compare each input $z(x_i)$ with all other $z(x_{j\ne i})$, in order to learn an abstract grouping which is then concretized into an estimation of the number of clusters (subnet (d)) and a cluster assignment (subnet (c)).

\begin{figure}[t!]
  \centering
  \includegraphics[width=0.5\columnwidth]{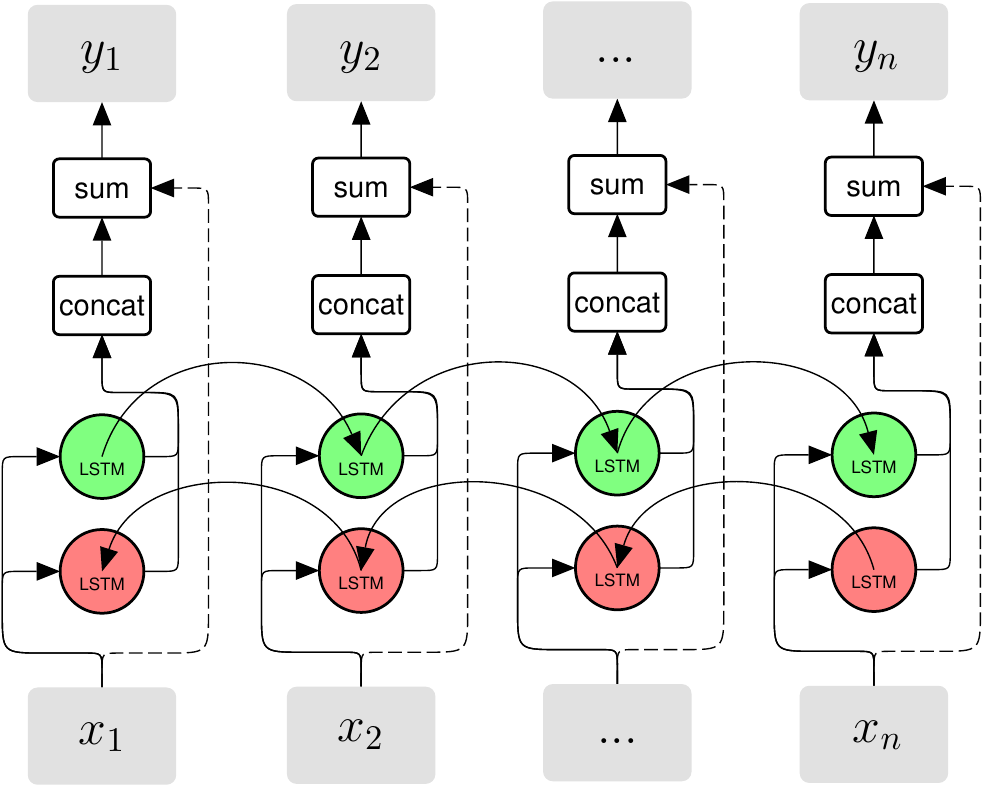}
  \caption{$\mathrm{RBDLSTM}$-layer: A $\mathrm{BDLSTM}$ with residual connections (dashed lines). The variables $x_i$ and $y_i$ are named independently from the notation in Fig. \ref{sec:our_model}.}
  \label{fig:rbdlstm}
\end{figure}

To be able to process a non-fixed number of examples $n$ as input, we use a recurrent neural network. Specifically, we use stacked residual bi-directional LSTM-layers ($\mathrm{RBDLSTM}$), which are similar to the cells described in \cite{wu2016google} and visualized in Fig. \ref{fig:rbdlstm}. The residual connections allow a much more effective gradient flow during training and mitigate the problem of vanishing gradients. Additionally, the network can learn to use or bypass certain layers using the residual connections, thus reducing the architectural decision on the number of recurrent layers to the simpler one of finding a reasonable upper bound.

The first of overall two outputs is modeled by the cluster assignment network \mbox{(c)}. It contains a $\mathrm{softmax}$-layer to produce $P(\ell\mid x_i,k)$, which assigns a cluster index $\ell$ to each input $x_i$, given $k$ clusters (i.e., we get a distribution over possible cluster assignments for each input and every possible number of clusters). The second output, produced by the cluster-count estimating network \mbox{(d)}, is built from another $\mathrm{BDLSTM}$-layer. Due to the bi-directionality of the network, we concatenate its first and the last output vector into a fully connected layer of twice as many units using again $\mathrm{LeakyReLUs}$. The subsequent $\mathrm{softmax}$-activation finally models the distribution $P(k)$ for $1\leq k\leq k_\mathrm{max}$. The next subsection shows how this neural network learns to approximate these two complicated probability distributions \cite{Lee2017OnTA} purely from pairwise constraints on data that is completely separate from any dataset to be clustered. No labels for clustering are needed.

\subsection{Training and loss}
In order to define a suitable loss-function, we first define an approximation (assuming independence) of the probability that $x_i$ and $x_j$ are assigned to the same cluster for a given $k$ as
\begin{equation}
  P_{ij}(k)=\sum_{\ell=1}^{k} P(\ell\mid x_i,k)P(\ell\mid x_j,k).
\end{equation}
By marginalizing over $k$, we obtain $P_{ij}$, the probability that $x_i$ and $x_j$ belong to the same cluster:
\begin{equation}
   P_{ij}=\sum_{k=1}^{k_{\mathrm{max}}} P(k) \sum_{\ell=1}^{k} P(\ell\mid x_i,k)P(\ell\mid x_j,k).
\end{equation}
Let $y_{ij}=1$ if $x_i$ and $x_j$ are from the same cluster (e.g., have the same group label) and $0$ otherwise. The loss component for \emph{cluster assignments}, $L_\mathrm{ca}$, is then given by the weighted binary cross entropy as
\begin{equation}
  L_{\mathrm{ca}}=\frac{-2}{n(n-1)}\sum_{i<j}{\left(\varphi_1 y_{ij} \log(P_{ij})+\varphi_2 (1-y_{ij}) \log(1-P_{ij})\right)}
\end{equation}
with weights $\varphi_1$ and $\varphi_2$. The idea behind the weighting is to account for the imbalance in the data due to there being more dissimilar than similar pairs $(x_i,x_j)$ as the number of clusters in the mini batch exceeds $2$. Hence, the weighting is computed using $\varphi_1=c\sqrt{1-\varphi}$ and $\varphi_2=c\sqrt{\varphi}$, with $\varphi$ being the expected value of $y_{ij}$ (i.e., the a priori probability of any two samples in a mini batch coming from the same cluster), and $c$ a normalization factor so that $\varphi_1 + \varphi_2 = 2$. The value $\varphi$ is computed over all possible cluster counts for a fixed input example count $n$, as during training, the cluster count is randomly chosen for each mini batch according to a uniform distribution. The weighting of the cross entropy given by $\varphi$ is then used to make sure that the network does not converge to a sub-optimal and trivial minimum. Intuitively, we thus account for permutations in the sequence of examples by checking rather for pairwise correctness (probability of same/different cluster) than specific indices.

The second loss term, $L_\mathrm{cc}$, penalizes a wrong \emph{number of clusters} and is given by the categorical cross entropy of $P(k)$ for the true number of clusters $k$ in the current mini batch:
\begin{equation}
  L_\mathrm{cc} = -\log(P(k)).
\end{equation}

The complete loss is given by $L_{\mathrm{tot}}=L_{\mathrm{cc}}+ \lambda L_{\mathrm{ca}}$. During training, we prepare each mini batch with $N$ sets of $n$ input examples, each set with $k=1\ldots k_\mathrm{max}$ clusters chosen uniformly. Note that this training procedure requires only the knowledge of $y_{ij}$ and is thus also possible for weakly labeled data. All input examples are randomly shuffled for training and testing to avoid that the network learns a bias w.r.t. the input order. To demonstrate that the network really learns an intra-class distance and not just classifies objects of a fixed set of classes, it is applied on totally different clusters at evaluation time than seen during training.

\subsection{Implicit vs. explicit distance learning}
\label{sec:metric_learning}
To elucidate the importance and validity of the implicit learning of distances in our subnetwork (b), we also provide a modified version of our network architecture for comparison, in which the calculation of the distances is done explicitly. Therefore, we add an extra component to the network before the RBDLSTM layers, as can be seen in Figure \ref{fig:clustering}: the optional metric learning block receives the fixed-size embeddings from the fully connected layer after the embedding network (a) as input and outputs the pairwise distances of the data points. The recurrent layers in block (b) then subsequently cluster the data points based on this pairwise distance information \cite{chin2010novel,arias2011clustering} provided by the metric learning block. 

We construct a novel metric learning block inspired by the work of \emph{Xing et al.} \cite{xing2003distance}. In contrast to their work, we optimize it end-to-end with backpropagation. This has been proposed in \cite{schwenker2001three} for classification alone; we do it here for a clustering task, for the whole covariance matrix, and jointly with the rest of our network. We construct the non-symmetric, non-negative dissimilarity measure $d^2_A$ between two data points $x_i$ and $x_j$ as
\begin{equation}
  d^2_A(x_i, x_j) =  (x_i-x_j)^{T}A(x_i-x_j)
  \label{eq:dist}
\end{equation}
and let the neural network training optimize $A$ through $L_{\mathrm{tot}}$ without intermediate losses. The matrix $A$ as used in $d^2_A$ can be thought of as a trainable distance metric. In every training step, it is projected into the space of positive semidefinite matrices.

\section{Experimental results}
\label{sec:experiments}

To assess the quality of our model, we perform clustering on three different datasets: for a proof of concept, we test on a set of generated \emph{2D points} with a high variety of shapes, coming from different distributions. For speaker clustering, we use the \emph{TIMIT} \cite{timit} corpus, a dataset of studio-quality speech recordings frequently used for pure speaker clustering in related work. For image clustering, we test on the \emph{COIL-100} \cite{nayar1996columbia} dataset, a collection of different isolated objects in various orientations. To compare to related work, we measure the performance with the standard evaluation scores misclassification rate (MR) \cite{liu2003online} and normalized mutual information (NMI) \cite{DBLP:journals/corr/abs-1110-2515}. Architecturally, we choose $m=14$ BDLSTM layers and $288$ units in the FC layer of subnetwork (b), $128$ units for the BDLSTM in subnetwork (d), and $\alpha=0.3$ for all $\mathrm{LeakyReLUs}$ in the experiments below. All hyperparameters where chosen based on preliminary experiments to achieve reasonable performance, but not tested nor tweaked extensively.

\begin{figure}[t!]
  \centering
  \includegraphics[width=\columnwidth]{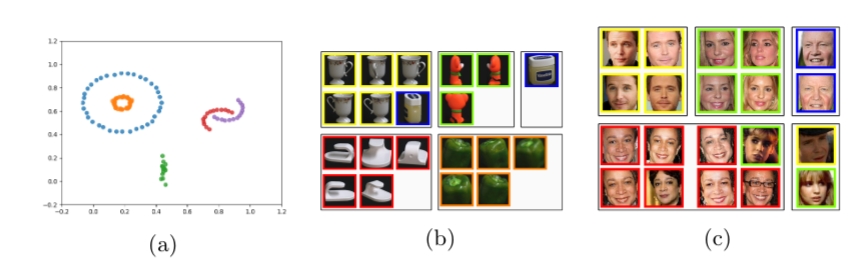}
  \caption{Clustering results for (a) 2D point data, (b) COIL-100 objects, and (c)faces from FaceScrub (for illustrative purposes). The color of points / colored borders of images depict true cluster membership.}
  \label{fig:output_2d_point_clustering}
\end{figure}

We set $k_\mathrm{max}=5$ and $\lambda=5$ for all experiments. For the 2D point data, we use $n=72$ inputs and a batch-size of $N=200$ (We used the batch size of $N=50$ for metric learning with 2D points). For TIMIT, the network input consists of $n=20$ audio snippets with a length of $1.28$ seconds, encoded as mel-spectrograms with $128\times 128$ pixels (identical to \cite{lukic2017speaker}). For COIL-100, we use $n=20$ inputs with a dimension of $128\times 128\times 3$. For TIMIT and \mbox{COIL-100}, a simple CNN with 3 conv/max-pooling layers is used as subnetwork (a). For TIMIT, we use $430$ of the $630$ available speakers for training (and $100$ of the remaining ones each for validation and evaluation). For \mbox{COIL-100}, we train on $80$ of the $100$ classes ($10$ for validation, $10$ for evaluation). For all runs, we optimize using Adadelta \cite{DBLP:journals/corr/abs-1212-5701} with a learning rate of $5.0$. Example of clustering are shown in Fig. \ref{fig:output_2d_point_clustering}. For all configurations, the used hardware set the limit on parameter values: we used the maximum possible batch size and values for $n$ and $k_\mathrm{max}$ that allow reasonable training times. However, values of $n\ge 1000$ where tested and lead to a large decrease in model accuracy. This is a major issue for future work.

The results on 2D data as presented in Fig. \ref{fig:output_2d_point_clustering} demonstrate that our method is able to learn specific and diverse characteristics of intuitive groupings. This is superior to any single traditional method, which only detects a certain class of cluster structure (e.g., defined by distance from a central point). Although \cite{lukic2017speaker} reach moderately better scores for the speaker clustering task and \cite{DBLP:conf/cvpr/YangPB16} reach a superior $\mathrm{NMI}$ for \mbox{COIL-100}, our method finds reasonable clusterings, is more flexible through end-to-end training and is not tuned to a specific kind of data. Hence, we assume, backed by the additional experiments to be found online, that our model works well also for other data types and datasets, given a suitable embedding network. \mbox{Tab. \ref{tbl:results}} gives the numerical results for said datasets in the row called ``L2C'' without using the explicit metric learning block. Extensive preliminary experiments on other public datasets like e.g. FaceScrub \cite{ng2014data} confirm these results: learning to cluster reaches promising performance while not yet being on par with tailor-made state-of-the-art approaches.

\begin{table}[t!]
  \scriptsize
  \centering
  \caption{$\mathrm{NMI} \in [0,1]$ and $\mathrm{MR} \in [0,1]$ averaged over $300$ evaluations of a trained network. We abbreviate our ``learning to cluster'' method as ``L2C''.}
  \label{tbl:results}
  \begin{tabular}{|l|l|l|l|l|l|l|}
  \hline
             & \multicolumn{2}{l|}{\textbf{2D Points (self generated)}} & \multicolumn{2}{l|}{\textbf{TIMIT}} & \multicolumn{2}{l|}{\textbf{COIL-100}} \\ \hline
             & \textbf{MR}        & \textbf{NMI}       & \textbf{MR}      & \textbf{NMI}     & \textbf{MR}       & \textbf{NMI}       \\ \hline
  L2C ($=$our method) & \textbf{0.004} & \textbf{0.993} & $0.060$ & $0.928$ & $0.116$ & $0.867$ \\ \hline
  L2C + Euclidean & $0.177$ & $0.730$ & $0.093$  & $0.883$ & $0.123$  & $0.884$  \\ \hline
  L2C + Mahalanobis & $0.185$ & $0.725$ & $0.104$ & $0.882$ & $0.093$ & $0.890$ \\ \hline
  L2C + Metric Learning & $0.165$ & $0.740$ & $0.101$ & $0.880$ & $0.100$ & $0.880$ \\ \hline
  Random cluster assignment & $0.485$ & $0.232$ & $0.435$ & $0.346$ & $0.435$ & $0.346$ \\   \hline
  K-Means & $0.178$ & $0.796$ & - & - & - & - \\ \hline
  DBSCAN & $0.265$ & $0.676$ & - & - & - & - \\ \hline
  Lukic et al. \cite{lukic2017speaker} & - & - & \textbf{0} & - & - & - \\ \hline
  Yang et al. \cite{DBLP:conf/cvpr/YangPB16} & - & - & - & - & - & \textbf{0.985} \\ \hline
  \end{tabular}
\end{table}

We compare the performance of our implicit distance metric learning method to versions enhanced by  different explicit schemes for pairwise similarity computation prior to clustering. Specifically, three implementations of the optional metric learning block in subnetwork (b) are evaluated: using a fixed  diagonal matrix $A$ (resembling the Euclidean distance), training a diagonal $A$ (resembling Mahalanobis distance), and learning the entire coefficients of the distance matrix $A$. Since we argue above that our approach combines \emph{implicit} deep metric embedding with clustering in an end-to-end architecture, one would not expect that adding \emph{explicit} metric computation changes the results by a large extend. This assumption is largely confirmed by the results in the ``L2C$+$\dots'' rows in \mbox{Tab. \ref{tbl:results}}: for COIL-100, Euclidean gives slightly worse, and the other two slightly better results than L2C alone; for TIMIT, all results are worse but still reasonable. We attribute the considerable performance drop on 2D points using all three explicit schemes to the fact that in this case much more instances are to be compared with each other (as each instance is smaller than e.g. an image, $n$ is larger). This might have needed further adaptations like e.g. larger batch sizes (reduced here to $N=50$ for computational reasons) and longer training times.

\section{Discussion and conclusions}
\label{sec:conclusions}

We have presented a novel approach to learn neural models that directly output a probabilistic clustering on previously unseen groups of data; this includes a solution to the problem of outputting similar but unspecific ``labels'' for similar objects of unseen ``classes''. A trained model is able to cluster different data types with promising results. This is a complete end-to-end approach to clustering that learns both the relevant features and the ``algorithm'' by which to produce the clustering itself. It outputs probabilities for cluster membership of all inputs as well as the number of clusters in test data. The learning phase only requires pairwise labels between examples from a separate training set, and no explicit similarity measure needs to be provided. This is especially useful for high-dimensional, perceptual data like images and audio, where similarity is usually semantically defined by humans. Our experiments confirm that our algorithm is able to implicitly learn a metric and directly use it for the included clustering. This is similar in spirit to the very recent work of \emph{Hsu et al.} \cite{hsu2018learning}, but does not need and optimization on the test (clustering) set and finds $k$ autonomously. It is a novel approach to \emph{learn to cluster}, introducing a novel architecture and loss design.

We observe that the clustering accuracy depends on the availability of a large number of different classes during training. We attribute this to the fact that the network needs to learn intra-class distances, a task inherently more difficult than just to distinguish between objects of a fixed amount of classes like in classification problems. We understand the presented work as an early investigation into the new paradigm of learning to cluster by perceptual similarity specified through examples. It is inspired by our work on speaker clustering with deep neural networks, where we increasingly observe the need to go beyond surrogate tasks for learning, training end-to-end specifically for clustering to close a performance leak. While this works satisfactory for initial results, points for improvement revolve around scaling the approach to practical applicability, which foremost means to get rid of the dependency on $n$ for the partition size.

The number $n$ of input examples to assess simultaneously is very relevant in practice: if an input data set has thousands of examples, incoherent single clusterings of subsets of $n$ points would be required to be merged to produce a clustering of the whole dataset based on our model. As the (RBD)LSTM layers responsible for assessing points simultaneously in principle have a long, but still local (short-term) horizon, they are not apt to grasp similarities of thousands of objects. Several ideas exist to change the architecture, including to replace recurrent layers with temporal convolutions, or using our approach to seed some sort of differentiable K-means or EM layer on top of it. Increasing $n$ is a prerequisite to also increase the maximum number of clusters $k$, as $k\ll n$. For practical applicability, $k$ needs to be increased by an order of magnitude; we plan to do this in the future. This might open up novel applications of our model in the area of transfer learning and domain adaptation.

\addtocontents{toc}{\vspace{2em}}  % Add a gap in the Contents, for aesthetics
\backmatter

\label{Bibliography}
\lhead{\emph{Bibliography}} 
\bibliographystyle{apa}
\bibliography{Bibliography2}  

\begin{thebibliography}{}

\bibitem[\protect\astroncite{Alemu and
  Pelillo}{2020}]{DBLP:journals/corr/abs-1808-05075}
Alemu, L.~T. and Pelillo, M. (2020).
\newblock Multi-feature fusion for image retrieval using constrained dominant
  sets.
\newblock {\em Image Vis. Comput.}, 94:103862.

\bibitem[\protect\astroncite{Alemu
  et~al.}{2019}]{DBLP:journals/corr/abs-1904-11397}
Alemu, L.~T., Shah, M., and Pelillo, M. (2019).
\newblock Deep constrained dominant sets for person re-identification.
\newblock In {\em 2019 {IEEE/CVF} International Conference on Computer Vision,
  {ICCV} 2019, Seoul, Korea (South), October 27 - November 2, 2019}, pages
  9854--9863. {IEEE}.

\bibitem[\protect\astroncite{Aljalbout
  et~al.}{2018}]{DBLP:journals/corr/abs-1801-07648}
Aljalbout, E., Golkov, V., Siddiqui, Y., and Cremers, D. (2018).
\newblock Clustering with deep learning: Taxonomy and new methods.
\newblock {\em CoRR}, abs/1801.07648.

\bibitem[\protect\astroncite{Amodei et~al.}{2016}]{amodei2016deep}
Amodei, D., Ananthanarayanan, S., Anubhai, R., Bai, J., Battenberg, E., Case,
  C., Casper, J., Catanzaro, B., Chen, J., Chrzanowski, M., Coates, A., Diamos,
  G., Elsen, E., Engel, J.~H., Fan, L., Fougner, C., Hannun, A.~Y., Jun, B.,
  Han, T., LeGresley, P., Li, X., Lin, L., Narang, S., Ng, A.~Y., Ozair, S.,
  Prenger, R., Qian, S., Raiman, J., Satheesh, S., Seetapun, D., Sengupta, S.,
  Wang, C., Wang, Y., Wang, Z., Xiao, B., Xie, Y., Yogatama, D., Zhan, J., and
  Zhu, Z. (2016).
\newblock Deep speech 2 : End-to-end speech recognition in english and
  mandarin.
\newblock In Balcan, M. and Weinberger, K.~Q., editors, {\em Proceedings of the
  33nd International Conference on Machine Learning, {ICML} 2016, New York
  City, NY, USA, June 19-24, 2016}, volume~48 of {\em {JMLR} Workshop and
  Conference Proceedings}, pages 173--182. JMLR.org.

\bibitem[\protect\astroncite{Arias{-}Castro}{2011}]{arias2011clustering}
Arias{-}Castro, E. (2011).
\newblock Clustering based on pairwise distances when the data is of mixed
  dimensions.
\newblock {\em {IEEE} Trans. Inf. Theory}, 57(3):1692--1706.

\bibitem[\protect\astroncite{Avelar
  et~al.}{2019}]{DBLP:journals/corr/abs-1901-07984}
Avelar, P. H.~C., Lemos, H., Prates, M. O.~R., Gori, M., and Lamb, L.~C.
  (2019).
\newblock Typed graph networks.
\newblock {\em CoRR}, abs/1901.07984.

\bibitem[\protect\astroncite{Bai and Urtasun}{2017}]{DBLP:conf/cvpr/BaiU17}
Bai, M. and Urtasun, R. (2017).
\newblock Deep watershed transform for instance segmentation.
\newblock In {\em 2017 {IEEE} Conference on Computer Vision and Pattern
  Recognition, {CVPR} 2017, Honolulu, HI, USA, July 21-26, 2017}, pages
  2858--2866. {IEEE} Computer Society.

\bibitem[\protect\astroncite{Bainbridge and Bell}{2001}]{omr}
Bainbridge, D. and Bell, T. (2001).
\newblock The challenge of optical music recognition.
\newblock {\em Comput. Humanit.}, 35(2):95--121.

\bibitem[\protect\astroncite{Baro et~al.}{2016}]{DBLP:conf/icfhr/BaroRF16}
Baro, A., Riba, P., and Forn{\'{e}}s, A. (2016).
\newblock Towards the recognition of compound music notes in handwritten music
  scores.
\newblock In {\em 15th International Conference on Frontiers in Handwriting
  Recognition, {ICFHR} 2016, Shenzhen, China, October 23-26, 2016}, pages
  465--470. {IEEE} Computer Society.

\bibitem[\protect\astroncite{Basu
  et~al.}{2002}]{Basu02semi-supervisedclustering}
Basu, S., Banerjee, A., and Mooney, R.~J. (2002).
\newblock Semi-supervised clustering by seeding.
\newblock In Sammut, C. and Hoffmann, A.~G., editors, {\em Machine Learning,
  Proceedings of the Nineteenth International Conference {(ICML} 2002),
  University of New South Wales, Sydney, Australia, July 8-12, 2002}, pages
  27--34. Morgan Kaufmann.

\bibitem[\protect\astroncite{Battaglia
  et~al.}{2018}]{DBLP:journals/corr/abs-1806-01261}
Battaglia, P.~W., Hamrick, J.~B., Bapst, V., Sanchez{-}Gonzalez, A., Zambaldi,
  V.~F., Malinowski, M., Tacchetti, A., Raposo, D., Santoro, A., Faulkner, R.,
  G{\"{u}}l{\c{c}}ehre, {\c{C}}., Song, H.~F., Ballard, A.~J., Gilmer, J.,
  Dahl, G.~E., Vaswani, A., Allen, K.~R., Nash, C., Langston, V., Dyer, C.,
  Heess, N., Wierstra, D., Kohli, P., Botvinick, M., Vinyals, O., Li, Y., and
  Pascanu, R. (2018).
\newblock Relational inductive biases, deep learning, and graph networks.
\newblock {\em CoRR}, abs/1806.01261.

\bibitem[\protect\astroncite{Bengio
  et~al.}{1994}]{DBLP:journals/tnn/BengioSF94}
Bengio, Y., Simard, P.~Y., and Frasconi, P. (1994).
\newblock Learning long-term dependencies with gradient descent is difficult.
\newblock {\em {IEEE} Trans. Neural Networks}, 5(2):157--166.

\bibitem[\protect\astroncite{Bergstra and
  Bengio}{2012}]{DBLP:journals/jmlr/BergstraB12}
Bergstra, J. and Bengio, Y. (2012).
\newblock Random search for hyper-parameter optimization.
\newblock {\em J. Mach. Learn. Res.}, 13:281--305.

\bibitem[\protect\astroncite{Berthelot
  et~al.}{2019}]{DBLP:conf/nips/BerthelotCGPOR19}
Berthelot, D., Carlini, N., Goodfellow, I.~J., Papernot, N., Oliver, A., and
  Raffel, C. (2019).
\newblock Mixmatch: {A} holistic approach to semi-supervised learning.
\newblock In Wallach, H.~M., Larochelle, H., Beygelzimer, A.,
  d'Alch{\'{e}}{-}Buc, F., Fox, E.~B., and Garnett, R., editors, {\em Advances
  in Neural Information Processing Systems 32: Annual Conference on Neural
  Information Processing Systems 2019, NeurIPS 2019, 8-14 December 2019,
  Vancouver, BC, Canada}, pages 5050--5060.

\bibitem[\protect\astroncite{Beucher et~al.}{1992}]{beucher1992watershed}
Beucher, S. et~al. (1992).
\newblock The watershed transformation applied to image segmentation.
\newblock {\em SCANNING MICROSCOPY-SUPPLEMENT-}, pages 299--299.

\bibitem[\protect\astroncite{Branson
  et~al.}{2014}]{DBLP:journals/ijcv/BransonHWPB14}
Branson, S., Horn, G.~V., Wah, C., Perona, P., and Belongie, S.~J. (2014).
\newblock The ignorant led by the blind: {A} hybrid human-machine vision system
  for fine-grained categorization.
\newblock {\em Int. J. Comput. Vis.}, 108(1-2):3--29.

\bibitem[\protect\astroncite{Bras{\'{o}} and
  Leal{-}Taix{\'{e}}}{2020}]{DBLP:journals/corr/abs-1912-07515}
Bras{\'{o}}, G. and Leal{-}Taix{\'{e}}, L. (2020).
\newblock Learning a neural solver for multiple object tracking.
\newblock In {\em 2020 {IEEE} Conference on Computer Vision and Pattern
  Recognition, {CVPR} 2020}. {IEEE} Computer Society.

\bibitem[\protect\astroncite{Bromley et~al.}{1993}]{bromley1994signature}
Bromley, J., Guyon, I., LeCun, Y., S{\"{a}}ckinger, E., and Shah, R. (1993).
\newblock Signature verification using a siamese time delay neural network.
\newblock In Cowan, J.~D., Tesauro, G., and Alspector, J., editors, {\em
  Advances in Neural Information Processing Systems 6, [7th {NIPS} Conference,
  Denver, Colorado, USA, 1993]}, pages 737--744. Morgan Kaufmann.

\bibitem[\protect\astroncite{Bruna
  et~al.}{2014}]{DBLP:journals/corr/BrunaZSL13}
Bruna, J., Zaremba, W., Szlam, A., and LeCun, Y. (2014).
\newblock Spectral networks and locally connected networks on graphs.
\newblock In Bengio, Y. and LeCun, Y., editors, {\em 2nd International
  Conference on Learning Representations, {ICLR} 2014, Banff, AB, Canada, April
  14-16, 2014, Conference Track Proceedings}.

\bibitem[\protect\astroncite{{\c{C}}akir
  et~al.}{2019}]{DBLP:conf/cvpr/Cakir0XKS19}
{\c{C}}akir, F., He, K., Xia, X., Kulis, B., and Sclaroff, S. (2019).
\newblock Deep metric learning to rank.
\newblock In {\em {IEEE} Conference on Computer Vision and Pattern Recognition,
  {CVPR} 2019, Long Beach, CA, USA, June 16-20, 2019}, pages 1861--1870.
  Computer Vision Foundation / {IEEE}.

\bibitem[\protect\astroncite{Calvo{-}Zaragoza and Oncina}{2014}]{Zaragoza2014}
Calvo{-}Zaragoza, J. and Oncina, J. (2014).
\newblock Recognition of pen-based music notation: The {HOMUS} dataset.
\newblock In {\em 22nd International Conference on Pattern Recognition, {ICPR}
  2014, Stockholm, Sweden, August 24-28, 2014}, pages 3038--3043. {IEEE}
  Computer Society.

\bibitem[\protect\astroncite{Calvo{-}Zaragoza et~al.}{2016}]{Zaragoza2016}
Calvo{-}Zaragoza, J., Rizo, D., and Quereda, J. M.~I. (2016).
\newblock Two (note) heads are better than one: Pen-based multimodal
  interaction with music scores.
\newblock In Mandel, M.~I., Devaney, J., Turnbull, D., and Tzanetakis, G.,
  editors, {\em Proceedings of the 17th International Society for Music
  Information Retrieval Conference, {ISMIR} 2016, New York City, United States,
  August 7-11, 2016}, pages 509--514.

\bibitem[\protect\astroncite{Calvo{-}Zaragoza
  et~al.}{2017}]{DBLP:conf/ismir/Calvo-ZaragozaV17}
Calvo{-}Zaragoza, J., Valero{-}Mas, J.~J., and Pertusa, A. (2017).
\newblock End-to-end optical music recognition using neural networks.
\newblock In Cunningham, S.~J., Duan, Z., Hu, X., and Turnbull, D., editors,
  {\em Proceedings of the 18th International Society for Music Information
  Retrieval Conference, {ISMIR} 2017, Suzhou, China, October 23-27, 2017},
  pages 472--477.

\bibitem[\protect\astroncite{Chin et~al.}{2010}]{chin2010novel}
Chin, C., Shih, A.~C., and Fan, K. (2010).
\newblock A novel spectral clustering method based on pairwise distance matrix.
\newblock {\em J. Inf. Sci. Eng.}, 26(2):649--658.

\bibitem[\protect\astroncite{Choi et~al.}{2020}]{jiwoong}
Choi, J., Elezi, I., Lee, H.-J., Farabet, C., and Alvarez, J. (2020).
\newblock Deep active learning for object detection with mixture density
  networks.

\bibitem[\protect\astroncite{Choi et~al.}{2017}]{DBLP:conf/icdar/ChoiCRZ17}
Choi, K., Co{\"{u}}asnon, B., Ricquebourg, Y., and Zanibbi, R. (2017).
\newblock Bootstrapping samples of accidentals in dense piano scores for
  cnn-based detection.
\newblock In {\em 12th International Workshop on Graphics Recognitio, 14th
  {IAPR} International Conference on Document Analysis and Recognition,
  GREC@ICDAR 2017, Kyoto, Japan, November 9-15, 2017}, pages 19--20. {IEEE}.

\bibitem[\protect\astroncite{Chopra et~al.}{2005}]{DBLP:conf/cvpr/ChopraHL05}
Chopra, S., Hadsell, R., and LeCun, Y. (2005).
\newblock Learning a similarity metric discriminatively, with application to
  face verification.
\newblock In {\em 2005 {IEEE} Computer Society Conference on Computer Vision
  and Pattern Recognition {(CVPR} 2005), 20-26 June 2005, San Diego, CA,
  {USA}}, pages 539--546. {IEEE} Computer Society.

\bibitem[\protect\astroncite{Ciresan et~al.}{2012a}]{Ciresan2012isbi}
Ciresan, D.~C., Giusti, A., Gambardella, L.~M., and Schmidhuber, J. (2012a).
\newblock Neural networks for segmenting neuronal structures in {EM} stacks.
\newblock In {\em ISBI Segmentation Challenge Competition: Abstracts}.

\bibitem[\protect\astroncite{Ciresan
  et~al.}{2012b}]{DBLP:conf/cvpr/CiresanMS12}
Ciresan, D.~C., Meier, U., and Schmidhuber, J. (2012b).
\newblock Multi-column deep neural networks for image classification.
\newblock In {\em 2012 {IEEE} Conference on Computer Vision and Pattern
  Recognition, Providence, RI, USA, June 16-21, 2012}, pages 3642--3649. {IEEE}
  Computer Society.

\bibitem[\protect\astroncite{Cortes and
  Vapnik}{1995}]{DBLP:journals/ml/CortesV95}
Cortes, C. and Vapnik, V. (1995).
\newblock Support-vector networks.
\newblock {\em Mach. Learn.}, 20(3):273--297.

\bibitem[\protect\astroncite{Deng et~al.}{2009}]{DBLP:conf/cvpr/DengDSLL009}
Deng, J., Dong, W., Socher, R., Li, L., Li, K., and Li, F. (2009).
\newblock Imagenet: {A} large-scale hierarchical image database.
\newblock In {\em 2009 {IEEE} Computer Society Conference on Computer Vision
  and Pattern Recognition {(CVPR} 2009), 20-25 June 2009, Miami, Florida,
  {USA}}, pages 248--255. {IEEE} Computer Society.

\bibitem[\protect\astroncite{Duan et~al.}{2019}]{DDBLP:conf/cvpr/Duan2019}
Duan, Y., Chen, L., Lu, J., and Zhou, J. (2019).
\newblock Deep embedding learning with discriminative sampling policy.
\newblock In {\em {IEEE} Conference on Computer Vision and Pattern Recognition,
  {CVPR} 2019, Long Beach, CA, USA, June 16-20, 2019}, pages 4964--4973.
  Computer Vision Foundation / {IEEE}.

\bibitem[\protect\astroncite{Duvenaud
  et~al.}{2015}]{DBLP:conf/nips/DuvenaudMABHAA15}
Duvenaud, D., Maclaurin, D., Aguilera{-}Iparraguirre, J.,
  G{\'{o}}mez{-}Bombarelli, R., Hirzel, T., Aspuru{-}Guzik, A., and Adams,
  R.~P. (2015).
\newblock Convolutional networks on graphs for learning molecular fingerprints.
\newblock In Cortes, C., Lawrence, N.~D., Lee, D.~D., Sugiyama, M., and
  Garnett, R., editors, {\em Advances in Neural Information Processing Systems
  28: Annual Conference on Neural Information Processing Systems 2015, December
  7-12, 2015, Montreal, Quebec, Canada}, pages 2224--2232.

\bibitem[\protect\astroncite{Elezi et~al.}{2018a}]{DBLP:conf/icpr/EleziTVP18}
Elezi, I., Torcinovich, A., Vascon, S., and Pelillo, M. (2018a).
\newblock Transductive label augmentation for improved deep network learning.
\newblock In {\em 24th International Conference on Pattern Recognition, {ICPR}
  2018, Beijing, China, August 20-24, 2018}, pages 1432--1437. {IEEE} Computer
  Society.

\bibitem[\protect\astroncite{Elezi
  et~al.}{2018b}]{DBLP:journals/corr/abs-1810-05423}
Elezi, I., Tuggener, L., Pelillo, M., and Stadelmann, T. (2018b).
\newblock Deepscores and deep watershed detection: current state and open
  issues.
\newblock {\em CoRR}, abs/1810.05423.

\bibitem[\protect\astroncite{Elezi et~al.}{2019}]{DBLP:journals/corr/groupLoss}
Elezi, I., Vascon, S., Torcinovich, A., Pelillo, M., and Leal{-}Taix{\'{e}}, L.
  (2019).
\newblock The group loss for deep metric learning.
\newblock {\em CoRR}, abs/1912.00385.

\bibitem[\protect\astroncite{Elman}{1990}]{DBLP:journals/cogsci/Elman90}
Elman, J.~L. (1990).
\newblock Finding structure in time.
\newblock {\em Cogn. Sci.}, 14(2):179--211.

\bibitem[\protect\astroncite{Erdem and
  Pelillo}{2012}]{DBLP:journals/neco/ErdemP12}
Erdem, A. and Pelillo, M. (2012).
\newblock Graph transduction as a noncooperative game.
\newblock {\em Neural Computation}, 24(3):700--723.

\bibitem[\protect\astroncite{Ester et~al.}{1996}]{ester1996density}
Ester, M., Kriegel, H., Sander, J., and Xu, X. (1996).
\newblock A density-based algorithm for discovering clusters in large spatial
  databases with noise.
\newblock In Simoudis, E., Han, J., and Fayyad, U.~M., editors, {\em
  Proceedings of the Second International Conference on Knowledge Discovery and
  Data Mining (KDD-96), Portland, Oregon, {USA}}, pages 226--231. {AAAI} Press.

\bibitem[\protect\astroncite{Everingham
  et~al.}{2010}]{DBLP:journals/ijcv/EveringhamGWWZ10}
Everingham, M., Gool, L.~V., Williams, C. K.~I., Winn, J.~M., and Zisserman, A.
  (2010).
\newblock The pascal visual object classes {(VOC)} challenge.
\newblock {\em Int. J. Comput. Vis.}, 88(2):303--338.

\bibitem[\protect\astroncite{Forn{\'{e}}s et~al.}{2012}]{Fornes2012}
Forn{\'{e}}s, A., Dutta, A., Gordo, A., and Llad{\'{o}}s, J. (2012).
\newblock {CVC-MUSCIMA:} a ground truth of handwritten music score images for
  writer identification and staff removal.
\newblock {\em {IJDAR}}, 15(3):243--251.

\bibitem[\protect\astroncite{Frasconi
  et~al.}{1998}]{DBLP:journals/tnn/FrasconiGS98}
Frasconi, P., Gori, M., and Sperduti, A. (1998).
\newblock A general framework for adaptive processing of data structures.
\newblock {\em {IEEE} Trans. Neural Networks}, 9(5):768--786.

\bibitem[\protect\astroncite{Fukushima and
  Miyake}{1982}]{DBLP:journals/pr/FukushimaM82}
Fukushima, K. and Miyake, S. (1982).
\newblock Neocognitron: {A} new algorithm for pattern recognition tolerant of
  deformations and shifts in position.
\newblock {\em Pattern Recognit.}, 15(6):455--469.

\bibitem[\protect\astroncite{Gal and Ghahramani}{2016}]{gal2016dropout}
Gal, Y. and Ghahramani, Z. (2016).
\newblock Dropout as a bayesian approximation: Representing model uncertainty
  in deep learning.
\newblock In Balcan, M. and Weinberger, K.~Q., editors, {\em Proceedings of the
  33nd International Conference on Machine Learning, {ICML} 2016, New York
  City, NY, USA, June 19-24, 2016}, volume~48 of {\em {JMLR} Workshop and
  Conference Proceedings}, pages 1050--1059. JMLR.org.

\bibitem[\protect\astroncite{Gallego and
  Calvo{-}Zaragoza}{2017}]{DBLP:journals/eswa/GallegoC17}
Gallego, A. and Calvo{-}Zaragoza, J. (2017).
\newblock Staff-line removal with selectional auto-encoders.
\newblock {\em Expert Syst. Appl.}, 89:138--148.

\bibitem[\protect\astroncite{Garofolo et~al.}{1993}]{timit}
Garofolo, J.~S., Lamel, L.~F., Fisher, W.~M., Fiscus, J.~G., Pallett, D.~S.,
  and Dahlgren, N.~L. (1993).
\newblock Darpa timit acoustic phonetic continuous speech corpus {CDROM}.

\bibitem[\protect\astroncite{Ge et~al.}{2018}]{DBLP:conf/eccv/GeHDS18}
Ge, W., Huang, W., Dong, D., and Scott, M.~R. (2018).
\newblock Deep metric learning with hierarchical triplet loss.
\newblock In Ferrari, V., Hebert, M., Sminchisescu, C., and Weiss, Y., editors,
  {\em Computer Vision - {ECCV} 2018 - 15th European Conference, Munich,
  Germany, September 8-14, 2018, Proceedings, Part {VI}}, volume 11210 of {\em
  Lecture Notes in Computer Science}, pages 272--288. Springer.

\bibitem[\protect\astroncite{Gebru et~al.}{2017}]{DBLP:conf/iccv/GebruHF17}
Gebru, T., Hoffman, J., and Fei{-}Fei, L. (2017).
\newblock Fine-grained recognition in the wild: {A} multi-task domain
  adaptation approach.
\newblock In {\em {IEEE} International Conference on Computer Vision, {ICCV}
  2017, Venice, Italy, October 22-29, 2017}, pages 1358--1367. {IEEE} Computer
  Society.

\bibitem[\protect\astroncite{Gilmer et~al.}{2017}]{DBLP:conf/icml/GilmerSRVD17}
Gilmer, J., Schoenholz, S.~S., Riley, P.~F., Vinyals, O., and Dahl, G.~E.
  (2017).
\newblock Neural message passing for quantum chemistry.
\newblock In Precup, D. and Teh, Y.~W., editors, {\em Proceedings of the 34th
  International Conference on Machine Learning, {ICML} 2017, Sydney, NSW,
  Australia, 6-11 August 2017}, volume~70 of {\em Proceedings of Machine
  Learning Research}, pages 1263--1272. {PMLR}.

\bibitem[\protect\astroncite{Girshick et~al.}{2018}]{Detectron2018}
Girshick, R., Radosavovic, I., Gkioxari, G., Doll\'{a}r, P., and He, K. (2018).
\newblock Detectron.
\newblock \url{https://github.com/facebookresearch/detectron}.

\bibitem[\protect\astroncite{Girshick
  et~al.}{2014}]{DBLP:conf/cvpr/GirshickDDM14}
Girshick, R.~B., Donahue, J., Darrell, T., and Malik, J. (2014).
\newblock Rich feature hierarchies for accurate object detection and semantic
  segmentation.
\newblock In {\em 2014 {IEEE} Conference on Computer Vision and Pattern
  Recognition, {CVPR} 2014, Columbus, OH, USA, June 23-28, 2014}, pages
  580--587. {IEEE} Computer Society.

\bibitem[\protect\astroncite{Goodfellow
  et~al.}{2016}]{DBLP:books/daglib/0040158}
Goodfellow, I.~J., Bengio, Y., and Courville, A.~C. (2016).
\newblock {\em Deep Learning}.
\newblock Adaptive computation and machine learning. {MIT} Press.

\bibitem[\protect\astroncite{Goodfellow et~al.}{2014}]{Goodfellow2013}
Goodfellow, I.~J., Bulatov, Y., Ibarz, J., Arnoud, S., and Shet, V.~D. (2014).
\newblock Multi-digit number recognition from street view imagery using deep
  convolutional neural networks.
\newblock In Bengio, Y. and LeCun, Y., editors, {\em 2nd International
  Conference on Learning Representations, {ICLR} 2014, Banff, AB, Canada, April
  14-16, 2014, Conference Track Proceedings}.

\bibitem[\protect\astroncite{Greff et~al.}{2017}]{DBLP:conf/nips/GreffSS17}
Greff, K., van Steenkiste, S., and Schmidhuber, J. (2017).
\newblock Neural expectation maximization.
\newblock In Guyon, I., von Luxburg, U., Bengio, S., Wallach, H.~M., Fergus,
  R., Vishwanathan, S. V.~N., and Garnett, R., editors, {\em Advances in Neural
  Information Processing Systems 30: Annual Conference on Neural Information
  Processing Systems 2017, 4-9 December 2017, Long Beach, CA, {USA}}, pages
  6691--6701.

\bibitem[\protect\astroncite{Gregory et~al.}{2007}]{caltech}
Gregory, G., Alex, H., and Pietro, P. (2007).
\newblock Caltech-256 object category dataset.
\newblock {\em Technical Report - California Institute of Technology}.

\bibitem[\protect\astroncite{Guo et~al.}{2017}]{DBLP:conf/icml/GuoPSW17}
Guo, C., Pleiss, G., Sun, Y., and Weinberger, K.~Q. (2017).
\newblock On calibration of modern neural networks.
\newblock In Precup, D. and Teh, Y.~W., editors, {\em Proceedings of the 34th
  International Conference on Machine Learning, {ICML} 2017, Sydney, NSW,
  Australia, 6-11 August 2017}, volume~70 of {\em Proceedings of Machine
  Learning Research}, pages 1321--1330. {PMLR}.

\bibitem[\protect\astroncite{Hajic and
  Pecina}{2017}]{DBLP:conf/icdar/HajicP17b}
Hajic, J. and Pecina, P. (2017).
\newblock The {MUSCIMA++} dataset for handwritten optical music recognition.
\newblock In {\em 14th {IAPR} International Conference on Document Analysis and
  Recognition, {ICDAR} 2017, Kyoto, Japan, November 9-15, 2017}, pages 39--46.
  {IEEE}.

\bibitem[\protect\astroncite{H{\"{a}}usser
  et~al.}{2017}]{DBLP:conf/cvpr/HausserMC17}
H{\"{a}}usser, P., Mordvintsev, A., and Cremers, D. (2017).
\newblock Learning by association - {A} versatile semi-supervised training
  method for neural networks.
\newblock In {\em 2017 {IEEE} Conference on Computer Vision and Pattern
  Recognition, {CVPR} 2017, Honolulu, HI, USA, July 21-26, 2017}, pages
  626--635. {IEEE} Computer Society.

\bibitem[\protect\astroncite{He et~al.}{2018a}]{DBLP:conf/cvpr/0003CBS18}
He, K., {\c{C}}akir, F., Bargal, S.~A., and Sclaroff, S. (2018a).
\newblock Hashing as tie-aware learning to rank.
\newblock In {\em 2018 {IEEE} Conference on Computer Vision and Pattern
  Recognition, {CVPR} 2018, Salt Lake City, UT, USA, June 18-22, 2018}, pages
  4023--4032. {IEEE} Computer Society.

\bibitem[\protect\astroncite{He et~al.}{2017}]{DBLP:conf/iccv/HeGDG17}
He, K., Gkioxari, G., Doll{\'{a}}r, P., and Girshick, R.~B. (2017).
\newblock Mask {R-CNN}.
\newblock In {\em {IEEE} International Conference on Computer Vision, {ICCV}
  2017, Venice, Italy, October 22-29, 2017}, pages 2980--2988. {IEEE} Computer
  Society.

\bibitem[\protect\astroncite{He et~al.}{2018b}]{DBLP:conf/cvpr/0003LS18}
He, K., Lu, Y., and Sclaroff, S. (2018b).
\newblock Local descriptors optimized for average precision.
\newblock In {\em 2018 {IEEE} Conference on Computer Vision and Pattern
  Recognition, {CVPR} 2018, Salt Lake City, UT, USA, June 18-22, 2018}, pages
  596--605. {IEEE} Computer Society.

\bibitem[\protect\astroncite{He et~al.}{2016}]{DBLP:conf/cvpr/HeZRS16}
He, K., Zhang, X., Ren, S., and Sun, J. (2016).
\newblock Deep residual learning for image recognition.
\newblock In {\em 2016 {IEEE} Conference on Computer Vision and Pattern
  Recognition, {CVPR} 2016, Las Vegas, NV, USA, June 27-30, 2016}, pages
  770--778. {IEEE} Computer Society.

\bibitem[\protect\astroncite{Hochreiter}{1990}]{Hochreiter:90}
Hochreiter, S. (1990).
\newblock Implementierung und anwendung eines `neuronalen'
  echtzeit-lernalgorithmus f\"{u}r reaktive umgebungen.
  fortgeschrittenenpraktikum, institut f\"{u}r informatik, lehrstuhl prof.
  brauer, technische universit\"{a}t m\"{u}nchen.

\bibitem[\protect\astroncite{Hochreiter and
  Schmidhuber}{1997}]{hochreiter1997long}
Hochreiter, S. and Schmidhuber, J. (1997).
\newblock Long short-term memory.
\newblock {\em Neural Computation}, 9(8):1735--1780.

\bibitem[\protect\astroncite{Hoffer and Ailon}{2015}]{hoffer2015deep}
Hoffer, E. and Ailon, N. (2015).
\newblock Deep metric learning using triplet network.
\newblock In Feragen, A., Pelillo, M., and Loog, M., editors, {\em
  Similarity-Based Pattern Recognition - Third International Workshop, {SIMBAD}
  2015, Copenhagen, Denmark, October 12-14, 2015, Proceedings}, volume 9370 of
  {\em Lecture Notes in Computer Science}, pages 84--92. Springer.

\bibitem[\protect\astroncite{Hornik}{1991}]{Hornik}
Hornik, K. (1991).
\newblock Approximation capabilities of multilayer feedforward networks.
\newblock {\em Neural Networks}, 4(2):251--257.

\bibitem[\protect\astroncite{Hsu et~al.}{2018}]{hsu2018learning}
Hsu, Y., Lv, Z., and Kira, Z. (2018).
\newblock Learning to cluster in order to transfer across domains and tasks.
\newblock In {\em 6th International Conference on Learning Representations,
  {ICLR} 2018, Vancouver, BC, Canada, April 30 - May 3, 2018, Conference Track
  Proceedings}. OpenReview.net.

\bibitem[\protect\astroncite{Huang et~al.}{2017}]{DBLP:conf/cvpr/HuangLMW17}
Huang, G., Liu, Z., van~der Maaten, L., and Weinberger, K.~Q. (2017).
\newblock Densely connected convolutional networks.
\newblock In {\em 2017 {IEEE} Conference on Computer Vision and Pattern
  Recognition, {CVPR} 2017, Honolulu, HI, USA, July 21-26, 2017}, pages
  2261--2269. {IEEE} Computer Society.

\bibitem[\protect\astroncite{Hummel and
  Zucker}{1983}]{DBLP:journals/pami/HummelZ83}
Hummel, R.~A. and Zucker, S.~W. (1983).
\newblock On the foundations of relaxation labeling processes.
\newblock {\em {IEEE} Trans. Pattern Anal. Mach. Intell.}, 5(3):267--287.

\bibitem[\protect\astroncite{hyun Lee}{2013}]{Lee_pseudo-label:the}
hyun Lee, D. (2013).
\newblock Pseudo-label: The simple and efficient semi-supervised learning
  method for deep neural networks.
\newblock In {\em Workshop on Challenges in Representation Learning (ICML)},
  volume~2, page~3.

\bibitem[\protect\astroncite{Ioffe and Szegedy}{2015}]{DBLP:conf/icml/IoffeS15}
Ioffe, S. and Szegedy, C. (2015).
\newblock Batch normalization: Accelerating deep network training by reducing
  internal covariate shift.
\newblock In Bach, F.~R. and Blei, D.~M., editors, {\em Proceedings of the 32nd
  International Conference on Machine Learning, {ICML} 2015, Lille, France,
  6-11 July 2015}, volume~37 of {\em {JMLR} Workshop and Conference
  Proceedings}, pages 448--456. JMLR.org.

\bibitem[\protect\astroncite{J{\'{e}}gou
  et~al.}{2011}]{DBLP:journals/pami/JegouDS11}
J{\'{e}}gou, H., Douze, M., and Schmid, C. (2011).
\newblock Product quantization for nearest neighbor search.
\newblock {\em {IEEE} Trans. Pattern Anal. Mach. Intell.}, 33(1):117--128.

\bibitem[\protect\astroncite{Jr. and
  Pecina}{2017a}]{DBLP:journals/corr/abs-1708-01806}
Jr., J.~H. and Pecina, P. (2017a).
\newblock Detecting noteheads in handwritten scores with convnets and bounding
  box regression.
\newblock {\em CoRR}, abs/1708.01806.

\bibitem[\protect\astroncite{Jr. and Pecina}{2017b}]{Hajic2017}
Jr., J.~H. and Pecina, P. (2017b).
\newblock In search of a dataset for handwritten optical music recognition:
  Introducing {MUSCIMA++}.
\newblock {\em CoRR}, abs/1703.04824.

\bibitem[\protect\astroncite{Kadar and
  Ben{-}Shahar}{2014}]{DBLP:conf/eccv/KadarB14}
Kadar, I. and Ben{-}Shahar, O. (2014).
\newblock Scenenet: {A} perceptual ontology for scene understanding.
\newblock In Agapito, L., Bronstein, M.~M., and Rother, C., editors, {\em
  Computer Vision - {ECCV} 2014 Workshops - Zurich, Switzerland, September 6-7
  and 12, 2014, Proceedings, Part {II}}, volume 8926 of {\em Lecture Notes in
  Computer Science}, pages 385--400. Springer.

\bibitem[\protect\astroncite{Kampffmeyer et~al.}{2017}]{kampffmeyer2017}
Kampffmeyer, M., L{\o}kse, S., Bianchi, F.~M., Livi, L., Salberg, A., and
  Jenssen, R. (2017).
\newblock Deep divergence-based clustering.
\newblock In Ueda, N., Watanabe, S., Matsui, T., Chien, J., and Larsen, J.,
  editors, {\em 27th {IEEE} International Workshop on Machine Learning for
  Signal Processing, {MLSP} 2017, Tokyo, Japan, September 25-28, 2017}, pages
  1--6. {IEEE}.

\bibitem[\protect\astroncite{Karpathy}{2016}]{KarpathyThesis}
Karpathy, A. (2016).
\newblock {\em Connecting images and natural language}.
\newblock PhD thesis, Stanford, CA, USA.

\bibitem[\protect\astroncite{Kaufman and Rousseeuw}{1990}]{kaufman1990finding}
Kaufman, L. and Rousseeuw, P.~J. (1990).
\newblock {\em Finding Groups in Data: An Introduction to Cluster Analysis}.
\newblock John Wiley.

\bibitem[\protect\astroncite{Keskar et~al.}{2017}]{DBLP:conf/iclr/KeskarMNST17}
Keskar, N.~S., Mudigere, D., Nocedal, J., Smelyanskiy, M., and Tang, P. T.~P.
  (2017).
\newblock On large-batch training for deep learning: Generalization gap and
  sharp minima.
\newblock In {\em 5th International Conference on Learning Representations,
  {ICLR} 2017, Toulon, France, April 24-26, 2017, Conference Track
  Proceedings}. OpenReview.net.

\bibitem[\protect\astroncite{Kiefer and Wolfowitz}{1952}]{Kiefer}
Kiefer, J. and Wolfowitz, J. (1952).
\newblock Stochastic estimation of the maximum of a regression function.
\newblock {\em Ann. Math. Statist}, 23(3):462--466.

\bibitem[\protect\astroncite{Kim and Song}{2011}]{jin2011expectation}
Kim, H. and Song, H.~Y. (2011).
\newblock Daily life mobility of a student: From position data to human
  mobility model through expectation maximization clustering.
\newblock In Kim, T., Adeli, H., Grosky, W.~I., Pissinou, N., Shih, T.~K.,
  Rothwell, E.~J., Kang, B.~H., and Shin, S., editors, {\em Multimedia,
  Computer Graphics and Broadcasting - International Conference, MulGraB 2011,
  Held as Part of the Future Generation Information Technology Conference,
  {FGIT} 2011, in Conjunction with {GDC} 2011, Jeju Island, Korea, December
  8-10, 2011. Proceedings, Part {II}}, volume 263 of {\em Communications in
  Computer and Information Science}, pages 88--97. Springer.

\bibitem[\protect\astroncite{Kim et~al.}{2018}]{DBLP:conf/eccv/KimGCLK18}
Kim, W., Goyal, B., Chawla, K., Lee, J., and Kwon, K. (2018).
\newblock Attention-based ensemble for deep metric learning.
\newblock In Ferrari, V., Hebert, M., Sminchisescu, C., and Weiss, Y., editors,
  {\em Computer Vision - {ECCV} 2018 - 15th European Conference, Munich,
  Germany, September 8-14, 2018, Proceedings, Part {I}}, volume 11205 of {\em
  Lecture Notes in Computer Science}, pages 760--777. Springer.

\bibitem[\protect\astroncite{Kingma and Ba}{2015}]{DBLP:conf/iclr/Kingma14}
Kingma, D.~P. and Ba, J. (2015).
\newblock Adam: {A} method for stochastic optimization.
\newblock In Bengio, Y. and LeCun, Y., editors, {\em 3rd International
  Conference on Learning Representations, {ICLR} 2015, San Diego, CA, USA, May
  7-9, 2015, Conference Track Proceedings}.

\bibitem[\protect\astroncite{Kingma et~al.}{2014}]{DBLP:conf/nips/KingmaMRW14}
Kingma, D.~P., Mohamed, S., Rezende, D.~J., and Welling, M. (2014).
\newblock Semi-supervised learning with deep generative models.
\newblock In Ghahramani, Z., Welling, M., Cortes, C., Lawrence, N.~D., and
  Weinberger, K.~Q., editors, {\em Advances in Neural Information Processing
  Systems 27: Annual Conference on Neural Information Processing Systems 2014,
  December 8-13 2014, Montreal, Quebec, Canada}, pages 3581--3589.

\bibitem[\protect\astroncite{Kipf and Welling}{2017}]{DBLP:conf/iclr/KipfW17}
Kipf, T.~N. and Welling, M. (2017).
\newblock Semi-supervised classification with graph convolutional networks.
\newblock In {\em 5th International Conference on Learning Representations,
  {ICLR} 2017, Toulon, France, April 24-26, 2017, Conference Track
  Proceedings}. OpenReview.net.

\bibitem[\protect\astroncite{Krause
  et~al.}{2013}]{KrauseStarkDengFei-Fei_3DRR2013}
Krause, J., Stark, M., Deng, J., and Fei{-}Fei, L. (2013).
\newblock 3d object representations for fine-grained categorization.
\newblock In {\em 2013 {IEEE} International Conference on Computer Vision
  Workshops, {ICCV} Workshops 2013, Sydney, Australia, December 1-8, 2013},
  pages 554--561. {IEEE} Computer Society.

\bibitem[\protect\astroncite{Krizhevsky and Hinton}{2009}]{Krizhevsky2009}
Krizhevsky, A. and Hinton, G. (2009).
\newblock Learning multiple layers of features from tiny images.

\bibitem[\protect\astroncite{Krizhevsky
  et~al.}{2012}]{DBLP:conf/nips/KrizhevskySH12}
Krizhevsky, A., Sutskever, I., and Hinton, G.~E. (2012).
\newblock Imagenet classification with deep convolutional neural networks.
\newblock In Bartlett, P.~L., Pereira, F. C.~N., Burges, C. J.~C., Bottou, L.,
  and Weinberger, K.~Q., editors, {\em Advances in Neural Information
  Processing Systems 25: 26th Annual Conference on Neural Information
  Processing Systems 2012. Proceedings of a meeting held December 3-6, 2012,
  Lake Tahoe, Nevada, United States}, pages 1106--1114.

\bibitem[\protect\astroncite{Laine and Aila}{2017}]{DBLP:conf/iclr/LaineA17}
Laine, S. and Aila, T. (2017).
\newblock Temporal ensembling for semi-supervised learning.
\newblock In {\em 5th International Conference on Learning Representations,
  {ICLR} 2017, Toulon, France, April 24-26, 2017, Conference Track
  Proceedings}. OpenReview.net.

\bibitem[\protect\astroncite{Law et~al.}{2017}]{DBLP:conf/icml/LawUZ17}
Law, M.~T., Urtasun, R., and Zemel, R.~S. (2017).
\newblock Deep spectral clustering learning.
\newblock In Precup, D. and Teh, Y.~W., editors, {\em Proceedings of the 34th
  International Conference on Machine Learning, {ICML} 2017, Sydney, NSW,
  Australia, 6-11 August 2017}, volume~70 of {\em Proceedings of Machine
  Learning Research}, pages 1985--1994. {PMLR}.

\bibitem[\protect\astroncite{Le et~al.}{2012}]{DBLP:conf/icml/LeRMDCCDN12}
Le, Q.~V., Ranzato, M., Monga, R., Devin, M., Corrado, G., Chen, K., Dean, J.,
  and Ng, A.~Y. (2012).
\newblock Building high-level features using large scale unsupervised learning.
\newblock In {\em Proceedings of the 29th International Conference on Machine
  Learning, {ICML} 2012, Edinburgh, Scotland, UK, June 26 - July 1, 2012}.
  icml.cc / Omnipress.

\bibitem[\protect\astroncite{LeCun
  et~al.}{2015}]{DBLP:journals/nature/LeCunBH15}
LeCun, Y., Bengio, Y., and Hinton, G.~E. (2015).
\newblock Deep learning.
\newblock {\em Nat.}, 521(7553):436--444.

\bibitem[\protect\astroncite{LeCun
  et~al.}{1989}]{DBLP:journals/neco/LeCunBDHHHJ89}
LeCun, Y., Boser, B.~E., Denker, J.~S., Henderson, D., Howard, R.~E., Hubbard,
  W.~E., and Jackel, L.~D. (1989).
\newblock Backpropagation applied to handwritten zip code recognition.
\newblock {\em Neural Computation}, 1(4):541--551.

\bibitem[\protect\astroncite{LeCun et~al.}{1990}]{LeCun1990}
LeCun, Y., Boser, B.~E., Denker, J.~S., Henderson, D., Howard, R.~E., Hubbard,
  W.~E., and Jackel, L.~D. (1990).
\newblock Handwritten digit recognition with a back-propagation network.
\newblock {\em In Advances in neural information processing systems}, pages
  396--404.

\bibitem[\protect\astroncite{LeCun et~al.}{1998a}]{lecun1998gradient}
LeCun, Y., Bottou, L., Bengio, Y., and Haffner, P. (1998a).
\newblock Gradient-based learning applied to document recognition.
\newblock {\em Proceedings of the IEEE}, 86(11):2278--2324.

\bibitem[\protect\astroncite{LeCun et~al.}{2012}]{DBLP:series/lncs/LeCunBOM12}
LeCun, Y., Bottou, L., Orr, G.~B., and M{\"{u}}ller, K. (2012).
\newblock Efficient backprop.
\newblock In Montavon, G., Orr, G.~B., and M{\"{u}}ller, K., editors, {\em
  Neural Networks: Tricks of the Trade - Second Edition}, volume 7700 of {\em
  Lecture Notes in Computer Science}, pages 9--48. Springer.

\bibitem[\protect\astroncite{LeCun et~al.}{1998b}]{Lecun1998}
LeCun, Y., Cortes, C., and Burges, C.~J. (1998b).
\newblock The mnist database of handwritten digits.

\bibitem[\protect\astroncite{Lee and Osindero}{2016}]{Lee16}
Lee, C. and Osindero, S. (2016).
\newblock Recursive recurrent nets with attention modeling for {OCR} in the
  wild.
\newblock In {\em 2016 {IEEE} Conference on Computer Vision and Pattern
  Recognition, {CVPR} 2016, Las Vegas, NV, USA, June 27-30, 2016}, pages
  2231--2239. {IEEE} Computer Society.

\bibitem[\protect\astroncite{Lee et~al.}{2017}]{Lee2017OnTA}
Lee, H., Ge, R., Ma, T., Risteski, A., and Arora, S. (2017).
\newblock On the ability of neural nets to express distributions.
\newblock In Kale, S. and Shamir, O., editors, {\em Proceedings of the 30th
  Conference on Learning Theory, {COLT} 2017, Amsterdam, The Netherlands, 7-10
  July 2017}, volume~65 of {\em Proceedings of Machine Learning Research},
  pages 1271--1296. {PMLR}.

\bibitem[\protect\astroncite{Levine et~al.}{2016}]{levine2016end}
Levine, S., Finn, C., Darrell, T., and Abbeel, P. (2016).
\newblock End-to-end training of deep visuomotor policies.
\newblock {\em J. Mach. Learn. Res.}, 17:39:1--39:40.

\bibitem[\protect\astroncite{Lin et~al.}{2017a}]{DBLP:conf/cvpr/LinMSR17}
Lin, G., Milan, A., Shen, C., and Reid, I.~D. (2017a).
\newblock Refinenet: Multi-path refinement networks for high-resolution
  semantic segmentation.
\newblock In {\em 2017 {IEEE} Conference on Computer Vision and Pattern
  Recognition, {CVPR} 2017, Honolulu, HI, USA, July 21-26, 2017}, pages
  5168--5177. {IEEE} Computer Society.

\bibitem[\protect\astroncite{Lin et~al.}{2017b}]{DBLP:conf/iccv/LinGGHD17}
Lin, T., Goyal, P., Girshick, R.~B., He, K., and Doll{\'{a}}r, P. (2017b).
\newblock Focal loss for dense object detection.
\newblock In {\em {IEEE} International Conference on Computer Vision, {ICCV}
  2017, Venice, Italy, October 22-29, 2017}, pages 2999--3007. {IEEE} Computer
  Society.

\bibitem[\protect\astroncite{Lin et~al.}{2014}]{DBLP:conf/eccv/LinMBHPRDZ14}
Lin, T., Maire, M., Belongie, S.~J., Hays, J., Perona, P., Ramanan, D.,
  Doll{\'{a}}r, P., and Zitnick, C.~L. (2014).
\newblock Microsoft {COCO:} common objects in context.
\newblock In Fleet, D.~J., Pajdla, T., Schiele, B., and Tuytelaars, T.,
  editors, {\em Computer Vision - {ECCV} 2014 - 13th European Conference,
  Zurich, Switzerland, September 6-12, 2014, Proceedings, Part {V}}, volume
  8693 of {\em Lecture Notes in Computer Science}, pages 740--755. Springer.

\bibitem[\protect\astroncite{Linnainmaa}{1976}]{Linnainmaa76}
Linnainmaa, S. (1976).
\newblock Taylor expansion of the accumulated rounding error.
\newblock {\em BIT Numerical Mathematics}, 16 (2):146--160.

\bibitem[\protect\astroncite{Liu et~al.}{2011}]{Liu2011}
Liu, C., Yin, F., Wang, D., and Wang, Q. (2011).
\newblock {CASIA} online and offline chinese handwriting databases.
\newblock In {\em 2011 International Conference on Document Analysis and
  Recognition, {ICDAR} 2011, Beijing, China, September 18-21, 2011}, pages
  37--41. {IEEE} Computer Society.

\bibitem[\protect\astroncite{Liu and Kubala}{2003}]{liu2003online}
Liu, D. and Kubala, F. (2003).
\newblock Online speaker clustering.
\newblock In {\em 2003 {IEEE} International Conference on Acoustics, Speech,
  and Signal Processing, {ICASSP} '03, Hong Kong, April 6-10, 2003}, pages
  572--575. {IEEE}.

\bibitem[\protect\astroncite{Liu et~al.}{2016}]{DBLP:conf/eccv/LiuAESRFB16}
Liu, W., Anguelov, D., Erhan, D., Szegedy, C., Reed, S.~E., Fu, C., and Berg,
  A.~C. (2016).
\newblock {SSD:} single shot multibox detector.
\newblock In Leibe, B., Matas, J., Sebe, N., and Welling, M., editors, {\em
  Computer Vision - {ECCV} 2016 - 14th European Conference, Amsterdam, The
  Netherlands, October 11-14, 2016, Proceedings, Part {I}}, volume 9905 of {\em
  Lecture Notes in Computer Science}, pages 21--37. Springer.

\bibitem[\protect\astroncite{Long et~al.}{2015}]{Long2015}
Long, J., Shelhamer, E., and Darrell, T. (2015).
\newblock Fully convolutional networks for semantic segmentation.
\newblock In {\em {IEEE} Conference on Computer Vision and Pattern Recognition,
  {CVPR} 2015, Boston, MA, USA, June 7-12, 2015}, pages 3431--3440. {IEEE}
  Computer Society.

\bibitem[\protect\astroncite{Loshchilov and
  Hutter}{2019}]{DBLP:conf/iclr/LoshchilovH19}
Loshchilov, I. and Hutter, F. (2019).
\newblock Decoupled weight decay regularization.
\newblock In {\em 7th International Conference on Learning Representations,
  {ICLR} 2019, New Orleans, LA, USA, May 6-9, 2019}. OpenReview.net.

\bibitem[\protect\astroncite{Lowe}{2004}]{DBLP:journals/ijcv/Lowe04}
Lowe, D.~G. (2004).
\newblock Distinctive image features from scale-invariant keypoints.
\newblock {\em Int. J. Comput. Vis.}, 60(2):91--110.

\bibitem[\protect\astroncite{Lukic et~al.}{2017}]{lukic2017speaker}
Lukic, Y., Vogt, C., Durr, O., and Stadelmann, T. (2017).
\newblock Learning embeddings for speaker clustering based on voice equality.
\newblock In Ueda, N., Watanabe, S., Matsui, T., Chien, J., and Larsen, J.,
  editors, {\em 27th {IEEE} International Workshop on Machine Learning for
  Signal Processing, {MLSP} 2017, Tokyo, Japan, September 25-28, 2017}, pages
  1--6. {IEEE}.

\bibitem[\protect\astroncite{MacQueen}{1967}]{MacQueen}
MacQueen, J. (1967).
\newblock Some methods for classification and analysis of multivariate
  observations.
\newblock In {\em Proc. Fifth Berkeley Symp. on Math. Statist. and Prob., Vol.
  1}, pages 281--297.

\bibitem[\protect\astroncite{Manmatha
  et~al.}{2017}]{DBLP:conf/iccv/ManmathaWSK17}
Manmatha, R., Wu, C., Smola, A.~J., and Kr{\"{a}}henb{\"{u}}hl, P. (2017).
\newblock Sampling matters in deep embedding learning.
\newblock In {\em {IEEE} International Conference on Computer Vision, {ICCV}
  2017, Venice, Italy, October 22-29, 2017}, pages 2859--2867. {IEEE} Computer
  Society.

\bibitem[\protect\astroncite{Maximov
  et~al.}{2020}]{DBLP:journals/corr/abs-2005-09544}
Maximov, M., Elezi, I., and Leal{-}Taix{\'{e}}, L. (2020).
\newblock {CIAGAN:} conditional identity anonymization generative adversarial
  networks.
\newblock In {\em 2020 {IEEE} Conference on Computer Vision and Pattern
  Recognition, {CVPR} 2020}. {IEEE} Computer Society.

\bibitem[\protect\astroncite{McDaid
  et~al.}{2011}]{DBLP:journals/corr/abs-1110-2515}
McDaid, A.~F., Greene, D., and Hurley, N.~J. (2011).
\newblock Normalized mutual information to evaluate overlapping community
  finding algorithms.
\newblock {\em CoRR}, abs/1110.2515.

\bibitem[\protect\astroncite{Meier et~al.}{2018}]{DBLP:conf/annpr/MeierEADS18}
Meier, B.~B., Elezi, I., Amirian, M., D{\"{u}}rr, O., and Stadelmann, T.
  (2018).
\newblock Learning neural models for end-to-end clustering.
\newblock In Pancioni, L., Schwenker, F., and Trentin, E., editors, {\em
  Artificial Neural Networks in Pattern Recognition - 8th {IAPR} {TC3}
  Workshop, {ANNPR} 2018, Siena, Italy, September 19-21, 2018, Proceedings},
  volume 11081 of {\em Lecture Notes in Computer Science}, pages 126--138.
  Springer.

\bibitem[\protect\astroncite{Mequanint
  et~al.}{2019}]{DBLP:journals/pami/MequanintAP19}
Mequanint, E.~Z., Alemu, L.~T., and Pelillo, M. (2019).
\newblock Dominant sets for "constrained" image segmentation.
\newblock {\em {IEEE} Trans. Pattern Anal. Mach. Intell.}, 41(10):2438--2451.

\bibitem[\protect\astroncite{Mikolov et~al.}{2013}]{mikolov2013efficient}
Mikolov, T., Chen, K., Corrado, G., and Dean, J. (2013).
\newblock Efficient estimation of word representations in vector space.
\newblock In Bengio, Y. and LeCun, Y., editors, {\em 1st International
  Conference on Learning Representations, {ICLR} 2013, Scottsdale, Arizona,
  USA, May 2-4, 2013, Workshop Track Proceedings}.

\bibitem[\protect\astroncite{Miller and Zucker}{1991}]{miller1991copositive}
Miller, D.~A. and Zucker, S.~W. (1991).
\newblock Copositive-plus lemke algorithm solves polymatrix games.
\newblock {\em Oper. Res. Lett.}, 10(5):285--290.

\bibitem[\protect\astroncite{Miyato
  et~al.}{2019}]{DBLP:journals/pami/MiyatoMKI19}
Miyato, T., Maeda, S., Koyama, M., and Ishii, S. (2019).
\newblock Virtual adversarial training: {A} regularization method for
  supervised and semi-supervised learning.
\newblock {\em {IEEE} Trans. Pattern Anal. Mach. Intell.}, 41(8):1979--1993.

\bibitem[\protect\astroncite{Mori et~al.}{1999}]{mori1999}
Mori, S., Nishida, H., and Yamada, H. (1999).
\newblock {\em Optical character recognition}.
\newblock John Wiley \& Sons, Inc.

\bibitem[\protect\astroncite{Movshovitz{-}Attias
  et~al.}{2017}]{DBLP:conf/iccv/Movshovitz-Attias17}
Movshovitz{-}Attias, Y., Toshev, A., Leung, T.~K., Ioffe, S., and Singh, S.
  (2017).
\newblock No fuss distance metric learning using proxies.
\newblock In {\em {IEEE} International Conference on Computer Vision, {ICCV}
  2017, Venice, Italy, October 22-29, 2017}, pages 360--368. {IEEE} Computer
  Society.

\bibitem[\protect\astroncite{Murtagh}{1983}]{murtagh1983survey}
Murtagh, F. (1983).
\newblock A survey of recent advances in hierarchical clustering algorithms.
\newblock {\em Comput. J.}, 26(4):354--359.

\bibitem[\protect\astroncite{Nair and Hinton}{2010}]{DBLP:conf/icml/NairH10}
Nair, V. and Hinton, G.~E. (2010).
\newblock Rectified linear units improve restricted boltzmann machines.
\newblock In F{\"{u}}rnkranz, J. and Joachims, T., editors, {\em Proceedings of
  the 27th International Conference on Machine Learning (ICML-10), June 21-24,
  2010, Haifa, Israel}, pages 807--814. Omnipress.

\bibitem[\protect\astroncite{Nash}{1951}]{Nash1951}
Nash, J. (1951).
\newblock Non-cooperative games.
\newblock {\em Annals of Mathematics}, pages 286--295.

\bibitem[\protect\astroncite{Nayar et~al.}{1996}]{nayar1996columbia}
Nayar, S., Nene, S., and Murase, H. (1996).
\newblock Columbia object image library ({COIL} 100).
\newblock {\em Department of Comp. Science, Columbia University, Tech. Rep.
  CUCS-006-96}.

\bibitem[\protect\astroncite{Nesterov}{1983}]{Nesterov}
Nesterov, Y.~E. (1983).
\newblock A method for solving the convex programming problem with convergence
  rate $o (1/k^ 2)$.
\newblock {\em Dokl. akad. nauk Sssr.}, 269:543--547.

\bibitem[\protect\astroncite{Netzer et~al.}{2011}]{Netzer2011}
Netzer, Y., Wang, T., Coates, A., Bissacco, A., Wu, B., and Ng, A.~Y. (2011).
\newblock Reading digits in natural images with unsupervised feature learning.
\newblock {\em In NIPS workshop on deep learning and unsupervised feature
  learning (Vol. 2011, No. 2, p. 5)}.

\bibitem[\protect\astroncite{Ng}{2018}]{ng2018yearning}
Ng, A. (2018).
\newblock {\em Machine Learning Yearning - Technical Strategy for AI Engineers
  in the Era of Deep Learning}.
\newblock [to appear].

\bibitem[\protect\astroncite{Ng and Winkler}{2014}]{ng2014data}
Ng, H. and Winkler, S. (2014).
\newblock A data-driven approach to cleaning large face datasets.
\newblock In {\em 2014 {IEEE} International Conference on Image Processing,
  {ICIP} 2014, Paris, France, October 27-30, 2014}, pages 343--347. {IEEE}.

\bibitem[\protect\astroncite{Oh and Jung}{2004}]{gpu2004}
Oh, K. and Jung, K. (2004).
\newblock {GPU} implementation of neural networks.
\newblock {\em Pattern Recognit.}, 37(6):1311--1314.

\bibitem[\protect\astroncite{Olah}{2015}]{colah}
Olah, C. (2015).
\newblock Understanding lstm networks.
\newblock {\em https://colah.github.io/posts/2015-08-Understanding-LSTMs/}.

\bibitem[\protect\astroncite{Oliver et~al.}{2018}]{DBLP:conf/nips/OliverORCG18}
Oliver, A., Odena, A., Raffel, C., Cubuk, E.~D., and Goodfellow, I.~J. (2018).
\newblock Realistic evaluation of deep semi-supervised learning algorithms.
\newblock In {\em Advances in Neural Information Processing Systems 31: Annual
  Conference on Neural Information Processing Systems 2018, NeurIPS 2018, 3-8
  December 2018, Montr{\'{e}}al, Canada.}, pages 3239--3250.

\bibitem[\protect\astroncite{Opitz et~al.}{2017}]{DBLP:conf/iccv/OpitzWPB17}
Opitz, M., Waltner, G., Possegger, H., and Bischof, H. (2017).
\newblock {BIER} - boosting independent embeddings robustly.
\newblock In {\em {IEEE} International Conference on Computer Vision, {ICCV}
  2017, Venice, Italy, October 22-29, 2017}, pages 5199--5208. {IEEE} Computer
  Society.

\bibitem[\protect\astroncite{Opitz
  et~al.}{2020}]{DBLP:journals/corr/abs-1801-04815}
Opitz, M., Waltner, G., Possegger, H., and Bischof, H. (2020).
\newblock Deep metric learning with {BIER:} boosting independent embeddings
  robustly.
\newblock {\em {IEEE} Trans. Pattern Anal. Mach. Intell.}, 42(2):276--290.

\bibitem[\protect\astroncite{Pacha et~al.}{2018}]{DBLP:conf/iwdas/Pacha2018}
Pacha, A., Choi, K., Co{\"{u}}asnon, B., Ricquebourg, Y., Zanibbi, R., and
  Eidenberger, H. (2018).
\newblock Handwritten music object detection: Open issues and baseline results.
\newblock In {\em 13th {IAPR} International Workshop on Document Analysis
  Systems, {DAS} 2018, Vienna, Austria, April 24-27, 2018}, pages 163--168.
  {IEEE} Computer Society.

\bibitem[\protect\astroncite{Pacha and
  Eidenberger}{2017}]{DBLP:conf/icmla/PachaE17}
Pacha, A. and Eidenberger, H. (2017).
\newblock Towards self-learning optical music recognition.
\newblock In Chen, X., Luo, B., Luo, F., Palade, V., and Wani, M.~A., editors,
  {\em 16th {IEEE} International Conference on Machine Learning and
  Applications, {ICMLA} 2017, Cancun, Mexico, December 18-21, 2017}, pages
  795--800. {IEEE}.

\bibitem[\protect\astroncite{Pacha et~al.}{2017}]{DBLP:journals/as/Pacha18}
Pacha, A., Hajič, J., and Calvo-Zaragoza, J. (2017).
\newblock A baseline for general music object detection with deep learning.
\newblock {\em Applied Sciences}.

\bibitem[\protect\astroncite{Park et~al.}{2019}]{DBLP:conf/cvpr/Park2019}
Park, W., Kim, D., Lu, Y., and Cho, M. (2019).
\newblock Relational knowledge distillation.
\newblock In {\em {IEEE} Conference on Computer Vision and Pattern Recognition,
  {CVPR} 2019, Long Beach, CA, USA, June 16-20, 2019}, pages 3967--3976.
  Computer Vision Foundation / {IEEE}.

\bibitem[\protect\astroncite{Paszke et~al.}{2017}]{Paszke17}
Paszke, A., Gross, S., Chintala, S., Chanan, G., Yang, E., DeVito, Z., Lin, Z.,
  Desmaison, A., Antiga, L., and Lerer, A. (2017).
\newblock Automatic differentiation in pytorch.
\newblock {\em NIPS Workshops}.

\bibitem[\protect\astroncite{Pearson}{1895}]{DBLP:journals/rsl/Pearson95}
Pearson, K. (1895).
\newblock Notes on regression and inheritance in the case of two parents.
\newblock {\em Proceedings of the Royal Society of London}, 58:240--242.

\bibitem[\protect\astroncite{Pelillo}{1997}]{DBLP:journals/jmiv/Pelillo97}
Pelillo, M. (1997).
\newblock The dynamics of nonlinear relaxation labeling processes.
\newblock {\em J. Math. Imaging Vis.}, 7(4):309--323.

\bibitem[\protect\astroncite{Pelillo
  et~al.}{2017}]{DBLP:journals/prl/PelilloEF17}
Pelillo, M., Elezi, I., and Fiorucci, M. (2017).
\newblock Revealing structure in large graphs: Szemer{\'{e}}di's regularity
  lemma and its use in pattern recognition.
\newblock {\em Pattern Recognit. Lett.}, 87:4--11.

\bibitem[\protect\astroncite{Qian
  et~al.}{2019}]{DBLP:journals/corr/abs-1909-05235}
Qian, Q., Shang, L., Sun, B., Hu, J., Tacoma, T., Li, H., and Jin, R. (2019).
\newblock Softtriple loss: Deep metric learning without triplet sampling.
\newblock In {\em 2019 {IEEE/CVF} International Conference on Computer Vision,
  {ICCV} 2019, Seoul, Korea (South), October 27 - November 2, 2019}, pages
  6449--6457. {IEEE}.

\bibitem[\protect\astroncite{Quattoni and
  Torralba}{2009}]{DBLP:conf/cvpr/QuattoniT09}
Quattoni, A. and Torralba, A. (2009).
\newblock Recognizing indoor scenes.
\newblock In {\em 2009 {IEEE} Computer Society Conference on Computer Vision
  and Pattern Recognition {(CVPR} 2009), 20-25 June 2009, Miami, Florida,
  {USA}}, pages 413--420. {IEEE} Computer Society.

\bibitem[\protect\astroncite{Raina et~al.}{2009}]{Raina2009}
Raina, R., Madhavan, A., and Ng, A.~Y. (2009).
\newblock Large-scale deep unsupervised learning using graphics processors.
\newblock In Danyluk, A.~P., Bottou, L., and Littman, M.~L., editors, {\em
  Proceedings of the 26th Annual International Conference on Machine Learning,
  {ICML} 2009, Montreal, Quebec, Canada, June 14-18, 2009}, volume 382 of {\em
  {ACM} International Conference Proceeding Series}, pages 873--880. {ACM}.

\bibitem[\protect\astroncite{Rebelo
  et~al.}{2010}]{DBLP:journals/ijdar/RebeloCC10}
Rebelo, A., Capela, G.~A., and Cardoso, J.~S. (2010).
\newblock Optical recognition of music symbols - {A} comparative study.
\newblock {\em {IJDAR}}, 13(1):19--31.

\bibitem[\protect\astroncite{Rebelo et~al.}{2012}]{Rebelo2012}
Rebelo, A., Fujinaga, I., Paszkiewicz, F., Mar{\c{c}}al, A. R.~S., Guedes, C.,
  and Cardoso, J.~S. (2012).
\newblock Optical music recognition: state-of-the-art and open issues.
\newblock {\em Int. J. Multim. Inf. Retr.}, 1(3):173--190.

\bibitem[\protect\astroncite{Redmon et~al.}{2016}]{Redmon2016}
Redmon, J., Divvala, S.~K., Girshick, R.~B., and Farhadi, A. (2016).
\newblock You only look once: Unified, real-time object detection.
\newblock In {\em 2016 {IEEE} Conference on Computer Vision and Pattern
  Recognition, {CVPR} 2016, Las Vegas, NV, USA, June 27-30, 2016}, pages
  779--788. {IEEE} Computer Society.

\bibitem[\protect\astroncite{Redmon and
  Farhadi}{2017}]{DBLP:conf/cvpr/RedmonF17}
Redmon, J. and Farhadi, A. (2017).
\newblock {YOLO9000:} better, faster, stronger.
\newblock In {\em 2017 {IEEE} Conference on Computer Vision and Pattern
  Recognition, {CVPR} 2017, Honolulu, HI, USA, July 21-26, 2017}, pages
  6517--6525. {IEEE} Computer Society.

\bibitem[\protect\astroncite{Ren et~al.}{2015}]{DBLP:conf/nips/RenHGS15}
Ren, S., He, K., Girshick, R.~B., and Sun, J. (2015).
\newblock Faster {R-CNN:} towards real-time object detection with region
  proposal networks.
\newblock In {\em Advances in Neural Information Processing Systems 28: Annual
  Conference on Neural Information Processing Systems 2015, December 7-12,
  2015, Montreal, Quebec, Canada}, pages 91--99.

\bibitem[\protect\astroncite{Revaud
  et~al.}{2019}]{DBLP:journals/corr/abs-1906-07589}
Revaud, J., Almaz{\'{a}}n, J., Rezende, R.~S., and de~Souza, C.~R. (2019).
\newblock Learning with average precision: Training image retrieval with a
  listwise loss.
\newblock In {\em 2019 {IEEE/CVF} International Conference on Computer Vision,
  {ICCV} 2019, Seoul, Korea (South), October 27 - November 2, 2019}, pages
  5106--5115. {IEEE}.

\bibitem[\protect\astroncite{Ronneberger
  et~al.}{2015}]{DBLP:conf/miccai/RonnebergerFB15}
Ronneberger, O., Fischer, P., and Brox, T. (2015).
\newblock U-net: Convolutional networks for biomedical image segmentation.
\newblock In {\em Medical Image Computing and Computer-Assisted Intervention -
  {MICCAI} 2015 - 18th International Conference Munich, Germany, October 5 - 9,
  2015, Proceedings, Part {III}}, pages 234--241.

\bibitem[\protect\astroncite{Rosenfeld et~al.}{1976}]{RosHumZuc76}
Rosenfeld, A., Hummel, R.~A., and Zucker, S.~W. (1976).
\newblock Scene labeling by relaxation operations.
\newblock {\em IEEE Trans. Syst. Man Cybern.}, 6:420--433.

\bibitem[\protect\astroncite{Rosenfeld
  et~al.}{2018}]{DBLP:journals/corr/abs-1808-03305}
Rosenfeld, A., Zemel, R.~S., and Tsotsos, J.~K. (2018).
\newblock The elephant in the room.
\newblock {\em CoRR}, abs/1808.03305.

\bibitem[\protect\astroncite{Rossant and
  Bloch}{2007}]{DBLP:journals/ejasp/RossantB07}
Rossant, F. and Bloch, I. (2007).
\newblock Robust and adaptive {OMR} system including fuzzy modeling, fusion of
  musical rules, and possible error detection.
\newblock {\em {EURASIP} J. Adv. Signal Process.}, 2007.

\bibitem[\protect\astroncite{Rumelhart
  et~al.}{1986}]{Rumelhart:1986:PDP:104279}
Rumelhart, D.~E., McClelland, J.~L., and PDP Research~Group, C., editors
  (1986).
\newblock {\em Parallel Distributed Processing: Explorations in the
  Microstructure of Cognition, Vol. 1: Foundations}.
\newblock MIT Press.

\bibitem[\protect\astroncite{Russakovsky
  et~al.}{2015}]{DBLP:journals/corr/RussakovskyDSKSMHKKBBF14}
Russakovsky, O., Deng, J., Su, H., Krause, J., Satheesh, S., Ma, S., Huang, Z.,
  Karpathy, A., Khosla, A., Bernstein, M.~S., Berg, A.~C., and Li, F. (2015).
\newblock Imagenet large scale visual recognition challenge.
\newblock {\em Int. J. Comput. Vis.}, 115(3):211--252.

\bibitem[\protect\astroncite{Sabour et~al.}{2017}]{DBLP:conf/nips/SabourFH17}
Sabour, S., Frosst, N., and Hinton, G.~E. (2017).
\newblock Dynamic routing between capsules.
\newblock In Guyon, I., von Luxburg, U., Bengio, S., Wallach, H.~M., Fergus,
  R., Vishwanathan, S. V.~N., and Garnett, R., editors, {\em Advances in Neural
  Information Processing Systems 30: Annual Conference on Neural Information
  Processing Systems 2017, 4-9 December 2017, Long Beach, CA, {USA}}, pages
  3856--3866.

\bibitem[\protect\astroncite{Sanakoyeu
  et~al.}{2019}]{DDBLP:conf/cvpr/Sanakoyeu2019}
Sanakoyeu, A., Tschernezki, V., B{\"{u}}chler, U., and Ommer, B. (2019).
\newblock Divide and conquer the embedding space for metric learning.
\newblock In {\em {IEEE} Conference on Computer Vision and Pattern Recognition,
  {CVPR} 2019, Long Beach, CA, USA, June 16-20, 2019}, pages 471--480. Computer
  Vision Foundation / {IEEE}.

\bibitem[\protect\astroncite{Scarselli
  et~al.}{2009}]{DBLP:journals/tnn/ScarselliGTHM09}
Scarselli, F., Gori, M., Tsoi, A.~C., Hagenbuchner, M., and Monfardini, G.
  (2009).
\newblock The graph neural network model.
\newblock {\em {IEEE} Trans. Neural Networks}, 20(1):61--80.

\bibitem[\protect\astroncite{Schmidhuber}{2015}]{DBLP:journals/nn/Schmidhuber15}
Schmidhuber, J. (2015).
\newblock Deep learning in neural networks: An overview.
\newblock {\em Neural Networks}, 61:85--117.

\bibitem[\protect\astroncite{Schroff et~al.}{2015}]{DBLP:conf/cvpr/SchroffKP15}
Schroff, F., Kalenichenko, D., and Philbin, J. (2015).
\newblock Facenet: {A} unified embedding for face recognition and clustering.
\newblock In {\em {IEEE} Conference on Computer Vision and Pattern Recognition,
  {CVPR} 2015, Boston, MA, USA, June 7-12, 2015}, pages 815--823. {IEEE}
  Computer Society.

\bibitem[\protect\astroncite{Schultz and
  Joachims}{2003}]{DBLP:conf/nips/SchultzJ03}
Schultz, M. and Joachims, T. (2003).
\newblock Learning a distance metric from relative comparisons.
\newblock In Thrun, S., Saul, L.~K., and Sch{\"{o}}lkopf, B., editors, {\em
  Advances in Neural Information Processing Systems 16 [Neural Information
  Processing Systems, {NIPS} 2003, December 8-13, 2003, Vancouver and Whistler,
  British Columbia, Canada]}, pages 41--48. {MIT} Press.

\bibitem[\protect\astroncite{Schwenker et~al.}{2001}]{schwenker2001three}
Schwenker, F., Kestler, H.~A., and Palm, G. (2001).
\newblock Three learning phases for radial-basis-function networks.
\newblock {\em Neural Networks}, 14(4-5):439--458.

\bibitem[\protect\astroncite{Shi et~al.}{2017}]{shi2017end}
Shi, B., Bai, X., and Yao, C. (2017).
\newblock An end-to-end trainable neural network for image-based sequence
  recognition and its application to scene text recognition.
\newblock {\em {IEEE} Trans. Pattern Anal. Mach. Intell.}, 39(11):2298--2304.

\bibitem[\protect\astroncite{Sigtia et~al.}{2016}]{sigtia2016end}
Sigtia, S., Benetos, E., and Dixon, S. (2016).
\newblock An end-to-end neural network for polyphonic piano music
  transcription.
\newblock {\em {IEEE} {ACM} Trans. Audio Speech Lang. Process.},
  24(5):927--939.

\bibitem[\protect\astroncite{Simonyan and Zisserman}{2015}]{Simonyan2014}
Simonyan, K. and Zisserman, A. (2015).
\newblock Very deep convolutional networks for large-scale image recognition.
\newblock In Bengio, Y. and LeCun, Y., editors, {\em 3rd International
  Conference on Learning Representations, {ICLR} 2015, San Diego, CA, USA, May
  7-9, 2015, Conference Track Proceedings}.

\bibitem[\protect\astroncite{Smith}{1982}]{smith1982evolution}
Smith, J.~M. (1982).
\newblock {\em Evolution and the Theory of Games}.
\newblock Cambridge University Press.

\bibitem[\protect\astroncite{Sohn}{2016}]{DBLP:conf/nips/Sohn16}
Sohn, K. (2016).
\newblock Improved deep metric learning with multi-class n-pair loss objective.
\newblock In Lee, D.~D., Sugiyama, M., von Luxburg, U., Guyon, I., and Garnett,
  R., editors, {\em Advances in Neural Information Processing Systems 29:
  Annual Conference on Neural Information Processing Systems 2016, December
  5-10, 2016, Barcelona, Spain}, pages 1849--1857.

\bibitem[\protect\astroncite{Song et~al.}{2017}]{DBLP:conf/cvpr/SongJR017}
Song, H.~O., Jegelka, S., Rathod, V., and Murphy, K. (2017).
\newblock Deep metric learning via facility location.
\newblock In {\em 2017 {IEEE} Conference on Computer Vision and Pattern
  Recognition, {CVPR} 2017, Honolulu, HI, USA, July 21-26, 2017}, pages
  2206--2214. {IEEE} Computer Society.

\bibitem[\protect\astroncite{Song et~al.}{2016}]{DBLP:conf/cvpr/SongXJS16}
Song, H.~O., Xiang, Y., Jegelka, S., and Savarese, S. (2016).
\newblock Deep metric learning via lifted structured feature embedding.
\newblock In {\em 2016 {IEEE} Conference on Computer Vision and Pattern
  Recognition, {CVPR} 2016, Las Vegas, NV, USA, June 27-30, 2016}, pages
  4004--4012. {IEEE} Computer Society.

\bibitem[\protect\astroncite{Srivastava
  et~al.}{2014}]{DBLP:journals/jmlr/SrivastavaHKSS14}
Srivastava, N., Hinton, G.~E., Krizhevsky, A., Sutskever, I., and
  Salakhutdinov, R. (2014).
\newblock Dropout: a simple way to prevent neural networks from overfitting.
\newblock {\em J. Mach. Learn. Res.}, 15(1):1929--1958.

\bibitem[\protect\astroncite{Srivastava et~al.}{2015}]{greff2015nips}
Srivastava, R.~K., Greff, K., and Schmidhuber, J. (2015).
\newblock Training very deep networks.
\newblock In Cortes, C., Lawrence, N.~D., Lee, D.~D., Sugiyama, M., and
  Garnett, R., editors, {\em Advances in Neural Information Processing Systems
  28: Annual Conference on Neural Information Processing Systems 2015, December
  7-12, 2015, Montreal, Quebec, Canada}, pages 2377--2385.

\bibitem[\protect\astroncite{Stadelmann
  et~al.}{2018}]{DBLP:conf/annpr/StadelmannAAADE18}
Stadelmann, T., Amirian, M., Arabaci, I., Arnold, M., Duivesteijn, G.~F.,
  Elezi, I., Geiger, M., L{\"{o}}rwald, S., Meier, B.~B., Rombach, K., and
  Tuggener, L. (2018).
\newblock Deep learning in the wild.
\newblock In Pancioni, L., Schwenker, F., and Trentin, E., editors, {\em
  Artificial Neural Networks in Pattern Recognition - 8th {IAPR} {TC3}
  Workshop, {ANNPR} 2018, Siena, Italy, September 19-21, 2018, Proceedings},
  volume 11081 of {\em Lecture Notes in Computer Science}, pages 17--38.
  Springer.

\bibitem[\protect\astroncite{Stallkamp et~al.}{2011}]{Stallkamp2011}
Stallkamp, J., Schlipsing, M., Salmen, J., and Igel, C. (2011).
\newblock The german traffic sign recognition benchmark: {A} multi-class
  classification competition.
\newblock In {\em The 2011 International Joint Conference on Neural Networks,
  {IJCNN} 2011, San Jose, California, USA, July 31 - August 5, 2011}, pages
  1453--1460. {IEEE}.

\bibitem[\protect\astroncite{Sutskever}{2013}]{SutskeverThesis}
Sutskever, I. (2013).
\newblock {\em Training recurrent neural networks}.
\newblock PhD thesis, Toronto, Canada.

\bibitem[\protect\astroncite{Szegedy
  et~al.}{2015}]{DBLP:conf/cvpr/SzegedyLJSRAEVR15}
Szegedy, C., Liu, W., Jia, Y., Sermanet, P., Reed, S.~E., Anguelov, D., Erhan,
  D., Vanhoucke, V., and Rabinovich, A. (2015).
\newblock Going deeper with convolutions.
\newblock In {\em {IEEE} Conference on Computer Vision and Pattern Recognition,
  {CVPR} 2015, Boston, MA, USA, June 7-12, 2015}, pages 1--9. {IEEE} Computer
  Society.

\bibitem[\protect\astroncite{Szegedy et~al.}{2014}]{Szegedy2013}
Szegedy, C., Zaremba, W., Sutskever, I., Bruna, J., Erhan, D., Goodfellow,
  I.~J., and Fergus, R. (2014).
\newblock Intriguing properties of neural networks.
\newblock In Bengio, Y. and LeCun, Y., editors, {\em 2nd International
  Conference on Learning Representations, {ICLR} 2014, Banff, AB, Canada, April
  14-16, 2014, Conference Track Proceedings}.

\bibitem[\protect\astroncite{Tarvainen and
  Valpola}{2017}]{DBLP:conf/iclr/TarvainenV17}
Tarvainen, A. and Valpola, H. (2017).
\newblock Mean teachers are better role models: Weight-averaged consistency
  targets improve semi-supervised deep learning results.
\newblock In Guyon, I., von Luxburg, U., Bengio, S., Wallach, H.~M., Fergus,
  R., Vishwanathan, S. V.~N., and Garnett, R., editors, {\em Advances in Neural
  Information Processing Systems 30: Annual Conference on Neural Information
  Processing Systems 2017, 4-9 December 2017, Long Beach, CA, {USA}}, pages
  1195--1204.

\bibitem[\protect\astroncite{Tieleman and Hinton}{2012}]{rmsprop}
Tieleman, T. and Hinton, G.~E. (2012).
\newblock Lecture 6.5-rmsprop: Divide the gradient by a running average of its
  recent magnitude.
\newblock {\em COURSERA: Neural networks for machine learning 4.2}, pages
  26--31.

\bibitem[\protect\astroncite{Torralba and
  Efros}{2011}]{DBLP:conf/cvpr/TorralbaE11}
Torralba, A. and Efros, A.~A. (2011).
\newblock Unbiased look at dataset bias.
\newblock In {\em The 24th {IEEE} Conference on Computer Vision and Pattern
  Recognition, {CVPR} 2011, Colorado Springs, CO, USA, 20-25 June 2011}, pages
  1521--1528. {IEEE} Computer Society.

\bibitem[\protect\astroncite{Tripodi et~al.}{2016}]{DBLP:conf/icpr/TripodiVP16}
Tripodi, R., Vascon, S., and Pelillo, M. (2016).
\newblock Context aware nonnegative matrix factorization clustering.
\newblock In {\em 23rd International Conference on Pattern Recognition, {ICPR}
  2016, Canc{\'{u}}n, Mexico, December 4-8, 2016}, pages 1719--1724. {IEEE}.

\bibitem[\protect\astroncite{Tuggener
  et~al.}{2018a}]{DBLP:conf/icpr/TuggenerESPS18}
Tuggener, L., Elezi, I., Schmidhuber, J., Pelillo, M., and Stadelmann, T.
  (2018a).
\newblock Deepscores-a dataset for segmentation, detection and classification
  of tiny objects.
\newblock In {\em 24th International Conference on Pattern Recognition, {ICPR}
  2018, Beijing, China, August 20-24, 2018}, pages 3704--3709. {IEEE} Computer
  Society.

\bibitem[\protect\astroncite{Tuggener
  et~al.}{2018b}]{DBLP:conf/ismir/TuggenerESS18}
Tuggener, L., Elezi, I., Schmidhuber, J., and Stadelmann, T. (2018b).
\newblock Deep watershed detector for music object recognition.
\newblock In G{\'{o}}mez, E., Hu, X., Humphrey, E., and Benetos, E., editors,
  {\em Proceedings of the 19th International Society for Music Information
  Retrieval Conference, {ISMIR} 2018, Paris, France, September 23-27, 2018},
  pages 271--278.

\bibitem[\protect\astroncite{van~der Maaten and
  Hinton}{2012}]{DBLP:journals/ml/MaatenH12}
van~der Maaten, L. and Hinton, G.~E. (2012).
\newblock Visualizing non-metric similarities in multiple maps.
\newblock {\em Mach. Learn.}, 87(1):33--55.

\bibitem[\protect\astroncite{van~der Wel and Ullrich}{2017}]{vanderWel17}
van~der Wel, E. and Ullrich, K. (2017).
\newblock Optical music recognition with convolutional sequence-to-sequence
  models.
\newblock In Cunningham, S.~J., Duan, Z., Hu, X., and Turnbull, D., editors,
  {\em Proceedings of the 18th International Society for Music Information
  Retrieval Conference, {ISMIR} 2017, Suzhou, China, October 23-27, 2017},
  pages 731--737.

\bibitem[\protect\astroncite{Vapnik}{1998}]{vapnik}
Vapnik, V. (1998).
\newblock {\em Statistical learning theory}.
\newblock Wiley.

\bibitem[\protect\astroncite{Vascon et~al.}{2020}]{DBLP:journals/prl/Vascon18}
Vascon, S., Frasca, M., Tripodi, R., Valentini, G., and Pelillo, M. (2020).
\newblock Protein function prediction as a graph-transduction game.
\newblock {\em Pattern Recognit. Lett.}, 134:96--105.

\bibitem[\protect\astroncite{Vaswani
  et~al.}{2017}]{DBLP:conf/nips/VaswaniSPUJGKP17}
Vaswani, A., Shazeer, N., Parmar, N., Uszkoreit, J., Jones, L., Gomez, A.~N.,
  Kaiser, L., and Polosukhin, I. (2017).
\newblock Attention is all you need.
\newblock In Guyon, I., von Luxburg, U., Bengio, S., Wallach, H.~M., Fergus,
  R., Vishwanathan, S. V.~N., and Garnett, R., editors, {\em Advances in Neural
  Information Processing Systems 30: Annual Conference on Neural Information
  Processing Systems 2017, 4-9 December 2017, Long Beach, CA, {USA}}, pages
  5998--6008.

\bibitem[\protect\astroncite{Wah et~al.}{2011}]{WahCUB_200_2011}
Wah, C., Branson, S., Welinder, P., Perona, P., and Belongie, S. (2011).
\newblock {The Caltech-UCSD Birds-200-2011 Dataset}.
\newblock Technical Report CNS-TR-2011-001, California Institute of Technology.

\bibitem[\protect\astroncite{Wang et~al.}{2017}]{DBLP:conf/iccv/WangZWLL17}
Wang, J., Zhou, F., Wen, S., Liu, X., and Lin, Y. (2017).
\newblock Deep metric learning with angular loss.
\newblock In {\em {IEEE} International Conference on Computer Vision, {ICCV}
  2017, Venice, Italy, October 22-29, 2017}, pages 2612--2620. {IEEE} Computer
  Society.

\bibitem[\protect\astroncite{Wang et~al.}{2018}]{DBLP:conf/cvpr/WangYZ0L18}
Wang, K., Yan, X., Zhang, D., Zhang, L., and Lin, L. (2018).
\newblock Towards human-machine cooperation: Self-supervised sample mining for
  object detection.
\newblock In {\em 2018 {IEEE} Conference on Computer Vision and Pattern
  Recognition, {CVPR} 2018, Salt Lake City, UT, USA, June 18-22, 2018}, pages
  1605--1613. {IEEE} Computer Society.

\bibitem[\protect\astroncite{Wang et~al.}{2019a}]{DBLP:conf/iccv/WangLSC019}
Wang, W., Lu, X., Shen, J., Crandall, D., and Shao, L. (2019a).
\newblock Zero-shot video object segmentation via attentive graph neural
  networks.
\newblock In {\em 2019 {IEEE/CVF} International Conference on Computer Vision,
  {ICCV} 2019, Seoul, Korea (South), October 27 - November 2, 2019}, pages
  9235--9244. {IEEE}.

\bibitem[\protect\astroncite{Wang et~al.}{2019b}]{DDBLP:conf/cvpr/Wand2019}
Wang, X., Han, X., Huang, W., Dong, D., and Scott, M.~R. (2019b).
\newblock Multi-similarity loss with general pair weighting for deep metric
  learning.
\newblock In {\em {IEEE} Conference on Computer Vision and Pattern Recognition,
  {CVPR} 2019, Long Beach, CA, USA, June 16-20, 2019}, pages 5022--5030.
  Computer Vision Foundation / {IEEE}.

\bibitem[\protect\astroncite{Wang et~al.}{2019c}]{DBLP:conf/cvpr/WangHKHGR19}
Wang, X., Hua, Y., Kodirov, E., Hu, G., Garnier, R., and Robertson, N.~M.
  (2019c).
\newblock Ranked list loss for deep metric learning.
\newblock In {\em {IEEE} Conference on Computer Vision and Pattern Recognition,
  {CVPR} 2019, Long Beach, CA, USA, June 16-20, 2019}, pages 5207--5216.
  Computer Vision Foundation / {IEEE}.

\bibitem[\protect\astroncite{Weibull}{1997}]{weibull1997evolutionary}
Weibull, J. (1997).
\newblock {\em Evolutionary Game Theory}.
\newblock MIT Press.

\bibitem[\protect\astroncite{Weinberger and
  Saul}{2009}]{DBLP:journals/jmlr/WeinbergerS09}
Weinberger, K.~Q. and Saul, L.~K. (2009).
\newblock Distance metric learning for large margin nearest neighbor
  classification.
\newblock {\em J. Mach. Learn. Res.}, 10:207--244.

\bibitem[\protect\astroncite{Werbos}{1974}]{Werbos:74}
Werbos, P.~J. (1974).
\newblock {\em Beyond Regression: New Tools for Prediction and Analysis in the
  Behavioral Sciences}.
\newblock PhD thesis, Harvard University.

\bibitem[\protect\astroncite{Wu et~al.}{2009}]{Wu:2009:OTC:1529867.1529869}
Wu, K., Otoo, E.~J., and Suzuki, K. (2009).
\newblock Optimizing two-pass connected-component labeling algorithms.
\newblock {\em Pattern Anal. Appl.}, 12(2):117--135.

\bibitem[\protect\astroncite{Wu et~al.}{2016}]{wu2016google}
Wu, Y., Schuster, M., Chen, Z., Le, Q.~V., Norouzi, M., Macherey, W., Krikun,
  M., Cao, Y., Gao, Q., Macherey, K., Klingner, J., Shah, A., Johnson, M., Liu,
  X., Kaiser, L., Gouws, S., Kato, Y., Kudo, T., Kazawa, H., Stevens, K.,
  Kurian, G., Patil, N., Wang, W., Young, C., Smith, J., Riesa, J., Rudnick,
  A., Vinyals, O., Corrado, G., Hughes, M., and Dean, J. (2016).
\newblock Google's neural machine translation system: Bridging the gap between
  human and machine translation.
\newblock {\em CoRR}, abs/1609.08144.

\bibitem[\protect\astroncite{Xia
  et~al.}{2018}]{DBLP:journals/corr/abs-1711-10398}
Xia, G., Bai, X., Ding, J., Zhu, Z., Belongie, S.~J., Luo, J., Datcu, M.,
  Pelillo, M., and Zhang, L. (2018).
\newblock {DOTA:} {A} large-scale dataset for object detection in aerial
  images.
\newblock In {\em 2018 {IEEE} Conference on Computer Vision and Pattern
  Recognition, {CVPR} 2018, Salt Lake City, UT, USA, June 18-22, 2018}, pages
  3974--3983. {IEEE} Computer Society.

\bibitem[\protect\astroncite{Xiao et~al.}{2010}]{Xiao2010}
Xiao, J., Hays, J., Ehinger, K.~A., Oliva, A., and Torralba, A. (2010).
\newblock {SUN} database: Large-scale scene recognition from abbey to zoo.
\newblock In {\em The Twenty-Third {IEEE} Conference on Computer Vision and
  Pattern Recognition, {CVPR} 2010, San Francisco, CA, USA, 13-18 June 2010},
  pages 3485--3492. {IEEE} Computer Society.

\bibitem[\protect\astroncite{Xie et~al.}{2016}]{xie2016unsupervised}
Xie, J., Girshick, R.~B., and Farhadi, A. (2016).
\newblock Unsupervised deep embedding for clustering analysis.
\newblock In {\em Proceedings of the 33nd International Conference on Machine
  Learning, {ICML} 2016, New York City, NY, USA, June 19-24, 2016}, pages
  478--487.

\bibitem[\protect\astroncite{Xing et~al.}{2002}]{xing2003distance}
Xing, E.~P., Ng, A.~Y., Jordan, M.~I., and Russell, S.~J. (2002).
\newblock Distance metric learning with application to clustering with
  side-information.
\newblock In Becker, S., Thrun, S., and Obermayer, K., editors, {\em Advances
  in Neural Information Processing Systems 15 [Neural Information Processing
  Systems, {NIPS} 2002, December 9-14, 2002, Vancouver, British Columbia,
  Canada]}, pages 505--512. {MIT} Press.

\bibitem[\protect\astroncite{Xu et~al.}{2019}]{DDBLP:conf/cvpr/Xu2019}
Xu, X., Yang, Y., Deng, C., and Zheng, F. (2019).
\newblock Deep asymmetric metric learning via rich relationship mining.
\newblock In {\em {IEEE} Conference on Computer Vision and Pattern Recognition,
  {CVPR} 2019, Long Beach, CA, USA, June 16-20, 2019}, pages 4076--4085.
  Computer Vision Foundation / {IEEE}.

\bibitem[\protect\astroncite{Xuan et~al.}{2018}]{DBLP:conf/eccv/XuanSP18}
Xuan, H., Souvenir, R., and Pless, R. (2018).
\newblock Deep randomized ensembles for metric learning.
\newblock In Ferrari, V., Hebert, M., Sminchisescu, C., and Weiss, Y., editors,
  {\em Computer Vision - {ECCV} 2018 - 15th European Conference, Munich,
  Germany, September 8-14, 2018, Proceedings, Part {XVI}}, volume 11220 of {\em
  Lecture Notes in Computer Science}, pages 751--762. Springer.

\bibitem[\protect\astroncite{Yang et~al.}{2016}]{DBLP:conf/cvpr/YangPB16}
Yang, J., Parikh, D., and Batra, D. (2016).
\newblock Joint unsupervised learning of deep representations and image
  clusters.
\newblock In {\em 2016 {IEEE} Conference on Computer Vision and Pattern
  Recognition, {CVPR} 2016, Las Vegas, NV, USA, June 27-30, 2016}, pages
  5147--5156. {IEEE} Computer Society.

\bibitem[\protect\astroncite{Yosinski
  et~al.}{2014}]{DBLP:conf/nips/YosinskiCBL14}
Yosinski, J., Clune, J., Bengio, Y., and Lipson, H. (2014).
\newblock How transferable are features in deep neural networks?
\newblock In Ghahramani, Z., Welling, M., Cortes, C., Lawrence, N.~D., and
  Weinberger, K.~Q., editors, {\em Advances in Neural Information Processing
  Systems 27: Annual Conference on Neural Information Processing Systems 2014,
  December 8-13 2014, Montreal, Quebec, Canada}, pages 3320--3328.

\bibitem[\protect\astroncite{Yu et~al.}{2018}]{DBLP:conf/eccv/YuLGDT18}
Yu, B., Liu, T., Gong, M., Ding, C., and Tao, D. (2018).
\newblock Correcting the triplet selection bias for triplet loss.
\newblock In Ferrari, V., Hebert, M., Sminchisescu, C., and Weiss, Y., editors,
  {\em Computer Vision - {ECCV} 2018 - 15th European Conference, Munich,
  Germany, September 8-14, 2018, Proceedings, Part {VI}}, volume 11210 of {\em
  Lecture Notes in Computer Science}, pages 71--86. Springer.

\bibitem[\protect\astroncite{Yuan et~al.}{2017}]{DBLP:conf/iccv/YuanYZ17}
Yuan, Y., Yang, K., and Zhang, C. (2017).
\newblock Hard-aware deeply cascaded embedding.
\newblock In {\em {IEEE} International Conference on Computer Vision, {ICCV}
  2017, Venice, Italy, October 22-29, 2017}, pages 814--823. {IEEE} Computer
  Society.

\bibitem[\protect\astroncite{Zeiler}{2012}]{DBLP:journals/corr/abs-1212-5701}
Zeiler, M.~D. (2012).
\newblock {ADADELTA:} an adaptive learning rate method.
\newblock {\em CoRR}, abs/1212.5701.

\bibitem[\protect\astroncite{Zelnik{-}Manor and
  Perona}{2004}]{DBLP:conf/nips/Zelnik-ManorP04}
Zelnik{-}Manor, L. and Perona, P. (2004).
\newblock Self-tuning spectral clustering.
\newblock In {\em Advances in Neural Information Processing Systems 17 [Neural
  Information Processing Systems, {NIPS} 2004, December 13-18, 2004, Vancouver,
  British Columbia, Canada]}, pages 1601--1608.

\bibitem[\protect\astroncite{Zemene et~al.}{2016}]{DBLP:conf/icpr/ZemeneAP16}
Zemene, E., Alemu, L.~T., and Pelillo, M. (2016).
\newblock Constrained dominant sets for retrieval.
\newblock In {\em 23rd International Conference on Pattern Recognition, {ICPR}
  2016, Canc{\'{u}}n, Mexico, December 4-8, 2016}, pages 2568--2573. {IEEE}.

\bibitem[\protect\astroncite{Zhai and
  Wu}{2019}]{DBLP:journals/corr/abs-1811-12649}
Zhai, A. and Wu, H. (2019).
\newblock Classification is a strong baseline for deep metric learning.
\newblock In {\em 30th British Machine Vision Conference 2019, {BMVC} 2019,
  Cardiff, UK, September 9-12, 2019}, page~91. {BMVA} Press.

\bibitem[\protect\astroncite{Zhang et~al.}{2016}]{DBLP:conf/cvpr/ZhangZLZ16}
Zhang, X., Zhou, F., Lin, Y., and Zhang, S. (2016).
\newblock Embedding label structures for fine-grained feature representation.
\newblock In {\em 2016 {IEEE} Conference on Computer Vision and Pattern
  Recognition, {CVPR} 2016, Las Vegas, NV, USA, June 27-30, 2016}, pages
  1114--1123. {IEEE} Computer Society.

\bibitem[\protect\astroncite{Zhao et~al.}{2019}]{DBLP:conf/cvpr/ZhaoXC19}
Zhao, K., Xu, J., and Cheng, M. (2019).
\newblock Regularface: Deep face recognition via exclusive regularization.
\newblock In {\em {IEEE} Conference on Computer Vision and Pattern Recognition,
  {CVPR} 2019, Long Beach, CA, USA, June 16-20, 2019}, pages 1136--1144.
  Computer Vision Foundation / {IEEE}.

\bibitem[\protect\astroncite{Zheng et~al.}{2019}]{DBLP:conf/aaai/ZhengJSZWH19}
Zheng, X., Ji, R., Sun, X., Zhang, B., Wu, Y., and Huang, F. (2019).
\newblock Towards optimal fine grained retrieval via decorrelated centralized
  loss with normalize-scale layer.
\newblock In {\em The Thirty-Third {AAAI} Conference on Artificial
  Intelligence, {AAAI} 2019, The Thirty-First Innovative Applications of
  Artificial Intelligence Conference, {IAAI} 2019, The Ninth {AAAI} Symposium
  on Educational Advances in Artificial Intelligence, {EAAI} 2019, Honolulu,
  Hawaii, USA, January 27 - February 1, 2019}, pages 9291--9298. {AAAI} Press.

\bibitem[\protect\astroncite{Zhou et~al.}{2003}]{DBLP:conf/nips/ZhouBLWS03}
Zhou, D., Bousquet, O., Lal, T.~N., Weston, J., and Sch{\"{o}}lkopf, B. (2003).
\newblock Learning with local and global consistency.
\newblock In Thrun, S., Saul, L.~K., and Sch{\"{o}}lkopf, B., editors, {\em
  Advances in Neural Information Processing Systems 16 [Neural Information
  Processing Systems, {NIPS} 2003, December 8-13, 2003, Vancouver and Whistler,
  British Columbia, Canada]}, pages 321--328. {MIT} Press.

\bibitem[\protect\astroncite{Zhou and Sch{\"o}lkopf}{2004}]{zhou2004workshop}
Zhou, D. and Sch{\"o}lkopf, B. (2004).
\newblock A regularization framework for learning from graph data.
\newblock In {\em Workshop on Statistical Relational Learning at International
  Conference on Machine Learning}.

\bibitem[\protect\astroncite{Zhu}{2005}]{ZhuSemiSupervised}
Zhu, X. (2005).
\newblock {\em Semi-supervised Learning with Graphs}.
\newblock PhD thesis, Pittsburgh, PA, USA.

\bibitem[\protect\astroncite{Zhu and Ghahramani}{2002}]{zhu2002learning}
Zhu, X. and Ghahramani, Z. (2002).
\newblock Learning from labeled and unlabeled data with label propagation.

\bibitem[\protect\astroncite{Zhu et~al.}{2003}]{DBLP:conf/icml/ZhuGL03}
Zhu, X., Ghahramani, Z., and Lafferty, J.~D. (2003).
\newblock Semi-supervised learning using gaussian fields and harmonic
  functions.
\newblock In Fawcett, T. and Mishra, N., editors, {\em Machine Learning,
  Proceedings of the Twentieth International Conference {(ICML} 2003), August
  21-24, 2003, Washington, DC, {USA}}, pages 912--919. {AAAI} Press.

\end{thebibliography}

%\setboolean{@twoside}{false}
%\includepdf[pages=-, offset=75 -75]{Liberatoria_signed.pdf}

\end{document}